%% file: main.tex
\documentclass[acmtog,nonacm]{acmart}

\usepackage{booktabs} 
\usepackage{caption}
\usepackage{subcaption}
\usepackage{multirow}

\usepackage[ruled]{algorithm2e} 

\SetAlFnt{\small}
\SetAlCapFnt{\small}
\SetAlCapNameFnt{\small}
\SetAlCapHSkip{0pt}

\input{shortcuts}

\graphicspath{{./gfx/}}

\begin{document}
\title{PASTA: Controllable Part-Aware Shape Generation with Autoregressive Transformers}

\author{Songlin Li}
\affiliation{%
 \institution{Stanford University}
 \streetaddress{353 Jane Stanford Way}
 \city{Stanford}
 \state{CA}
 \postcode{94305}
 \country{USA}}
\email{svli97@stanford.edu}
\author{Despoina Paschalidou}
\affiliation{%
 \institution{Stanford University}
 \streetaddress{353 Jane Stanford Way}
 \city{Stanford}
 \state{CA}
 \postcode{94305}
 \country{USA}
}
\email{paschald@stanford.edu}
\author{Leonidas Guibas}
\affiliation{%
 \institution{Stanford University}
 \streetaddress{353 Jane Stanford Way}
 \city{Stanford}
 \state{CA}
 \postcode{94305}
 \country{USA}
 }
\email{guibas@stanford.edu}


%
%

%
%


\input{sec_abstract}

\maketitle

\input{sec_intro}
\input{sec_related}
\input{sec_method}
\input{sec_results}
\input{sec_conclusion}

\bibliographystyle{ACM-Reference-Format}
\bibliography{bibliography_long,bibliography,bibliography_custom}
\input{fig/shape_completion_diversity}
\input{fig/shape_generation_image_guided}
\input{fig/shape_generation_qualitative_lamps}
\input{fig/shape_generation_qualitative_all}
\input{fig/shape_completion_qualitative_lamps}
\input{fig/shape_completion_qualitative_tables}

\cleardoublepage
\input{supp_abstract}
\input{supp_implementation_details}
\input{supp_data_processing}
\input{supp_ablations}
\input{supp_additional_results}
\input{supp_discussion}
\end{document}

%% file: shortcuts.tex
\newcommand{\bB}{\mathbf{B}}
\newcommand{\bc}{\mathbf{c}}

\newcommand{\bF}{\mathbf{F}} 

\newcommand{\bo}{\mathbf{o}}

\newcommand{\bq}{\mathbf{q}}

\newcommand{\bs}{\mathbf{s}}
\newcommand{\bt}{\mathbf{t}}

\newcommand{\bx}{\mathbf{x}}
\newcommand{\by}{\mathbf{y}}
\newcommand{\bz}{\mathbf{z}}

\newcommand{\cD}{\mathcal{D}}

\newcommand{\cL}{\mathcal{L}}

\newcommand{\cP}{\mathcal{P}}

\newcommand{\cS}{\mathcal{S}}

\newcommand{\cX}{\mathcal{X}}
\newcommand{\cY}{\mathcal{Y}}

\DeclareMathOperator*{\argmin}{argmin~}

\makeatletter
\DeclareRobustCommand\onedot{\futurelet\@let@token\@onedot}
\def\@onedot{\ifx\@let@token.\else.\null\fi\xspace}
\def\eg{e.g\onedot} 
\def\ie{i.e\onedot} 
 
\def\etc{etc\onedot}

\def\wrt{wrt\onedot}

\def\etal{et~al\onedot}

\makeatother

\newcommand{\boldparagraph}[1]{\vspace{0.2cm}\noindent{\bf #1:} }

\newcommand{\figref}[1]{Fig.~\ref{#1}}
\newcommand{\tabref}[1]{Tab.~\ref{#1}}
\newcommand{\secref}[1]{Sec.~\ref{#1}}

\definecolor{darkred}{rgb}{0.7,0.2,0.1}
\definecolor{darkgreen}{rgb}{0,0.7,0}
\definecolor{orange}{RGB}{255,127,0}
\definecolor{ourpurple}{RGB}{127,127,204}
\definecolor{palgreen}{RGB}{51,179,179}
\definecolor{magenta}{RGB}{199,21,133}


%% file: sec_abstract.tex
\begin{abstract}
The increased demand for tools that automate the 3D content creation process 
led to tremendous progress in deep generative models that can generate diverse
3D objects of high fidelity. In this paper, we present PASTA, an autoregressive
transformer architecture for generating high quality 3D shapes. PASTA comprises two
main components: An autoregressive transformer that generates objects as a
sequence of cuboidal primitives and a blending network, implemented with a
transformer decoder that composes the sequences of cuboids and synthesizes high
quality meshes for each object. Our model is trained in two stages: First we train
our autoregressive generative model using only annotated cuboidal parts as
supervision and next, we train our blending network using explicit 3D
supervision, in the form of watertight meshes.  Evaluations on various ShapeNet
objects showcase the ability of our model to perform shape generation from
diverse inputs \eg from scratch, from a partial object, from text and images,
as well size-guided generation, by explicitly
conditioning on a bounding box that defines the object's boundaries. Moreover, as our model
considers the underlying part-based structure of a 3D object, we are able to select a
specific part and produce shapes with meaningful variations of this part.
As evidenced by our experiments, our model generates 3D shapes that are both
more realistic and diverse than existing part-based and non part-based methods,
while at the same time is simpler to implement and train.
\end{abstract}

%% file: sec_intro.tex
\section{Introduction}
\label{sec:intro}

The ability to generate realistic and diverse 3D shapes
has the potential to significantly accommodate the workflow
of artists and content creators and potentially enable new levels of creativity
through "generative art" \cite{Bailey2020ArtInAmerica}.
The tremendous progress in generative modelling and implicit-based
representations gave rise to several works \cite{Schwarz2020NEURIPS,
Chan2021CVPR, Chan2022CVPR, Gu2022ICLR, Gao2022NEURIPS} that generate
objects with high realism in terms of geometric details and texture.
Nevertheless, as these pipelines represent objects
holistically, \ie without taking into consideration the underlying part-based
structure of each object, they only support few interactive applications that
typically require deep technical knowledge of each model. However, shape
editing and manipulation involves controlling what parts of the object need to
be changed. To enable this level of control, an active area of research proposes
to consider the decomposition of shapes into parts
\cite{Mo2019SIGGRAPH, Hao2020CVPR, Mo2020CVPR, Wu2020CVPR, Gadelha2020CVPR,
Deprelle2019NIPS, Li2021SIGGRAPH, Hertz2022SIGGRAPH}.

\begin{figure}
    \centering
    \includegraphics[width=\linewidth]{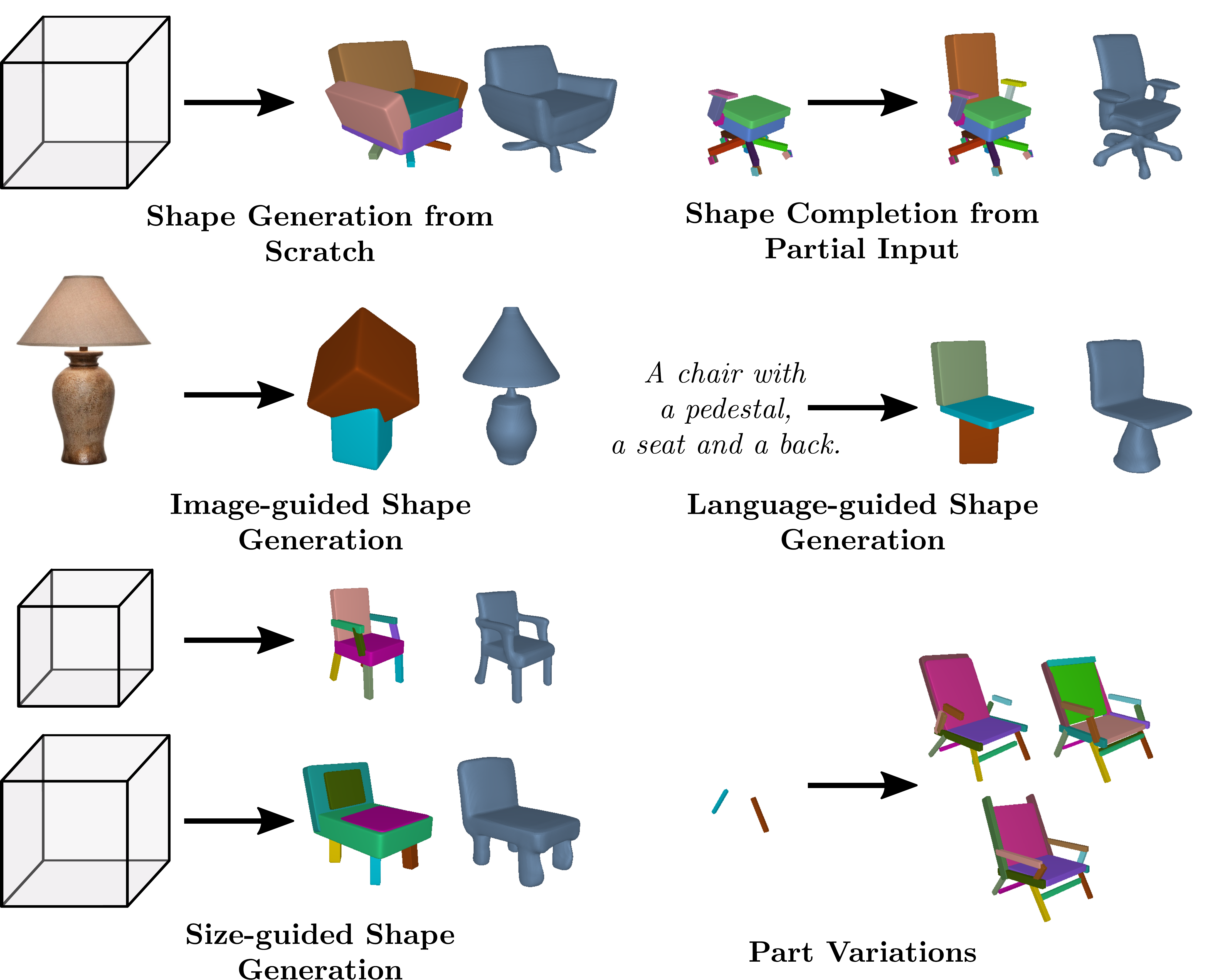}
    \vspace{-1.2em}
    \caption{{\bf Controllable Part-Aware 3D Shape Generation}. We
    propose a novel autoregressive architecture that can be used to perform
    several editing tasks, such as generating novel shapes
    from scratch, conditioned on a bounding box defining the object's
    boundaries, completing a 3D shape from a partial input, a text,
    an image or bounding boxes of different sizes, as well as generating
    plausible variations for specific parts of the object.
    }
    \label{fig:teaser}
    \vspace{-1.5em}
\end{figure}

Existing part-based generative models, represent 3D shapes as a collection of
simple shapes parametrized with cuboids \cite{Zou2017ICCV, Mo2019SIGGRAPH,
Wu2020CVPR, Mo2020CVPR}, spheres \cite{Hao2020CVPR}, implicit fields
\cite{Wu2020CVPR, Hertz2022SIGGRAPH} or more general handles
\cite{Gadelha2020CVPR, Liu2021CVPR}, and seek to synthesize new shapes in accordance to
their underlying structural understanding of the object.
For example, among the first to explore structure-based generative models were
\cite{Li2017SIGGRAPH, Mo2019SIGGRAPH} that utilized an autoencoder for
generating structured shapes. Despite their impressive capabilities on shape
generation and interpolation neither can perform shape completion from a
partial input or generate plausible part variations. Similarly, while
\cite{Zou2017ICCV, Wu2020CVPR} can generate 3D shapes as a sequences of parts,
they need to train a separate
model for different editing tasks, namely the same model cannot perform both
shape generation and shape completion, which makes their approach impractical.

To address these limitations, we devise PASTA, a novel part-aware generative model for 3D shapes.
PASTA comprises
two main components: An autoregressive transformer encoder that generates
shapes as unordered sets of parts and a blending network, implemented as a transformer decoder that
combines the part sequences and produces high-quality meshes. Each
component of our architecture is trained independently. In particular, we optimize our
autoregressive transformer to maximize the log-likelihood of all part
arrangements in the dataset. Our supervision comes in the form of part labels and 3D
cuboids that specify the per-part size and pose. Unlike existing autoregressive
pipelines \cite{Tulsiani2021ICML, Paschalidou2021NEURIPS} that are trained
using teacher forcing, we train PASTA using scheduled sampling
\cite{Bengio2015NIPS, Mihaylova2019ARXIV} and showcase that it significantly
improves the generation performance of our model. To
train our blending network, we consider 3D supervision in the form of
watertight meshes and optimize it to reconstruct 3D shapes as implicit occupancy
fields \cite{Mescheder2019CVPR}.
We evaluate the performance of our model on several PartNet~\cite{Mo2019CVPR}
objects and demonstrate that our model can produce more realistic and diverse
3D objects in comparison to both part-based~\cite{Wu2020CVPR,
Paschalidou2021NEURIPS} and non part-based methods \cite{Chen2019CVPR}.
Furthermore, we showcase that our model can generate meaningful part
arrangements conditioned on versatile user input (\figref{fig:teaser}) including but
not limited to text and images.

In summary we make the following \textbf{contributions}: We propose the first
part-aware generative model using an autoregressive transformer architecture.
Our experiments on various PartNet objects
\cite{Mo2019CVPR} demonstrate that our model generates more diverse and plausible 3D
shapes in comparison to part-based~\cite{Wu2020CVPR,
Paschalidou2021NEURIPS} and non part-based methods \cite{Chen2019CVPR}.
Furthermore, our simple, yet effective architecture allows training a single
model capable of performing several editing operations, such as generating new
objects from scratch, generating part variations or completing partial shapes.

%% file: sec_related.tex
\section{Related Work}
\label{sec:related}

\emph{3D Representations }
Learning-based approaches for 3D reconstruction employ a neural network that
learns a function from the input to a mesh \cite{Liao2018CVPR,
Groueix2018CVPR, Kanazawa2018ECCV, Wang2018ECCV, Yang2019ICCV, Pan2019ICCV}, a
pointcloud \cite{Fan2017CVPR, Qi2017NIPS, Achlioptas2018ICML, Jiang2018ECCV,
Thomas2019ICCV, Yang2019ICCV}, a voxel grid \cite{Brock2016ARXIV, Choy2016ECCV,
Gadelha2017THREEDV, Rezende2016NIPS, Riegler2017CVPR, Stutz2018CVPR,
Xie2019ICCV} or an implicit surface \cite{Mescheder2019CVPR, Chen2019CVPR,
Park2019CVPR, Saito2019ICCV, Xu2019NIPS, Michalkiewicz2019ICCV}. Unlike
explicit representations that discretize the output space,
using voxels, points or mesh vertices,
implicit representations represent shapes
in the weights of a neural network that learns a mapping between
a query point and a context vector to a signed distance value
\cite{Park2019CVPR, Michalkiewicz2019ICCV, Atzmon2020CVPR, Gropp2020ICML,
Takikawa2021CVPR} or a binary occupancy value \cite{Mescheder2019CVPR,
Chen2019CVPR}. As these methods require 3D supervision, several works
propose combining them with surface 
\cite{Niemeyer2020CVPR, Yariv2020NIPS} or volumetric \cite{Mildenhall2020ECCV}
rendering to learn the 3D object geometry and texture directly from
images. In this work, we introduce a part-aware generative model that
parametrizes shapes as an occupancy field \cite{Mescheder2019CVPR}.

\emph{Primitive-based Representations }
Shape abstraction techniques represent shapes using semantically consistent
part arrangements
and seek to recover 
the 3D geometry using simple shapes such as
cuboids \cite{Tulsiani2017CVPRa, Niu2018CVPR, Zou2017ICCV, Li2017SIGGRAPH,
Mo2019SIGGRAPH, Dubrovina2019ICCV}, superquadrics \cite{Paschalidou2019CVPR,
Paschalidou2020CVPR}, convex solids \cite{Deng2020CVPR, Chen2020CVPR,
Gadelha2020CVPR}, spheres \cite{Hao2020CVPR} and 3D Gaussians
\cite{Genova2019ICCV}. In recent work, \cite{Paschalidou2021CVPR} proposed to
represent objects as a family of homeomorphic mappings, parametrized with an
Invertible Neural Network (INN) \cite{Dinh2015ICLR}. Likewise,
\cite{Genova2020CVPR} suggested to represent 3D objects using a structured set
of implicit functions \cite{Genova2019ICCV}. Another line of research employs
primitives to recover the 3D geometry using CSG trees
\cite{Sharma2018CVPR} or shape programs \cite{Ellis2018NIPS, Tian2019ICLR,
Liu2019ICLR}. Very recently \cite{Yao2021ICCV, Yao2022ARXIV}, explored learning
primitive-based representations only from images.
Unlike these works that focus on recovering the object geometry as a
collection of parts, we introduce a part-aware generative model that
synthesizes novel objects as a set of cuboidal primitives.

\begin{figure}
    \centering
    \includegraphics[width=\linewidth]{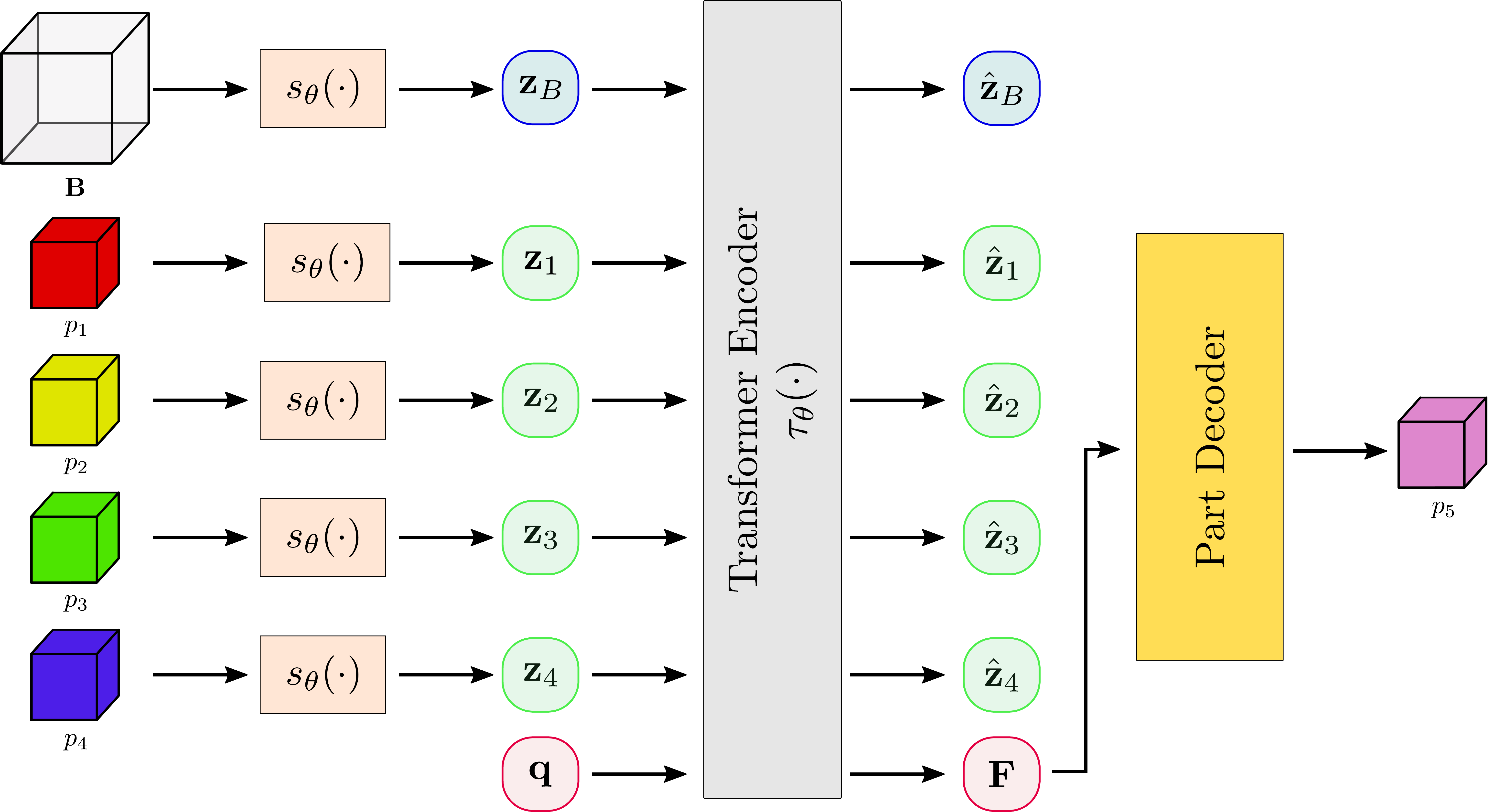}
    \caption{\small 
        {\bf{Object Generator.}} Given a sequence of $N$ parts and a bounding box $\bB$
        defining the object boundaries, the \emph{part encoder} $s_\theta(\cdot)$
        maps each part $p_j$ and the bounding box
        to an embedding vector. The bounding box's embedding vector $\bz_B$, the per-part
        embeddings $\{\bz_j\}_{i=1}^N$ and a learnable embedding vector $\bq$ are passed to the
        \emph{transformer decoder} that predicts a feature vector $\bF$ used to predict
        the attributes of the next part in the sequence. The \emph{part decoder}
        takes $\bF$ and autoregressively predicts the attribute distributions that are used
        to sample the attributes for the next part.
        }
    \label{fig:object_generator}
    \vspace{-1.75em}
\end{figure}

\emph{3D Generative Models }
Generative Adversarial Networks (GANS) \cite{Goodfellow2014NIPS} have
demonstrated impressive capabilities on several image synthesis
\cite{Brock2019ICLR, Choi2018CVPR, Huang2018ARXIV, Karras2019CVPR,
Karras2020CVPR} and editing \cite{Wang2018CVPRa, Shen2020CVPR, Alharbi2020NIPS,
Ling2021NIPS, Wang2021ICLR, Isola2017CVPR, Zhu2017ICCV, Choi2018CVPR} tasks.
However, adapting them to 3D content creation is non-trivial as they ignore
the 3D nature of the world and hence, lack an understanding of the
object's underlying geometry. To address this, 3D-aware GANs proposed to
incorporate 3D representations such as voxel grids \cite{Henzler2019ICCV,
Nguyen2019ICCV, Nguyen2020NeurIPS} in generative settings or combine them with
differentiable renderers \cite{Liao2020CVPRa, Zhang2021ICLR}. As an
alternative, several works explored generating 3D shapes as octrees
\cite{Ibing2021ARXIV}, pointclouds \cite{Achlioptas2018ICML, Yang2019ICCV,
Cai2020ECCV, Zhou2021ICCV, Luo2021CVPR, Li2021SIGGRAPH}, meshes
\cite{Nash2020ICML, Luo2021ICCV, Pavllo2021ICCV, Pavllo2020NEURIPS} and implicit
functions \cite{Mescheder2019CVPR, Chen2020CVPR}. While
these methods yield realistic geometries, they do not consider the part-based
object structure.
In contrast, we propose a part-aware generative model that generates shapes as
an unordered set of cuboids, which are then combined and synthesize
a high quality implicit shape.

\emph{Part-based Generative Models }
Our work is falls into the category of part-based generative models.
Zou \etal \cite{Zou2017ICCV} was among the first to
introduce a generative recurrent model, parametrized with LSTMs
\cite{Hochreiter1997NC} in combination with a Mixture Density Network (MDN) to
synthesize novel objects as a set of cuboids. Concurrently, Li \etal
\cite{Li2017SIGGRAPH} proposed to represent shapes using a symmetry hierarchy,
which defines how parts are recursively grouped by symmetry and assembled by
connectivity \cite{Wang2011EUROGRAPHICS}. In particular, they utilize an
RNN and generate objects as bounding box layouts, which
are then filled with voxelized parts. Likewise, StructureNet
\cite{Mo2019SIGGRAPH} utilizes a VAE~\cite{Kingma2015ICLR} and generates novel
shapes as n-ary graphs, where every node in the graph is associated with a
bounding box. Note that while \cite{Mo2019SIGGRAPH} can generate plausible new
shapes, perform shape and part interpolations, their model cannot be utilized for completion
from unconstrained inputs, \eg a partial object.
Furthermore, to perform shape editing, their model requires additional optimization steps in order to find a new shape in
the latent space that satisfies a specific edit. Concurrently,
\cite{Mo2020CVPR}, explored learning a latent space of structured shape
differences, using pairs of structured shapes demonstrating a specific edit.
Closely related to our work is
PQ-NET~\cite{Wu2020CVPR} that generates shapes autoregressively using an RNN
autoencoder. The RNN encoder takes a 3D shape segmented into parts and extracts
per-part features, which are then fed to the decoder that sequentially predicts
parts that reconstruct the input shape.
To be able to generate new shapes, they train a latent GAN
\cite{Achlioptas2018ICML} on the latent space of the autoencoder.
Moreover, to perform different editing tasks, they need to train different
variants of their model. Instead, our formulation allows applying a single model trained for
object completion on a variety of tasks. Furthermore, instead of using an RNN,
we employ an autoregressive transformer that synthesizes objects as a sequence
of cuboids, which are passed to our blending network to produce the final high-quality shape.

\begin{figure}
    \centering
    \includegraphics[width=\linewidth]{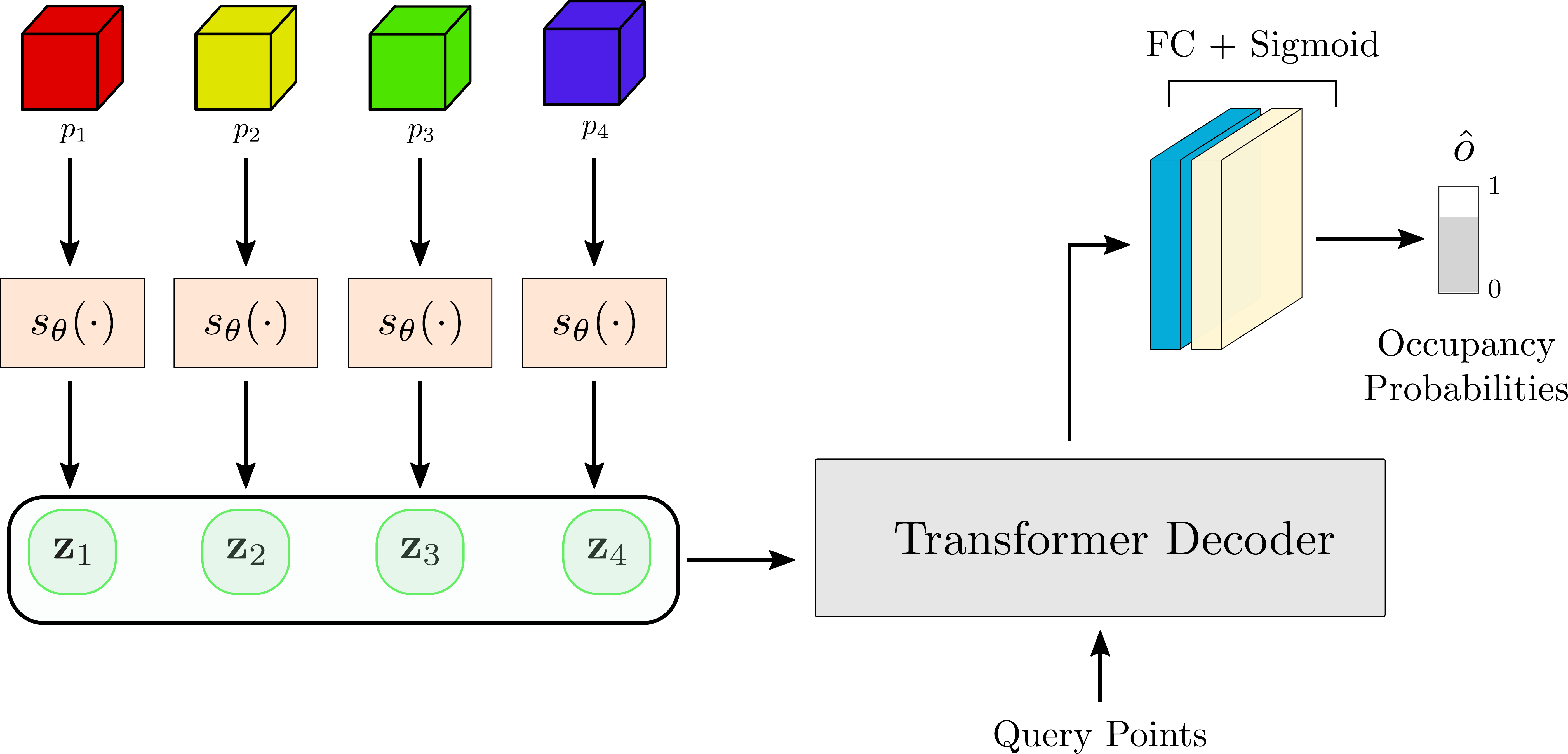}
    \caption{\small 
        {\bf{Blending Network.}} Given a sequence of $N$ parts,
        the \emph{part encoder} maps them into 
        embedding vectors $\{\bz_j\}_{i=1}^N$. We pass the per-part embedding
        vectors and a set of 3D query points $\cX$ to the 
        \emph{transformer decoder} that predicts the occupancy probabilities
        for the query points.}
    \label{fig:blending_network}
    \vspace{-1.5em}
\end{figure}
\emph{Autoregressive Transformers for Content Creation }
Transformer-based architectures \cite{Vaswani2017NIPS} have been extensively utilized
for various autoregressive tasks ranging from machine translation
\cite{Shaw2018NAACL, Ott2018WMT} to image \cite{Parmar2018ICML,
Katharopoulos2020ICML, Chen2020ICML, Esser2021CVPR, Tulsiani2021ICML} and music
\cite{Dhariwal2020ARXIV} generation, as well as
shape completion \cite{Yu2021ICCV, Mitta2022CVPR} and indoor scene synthesis
\cite{Wang2020ARXIV, Paschalidou2021NEURIPS}. Similar to ATISS
\cite{Paschalidou2021NEURIPS}, that is an autoregressive transformer for scene synthesis,
we pose object synthesis as an
autoregressive prediction problem and generate objects as a sequence of
cuboids. However, unlike ATISS that is trained with
teacher forcing, we train our model using scheduled sampling
\cite{Bengio2015NIPS, Mihaylova2019ARXIV} and showcase that it improves
the generation capabilities of our model.

%% file: sec_method.tex
\section{Method}
\label{sec:method}

PASTA is a part-aware generative model for synthesizing 3D shapes. Our
model comprises two main components that are trained independently: An
\emph{object generator} that sequentially generates objects as unordered
sequences of labelled parts, where each part is parametrized using a 3D
cuboidal primitive (\secref{subsec:objects_representation}), and a
\emph{blending network} that composes part sequences in a meaningful way and
synthesizes high quality implicit shapes. The \emph{object generator} is an
autoregressive model trained to maximize the log-likelihood of all possible
part arrangements in the dataset. We use part-level
supervision in the form of part labels and 3D cuboids that define the size and
the pose of each part (\secref{subsec:object_generator}). The
\emph{blending network} is an occupancy network \cite{Mescheder2019CVPR},
implemented with a transformer decoder that takes a sequence of cuboids and a
query 3D point and predicts whether this point is inside or outside the
surface boundaries (\secref{subsec:blending_network}). To train
the blending network, we assume explicit 3D supervision in the form of a
watertight mesh.

\subsection{Object Parametrization}
\label{subsec:objects_representation}

We define each object $\cS = \{\cP, \bB\}$ using a collection of $N$
parts, $\cP = \{p_1, \dots, p_N\}$, and a 3D bounding box $\bB$ that specifies
the object's boundaries.
Each part is represented with a 3D labelled cuboid and is parametrized using four
values describing its label, size, translation and rotation, $p_j =
\{\bc_j, \bs_j, \bt_j, \bo_j\}$. To model the label of each part, \ie the back
or the arm of a chair, we use a categorical distribution defined over the total
number of part labels in the dataset, whereas for the rest of the attributes we
use a mixture of logistic distributions \cite{Salimans2017ICLR, Oord2016SSW}.
In our setup, the rotation $\bo_j \in \mathbb{R}^6$ is the 6D representation
\cite{Zhou2019CVPR} of the 3D cuboid that contains the part. Likewise, the
translation $\bt_j \in \mathbb{R}^3$ is the center of the 3D cuboid containing
the part and the size $\bs_j \in \mathbb{R}^3$ is its width, height and depth.
Similarly, the bounding box $\bB$ is defined using $12$ parameters: three for its
size along each axis, three for the translation, which is the center of the box
and six for the rotation, which is a 6D representation.

Similar to ATISS~\cite{Paschalidou2021NEURIPS}, we predict the components of
$p_j$ autoregressively, namely part label first, followed by translation,
rotation and size. Hence, the probability of generating the $j$-th part,
conditioned on the previous parts and box $\bB$
becomes
\begin{equation}
    \begin{aligned}
    p_{\theta}(p_j \mid p_{<j}, \bB) =
        &p_{\theta}^c(\bc_j | p_{<j}, \bB)
        p_{\theta}^t(\bt_j |\bc_j, p_{<j}, \bB)\\
        &p_{\theta}^o(\bo_j |\bc_j, \bt_j, p_{<j}, \bB)
        p_{\theta}^s(\bs_j |\bc_j, \bt_j, \bo_j, p_{<j}, \bB),
    \end{aligned}
    \label{eq:part_likelihood}
\end{equation}
where $p_{\theta}^c$, $p_{\theta}^t$, $p_{\theta}^o$ and $p_{\theta}^s$
are the probabilities for the respective attributes.
Assuming a fixed ordering \wrt the part attributes is reasonable, as
we want our model to consider the part label before predicting its pose and
size. To compute the likelihood of generating an object $\cS$, we estimate
the likelihood of autoregressively generating its parts $\cP$ using any order,
as \cite{Paschalidou2021NEURIPS} demonstrated that not having a fixed ordering 
can be beneficial. Hence, the likelihood of generating an object $\cS$ conditioned on
a bounding box $\bB$ is
\begin{equation}
    p_{\theta}(\cS | \bB) = 
        \sum_{\hat{\cP}\in \pi(\cP)}
        \prod_{j\in \hat{\cP}}
            p_{\theta}(p_j \mid p_{<j}, \bB),
    \label{eq:object_probability}
\end{equation}
where $\pi(\cdot)$ is a permutation function that computes the set of
permutations of all object parts and $\hat{\cP}$ denotes an ordered part
sequence.

\subsection{Object Generator}
\label{subsec:object_generator}

The input to our \emph{object generator} is a set of objects in the form
of 3D labelled cuboids and their corresponding bounding boxes.
We implement our generator using an autoregressive transformer architecture
similar to ATISS.
The transformer model takes as input a sequence of embedding vectors that
represent the conditioning sequence and generates the features
$\bF$ that will be used to predict the attributes of the next part.
We map the per-part attributes to embedding vectors using a \emph{part encoder}
and the features $\bF$ to part attributes using a \emph{part decoder}.
Our model is illustrated in \figref{fig:object_generator}.

The part encoder network $s_\theta(\cdot)$ takes the attributes for each part $p_j=\{\bc_j,
\bs_j, \bt_j, \bo_j\}$ and maps them to an embedding vector $\bz_j$ 
\begin{equation}
    \bz_j = s_\theta \big( \left[ \lambda(\bc_j); \gamma(\bs_j); \gamma(\bt_j); \gamma(\bo_j) \right] \big),
    \label{eq:part_encoder}
\end{equation}
where $\lambda(\cdot)$ is a learnable embedding, $\gamma(\cdot)$ is a
positional encoding layer \cite{Vaswani2017NIPS} that is applied separately on
each attribute's dimension and $[\cdot\,;\cdot]$ denotes concatenation. To 
predict an embedding vector $\bz_B$ for $\bB$, we pass its attributes to $s_\theta(\cdot)$.

Similar to ATISS \cite{Paschalidou2021NEURIPS}, we implement our transformer
encoder $\tau_{\theta}(\cdot)$ as a multi-head attention transformer
without positional encoding \cite{Vaswani2017NIPS}, as we want to model objects
as unordered sets of parts. Our transformer encoder takes as
input $\{\bz_j\}_{j=1}^N$ the $N$ embeddings for all parts in the sequence,
$\bz_B$ the embedding for the bounding box and a learnable embedding
vector $\bq$, which is used to predict the feature vector $\bF$ that will be used to generate
the next part in the sequence. More formally,
\begin{equation}
    \bF = \tau_\theta \left(\bz_B, \{\bz_j\}_{j=1}^N, \bq \right).
    \label{eq:transformer_encoder}
\end{equation}

The last component of the object generator is the part decoder that takes as input the feature vector
$\bF$ and autoregressively predicts the attributes of the next part to be
generated. For the part label, we define a function $c_\theta(\cdot)$, implemented using a linear
projection layer, that takes $\bF$ and predicts the per-part label probability.
We predict the size, translation and rotation in two-stages.  First, we cluster
the values of each attribute from the training set into $20$ clusters using
K-Means. Subsequently, we predict a cluster for each attribute, which is then used
to predict the specific values.
More formally, for
the translation, we learn $t_{\theta}^{\text{coarse}}(\cdot)$ that predicts the
per-cluster probability from $\bF$ using a linear projection layer and
$t_{\theta}^{\text{fine}}(\cdot)$ that predicts the $7 \times K$ parameters
that define the mixture of logistics distribution for the translation. Specifically,
we have $K$ parameters for the mixing coefficients and $6 \times K$ for the means and variances.
In a similar manner, we define $o_{\theta}^{\text{coarse}}(\cdot)$
and $o_{\theta}^{\text{fine}}(\cdot)$ to predict the $13 \times K$ parameters
that define the mixture of logistics distribution for the
rotation and $s_{\theta}^{\text{coarse}}(\cdot)$ and $s_{\theta}^{\text{fine}}(\cdot)$
that predict the $7 \times K$ parameters that define the mixture of logistic distribution
for the size. To predict the part attributes in an autoregressive manner, we condition the prediction
of each attribute to the values of the previously predicted ones. In practice,
$t_\theta^{\text{coarse}}(\cdot)$, $t_\theta^{\text{fine}}(\cdot)$, $o_\theta^{\text{coarse}}(\cdot)$,
$o_\theta^{\text{fine}}(\cdot)$, $s_\theta^{\text{coarse}}(\cdot)$ and
$s_\theta^{\text{fine}}(\cdot)$ take as input $\bF$ concatenated with the
previously predicted attributes embedded by $\lambda(\cdot)$ and
$\gamma(\cdot)$ from \eqref{eq:part_encoder}.

\begin{figure}
    \centering
    \includegraphics[width=\linewidth]{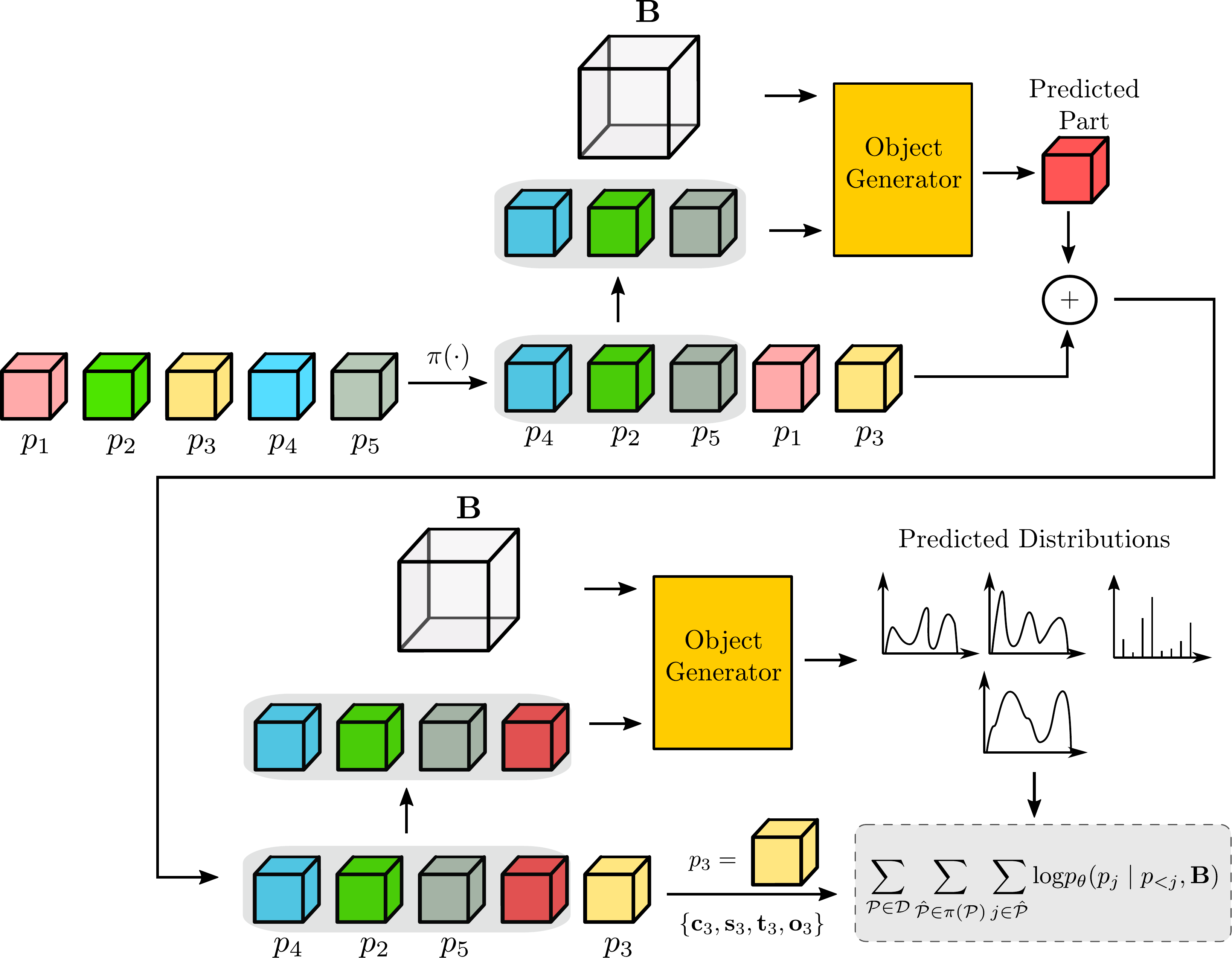}
    \caption{\small 
        {\bf{Scheduled Sampling.}} Given an object with $N$ parts, we
        first randomly permute them and keep the first $M$ parts (here $M=3$).
        We pass them to the object generator that predicts the next part to be generated
        (red cube). The newly generated cuboid is appended to the initial sequence
        with the $M$ objects and passed once again to the object generator, that
        predicts the next part to be generated. Our loss
        function from \eqref{eq:loss} is computed between the new part and the $M+2$
        part in the permutated sequence (yellow cube).
        }
    \label{fig:scheduled_sampling}
    \vspace{-1.5em}
\end{figure}

\subsection{Blending Network}
\label{subsec:blending_network}
The input to the \emph{blending network} is a sequence of labelled cuboids and
a set of 3D query points $\cX$, for which we want to predict their occupancy
probabilities, namely whether they lie inside or outside the surface
boundaries. In detail, our blending network consists of two main components:
(i) a \emph{part-encoder} that maps the part attributes into an embedding
vector, which is implemented as discussed in \secref{subsec:object_generator}
and (ii) a transformer decoder without self-attention that takes the part
embeddings and the query points and predicts whether they are inside or outside
the surface boundary (see \figref{fig:blending_network}).
The transformer comprises only cross attention layers and
MLPs, thus, each query 3D point attends to the per-part embeddings using cross
attention, without attending to the other points. The transformer output is
passed to a linear layer followed by a sigmoid non-linearity to get an
occupancy probability for each query point. We follow common practice and
before passing the query points to the decoder we map them to a higher
dimensional space using positional encoding~\cite{Vaswani2017NIPS}.

\input{fig/shape_generation_qualitative_chairs}
\input{tab/shapenet_generation_quantitative}
\subsection{Training and Inference}
\label{subsec:training_inference}

Unlike prior autoregressive models \cite{Ritchie2019CVPR,
Paschalidou2021NEURIPS} that are trained with teacher forced
embeddings, we train PASTA using scheduled sampling
\cite{Bengio2015NIPS}. The key idea is that during training we feed our model
with a mix of the teacher forced embeddings and the actual model's predictions
from the previous generation step. In particular, we
choose an object from the dataset and apply the permutation function
$\pi(\cdot)$ on its elements. Next, we randomly select the first $M$ objects
and pass them to the object generator that is used to predict 
the next part. The newly generated part is appended to the
initial sequence of $M$ parts and passed once again to the object generator to
predict the attribute distribution of the next part. Our model is trained to
maximize the likelihood of the $M+2$ object in the permuted sequence.
A pictorial representation of our scheduled sampling is provided in 
\figref{fig:scheduled_sampling}. We follow common practice and during training,
we train with both scheduled sampling and teacher forcing.

To train the object generator we follow \cite{Paschalidou2021NEURIPS} and maximize the likelihood
of generating all possible part sequences in the dataset $\cD$ in all possible
permutations, as follows
\begin{equation}
    \cL(\theta) =
        \sum_{\cP \in \cD}
        \sum_{\hat{\cP}\in \pi(\cP)}
        \sum_{j\in \hat{\cP}}
            \log p_{\theta}(p_j \mid p_{<j}, \bB).
    \label{eq:loss}
\end{equation}
For the mixture of logistic distributions, we use the discretized mixture of
logistics loss as defined in \cite{Oord2016SSW}.
To train the blending network, we assume 3D supervision in the form of a
watertight mesh, which we use to generate a set of occupancy pairs $\cX =
\{\{\bx_i, o_i\}\}_{i=1}^V$, namely a set of 3D points $\bx_i$ and their occupancy
labels $o_i$, denoting whether $\bx_i$ lies inside or outside the object. We
train the blending network using a classification loss between the predicted
and the target occupancies. Note that the blending network is trained
using ground-truth part sequences.

During inference, we start from a bounding box $\bB$ and
autoregressively sample the attribute values from the predicted distributions
for the next part to be generated. Once a new part is generated, it is used in
the next generation step until the \emph{end symbol} is predicted. To indicate
the end of sequence, we augment the part labels with an additional category,
which we refer to as \emph{end symbol}.  Once a complete sequence of parts is
generated, we pass it to the blending network that combines them into a single
implicit shape.

\subsection{Conditional Generation}
\label{subsec:inference}

Here, we discuss how PASTA can be used to perform language- and image-guided
generation.  Instead of conditioning the generation only on the bounding box
$B$, we now condition also on a textual description of the shape to be
generated. In particular, we utilize the pre-trained CLIP~\cite{Radford2021ICML}
model to extract text embeddings and pass them to the transformer encoder as an
additional input. Note that during training the pre-trained CLIP text encoder
remains frozen, namely is not optimized with the rest of our network. Once we
train PASTA with text embeddings from CLIP, we can use it, without any
re-training, also for image-guided generation.  This is possible because the
CLIP model has a joint latent space for text and images.  While, in our
experiments, we only demonstrate language- and image-guided generations, our
model can be extended to other types of conditioning such as depth maps or
pointclouds \etc by utilizing an appropriate encoder that generates embeddings
from the input.

%% file: fig/shape_generation_qualitative_chairs.tex
\begin{figure}
    \begin{subfigure}[t]{\linewidth}
    \centering
    \begin{subfigure}[b]{0.20\linewidth}
        \centering
	    IM-NET
    \end{subfigure}%
    \hfill%
    \begin{subfigure}[b]{0.20\linewidth}
	\centering
        PQ-NET
    \end{subfigure}%
    \hfill%
    \begin{subfigure}[b]{0.20\linewidth}
	\centering
        ATISS
    \end{subfigure}%
    \hfill%
    \begin{subfigure}[b]{0.20\linewidth}
        \centering
        Ours-Parts
    \end{subfigure}%
    \hfill%
    \begin{subfigure}[b]{0.20\linewidth}
        \centering
        Ours
    \end{subfigure}
    \end{subfigure}
    \vspace{-1.5em}
    \begin{subfigure}[b]{0.20\linewidth}
		\centering
		\includegraphics[width=\linewidth]{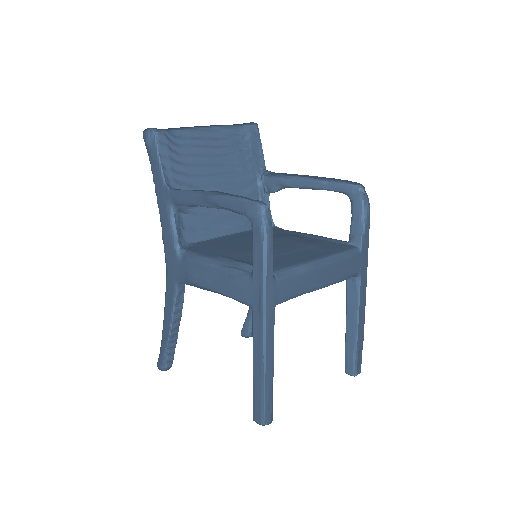}
    \end{subfigure}%
    \hfill%
    \begin{subfigure}[b]{0.20\linewidth}
		\centering
		\includegraphics[width=\linewidth]{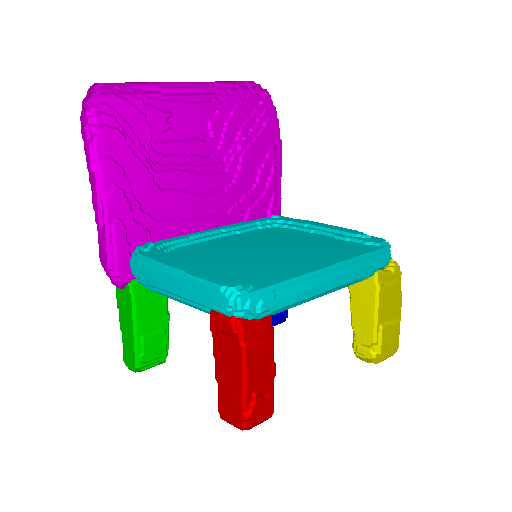}
    \end{subfigure}%
    \hfill%
    \begin{subfigure}[b]{0.20\linewidth}
		\centering
		\includegraphics[width=\linewidth]{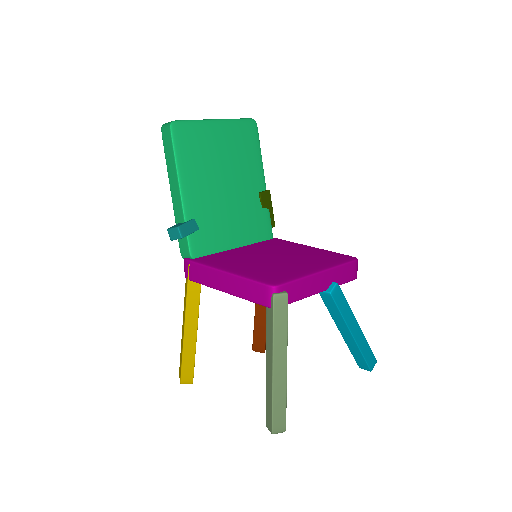}
    \end{subfigure}%
    \hfill%
    \begin{subfigure}[b]{0.20\linewidth}
		\centering
		\includegraphics[width=\linewidth]{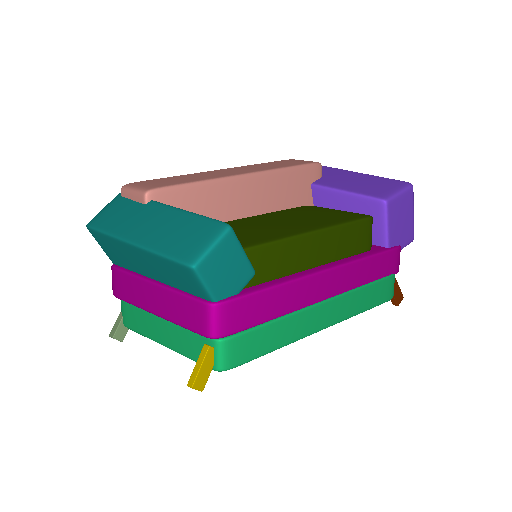}
    \end{subfigure}%
    \hfill%
    \begin{subfigure}[b]{0.20\linewidth}
		\centering
		\includegraphics[width=\linewidth]{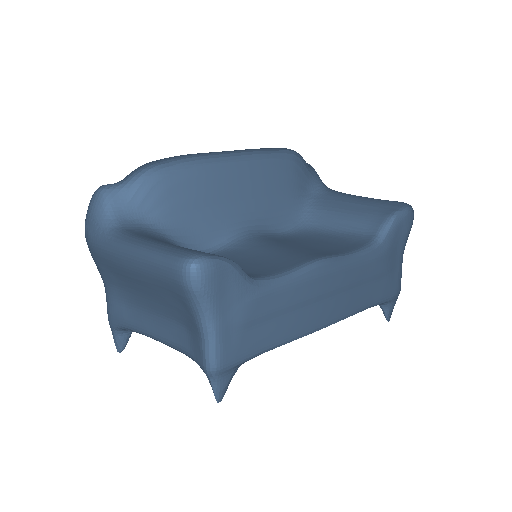}
    \end{subfigure}%
    \vskip\baselineskip%
    \vspace{-0.75em}
    \begin{subfigure}[b]{0.20\linewidth}
		\centering
		\includegraphics[width=\linewidth]{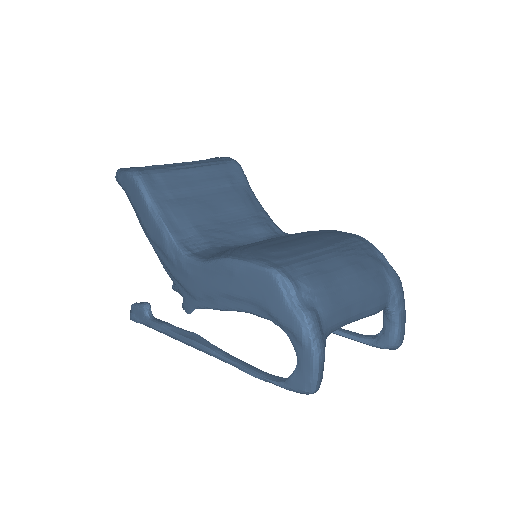}
    \end{subfigure}%
    \hfill%
    \begin{subfigure}[b]{0.20\linewidth}
		\centering
		\includegraphics[width=\linewidth]{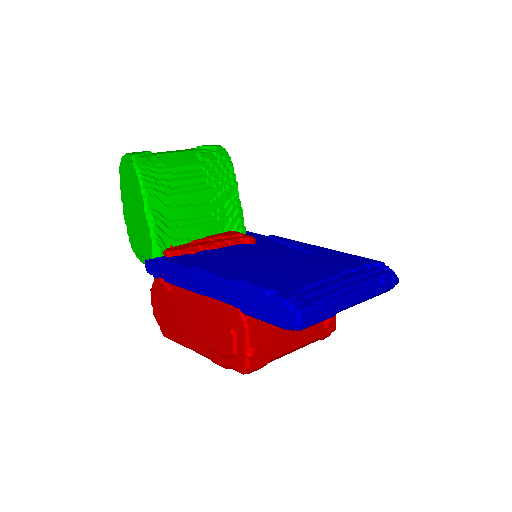}
    \end{subfigure}%
    \hfill%
    \begin{subfigure}[b]{0.20\linewidth}
		\centering
		\includegraphics[width=\linewidth]{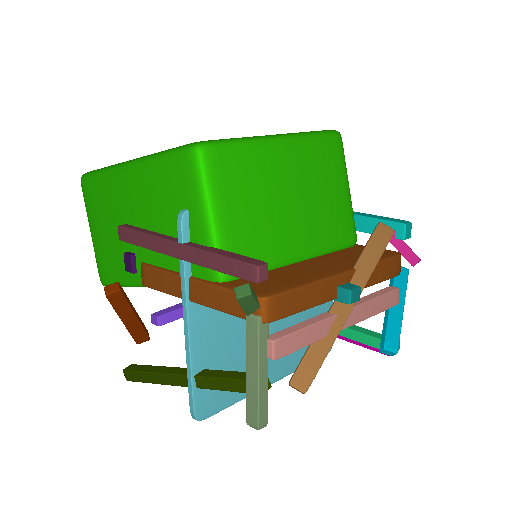}
    \end{subfigure}%
    \hfill%
    \begin{subfigure}[b]{0.20\linewidth}
		\centering
		\includegraphics[width=\linewidth]{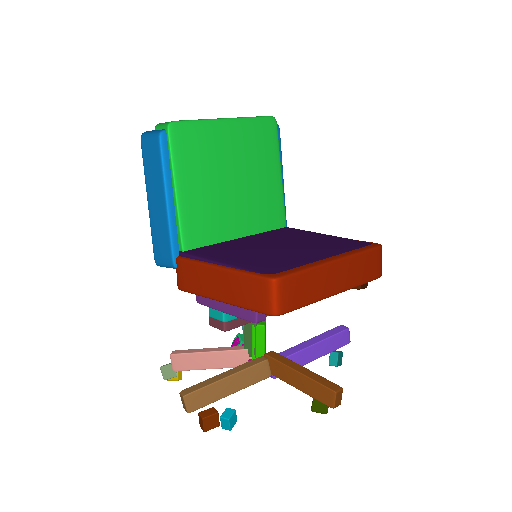}
    \end{subfigure}%
    \hfill%
    \begin{subfigure}[b]{0.20\linewidth}
		\centering
		\includegraphics[width=\linewidth]{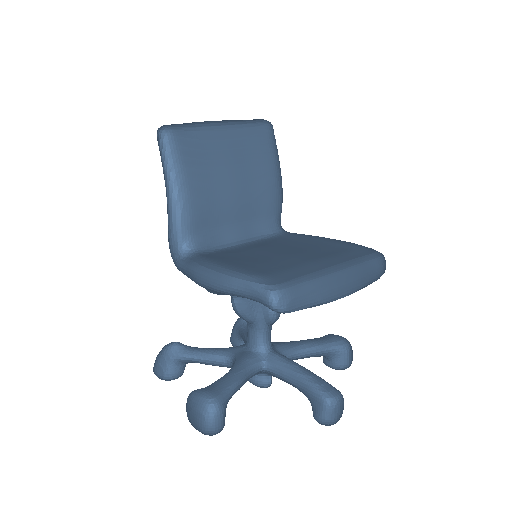}
    \end{subfigure}%
    \vskip\baselineskip%
    \vspace{-1.75em}
    \begin{subfigure}[b]{0.20\linewidth}
		\centering
		\includegraphics[width=\linewidth]{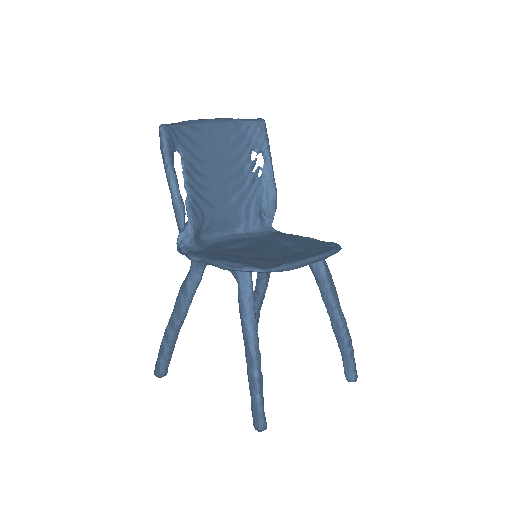}
    \end{subfigure}%
    \hfill%
    \begin{subfigure}[b]{0.20\linewidth}
		\centering
		\includegraphics[width=\linewidth]{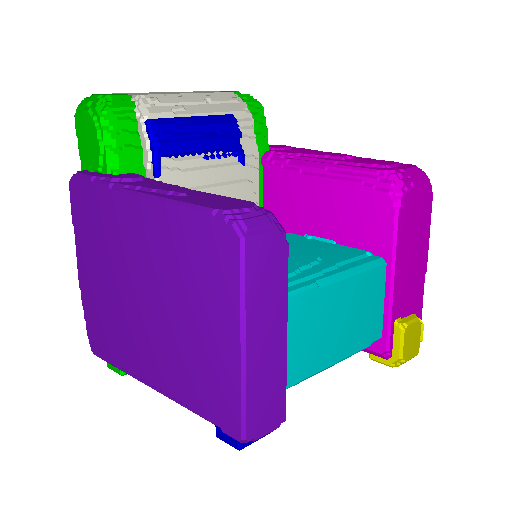}
    \end{subfigure}%
    \hfill%
    \begin{subfigure}[b]{0.20\linewidth}
		\centering
		\includegraphics[width=\linewidth]{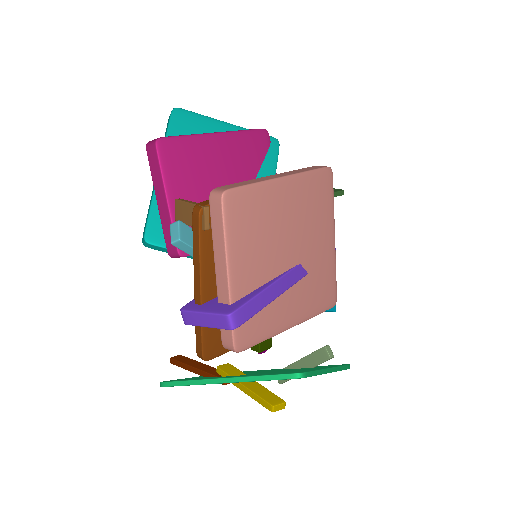}
    \end{subfigure}%
    \hfill%
    \begin{subfigure}[b]{0.20\linewidth}
		\centering
		\includegraphics[width=\linewidth]{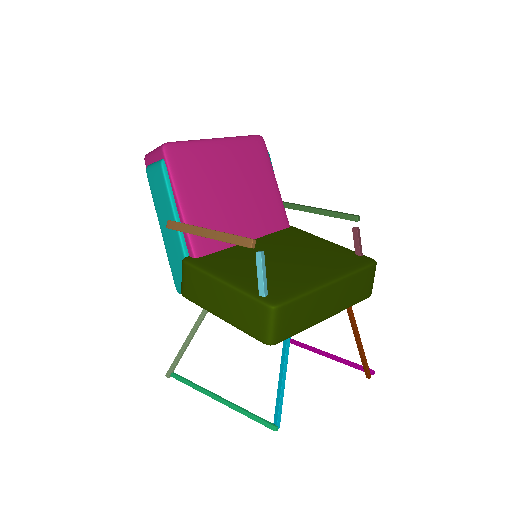}
    \end{subfigure}%
    \hfill%
    \begin{subfigure}[b]{0.20\linewidth}
		\centering
		\includegraphics[width=\linewidth]{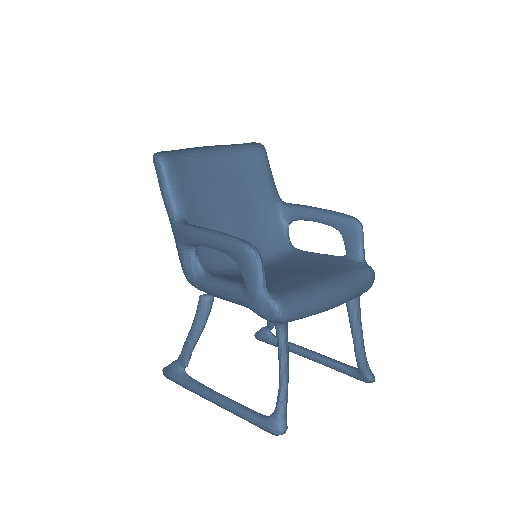}
    \end{subfigure}%
    \vskip\baselineskip%
   \vspace{-1.75em}
    \begin{subfigure}[b]{0.20\linewidth}
		\centering
		\includegraphics[width=\linewidth]{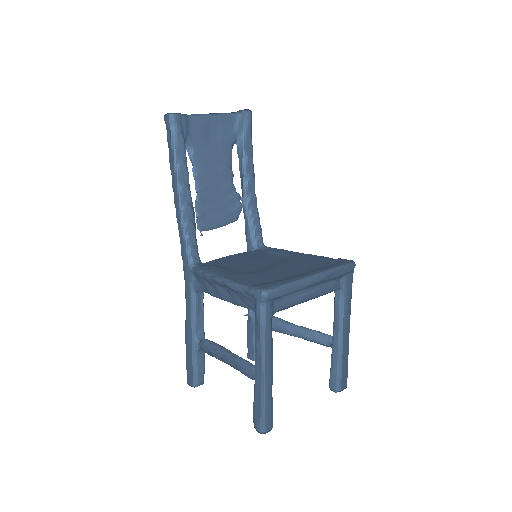}
    \end{subfigure}%
    \hfill%
    \begin{subfigure}[b]{0.20\linewidth}
		\centering
		\includegraphics[width=\linewidth]{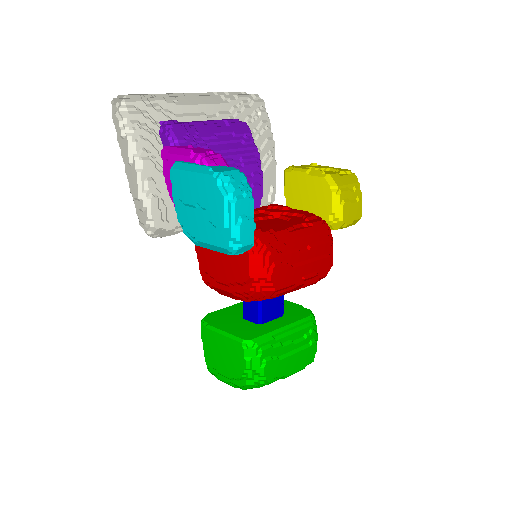}
    \end{subfigure}%
    \hfill%
    \begin{subfigure}[b]{0.20\linewidth}
		\centering
		\includegraphics[width=\linewidth]{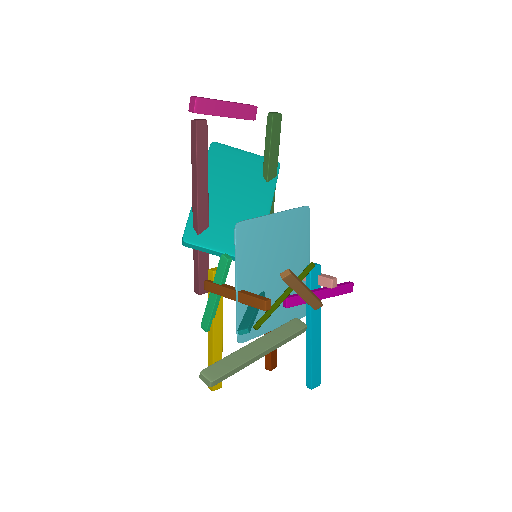}
    \end{subfigure}%
    \hfill%
    \begin{subfigure}[b]{0.20\linewidth}
		\centering
		\includegraphics[width=\linewidth]{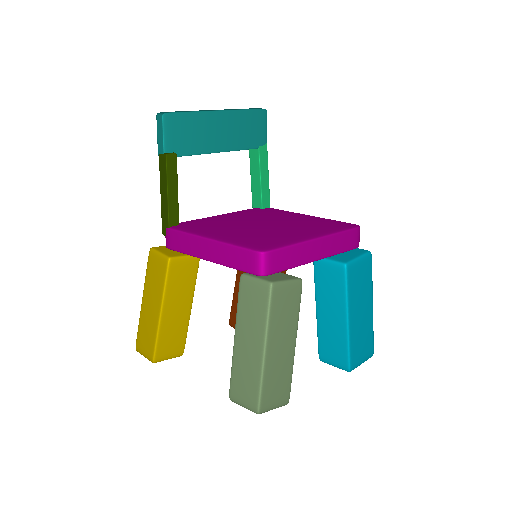}
    \end{subfigure}%
    \hfill%
    \begin{subfigure}[b]{0.20\linewidth}
		\centering
		\includegraphics[width=\linewidth]{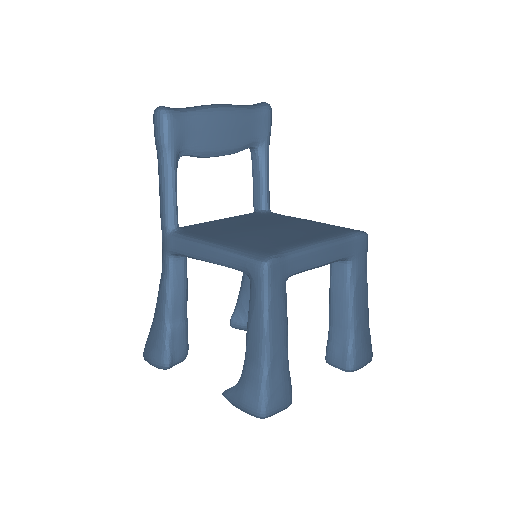}
    \end{subfigure}%
    \vskip\baselineskip%
   \vspace{-1.75em}
    \begin{subfigure}[b]{0.20\linewidth}
		\centering
		\includegraphics[width=\linewidth]{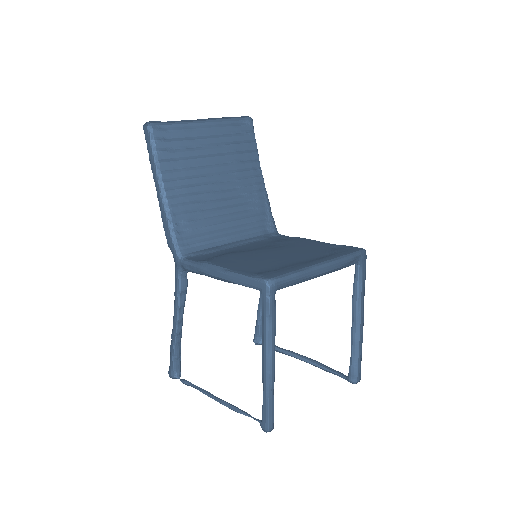}
    \end{subfigure}%
    \hfill%
    \begin{subfigure}[b]{0.20\linewidth}
		\centering
		\includegraphics[width=\linewidth]{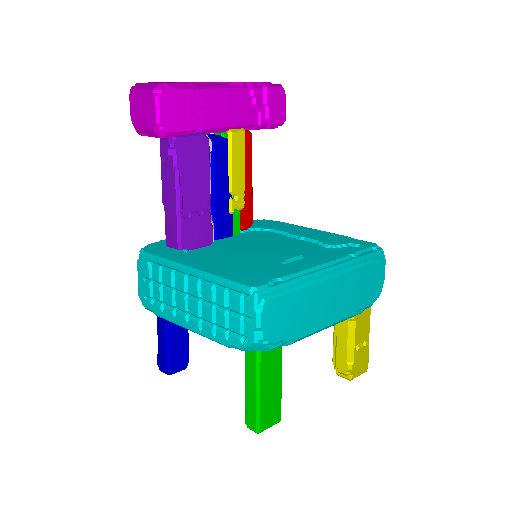}
    \end{subfigure}%
    \hfill%
    \begin{subfigure}[b]{0.20\linewidth}
		\centering
		\includegraphics[width=\linewidth]{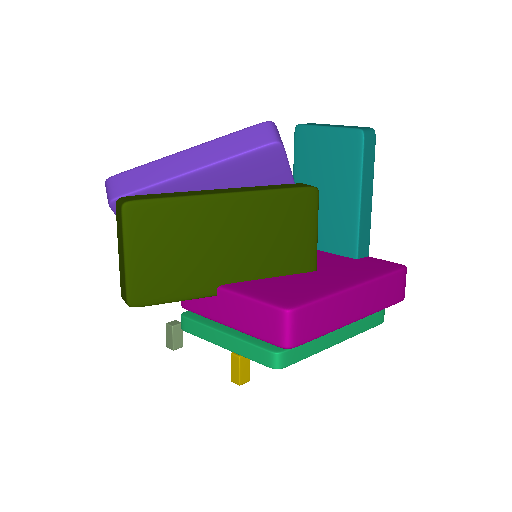}
    \end{subfigure}%
    \hfill%
    \begin{subfigure}[b]{0.20\linewidth}
		\centering
		\includegraphics[width=\linewidth]{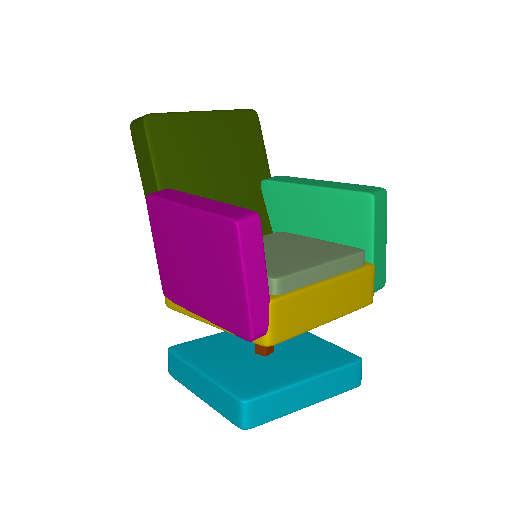}
    \end{subfigure}%
    \hfill%
    \begin{subfigure}[b]{0.20\linewidth}
		\centering
		\includegraphics[width=\linewidth]{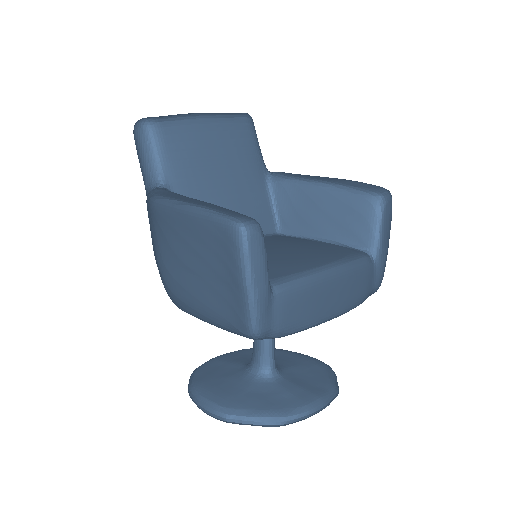}
    \end{subfigure}%
    \vskip\baselineskip%
    \vspace{-2.5em}
    \caption{{\bf Shape Generation Results on Chairs}. We show
    randomly generated chairs using our model,
    ATISS, PQ-NET and IM-NET.}
    \label{fig:shapenet_qualitative_comparison_chairs}
    \vspace{-1.2em}
\end{figure}

%% file: tab/shapenet_generation_quantitative.tex
\begin{table}
\resizebox{\columnwidth}{!}{%
    \centering
    \begin{tabular}{lc|cccc|cccc}
        \toprule
        \multirow{2}{*}{Method} & \multirow{2}{*}{Representation} &
        \multicolumn{4}{c}{MMD-CD ($\downarrow$)} & \multicolumn{4}{c}{COV-CD ($ \%, \uparrow$)} \\
        & & Chair & Table & Lamp & All & Chair & Table & Lamp & All\\
        \midrule
        IM-Net & Implicit & 3.49 & 2.65 & 4.07 & 4.74 & 55.76 & 54.71 & 82.66 & 38.25 \\
        \midrule
        PQ-Net & Implicit Parts & 4.49 & 3.94 & 3.73 & 4.82 & 48.93 & 46.44 & 77.56 & 40.41\\
        ATISS & Cuboids & 5.03 & 4.27 & 4.14 & 6.09 & 48.27 & 39.18 & \bf{91.56} & 35.03 \\
        \midrule
        Ours-Parts & Cuboids & 3.71 & 3.23 & 4.07 & 3.82 & 57.32 & \bf{57.14} & 81.87 & \bf{51.01} \\
        Ours & Implicit & \bf{3.21} & \bf{2.53} & \bf{2.93} & \bf{3.24} & \bf{57.73} & 56.80 & 88.88 & 50.89\\
        \bottomrule
    \end{tabular}
    }
    \caption{{\bf Shape Generation.} We perform
    category-specific training (3rd-5th and
    7th-9th columns) and joint training on multiple object categories (6th and 10th columns) and report
    the MMD-CD ($\downarrow$) and the COV-CD ($\uparrow$) between generated
    and real shapes from the test set.}
    \label{tab:shape_generation_quantitative}
    \vspace{-2.2em}
\end{table}

%% file: sec_results.tex
\section{Experimental Evaluation}
\label{sec:results}

In this section, we provide an extensive evaluation of our method comparing it
to relevant baselines. 
Additional results and implementation
details are provided in the supplementary.

\emph{Datasets }
We report results on three PartNet~\cite{Mo2019CVPR} categories: \emph{Chair,
Table} and \emph{Lamp}, which contain 4489, 5705, and 1554 shapes respectively.
For the \emph{Chair} category, there are 47 different types of parts,
while for the \emph{Lamp} and the \emph{Table} we have 32 and 43 respectively.
We train our model and our part-based baselines
using the part annotations and train/test splits
from PartNet. Moreover, for our model, we use the object bounding
boxes specified in PartNet.

\emph{Baselines }
In our evaluation, we include PQ-NET~\cite{Wu2020CVPR} that is a generative model
that generates 3D shapes using an RNN autoencoder, 
ATISS~\cite{Paschalidou2021NEURIPS}, which was originally introduced for
scene synthesis but can be easily adapted to part-based object generation. In particular,
instead of conditioning to a floor layout, we condition the generation
on the object's bounding box, as for our model. Finally,
we also compare
with IM-Net~\cite{Chen2019CVPR}, which
is an implicit-based generative model that does not reason about parts, hence
not enabling part-level control.

\emph{Metrics }
To evaluate the quality of the generated shapes, we report the Coverage Score
(COV) and the Minimum Matching Distance (MMD)
\cite{Achlioptas2018ICML} using the Chamfer-$L_2$ distance (CD) between points
sampled from real and generated shapes.
To compute these metrics \wrt
our part-based representation, we sample points on the surface of the union
of the generated cuboids.

\input{fig/shape_generation_qualitative_tables}
\subsection{Shape Generation}
We evaluate the performance of our model on the shape generation task
on chairs, tables
and lamps and perform category-specific training for our model and our baselines.
Conditioned on different bounding boxes, our model can successfully
synthesize meaningful part arrangements (see 4th column in
\figref{fig:shapenet_qualitative_comparison_chairs},
\figref{fig:shapenet_qualitative_comparison_tables} and
\figref{fig:shapenet_qualitative_comparison_lamps}), which are fed to
our blending network that combines them and 
yields plausible 3D meshes (see 5th column in
\figref{fig:shapenet_qualitative_comparison_chairs},
\figref{fig:shapenet_qualitative_comparison_tables} and
\figref{fig:shapenet_qualitative_comparison_lamps}).
We observe that PASTA consistently generates diverse and realistic part
arrangements (Ours-Parts) for all object categories, which are, in turn,
converted into meshes (Ours) that faithfully capture the initial
part representation with cuboids. On the contrary,
ATISS struggles to synthesize meaningful part sequences, \ie
the synthesized part arrangements consist
of parts positioned in unnatural positions, especially for the case of chairs
and tables. While
synthesized objects sampled from PQ-NET and IM-NET are more realistic than the
part arrangements produced by ATISS, they lack diversity, as indicated by the
coverage score in \tabref{tab:shape_generation_quantitative}. Note that 
our model, even without the blending network (see Ours-Parts in
\tabref{tab:shape_generation_quantitative}), outperforms both PQ-NET and ATISS
on chairs and tables on both metrics, while our complete architecture
(see Ours in \tabref{tab:shape_generation_quantitative}) 
outperforms also the non-part-based IM-NET on all object categories. 
For the case of lamps,
we observe that our model outperforms all baselines \wrt MMD-CD, while
performing on par with ATISS that achieves the highest score \wrt COV-CD. We hypothesize
that ATISS performs better on the lamps category, as they typically consist of fewer components,
hence making training and inference easier.
\input{fig/shape_completion_qualitative_chairs}
\input{tab/shapenet_completion_quantitative}
Next, we evaluate the ability of our model and our baselines to generate
plausible 3D shapes when jointly trained on multiple object categories, without
any class conditioning. Again our model consistently produces plausible part arrangements that are more
realistic and diverse than our baselines, as validated by our quantitative analysis in
\tabref{tab:shape_generation_quantitative} (see 6th and 10th column).
\figref{fig:shapenet_qualitative_comparison_all} provides a qualitative comparison
of six objects generated with our model and our baselines.

\input{fig/text_guided_generation_free_text}
\subsection{Shape Completion}
Starting from an incomplete sequence of parts, we evaluate whether our model
and our baselines can complete the input sequence in a meaningful
way. For this experiment, we only consider our part-based baselines
and all models are trained in a
category-specific manner. To ensure a fair
comparison to PQ-NET, which is trained with 3D parts of arbitrary
geometries, instead of conditioning their generation on the partial
set of cuboids (illustrated in the 1st column of
\figref{fig:shapenet_qualitative_completion_comparison_chairs},
\figref{fig:shapenet_qualitative_completion_comparison_tables} and
\figref{fig:shapenet_qualitative_completion_comparison_lamps}),
that is used for PASTA and ATISS, we utilize the corresponding
3D parts, that were used during PQ-NET's training.
From our qualitative evaluation, we observe
that both ATISS and PQ-NET tend to generate non-realistic part arrangements,
especially for the case of chairs (see unnatural back part for the 3rd and 6th chairs
in the 3rd column of
\figref{fig:shapenet_qualitative_completion_comparison_chairs}) and tables (see
missing leg in the 2nd table in the 2nd column of
\figref{fig:shapenet_qualitative_completion_comparison_tables}). For the easier case
of lamps (see \figref{fig:shapenet_qualitative_completion_comparison_lamps}),
we observe that all methods can complete the partial object in a meaningful
way. Note that, as PQ-NET relies on a
sequence-to-sequence autoencoder to learn the part arrangements, there is no guarantee
that the completed shape will contain the parts used for conditioning (see 3rd+4th row in
\figref{fig:shapenet_qualitative_completion_comparison_chairs}, 2nd
row in \figref{fig:shapenet_qualitative_completion_comparison_tables}).
Moreover, while for PASTA, the same model is used for both object completion
as well as object generation, PQ-NET requires training a different model
to perform the completion task.

To demonstrate that PASTA generates diverse part arrangements, we also visualize
three generated completions of our model conditioned on the same partial input
(see \figref{fig:shapenet_diversity}). We observe that our generations
are consistently valid and diverse. The quantitative results for
this experiment are summarized in
\tabref{tab:shape_completion_quantitative}. We note that our model
outperforms all baselines both on chairs and tables \wrt both metrics. For the
case of lamps, ATISS outperforms all methods \wrt COV-CD, while being worse
than all in terms of MMD-CD. 

\subsection{Applications}

In this section, we present several applications of our model, such as
conditional generation from text and images. In both experiments, our model is
trained in a category-specific manner.

\emph{Language-guided Generation }
For this experiment, we use the part labels provided in
PartNet~\cite{Mo2019CVPR} and generate utterances that describe the
part-based structure of each object. We train a variant of our model that
conditions on CLIP~\cite{Radford2021ICML} embeddings produced from our
textual descriptions in addition to the object bounding box 
as described in \secref{subsec:inference}. In \figref{fig:text_guided_generation},
we provide text-guided generations of our model and observe note that they
consistently match the input text (\eg the table with the two pedestals, or the
lamp connected to the ceiling).

\emph{Image-guided Generation:}
We now test the ability of our model to perform image-guided generation using
the same model that was trained for language-guided generation, without any
re-training. In particular, we take advantage of the CLIP's joint latent space,
and condition our model on image embeddings produced by CLIP's
image encoder.
\figref{fig:image_guided_generation} shows examples of image-guided synthesis.
While PASTA was never trained with images, we showcase that it generates
shapes of various object categories that faithfully match the input image. Notably,
the recovered parts capture fine geometric details such as the three legs of
the first table in the \figref{fig:image_guided_generation}.


%% file: fig/shape_generation_qualitative_tables.tex
\begin{figure}
    \begin{subfigure}[t]{\linewidth}
    \centering
    \begin{subfigure}[b]{0.20\linewidth}
        \centering
	IM-NET
    \end{subfigure}%
    \hfill%
    \begin{subfigure}[b]{0.20\linewidth}
	\centering
        PQ-NET
    \end{subfigure}%
    \hfill%
    \begin{subfigure}[b]{0.20\linewidth}
	\centering
        ATISS
    \end{subfigure}%
    \hfill%
    \begin{subfigure}[b]{0.20\linewidth}
        \centering
        Ours-Parts
    \end{subfigure}%
    \hfill%
    \begin{subfigure}[b]{0.20\linewidth}
        \centering
        Ours
    \end{subfigure}
    \end{subfigure}
    \vspace{-1.5em}
    \begin{subfigure}[b]{0.20\linewidth}
		\centering
		\includegraphics[width=\linewidth]{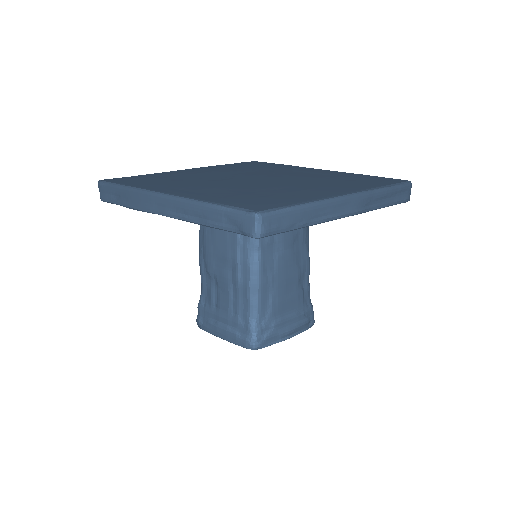}
    \end{subfigure}%
    \hfill%
    \begin{subfigure}[b]{0.20\linewidth}
		\centering
		\includegraphics[width=0.7\linewidth]{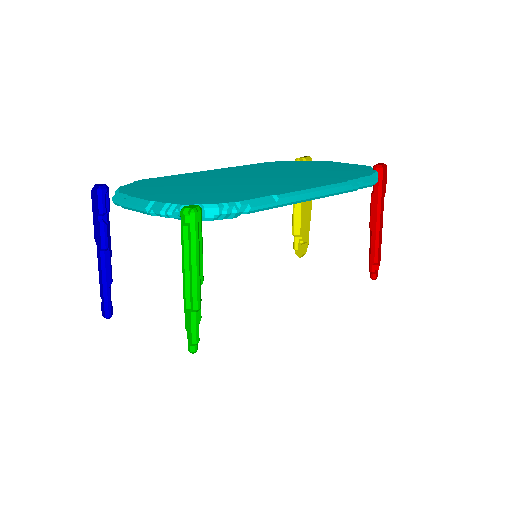}
    \end{subfigure}%
    \hfill%
    \begin{subfigure}[b]{0.20\linewidth}
		\centering
		\includegraphics[width=\linewidth]{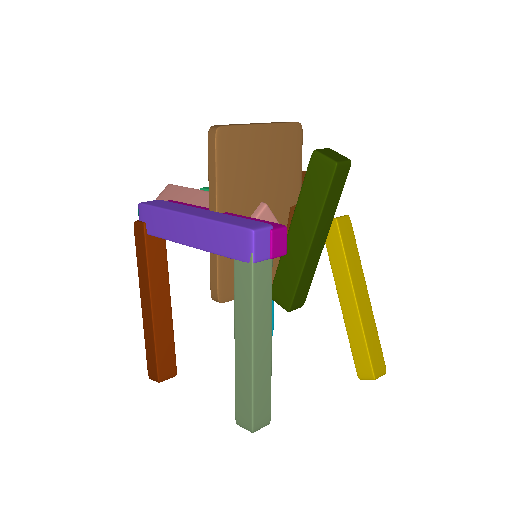}
    \end{subfigure}%
    \hfill%
    \begin{subfigure}[b]{0.20\linewidth}
		\centering
		\includegraphics[width=\linewidth]{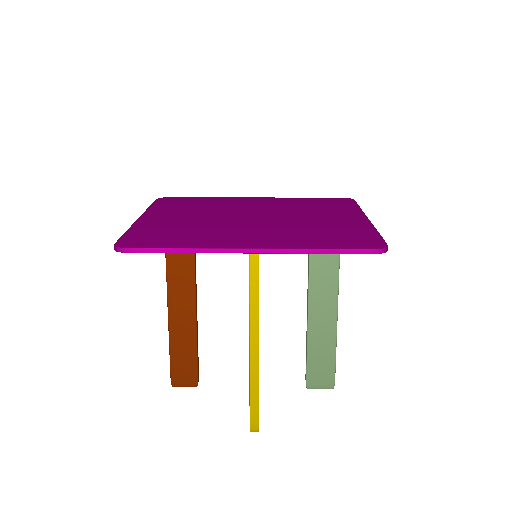}
    \end{subfigure}%
    \hfill%
    \begin{subfigure}[b]{0.20\linewidth}
		\centering
		\includegraphics[width=\linewidth]{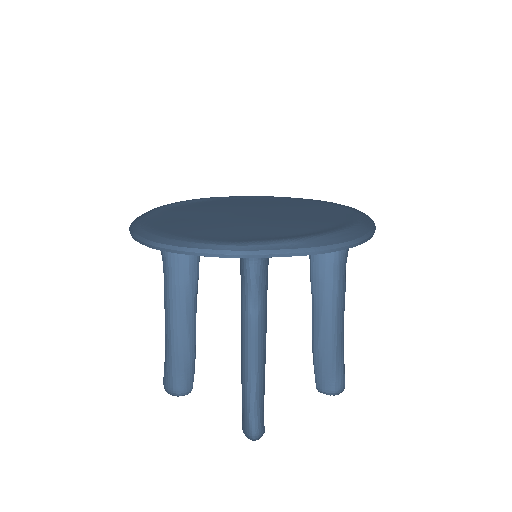}
    \end{subfigure}%
    \vskip\baselineskip%
    \vspace{-0.75em}
    \begin{subfigure}[b]{0.20\linewidth}
		\centering
		\includegraphics[width=\linewidth]{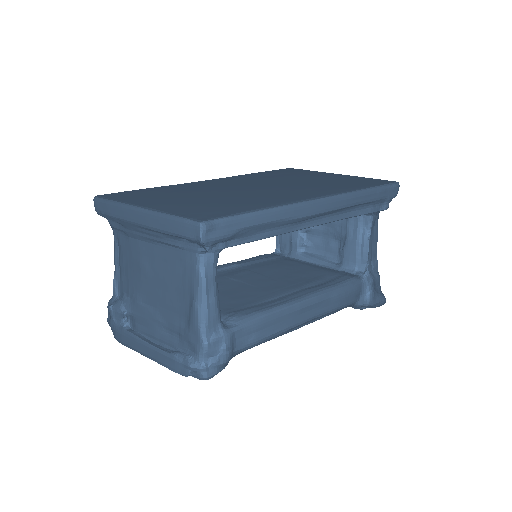}
    \end{subfigure}%
    \hfill%
    \begin{subfigure}[b]{0.20\linewidth}
		\centering
		\includegraphics[width=\linewidth]{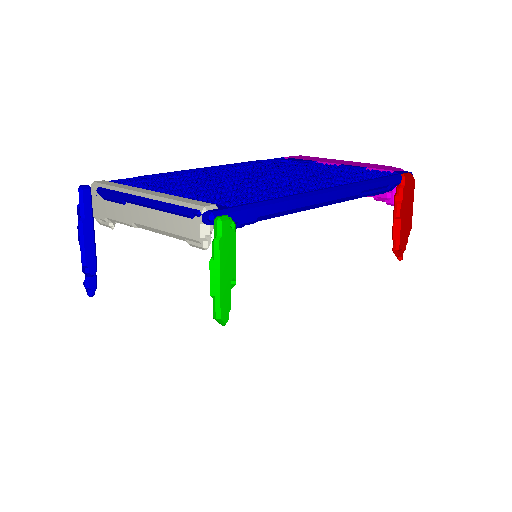}
    \end{subfigure}%
    \hfill%
    \begin{subfigure}[b]{0.20\linewidth}
		\centering
		\includegraphics[width=\linewidth]{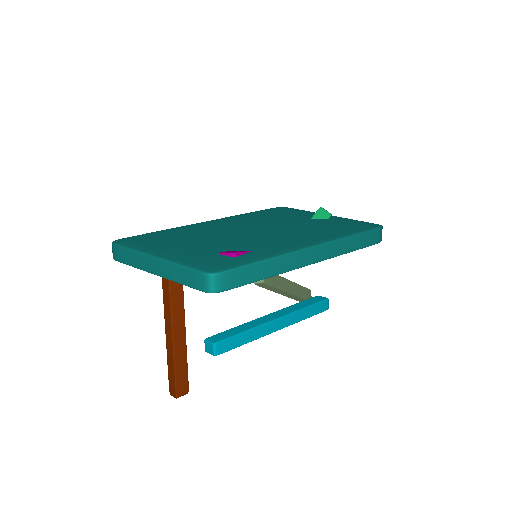}
    \end{subfigure}%
    \hfill%
    \begin{subfigure}[b]{0.20\linewidth}
		\centering
		\includegraphics[width=\linewidth]{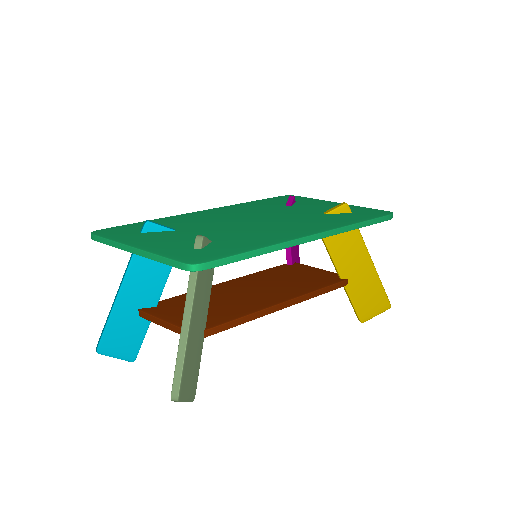}
    \end{subfigure}%
    \hfill%
    \begin{subfigure}[b]{0.20\linewidth}
		\centering
		\includegraphics[width=\linewidth]{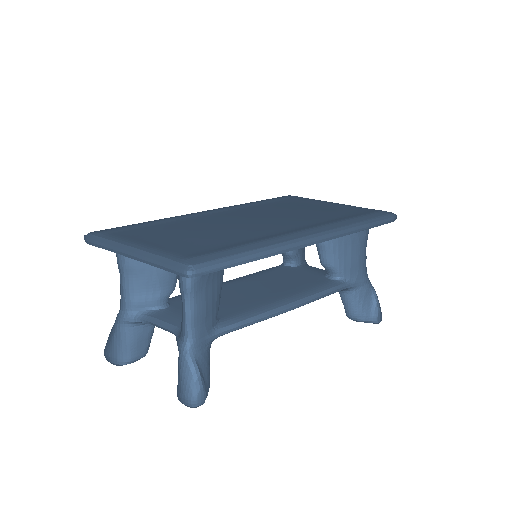}
    \end{subfigure}%
    \vskip\baselineskip%
    \vspace{-1.75em}
    \begin{subfigure}[b]{0.20\linewidth}
		\centering
		\includegraphics[width=\linewidth]{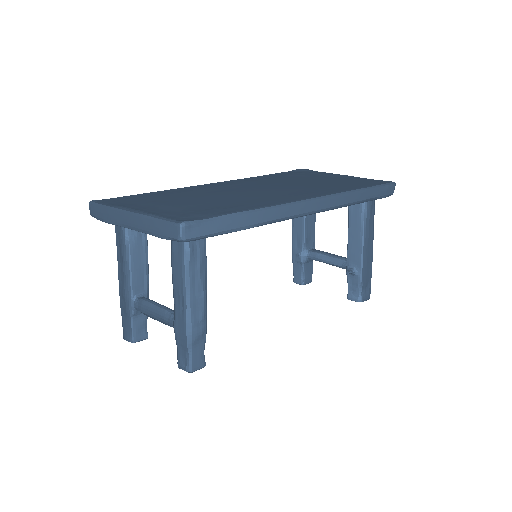}
    \end{subfigure}%
    \hfill%
    \begin{subfigure}[b]{0.20\linewidth}
		\centering
		\includegraphics[width=0.7\linewidth]{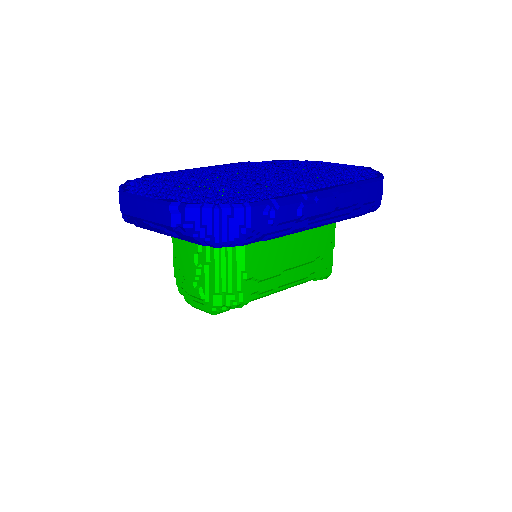}
    \end{subfigure}%
    \hfill%
    \begin{subfigure}[b]{0.20\linewidth}
		\centering
		\includegraphics[width=\linewidth]{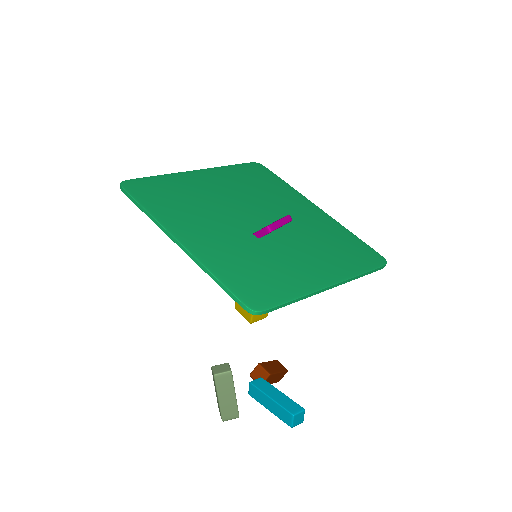}
    \end{subfigure}%
    \hfill%
    \begin{subfigure}[b]{0.20\linewidth}
		\centering
		\includegraphics[width=\linewidth]{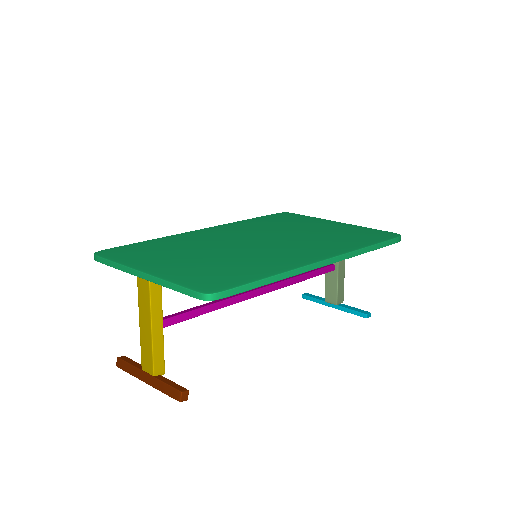}
    \end{subfigure}%
    \hfill%
    \begin{subfigure}[b]{0.20\linewidth}
		\centering
		\includegraphics[width=\linewidth]{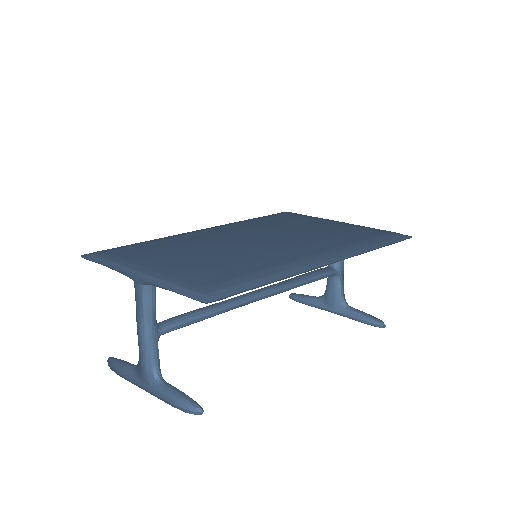}
    \end{subfigure}%
    \vskip\baselineskip%
    \vspace{-1.75em}
    \begin{subfigure}[b]{0.20\linewidth}
		\centering
		\includegraphics[width=\linewidth]{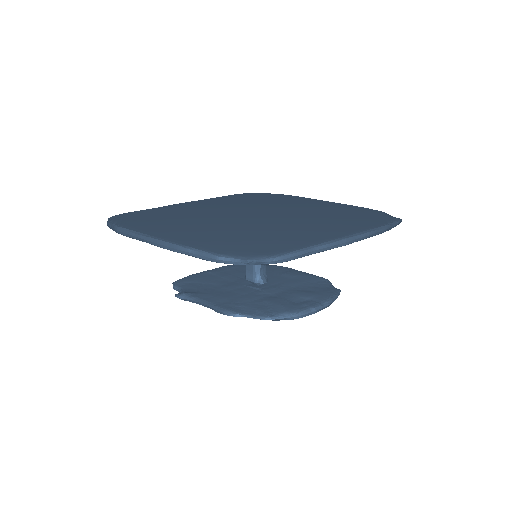}
    \end{subfigure}%
    \hfill%
    \begin{subfigure}[b]{0.20\linewidth}
		\centering
		\includegraphics[width=\linewidth]{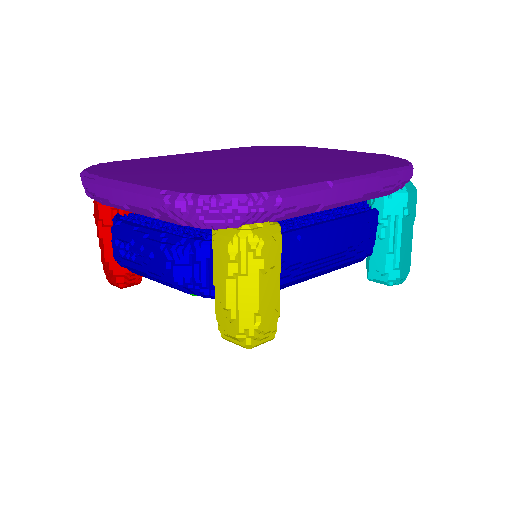}
    \end{subfigure}%
    \hfill%
    \begin{subfigure}[b]{0.20\linewidth}
		\centering
		\includegraphics[width=\linewidth]{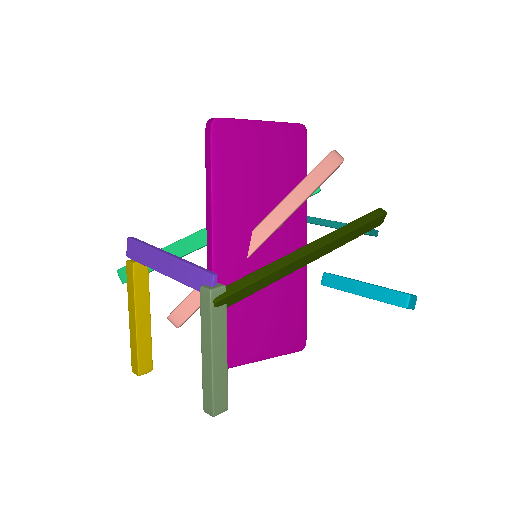}
    \end{subfigure}%
    \hfill%
    \begin{subfigure}[b]{0.20\linewidth}
		\centering
		\includegraphics[width=\linewidth]{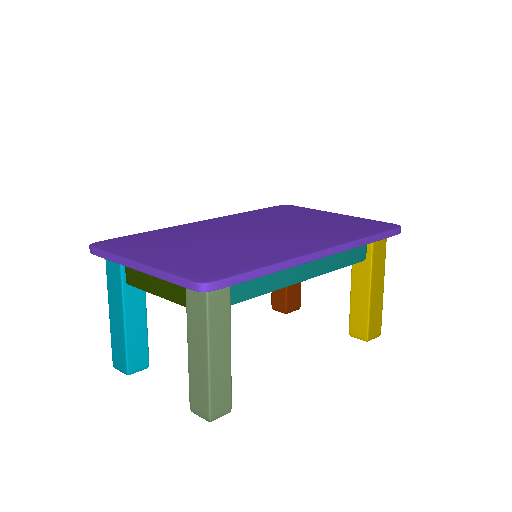}
    \end{subfigure}%
    \hfill%
    \begin{subfigure}[b]{0.20\linewidth}
		\centering
		\includegraphics[width=\linewidth]{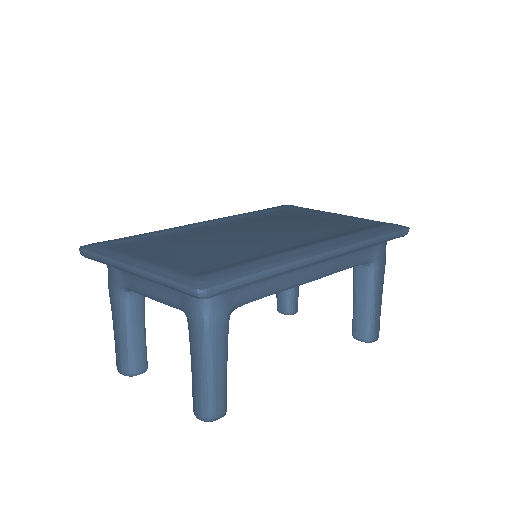}
    \end{subfigure}%
    \vskip\baselineskip%
    \vspace{-1.75em}
    \begin{subfigure}[b]{0.20\linewidth}
		\centering
		\includegraphics[width=\linewidth]{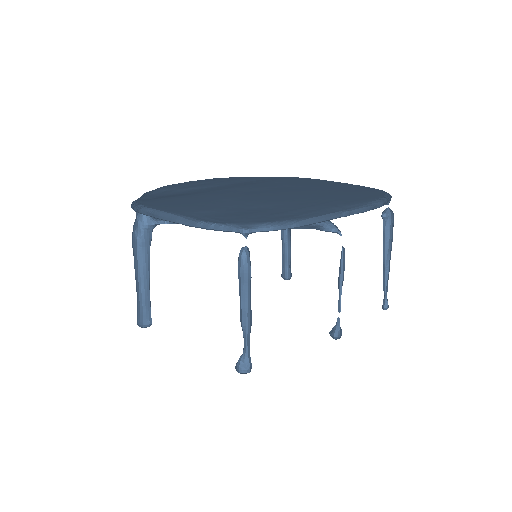}
    \end{subfigure}%
    \hfill%
    \begin{subfigure}[b]{0.20\linewidth}
		\centering
		\includegraphics[width=\linewidth]{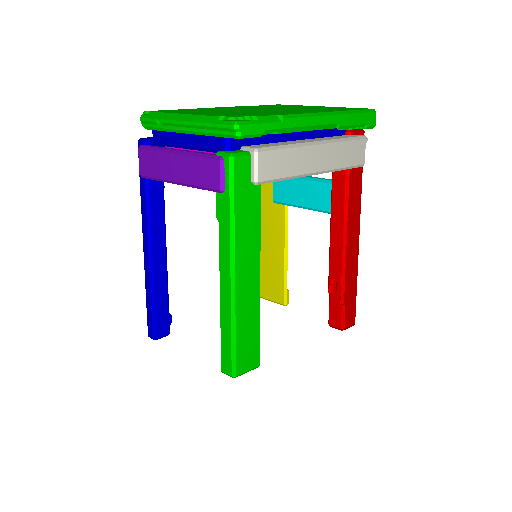}
    \end{subfigure}%
    \hfill%
    \begin{subfigure}[b]{0.20\linewidth}
		\centering
		\includegraphics[width=\linewidth]{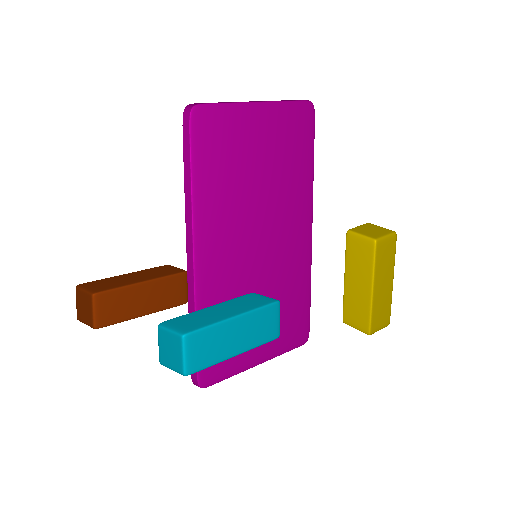}
    \end{subfigure}%
    \hfill%
    \begin{subfigure}[b]{0.20\linewidth}
		\centering
		\includegraphics[width=\linewidth]{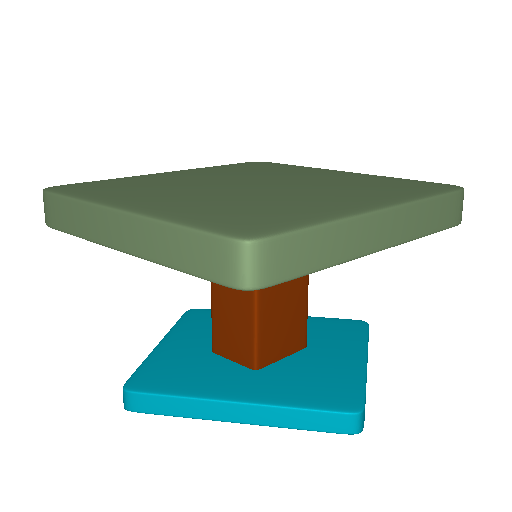}
    \end{subfigure}%
    \hfill%
    \begin{subfigure}[b]{0.20\linewidth}
		\centering
		\includegraphics[width=\linewidth]{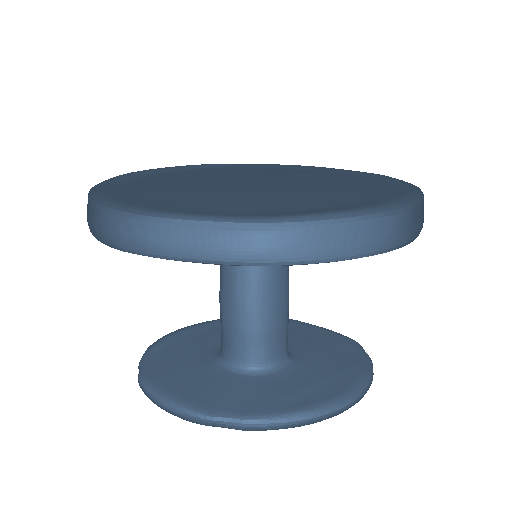}
    \end{subfigure}%
    \vskip\baselineskip%
    \vspace{-2.5em}
    \caption{{\bf Shape Generation Results on Tables}. We show
    randomly generated tables using our model, ATISS, PQ-NET and IM-NET.}
    \label{fig:shapenet_qualitative_comparison_tables}
    \vspace{-1.25em}
\end{figure}

%% file: fig/shape_completion_qualitative_chairs.tex
\begin{figure}
    \begin{subfigure}[t]{\linewidth}
    \centering
    \begin{subfigure}[b]{0.20\linewidth}
		\centering
		Partial Input
    \end{subfigure}%
    \hfill%
    \begin{subfigure}[b]{0.20\linewidth}
        \centering
        PQ-NET
    \end{subfigure}%
    \hfill%
    \begin{subfigure}[b]{0.20\linewidth}
		\centering
       ATISS 
    \end{subfigure}%
    \hfill%
    \begin{subfigure}[b]{0.20\linewidth}
        \centering
        Ours-Parts
    \end{subfigure}%
    \hfill%
    \begin{subfigure}[b]{0.20\linewidth}
        \centering
        Ours
    \end{subfigure}
    \end{subfigure}
    \vspace{-1.5em}
    \begin{subfigure}[b]{0.20\linewidth}
		\centering
            \includegraphics[width=\linewidth]{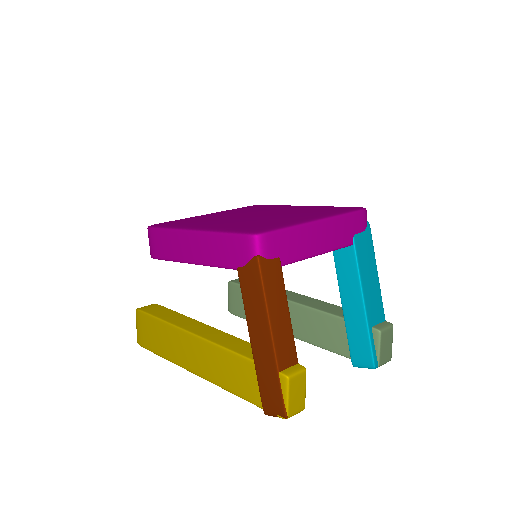}
    \end{subfigure}%
    \hfill%
    \begin{subfigure}[b]{0.20\linewidth}
		\centering
		\includegraphics[width=\linewidth]{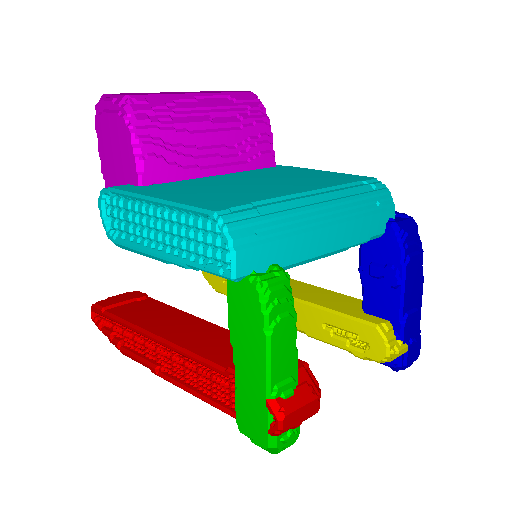}
    \end{subfigure}%
    \hfill%
    \begin{subfigure}[b]{0.20\linewidth}
		\centering
	      \includegraphics[width=\linewidth]{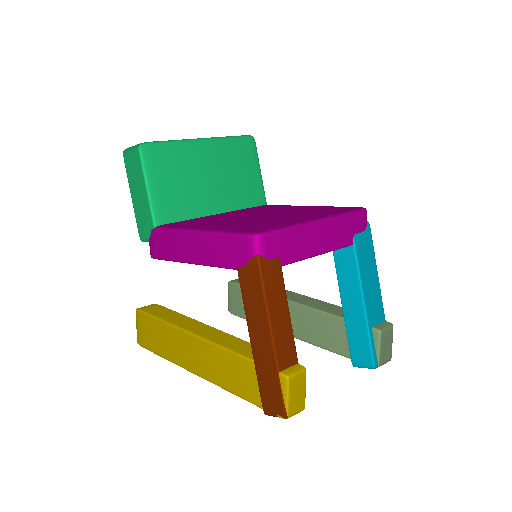}
    \end{subfigure}%
    \hfill%
    \begin{subfigure}[b]{0.20\linewidth}
		\centering
		\includegraphics[width=\linewidth]{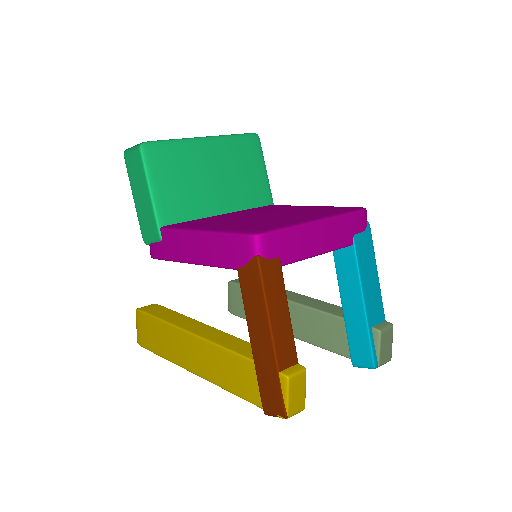}
    \end{subfigure}%
    \hfill%
    \begin{subfigure}[b]{0.20\linewidth}
		\centering
		\includegraphics[width=\linewidth]{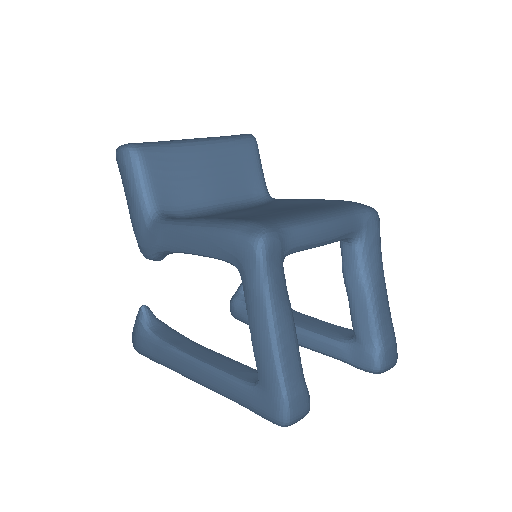}
    \end{subfigure}%
    \vskip\baselineskip%
    \vspace{-0.5em}
    \begin{subfigure}[b]{0.20\linewidth}
		\centering
           \includegraphics[width=\linewidth]{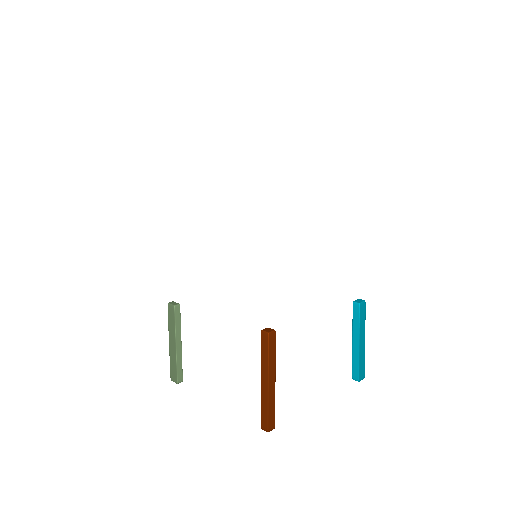}
    \end{subfigure}%
    \hfill%
    \begin{subfigure}[b]{0.20\linewidth}
		\centering
		\includegraphics[width=\linewidth]{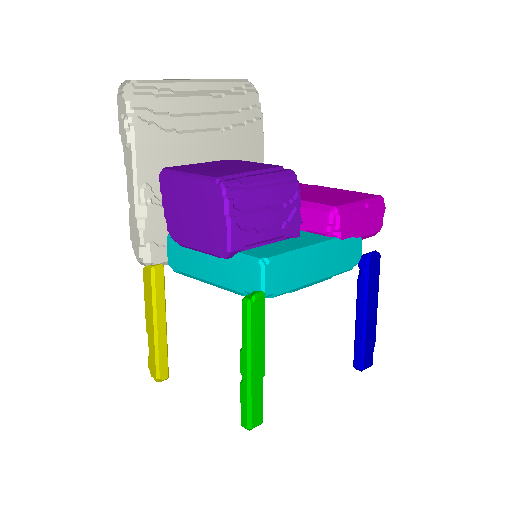}
    \end{subfigure}%
    \hfill%
    \begin{subfigure}[b]{0.20\linewidth}
		\centering
            \includegraphics[width=\linewidth]{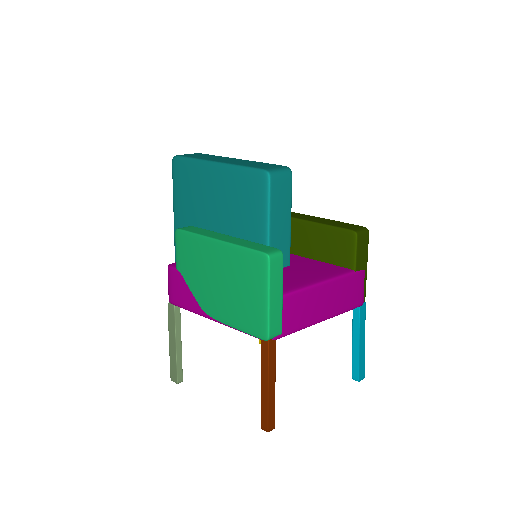}
    \end{subfigure}%
    \hfill%
    \begin{subfigure}[b]{0.20\linewidth}
		\centering
		\includegraphics[width=\linewidth]{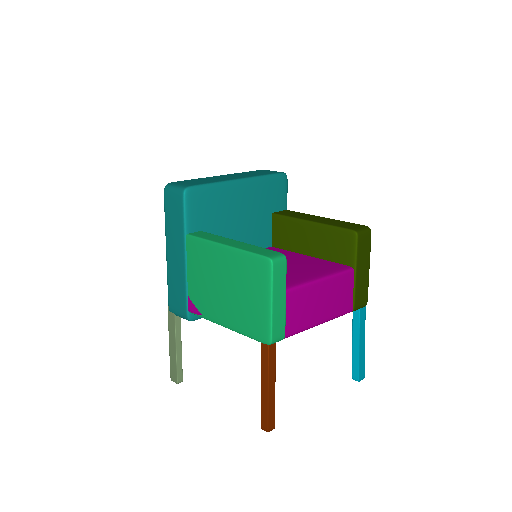}
    \end{subfigure}%
    \hfill%
    \begin{subfigure}[b]{0.20\linewidth}
		\centering
		\includegraphics[width=\linewidth]{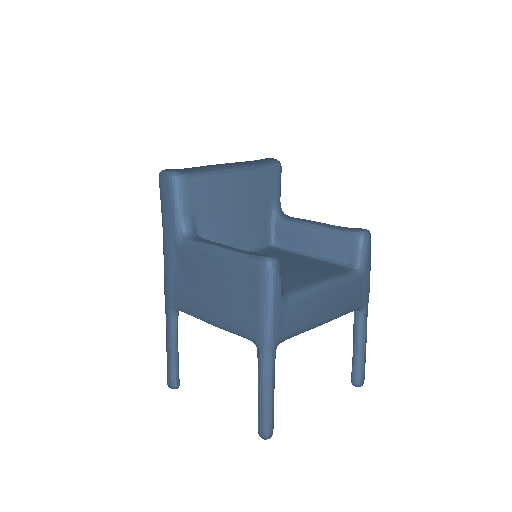}
    \end{subfigure}%
    \vskip\baselineskip%
    \vspace{-1.5em}
    \begin{subfigure}[b]{0.20\linewidth}
		\centering
            \includegraphics[width=\linewidth]{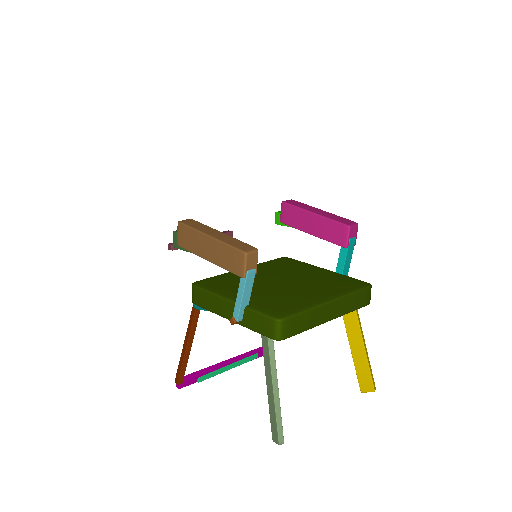}
    \end{subfigure}%
    \hfill%
    \begin{subfigure}[b]{0.20\linewidth}
		\centering
		\includegraphics[width=\linewidth]{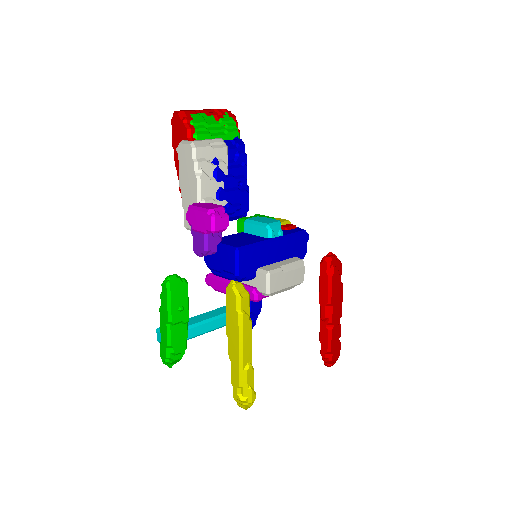}
    \end{subfigure}%
    \hfill%
    \begin{subfigure}[b]{0.20\linewidth}
		\centering
            \includegraphics[width=\linewidth]{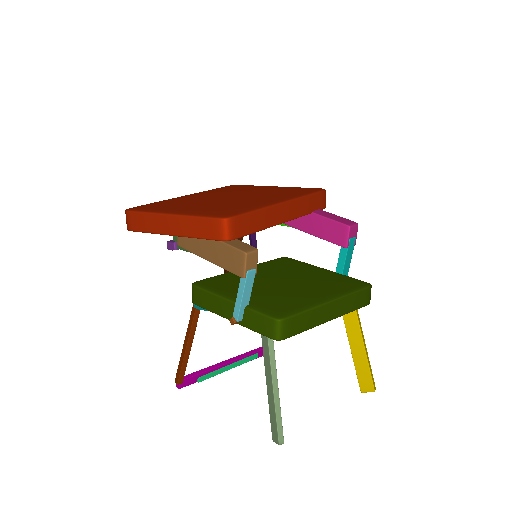}
    \end{subfigure}%
    \hfill%
    \begin{subfigure}[b]{0.20\linewidth}
		\centering
		\includegraphics[width=\linewidth]{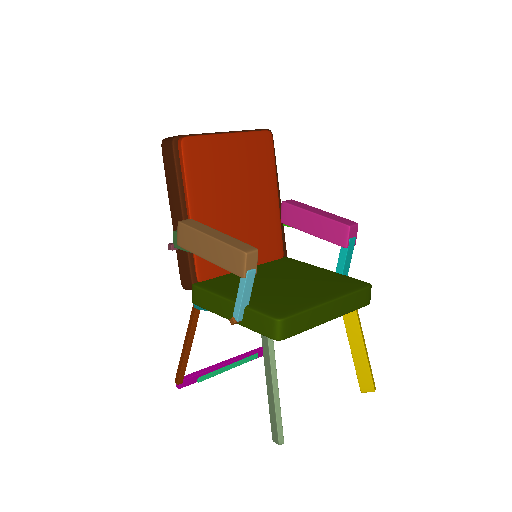}
    \end{subfigure}%
    \hfill%
    \begin{subfigure}[b]{0.20\linewidth}
		\centering
		\includegraphics[width=\linewidth]{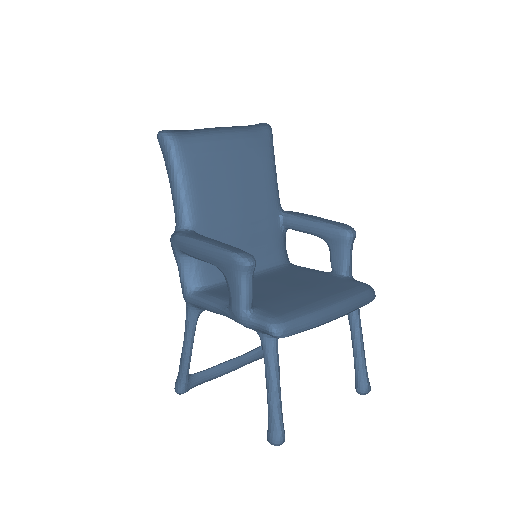}
    \end{subfigure}%
    \vskip\baselineskip%
   \vspace{-1.5em}
    \begin{subfigure}[b]{0.20\linewidth}
		\centering
  \includegraphics[width=\linewidth]{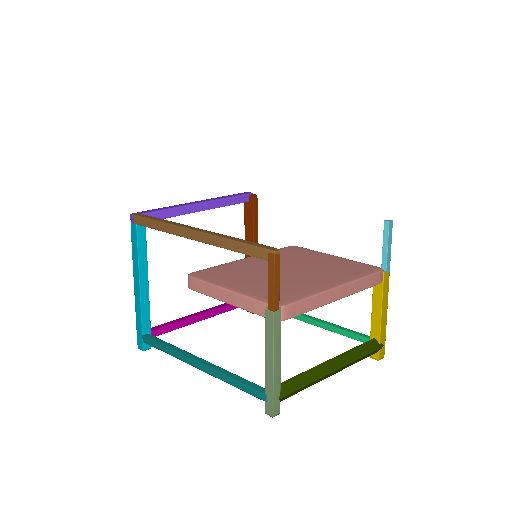}
    \end{subfigure}%
    \hfill%
    \begin{subfigure}[b]{0.20\linewidth}
		\centering
		\includegraphics[width=\linewidth]{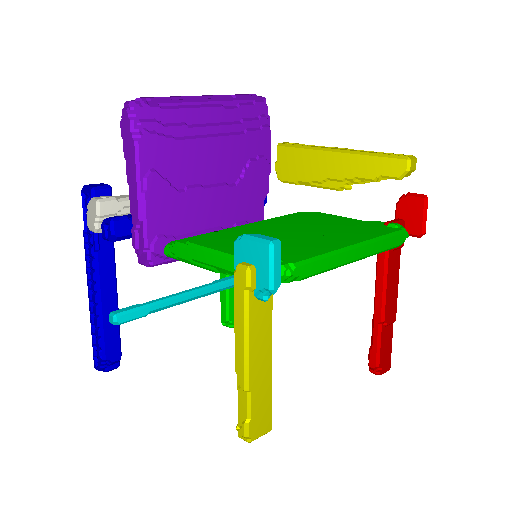}
    \end{subfigure}%
    \hfill%
    \begin{subfigure}[b]{0.20\linewidth}
		\centering
    	\includegraphics[width=\linewidth]{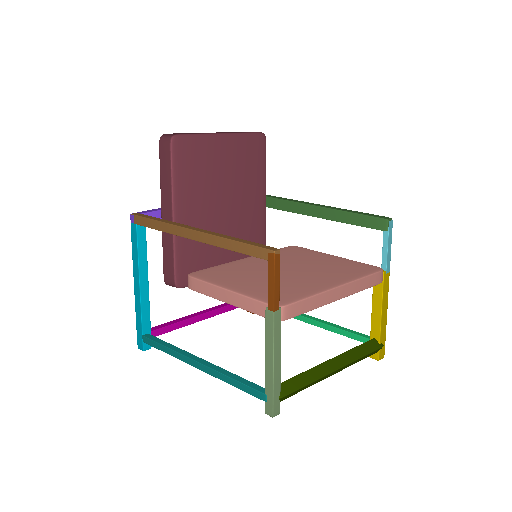}
    \end{subfigure}%
    \hfill%
    \begin{subfigure}[b]{0.20\linewidth}
		\centering
		\includegraphics[width=\linewidth]{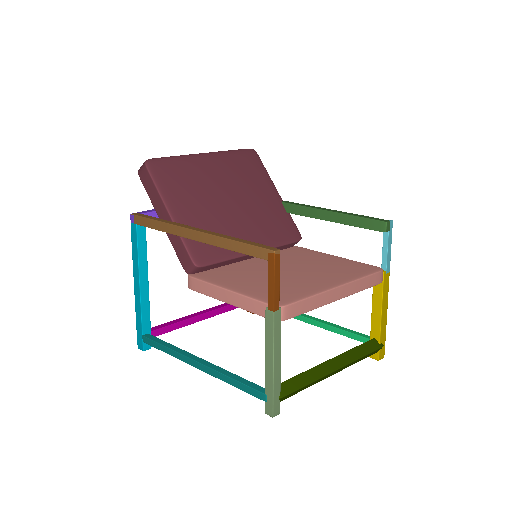}
    \end{subfigure}%
    \hfill%
    \begin{subfigure}[b]{0.20\linewidth}
		\centering
		\includegraphics[width=\linewidth]{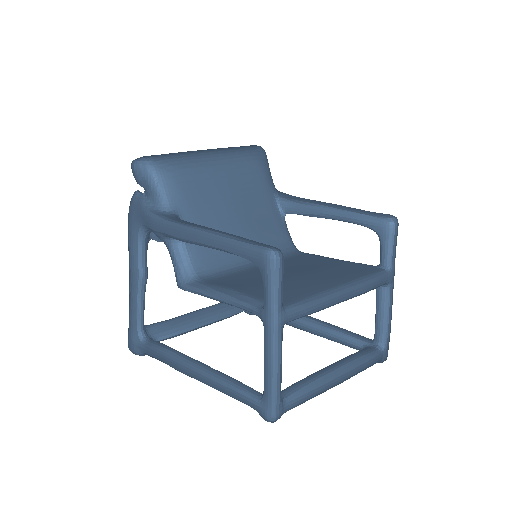}
    \end{subfigure}%
    \vskip\baselineskip%
   \vspace{-1.75em}
    \begin{subfigure}[b]{0.20\linewidth}
		\centering
  		\includegraphics[width=\linewidth]{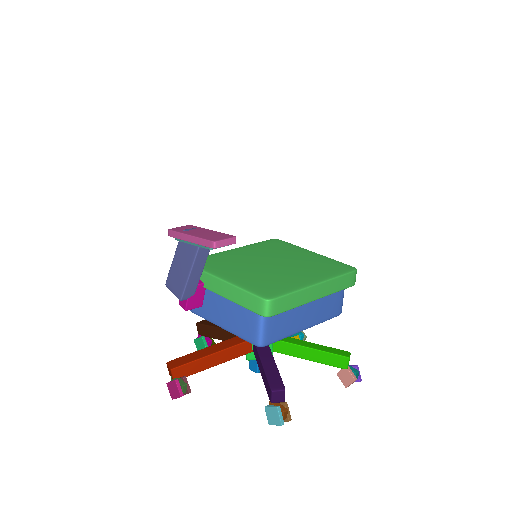}
    \end{subfigure}%
    \hfill%
    \begin{subfigure}[b]{0.20\linewidth}
		\centering
		\includegraphics[width=\linewidth]{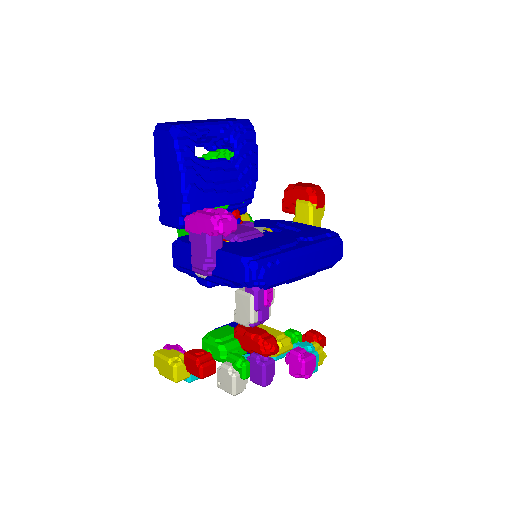}
    \end{subfigure}%
    \hfill%
    \begin{subfigure}[b]{0.20\linewidth}
		\centering
            \includegraphics[width=\linewidth]{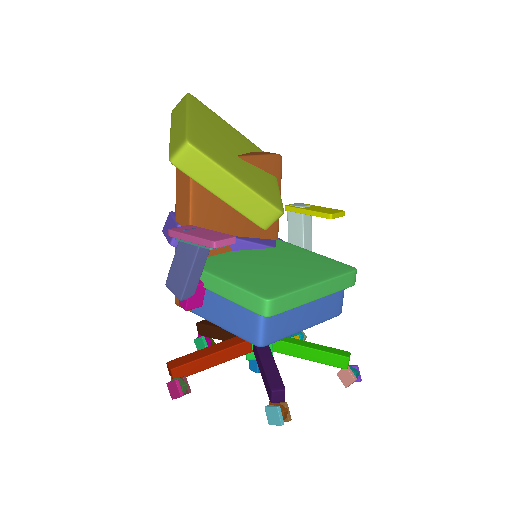}
    \end{subfigure}%
    \hfill%
    \begin{subfigure}[b]{0.20\linewidth}
		\centering
		\includegraphics[width=\linewidth]{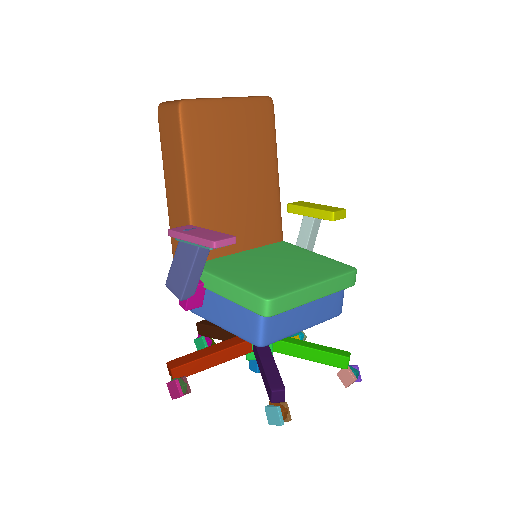}
    \end{subfigure}%
    \hfill%
    \begin{subfigure}[b]{0.20\linewidth}
		\centering
		\includegraphics[width=\linewidth]{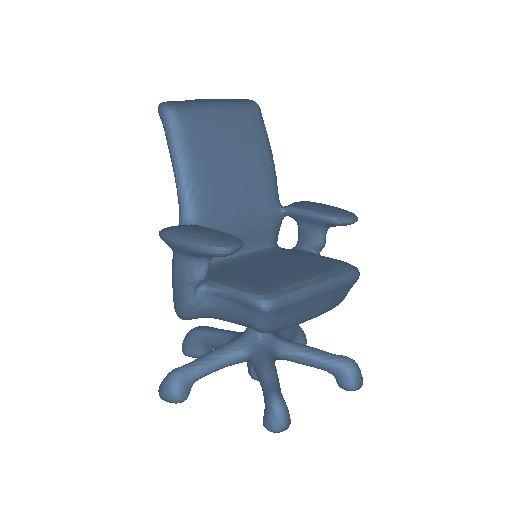}
    \end{subfigure}%
    \vskip\baselineskip%
    \vspace{-2.5em}
    \caption{{\bf Shape Completion Results on Chairs}. Starting
    from partial chairs, we show completions of our model,
    ATISS and PQ-NET.}
    \label{fig:shapenet_qualitative_completion_comparison_chairs}
    \vspace{-1.2em}
\end{figure}

%% file: tab/shapenet_completion_quantitative.tex
\begin{table}
\resizebox{\columnwidth}{!}{%
    \centering
    \begin{tabular}{lc|ccc|ccc}
        \toprule
        \multirow{2}{*}{Method} & \multirow{2}{*}{Representation} &
        \multicolumn{3}{c}{MMD-CD ($\downarrow$)} & \multicolumn{3}{c}{COV-CD ($ \%, \uparrow$)} \\
        & & Chair & Table & Lamp & Chair & Table & Lamp\\
        \toprule
        PQ-Net & Implicit Parts & 4.69 & 3.64 & \bf{4.55} & 35.77 & 42.31 & 53.33\\
        ATISS & Cuboids & 4.33 & 3.40 & 5.90 & 44.57 & 43.33 & \bf{60.88}\\
        \midrule
        Ours-Parts & Cuboids & 3.43 & 2.66 & 5.72 & \bf{50.49} & 56.13  & 57.78 \\
        Ours & Implicit & \bf{3.11} & \bf{2.33} & 5.58 & 49.00 & \bf{58.56} & 51.00\\
        \bottomrule
    \end{tabular}
    }
    \caption{{\bf Shape Completion.} We measure
    the MMD-CD ($\downarrow$) and the COV-CD ($\uparrow$) between the
    part-based representations of completed and real shapes from the test set.}
    \label{tab:shape_completion_quantitative}
    \vspace{-2.2em}
\end{table}

%% file: fig/text_guided_generation_free_text.tex
\begin{figure}
    \centering
    \begin{subfigure}[b]{0.31\linewidth}
	\centering
	\includegraphics[trim={2cm 2cm 2cm 2cm},clip,width=0.6\linewidth]{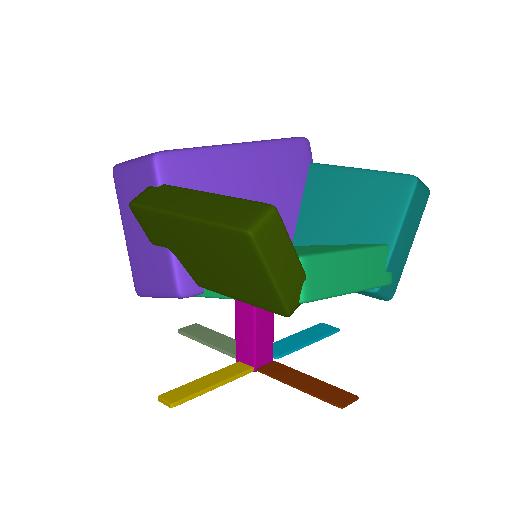}
    \end{subfigure}%
    \hfill%
     \begin{subfigure}[b]{0.31\linewidth}
	\centering
	\includegraphics[trim={2cm 2cm 2cm 2cm},clip,width=0.6\linewidth]{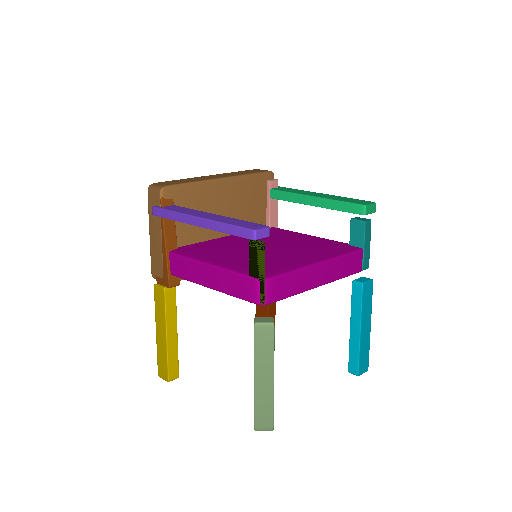}
    \end{subfigure}%
    \hfill%
    \begin{subfigure}[b]{0.31\linewidth}
	\centering
        \includegraphics[trim={2cm 2cm 2cm 2cm},clip,width=0.6\linewidth]{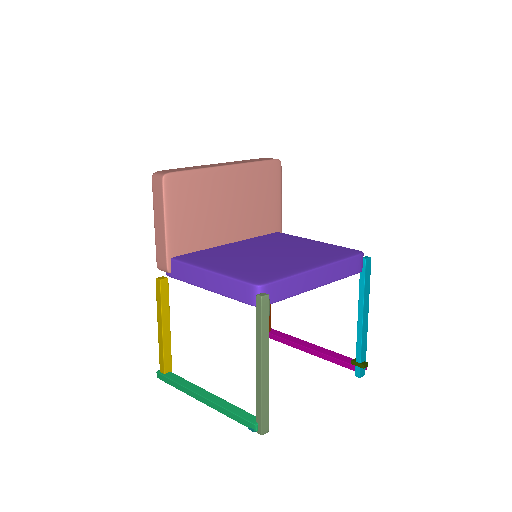}
    \end{subfigure}%
     \vskip\baselineskip%
    \vspace{-1.6em}
    \begin{subfigure}[b]{0.31\linewidth}
	\centering
	\includegraphics[trim={2cm 2cm 2cm 4cm},clip,width=0.6\linewidth]{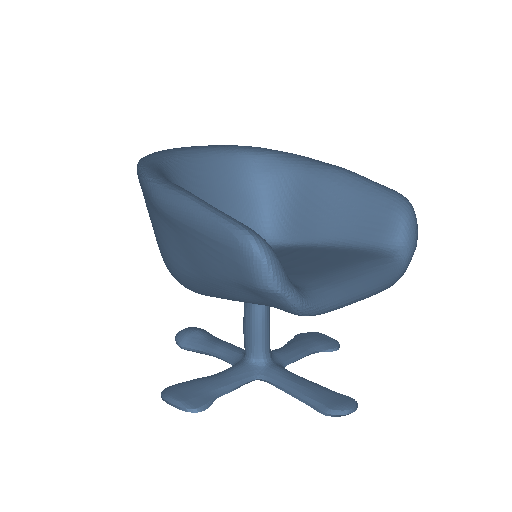}
    \end{subfigure}%
    \hfill%
     \begin{subfigure}[b]{0.31\linewidth}
	\centering
	\includegraphics[trim={2cm 2cm 2cm 4cm},clip,width=0.6\linewidth]{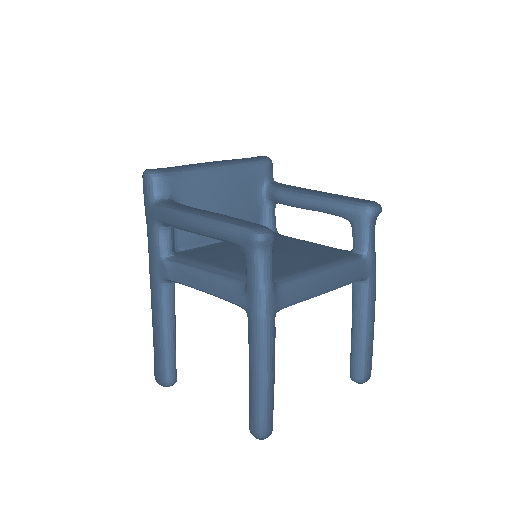}
    \end{subfigure}%
    \hfill%
    \begin{subfigure}[b]{0.31\linewidth}
	\centering
        \includegraphics[trim={2cm 2cm 2cm 4cm},clip,width=0.6\linewidth]{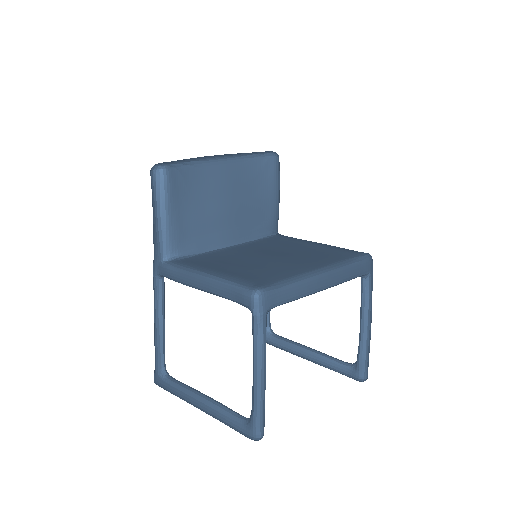}
    \end{subfigure}%
    \vskip\baselineskip%
    \vspace{-0.5em}
    \begin{subfigure}[t]{0.30\linewidth}
	\small{\textit{A chair with four legs, a central support and sofa-style arms.}}
    \end{subfigure}%
    \hfill%
     \begin{subfigure}[t]{0.30\linewidth}
	\small{\textit{A chair with four legs, two horizontal and two vertical bars.}}
    \end{subfigure}%
    \hfill%
    \begin{subfigure}[t]{0.30\linewidth}
        \small{\textit{A chair with four legs and two runners.}}
    \end{subfigure}%
    \vspace{-1.5em}
    \vskip\baselineskip%
    \begin{subfigure}[b]{0.31\linewidth}
	\centering
	\includegraphics[trim={2cm 2cm 2cm 2cm},clip,width=0.6\linewidth]{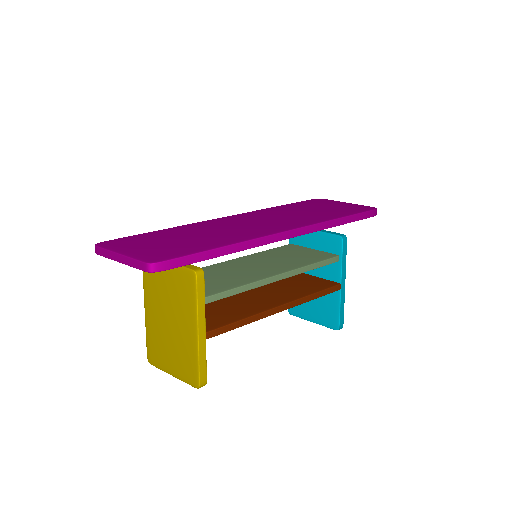}
    \end{subfigure}%
    \hfill%
     \begin{subfigure}[b]{0.31\linewidth}
	\centering
	\includegraphics[trim={2cm 2cm 2cm 2cm},clip,width=0.6\linewidth]{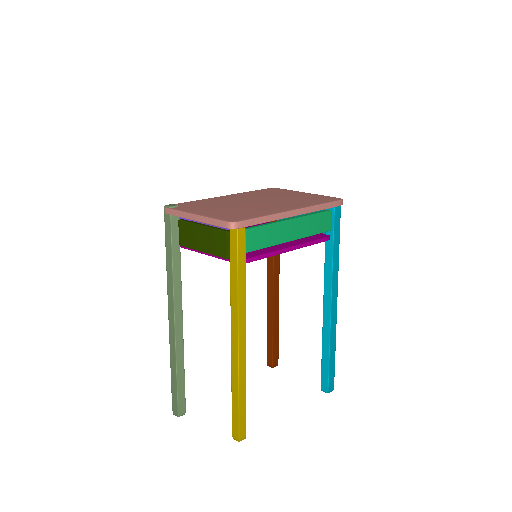}
    \end{subfigure}%
    \hfill%
    \begin{subfigure}[b]{0.31\linewidth}
	\centering
        \includegraphics[trim={2cm 2cm 2cm 2cm},clip,width=0.6\linewidth]{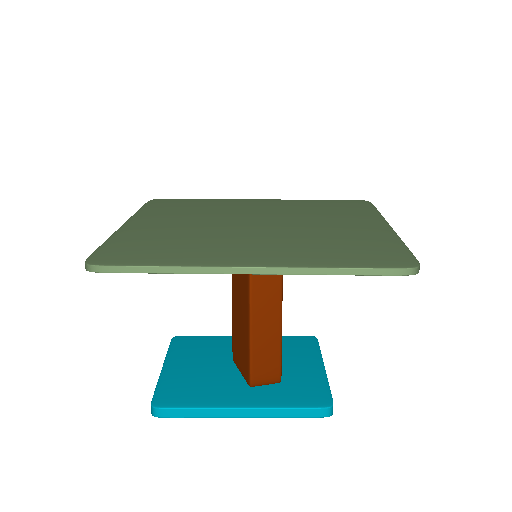}
    \end{subfigure}%
    \vskip\baselineskip%
    \vspace{-1.6em}
    \begin{subfigure}[b]{0.31\linewidth}
	\centering
	\includegraphics[trim={2cm 2cm 2cm 4cm},clip,width=0.6\linewidth]{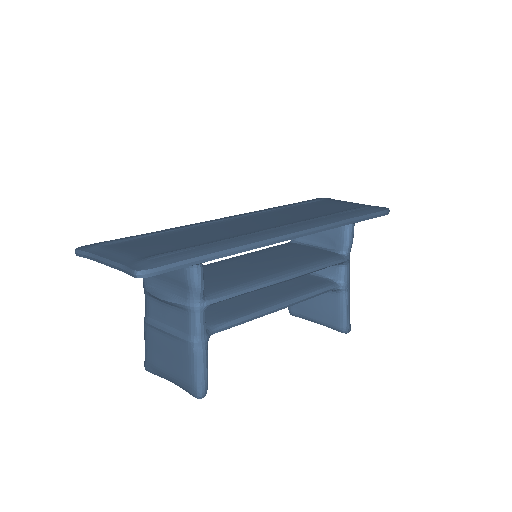}
    \end{subfigure}%
    \hfill%
     \begin{subfigure}[b]{0.31\linewidth}
	\centering
	\includegraphics[trim={2cm 2cm 2cm 4cm},clip,width=0.6\linewidth]{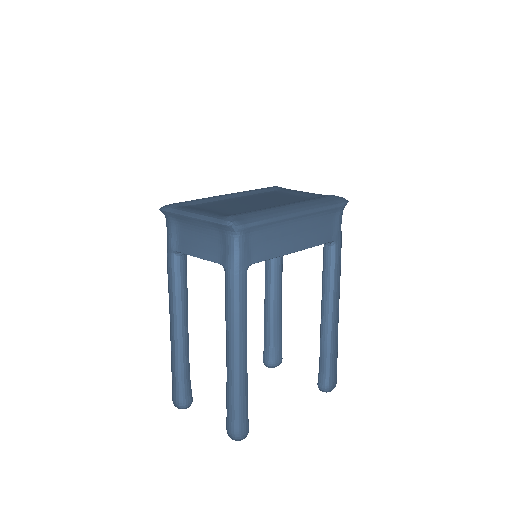}
    \end{subfigure}%
    \hfill%
    \begin{subfigure}[b]{0.31\linewidth}
	\centering
        \includegraphics[trim={2cm 2cm 2cm 4cm},clip,width=0.6\linewidth]{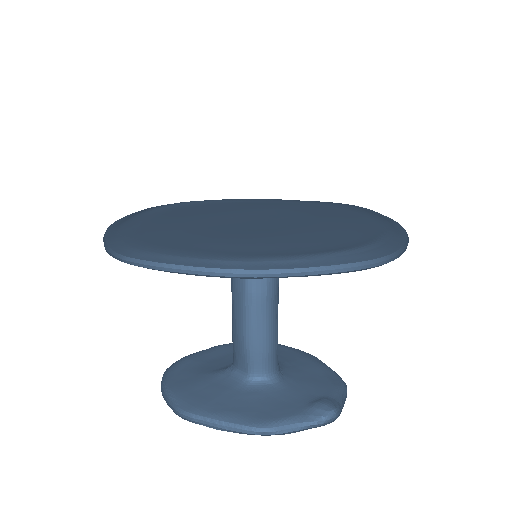}
    \end{subfigure}%
    \vskip\baselineskip%
    \vspace{-1.5em}
    \begin{subfigure}[t]{0.30\linewidth}
	\small{\textit{A table with two pedestals and a table top.}}
    \end{subfigure}%
    \hfill%
    \begin{subfigure}[t]{0.30\linewidth}
	\small{\textit{A table with four legs and a large drawer at the very top.}}
    \end{subfigure}%
    \hfill%
    \begin{subfigure}[t]{0.30\linewidth}
        \small{\textit{A table with a central support and a board.}}
    \end{subfigure}%
    \vskip\baselineskip%
    \vspace{-1.2em}
    \begin{subfigure}[b]{0.31\linewidth}
	\centering
	\includegraphics[trim={2cm 4cm 2cm 2cm},clip,width=0.6\linewidth]{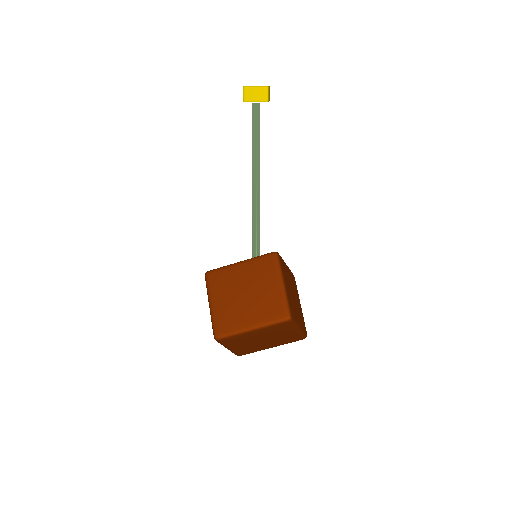}
    \end{subfigure}%
    \hfill%
     \begin{subfigure}[b]{0.31\linewidth}
	\centering
	\includegraphics[trim={2cm 4cm 2cm 2cm},clip,width=0.6\linewidth]{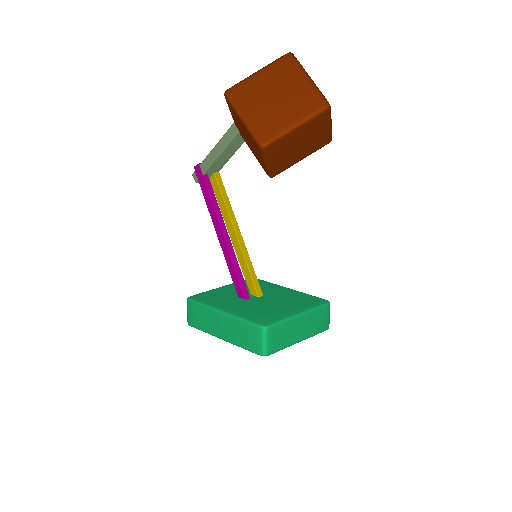}
    \end{subfigure}%
    \hfill%
    \begin{subfigure}[b]{0.31\linewidth}
	\centering
        \includegraphics[trim={2cm 2cm 2cm 2cm},clip,width=0.6\linewidth]{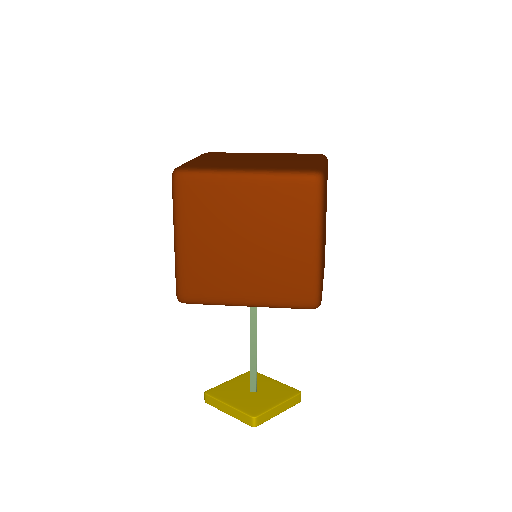}
    \end{subfigure}%
    \vskip\baselineskip%
    \vspace{-0.4em}
    \begin{subfigure}[b]{0.31\linewidth}
	\centering
	\includegraphics[trim={2cm 4cm 2cm 2cm},clip,width=0.6\linewidth]{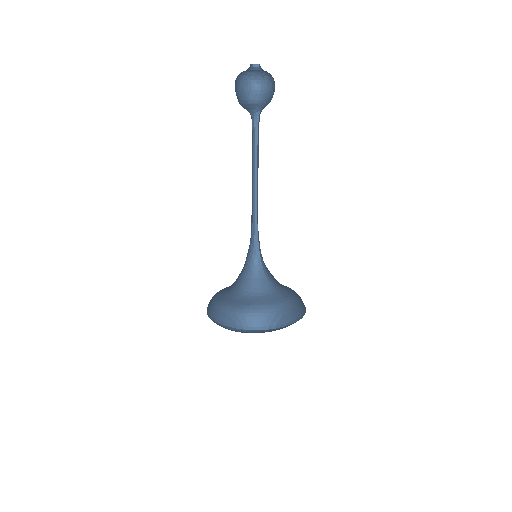}
    \end{subfigure}%
    \hfill%
     \begin{subfigure}[b]{0.31\linewidth}
	\centering
	\includegraphics[trim={2cm 4cm 2cm 2cm},clip,width=0.6\linewidth]{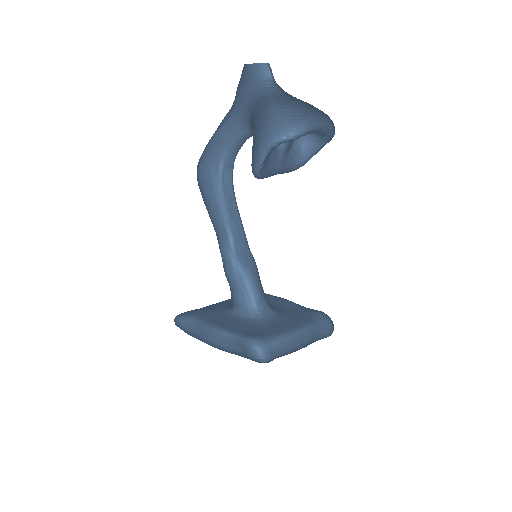}
    \end{subfigure}%
    \hfill%
    \begin{subfigure}[b]{0.31\linewidth}
	\centering
        \includegraphics[trim={2cm 2cm 2cm 3cm},clip,width=0.6\linewidth]{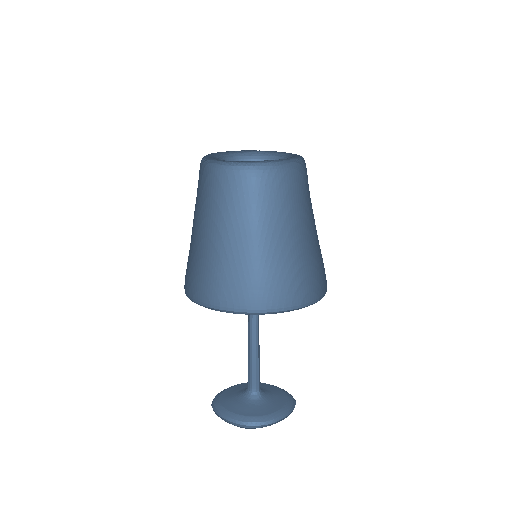}
    \end{subfigure}%
    \vskip\baselineskip%
    \vspace{-1.5em}
    \begin{subfigure}[t]{0.30\linewidth}
	\small{\textit{A lamp with a light bulb inside a lamp shade, connected to the ceiling.}}
    \end{subfigure}%
    \hfill%
    \begin{subfigure}[t]{0.30\linewidth}
	\small{\textit{A lamp with a light bulb held by an arm which is supported by a base and a body.}}
    \end{subfigure}%
    \hfill%
    \begin{subfigure}[t]{0.30\linewidth}
        \small{\textit{A lamp with a light bulb on top of a pole and a base.}}
    \end{subfigure}%
    \vskip\baselineskip%
    \vspace{-1.5em}
    \caption{{\bf Text-guided Shape Generation}. Given different text descriptions 
    our model can generate plausible 3D shapes of chairs, tables, and lamps.}
    \label{fig:text_guided_generation}
    \vspace{-1.5em}
\end{figure}

%% file: sec_conclusion.tex
\section{Conclusion}

In this paper, we introduced PASTA a part-aware generative model for 3D shapes.
Our architecture consists of two
main components: the \emph{object generator} that autoregressively generates
objects as sequences of labelled cuboids and the \emph{blending network}
that combines a sequence of cuboidal primitives and synthesizes a high-quality
implicit shape. Unlike traditional autoregressive models that are
trained with teacher forcing, we demonstrate that relying on scheduled sampling
\cite{Bengio2015NIPS} improves the generation
performance of our model. Our experiments, showcase that PASTA generates
more meaningful part arrangements and plausible 3D objects than both part-based
\cite{Wu2020CVPR, Paschalidou2021NEURIPS} and non part-based generative models
\cite{Chen2019CVPR}. In future work, we plan to extend our architecture to 
generate parts with textures. Note that this is a straight-forward extension of our method
if we simply replace our blending network with a NeRF-based
decoder~\cite{Mildenhall2020ECCV} that instead of predicting occupancies,
predicts colors and opacities. Another exciting direction for future research, is to
explore learning such part-based autoregressive models without explicit part annotations.

%% file: fig/shape_completion_diversity.tex
\begin{figure*}
    \centering
    \begin{subfigure}[b]{0.125\linewidth}
	\centering
        Partial Input
    \end{subfigure}%
    \hfill%
    \begin{subfigure}[b]{0.375\linewidth}
	\centering
        Object Completions
    \end{subfigure}%
    \hfill%
    \begin{subfigure}[b]{0.125\linewidth}
	\centering
        Partial Input
    \end{subfigure}%
    \hfill%
    \begin{subfigure}[b]{0.375\linewidth}
	\centering
        Object Completions
    \end{subfigure}%
    \vskip\baselineskip%
    \vspace{-1.2em}
    \begin{subfigure}[t]{0.125\linewidth}
	\centering
	\includegraphics[width=\linewidth]{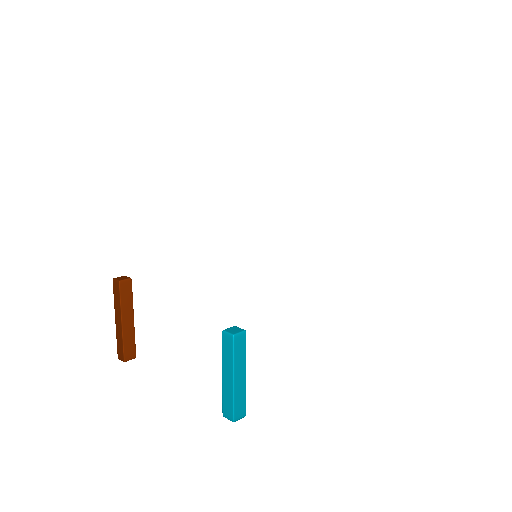}
    \end{subfigure}%
    \hfill%
    \begin{subfigure}[t]{0.375\linewidth}
	 \begin{subfigure}[b]{0.33\linewidth}
		\centering
		\includegraphics[width=\linewidth]{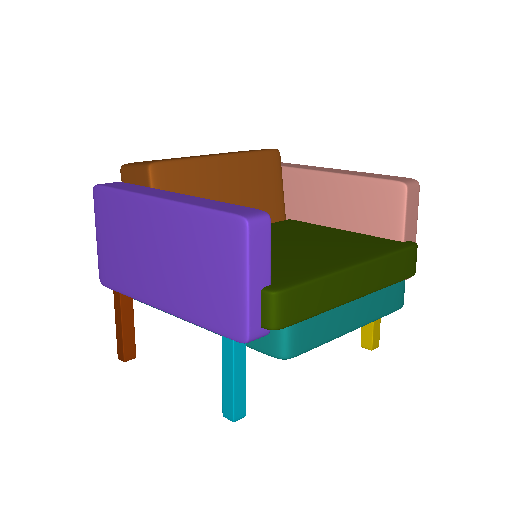}
    \end{subfigure}%
    \hfill%
    \begin{subfigure}[b]{0.33\linewidth}
		\centering
		\includegraphics[width=\linewidth]{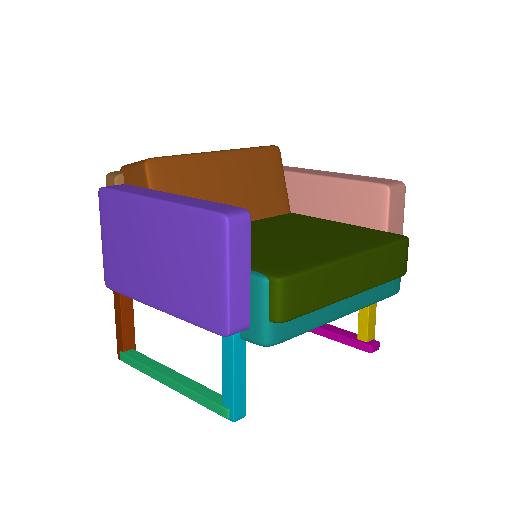}
    \end{subfigure}%
    \hfill%
    \begin{subfigure}[b]{0.33\linewidth}
		\centering
		\includegraphics[width=\linewidth]{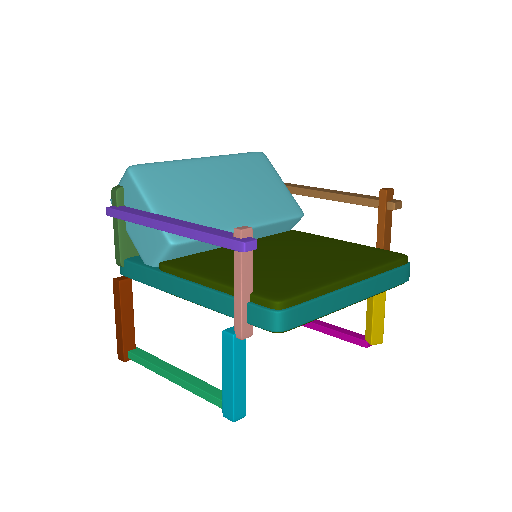}
    \end{subfigure}%
    \vskip\baselineskip%
    \vspace{-2.3em}
    \begin{subfigure}[b]{0.33\linewidth}
		\centering
		\includegraphics[width=\linewidth]{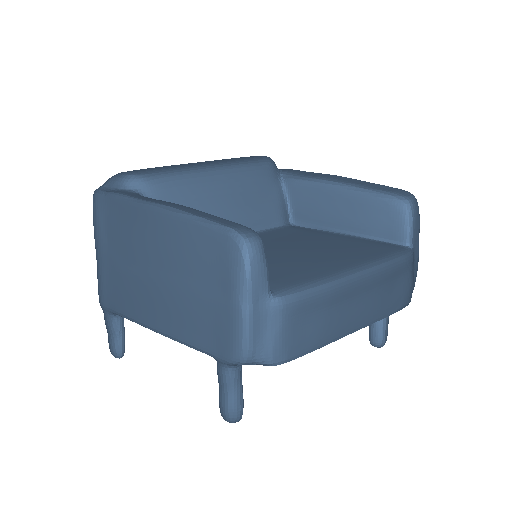}
    \end{subfigure}%
    \hfill%
    \begin{subfigure}[b]{0.33\linewidth}
		\centering
		\includegraphics[width=\linewidth]{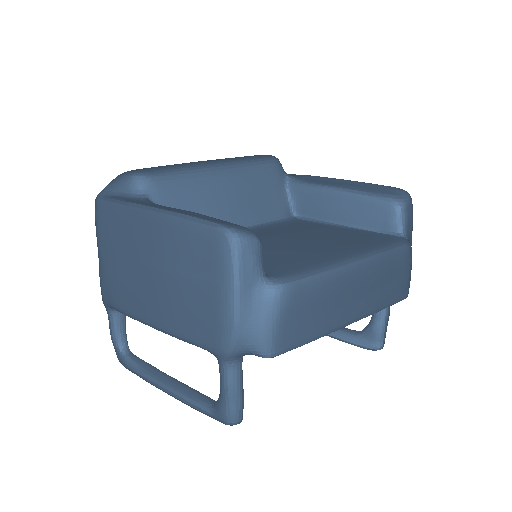}
    \end{subfigure}%
    \hfill%
    \begin{subfigure}[b]{0.33\linewidth}
		\centering
		\includegraphics[width=\linewidth]{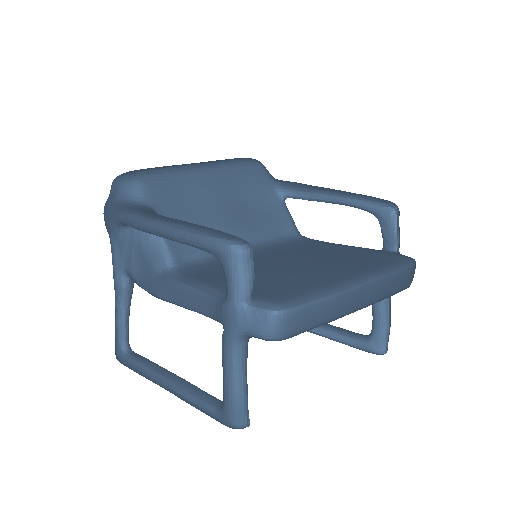}
    \end{subfigure}%
    \end{subfigure}%
    \hfill%
    \begin{subfigure}[t]{0.125\linewidth}
	\centering
	\includegraphics[width=\linewidth]{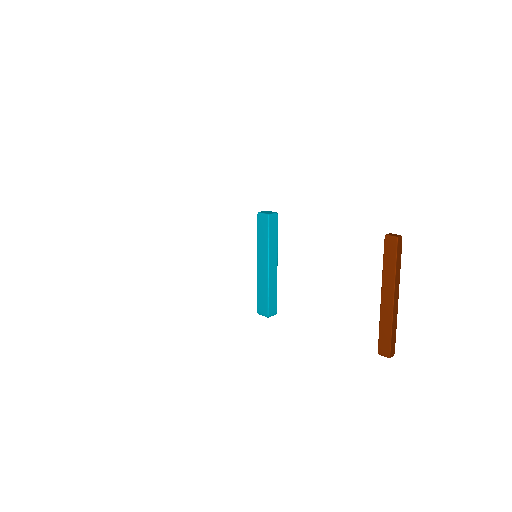}
    \end{subfigure}%
    \hfill%
    \begin{subfigure}[t]{0.375\linewidth}
    \begin{subfigure}[b]{0.33\linewidth}
		\centering
		\includegraphics[width=\linewidth]{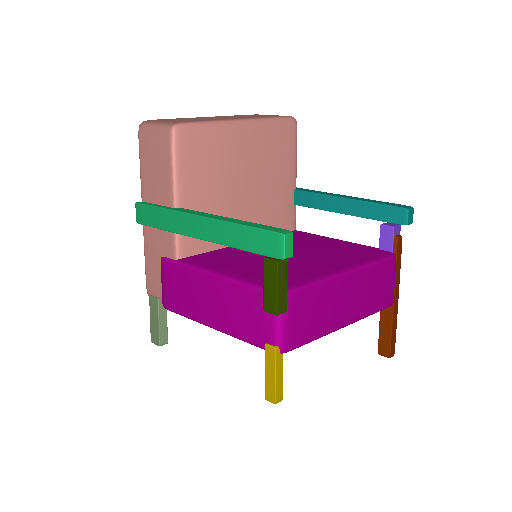}
    \end{subfigure}%
    \hfill%
    \begin{subfigure}[b]{0.33\linewidth}
		\centering
		\includegraphics[width=\linewidth]{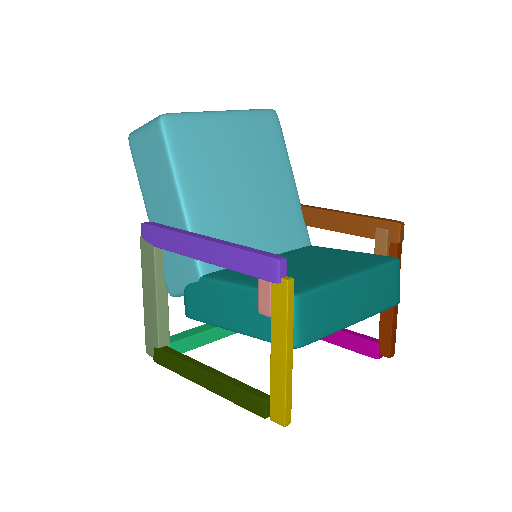}
    \end{subfigure}%
    \hfill%
    \begin{subfigure}[b]{0.33\linewidth}
		\centering
		\includegraphics[width=\linewidth]{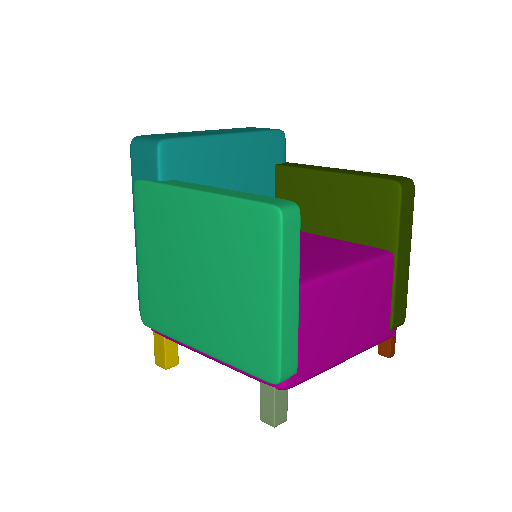}
    \end{subfigure}%
    \vskip\baselineskip%
    \vspace{-2.3em}
    \begin{subfigure}[b]{0.33\linewidth}
		\centering
		\includegraphics[width=\linewidth]{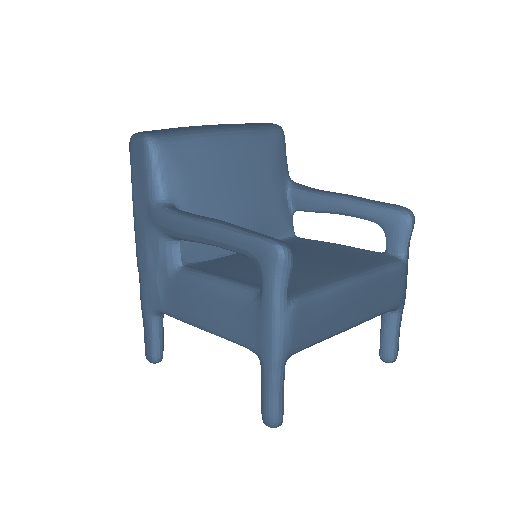}
    \end{subfigure}%
    \hfill%
    \begin{subfigure}[b]{0.33\linewidth}
		\centering
		\includegraphics[width=\linewidth]{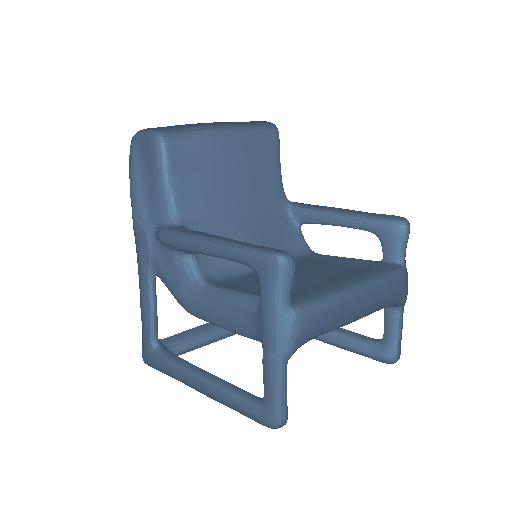}
    \end{subfigure}%
    \hfill%
    \begin{subfigure}[b]{0.33\linewidth}
		\centering
		\includegraphics[width=\linewidth]{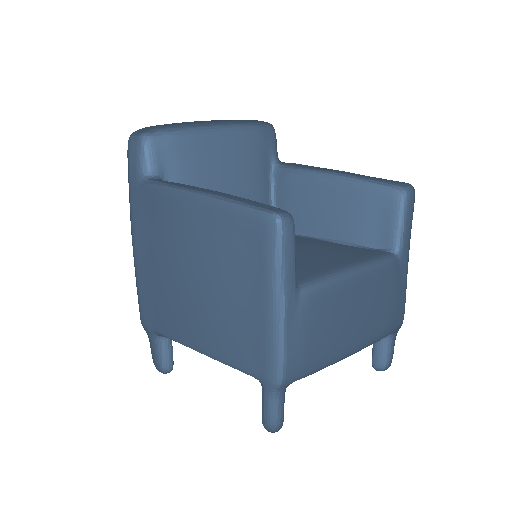}
    \end{subfigure}%
    \end{subfigure}%
    \vskip\baselineskip%
    \vspace{-2.3em}
    \begin{subfigure}[t]{0.125\linewidth}
	\centering
	\includegraphics[width=\linewidth]{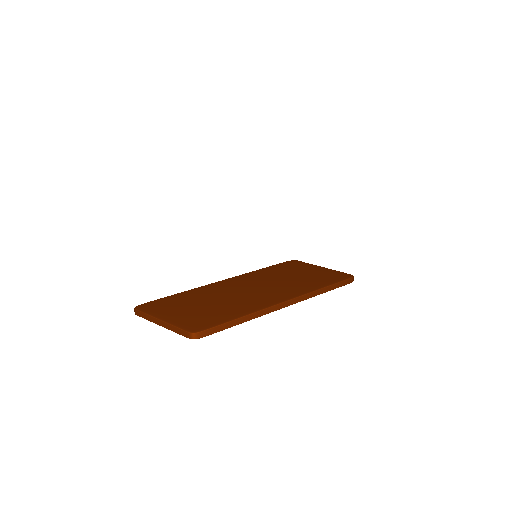}
    \end{subfigure}%
    \hfill%
    \begin{subfigure}[t]{0.375\linewidth}
	\begin{subfigure}[b]{0.33\linewidth}
		\centering
		\includegraphics[width=\linewidth]{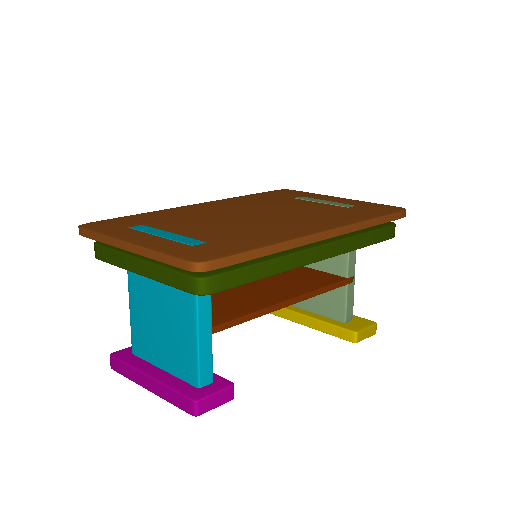}
    \end{subfigure}%
    \hfill%
    \begin{subfigure}[b]{0.33\linewidth}
		\centering
		\includegraphics[width=\linewidth]{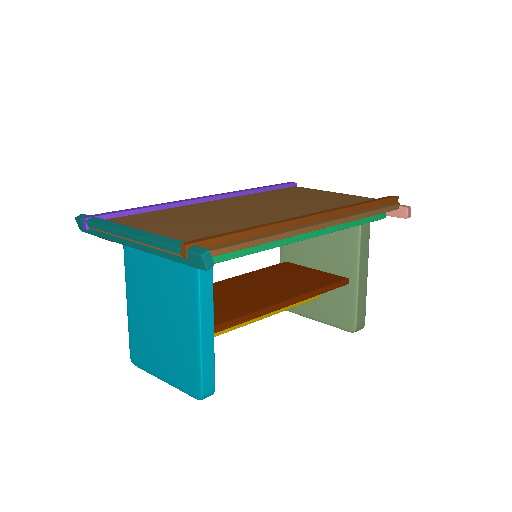}
    \end{subfigure}%
    \hfill%
    \begin{subfigure}[b]{0.33\linewidth}
		\centering
		\includegraphics[width=\linewidth]{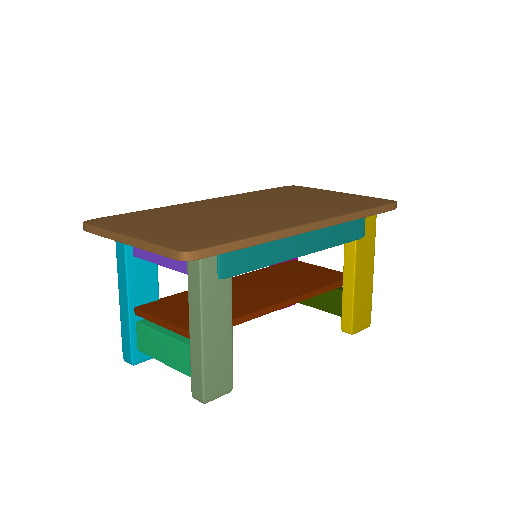}
    \end{subfigure}%
    \vskip\baselineskip%
    \vspace{-2.6em}
     \begin{subfigure}[b]{0.33\linewidth}
		\centering
		\includegraphics[width=\linewidth]{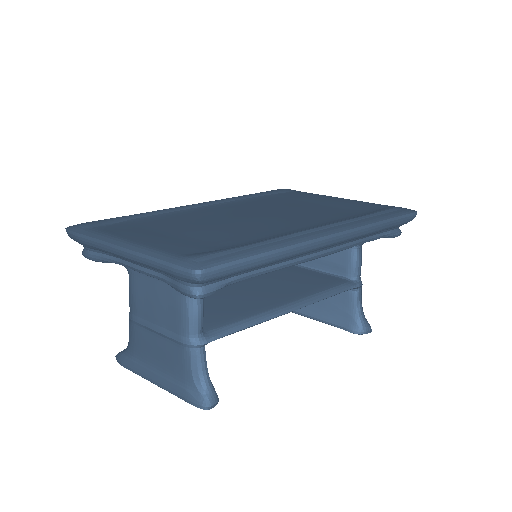}
    \end{subfigure}%
    \hfill%
    \begin{subfigure}[b]{0.33\linewidth}
		\centering
		\includegraphics[width=\linewidth]{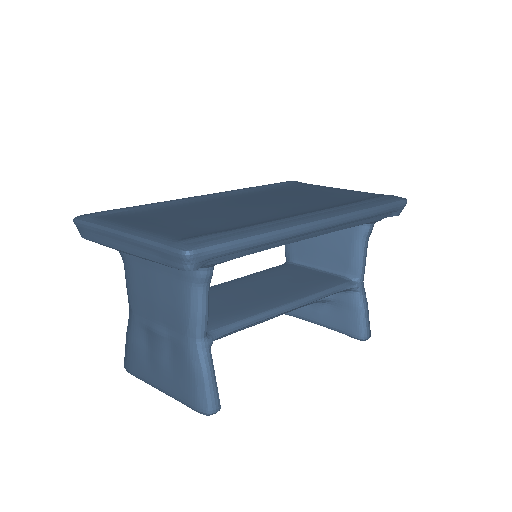}
    \end{subfigure}%
    \hfill%
    \begin{subfigure}[b]{0.33\linewidth}
		\centering
		\includegraphics[width=\linewidth]{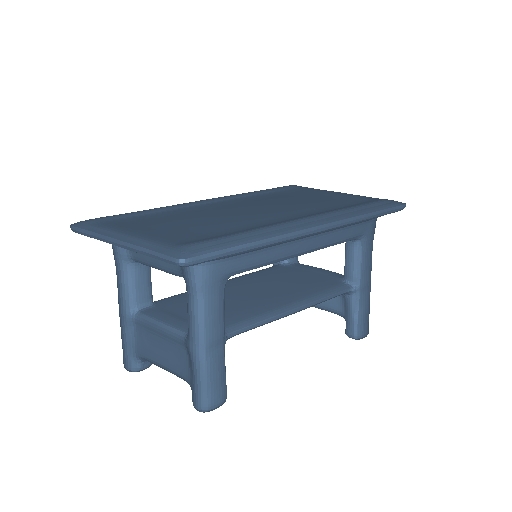}
    \end{subfigure}%
    \end{subfigure}%
    \hfill%
    \begin{subfigure}[t]{0.125\linewidth}
	\centering
	\includegraphics[width=\linewidth]{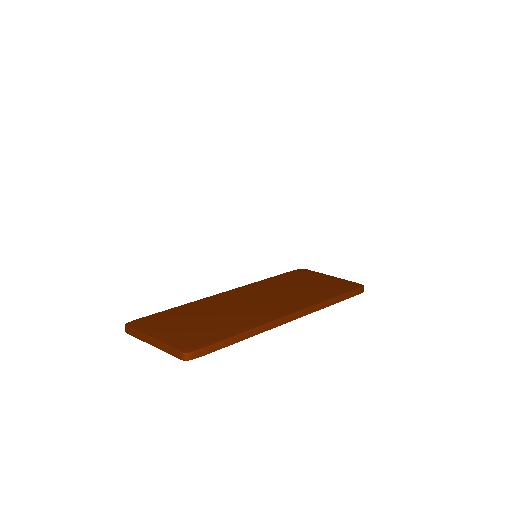}
    \end{subfigure}%
    \hfill%
    \begin{subfigure}[t]{0.375\linewidth}
	\begin{subfigure}[b]{0.33\linewidth}
		\centering
		\includegraphics[width=\linewidth]{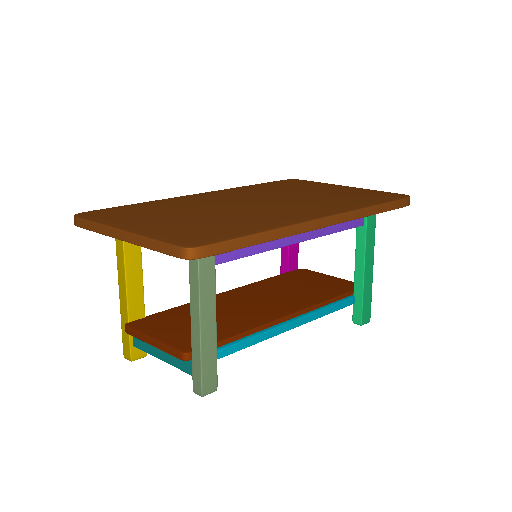}
    \end{subfigure}%
    \hfill%
    \begin{subfigure}[b]{0.33\linewidth}
		\centering
		\includegraphics[width=\linewidth]{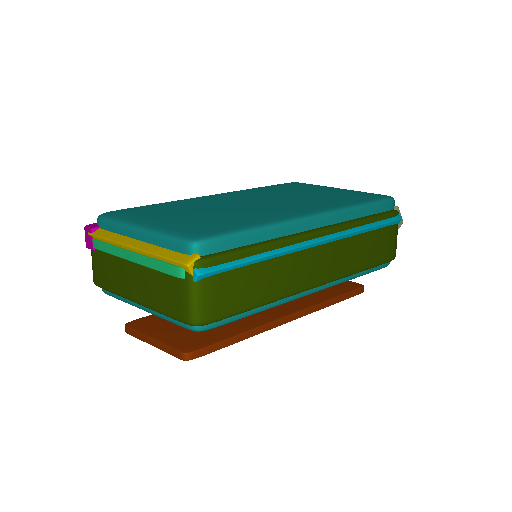}
    \end{subfigure}%
    \hfill%
    \begin{subfigure}[b]{0.33\linewidth}
		\centering
		\includegraphics[width=\linewidth]{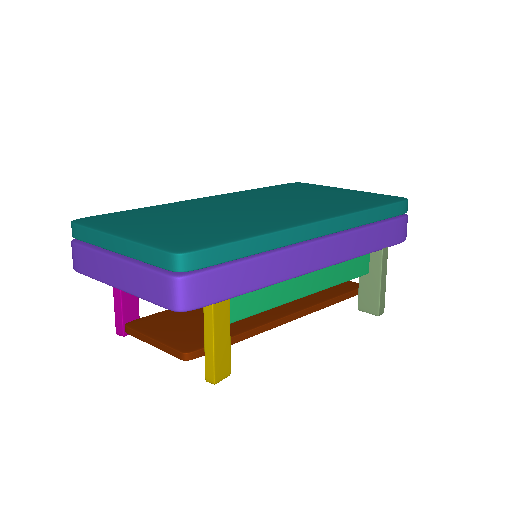}
    \end{subfigure}%
    \vskip\baselineskip%
    \vspace{-2.8em}
    \begin{subfigure}[b]{0.33\linewidth}
		\centering
		\includegraphics[width=\linewidth]{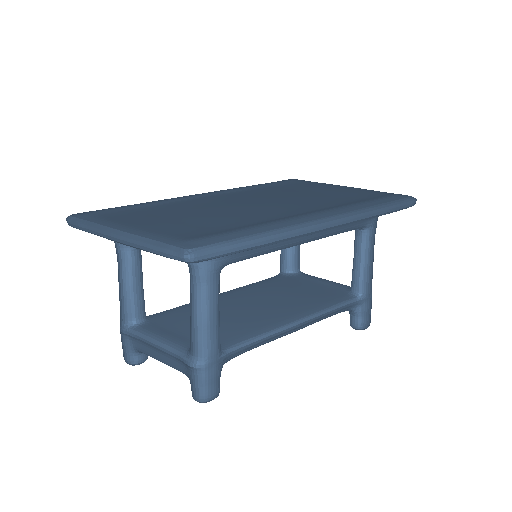}
    \end{subfigure}%
    \hfill%
    \begin{subfigure}[b]{0.33\linewidth}
		\centering
		\includegraphics[width=\linewidth]{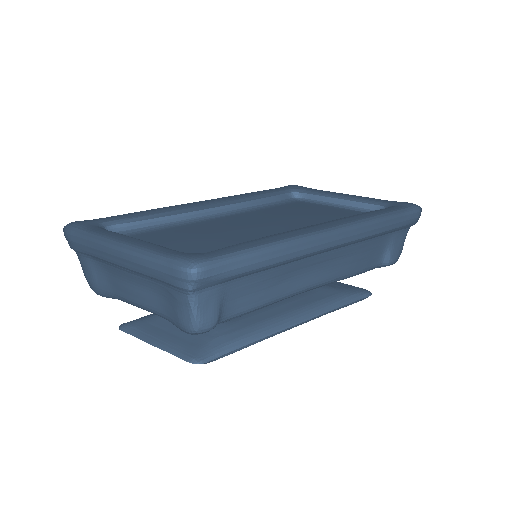}
    \end{subfigure}%
    \hfill%
    \begin{subfigure}[b]{0.33\linewidth}
		\centering
		\includegraphics[width=\linewidth]{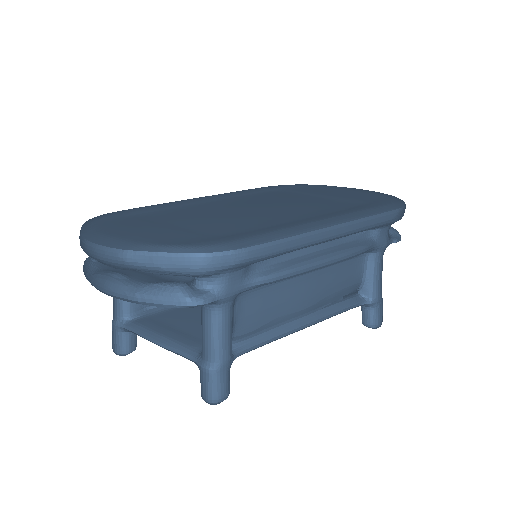}
    \end{subfigure}%
    \end{subfigure}%
    \vskip\baselineskip%
    \vspace{-2.3em}
    \begin{subfigure}[t]{0.125\linewidth}
	\centering
        \includegraphics[width=\linewidth]{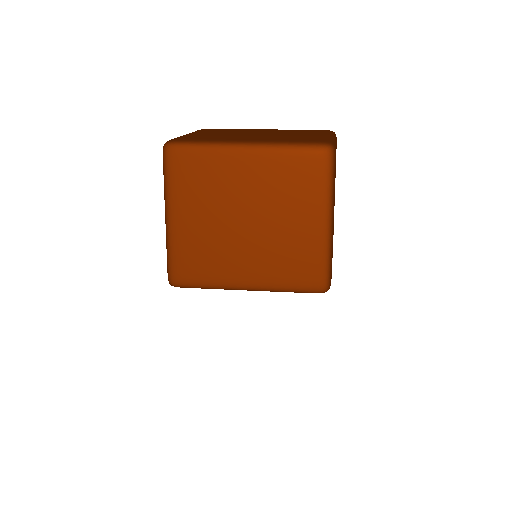}
    \end{subfigure}%
    \hfill%
    \begin{subfigure}[t]{0.375\linewidth}
    \begin{subfigure}[b]{0.33\linewidth}
		\centering
		\includegraphics[width=\linewidth]{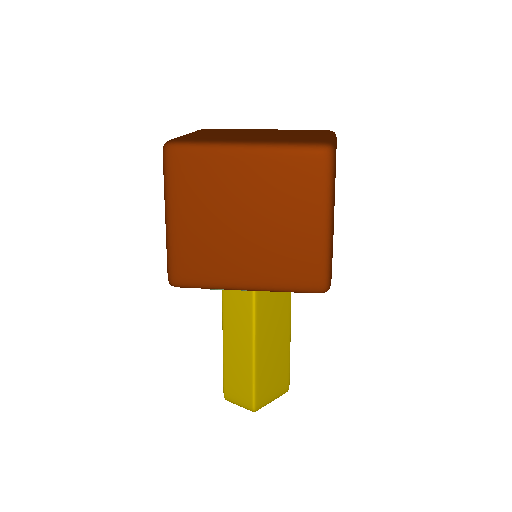}
    \end{subfigure}%
    \hfill%
    \begin{subfigure}[b]{0.33\linewidth}
		\centering
		\includegraphics[width=\linewidth]{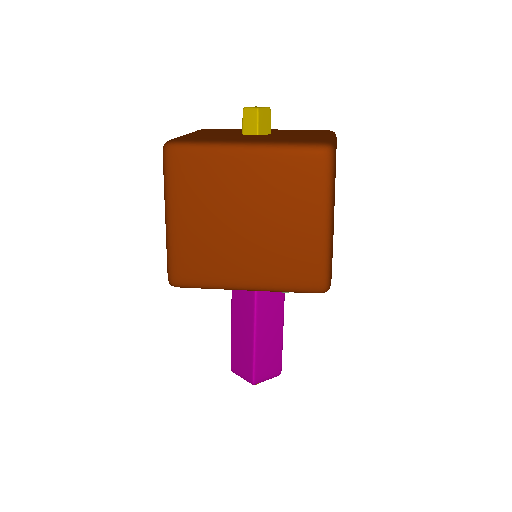}
    \end{subfigure}%
    \hfill%
    \begin{subfigure}[b]{0.33\linewidth}
		\centering
		\includegraphics[width=\linewidth]{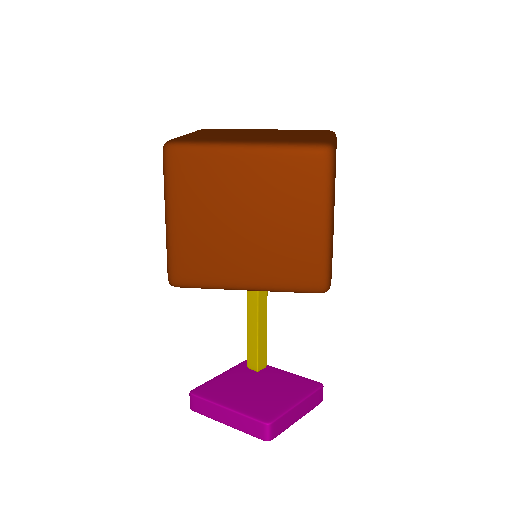}
    \end{subfigure}%
    \vskip\baselineskip%
    \vspace{-2.5em}
     \begin{subfigure}[b]{0.33\linewidth}
		\centering
		\includegraphics[width=\linewidth]{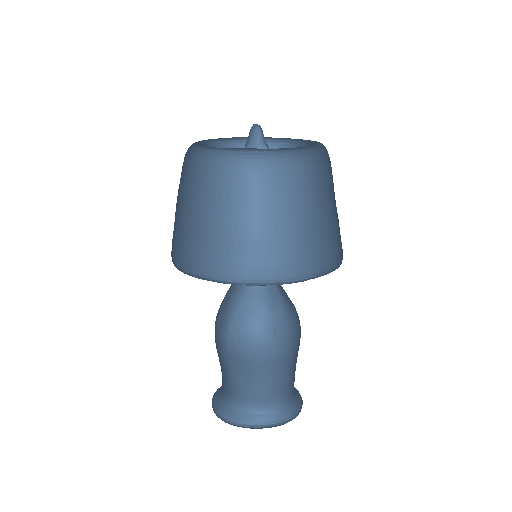}
    \end{subfigure}%
    \hfill%
    \begin{subfigure}[b]{0.33\linewidth}
		\centering
		\includegraphics[width=\linewidth]{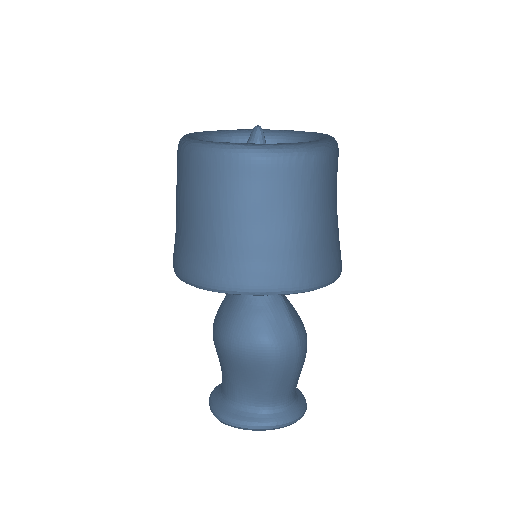}
    \end{subfigure}%
    \hfill%
    \begin{subfigure}[b]{0.33\linewidth}
		\centering
		\includegraphics[width=\linewidth]{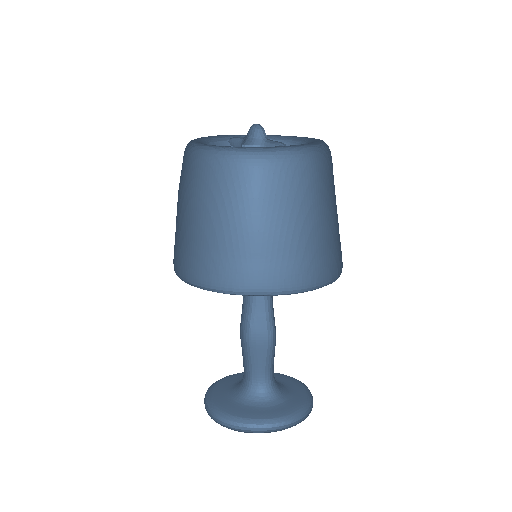}
    \end{subfigure}%
    \end{subfigure}%
    \hfill%
    \begin{subfigure}[t]{0.125\linewidth}
	\centering
        \includegraphics[width=\linewidth]{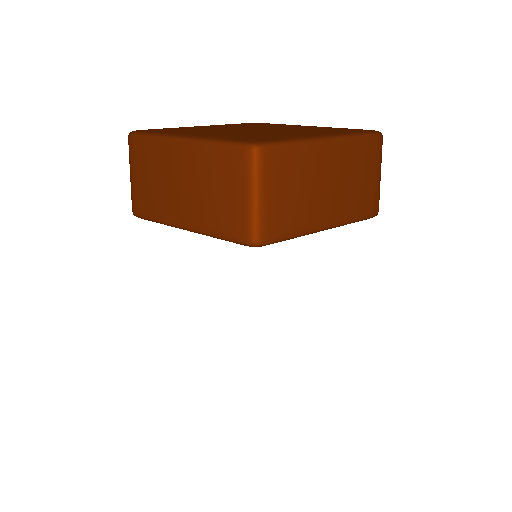}
    \end{subfigure}%
    \hfill%
    \begin{subfigure}[t]{0.375\linewidth}
   \begin{subfigure}[b]{0.33\linewidth}
		\centering
		\includegraphics[width=\linewidth]{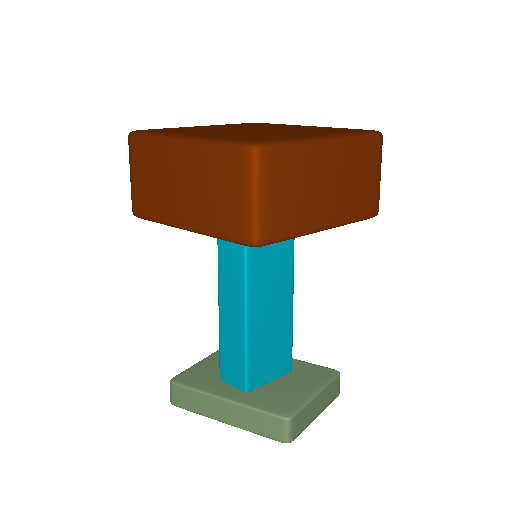}
    \end{subfigure}%
    \hfill%
    \begin{subfigure}[b]{0.33\linewidth}
		\centering
		\includegraphics[width=\linewidth]{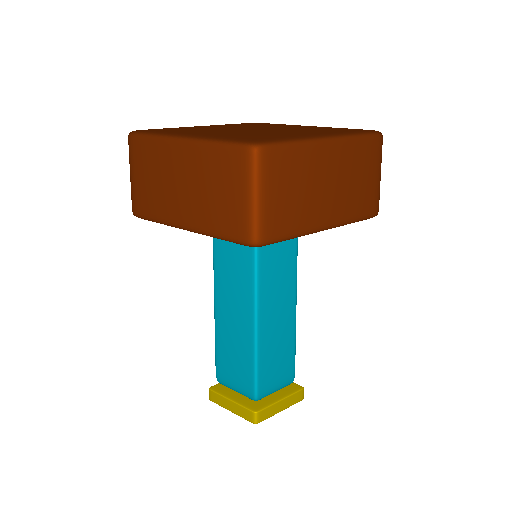}
    \end{subfigure}%
    \hfill%
    \begin{subfigure}[b]{0.33\linewidth}
		\centering
		\includegraphics[width=\linewidth]{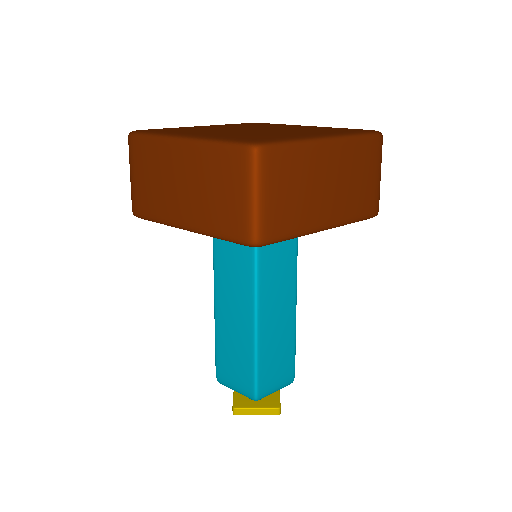}
    \end{subfigure}%
    \vskip\baselineskip%
    \vspace{-2.2em}
    \begin{subfigure}[b]{0.33\linewidth}
		\centering
		\includegraphics[width=\linewidth]{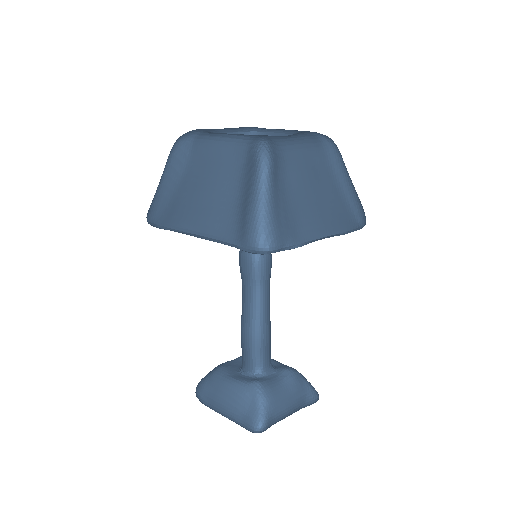}
    \end{subfigure}%
    \hfill%
    \begin{subfigure}[b]{0.33\linewidth}
		\centering
		\includegraphics[width=\linewidth]{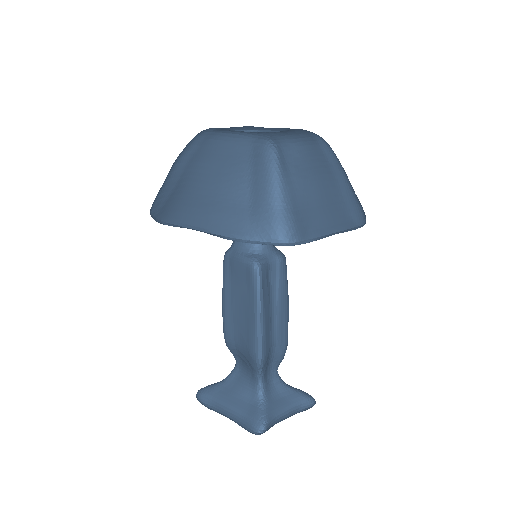}
    \end{subfigure}%
    \hfill%
    \begin{subfigure}[b]{0.33\linewidth}
		\centering
		\includegraphics[width=\linewidth]{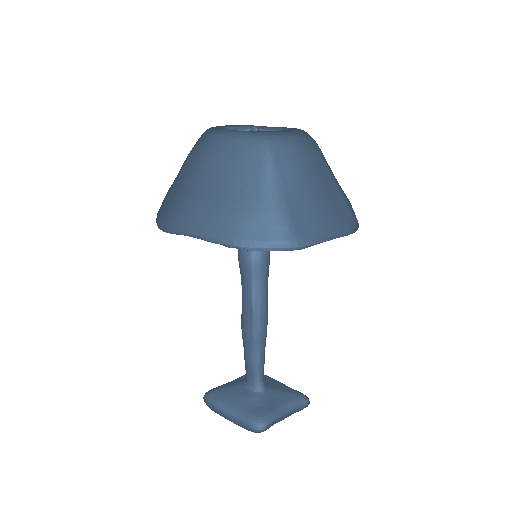}
    \end{subfigure}%
    \end{subfigure}%
    \vskip\baselineskip%
    \vspace{-2.2em}
    \caption{{\bf Diversity of Object Completions}. Starting from a partial object, we show several completions using our model.
    The first two rows show completion results for chairs, the third and fourth for tables and the last two for lamps.}
    \label{fig:shapenet_diversity}
\end{figure*}

%% file: fig/shape_generation_image_guided.tex
\begin{figure*}
    \vspace{1.75em}
    \centering
    \begin{subfigure}[b]{0.11\linewidth}
	\centering
        Input Image
    \end{subfigure}%
    \hfill%
    \begin{subfigure}[b]{0.11\linewidth}
	\centering
        Ours-Parts
    \end{subfigure}%
    \hfill%
    \begin{subfigure}[b]{0.11\linewidth}
	\centering
        Ours
    \end{subfigure}%
    \hfill%
     \begin{subfigure}[b]{0.11\linewidth}
	\centering
        Input Image
    \end{subfigure}%
    \hfill%
    \begin{subfigure}[b]{0.11\linewidth}
	\centering
        Ours-Parts
    \end{subfigure}%
    \hfill%
    \begin{subfigure}[b]{0.11\linewidth}
	\centering
        Ours
    \end{subfigure}%
    \hfill%
     \begin{subfigure}[b]{0.11\linewidth}
	\centering
        Input Image
    \end{subfigure}%
    \hfill%
    \begin{subfigure}[b]{0.11\linewidth}
	\centering
        Ours-Parts
    \end{subfigure}%
    \hfill%
    \begin{subfigure}[b]{0.11\linewidth}
	\centering
        Ours
    \end{subfigure}%
    \vskip\baselineskip%
    \vspace{-1.0em}
    \begin{subfigure}[b]{0.11\linewidth}
        \centering
	\includegraphics[width=\linewidth]{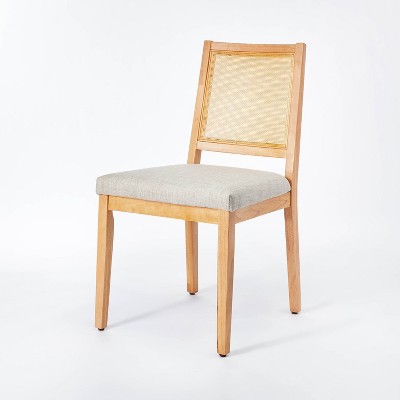}
    \end{subfigure}%
    \hfill%
    \begin{subfigure}[b]{0.11\linewidth}
        \centering
	\includegraphics[width=\linewidth]{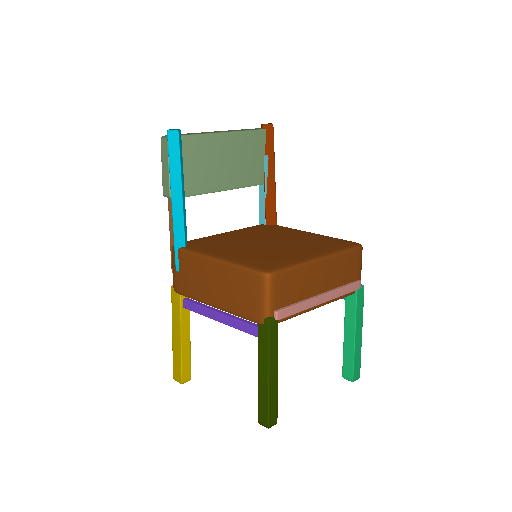}
    \end{subfigure}%
    \hfill%
    \begin{subfigure}[b]{0.11\linewidth}
	\centering
        \includegraphics[width=\linewidth]{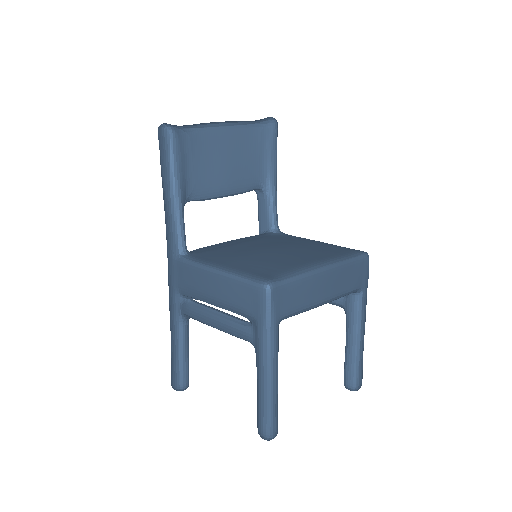}
    \end{subfigure}%
     \hfill%
    \begin{subfigure}[b]{0.11\linewidth}
        \centering
        \includegraphics[width=\linewidth]{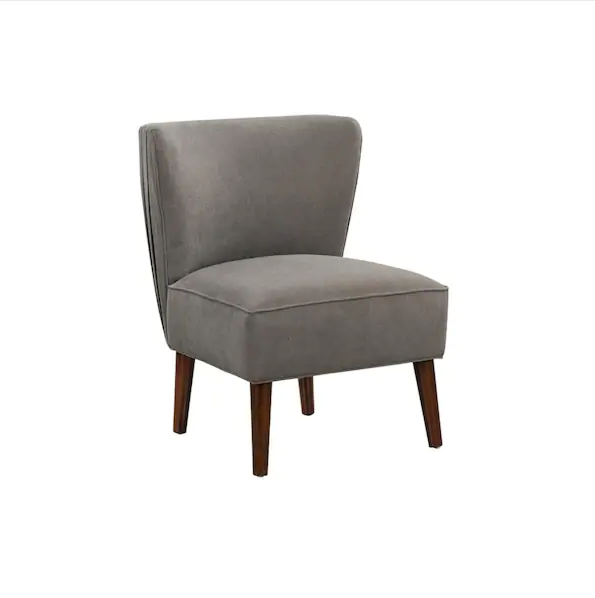}
    \end{subfigure}%
    \hfill%
    \begin{subfigure}[b]{0.11\linewidth}
        \centering
        \includegraphics[width=\linewidth]{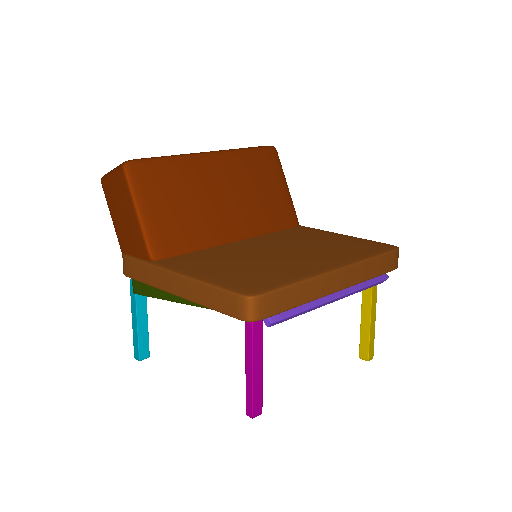}
    \end{subfigure}%
    \hfill%
    \begin{subfigure}[b]{0.11\linewidth}
	\centering
        \includegraphics[width=\linewidth]{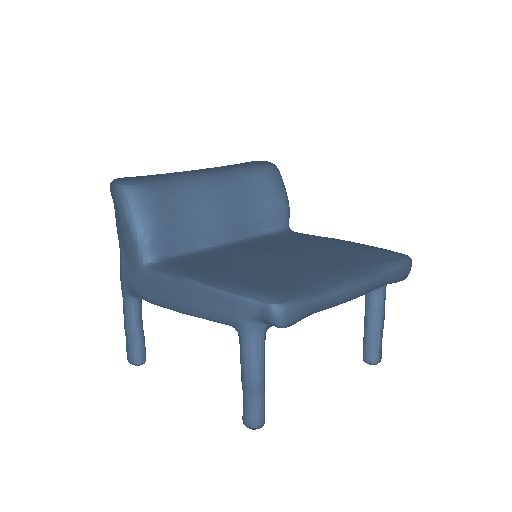}
    \end{subfigure}%
    \hfill%
    \begin{subfigure}[b]{0.11\linewidth}
        \centering
        \includegraphics[width=\linewidth]{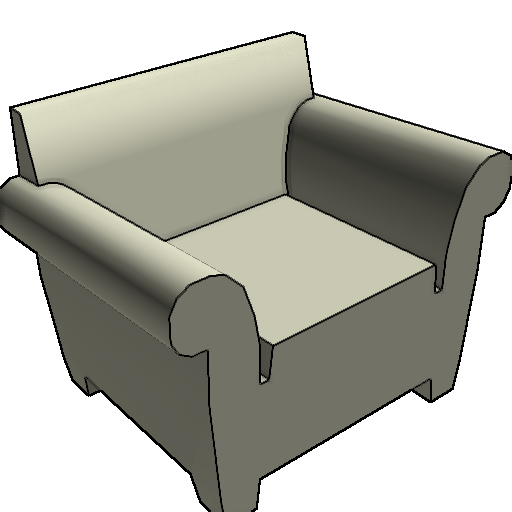}
    \end{subfigure}%
    \hfill%
    \begin{subfigure}[b]{0.11\linewidth}
        \centering
        \includegraphics[width=\linewidth]{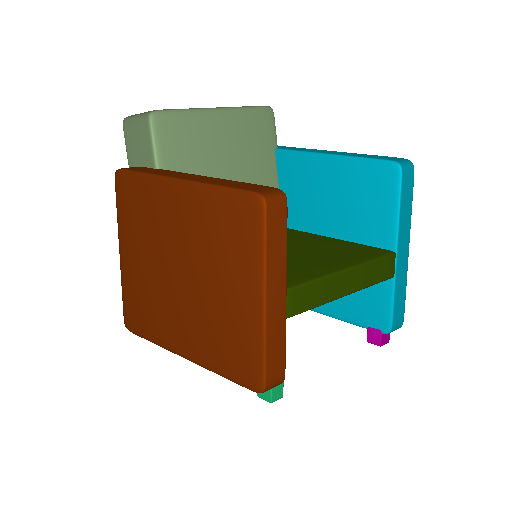}
    \end{subfigure}%
    \hfill%
    \begin{subfigure}[b]{0.11\linewidth}
	\centering
        \includegraphics[width=\linewidth]{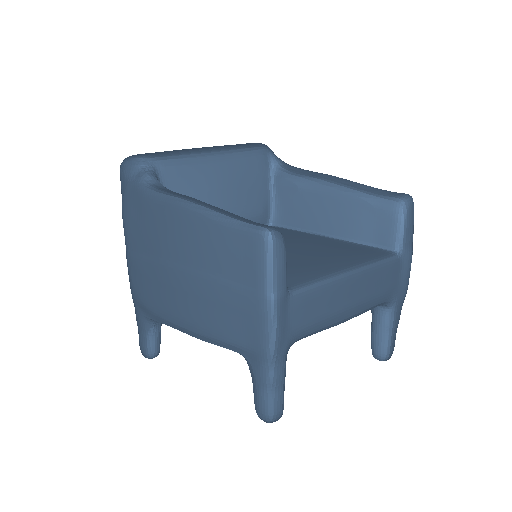}
    \end{subfigure}%
    \vskip\baselineskip%
    \vspace{-1.0em}
        \begin{subfigure}[b]{0.11\linewidth}
        \centering
	\includegraphics[width=\linewidth]{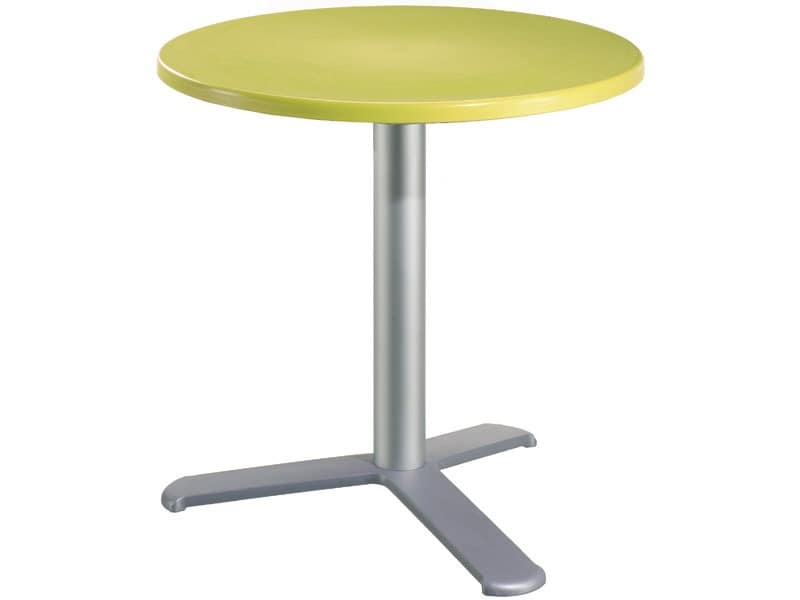}
    \end{subfigure}%
    \hfill%
    \begin{subfigure}[b]{0.11\linewidth}
        \centering
	\includegraphics[width=\linewidth]{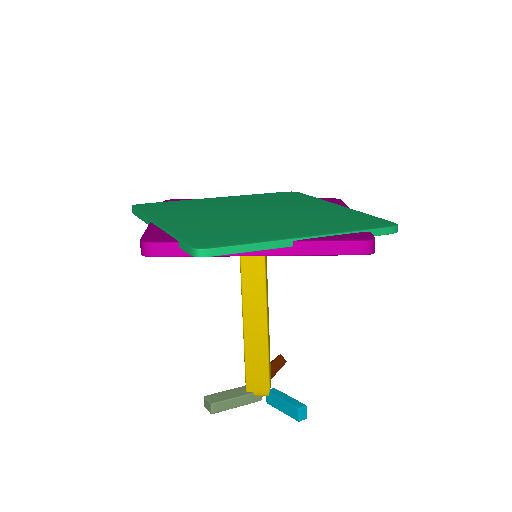}
    \end{subfigure}%
    \hfill%
    \begin{subfigure}[b]{0.11\linewidth}
	\centering
	\includegraphics[width=\linewidth]{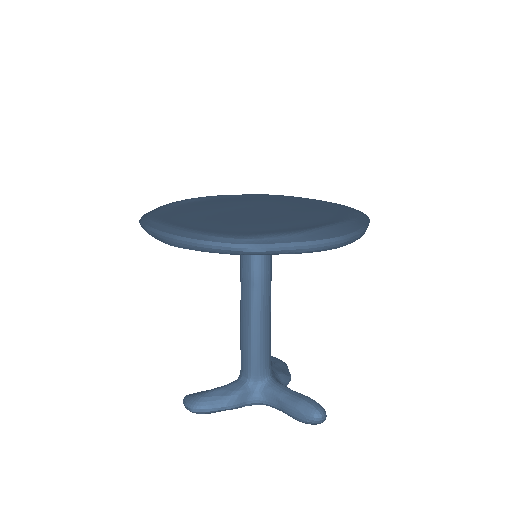}
    \end{subfigure}%
    \hfill%
    \begin{subfigure}[b]{0.11\linewidth}
	\centering
        \includegraphics[width=\linewidth]{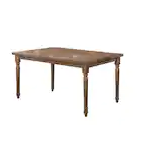}
    \end{subfigure}%
    \hfill%
    \begin{subfigure}[b]{0.11\linewidth}
        \centering
     \includegraphics[width=\linewidth]{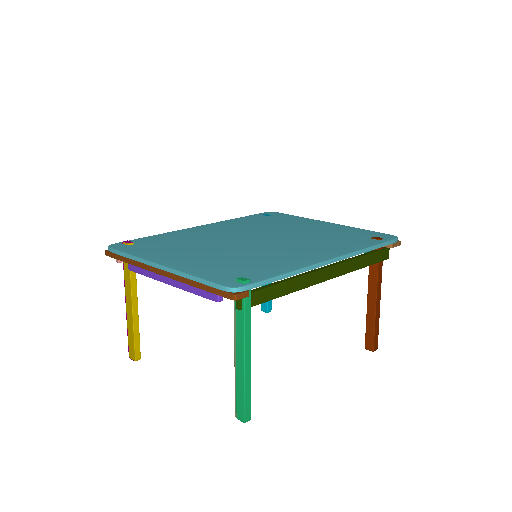}
    \end{subfigure}%
    \hfill%
    \begin{subfigure}[b]{0.11\linewidth}
	\centering
        \includegraphics[width=\linewidth]{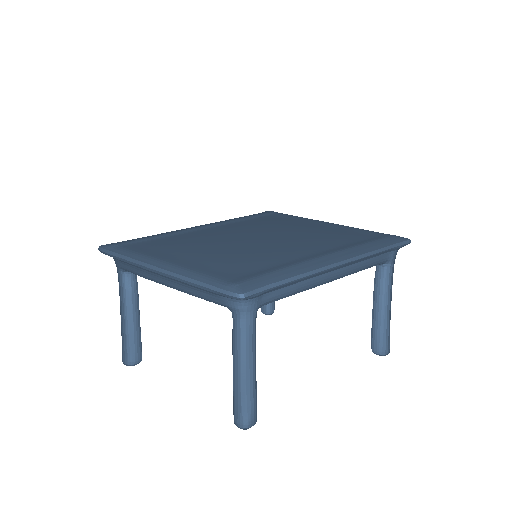}
    \end{subfigure}%
    \hfill%
    \begin{subfigure}[b]{0.11\linewidth}
	\centering
        \includegraphics[width=\linewidth]{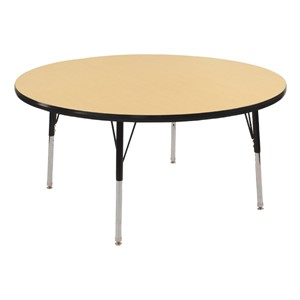}
    \end{subfigure}%
    \hfill%
    \begin{subfigure}[b]{0.11\linewidth}
        \centering
     \includegraphics[width=\linewidth]{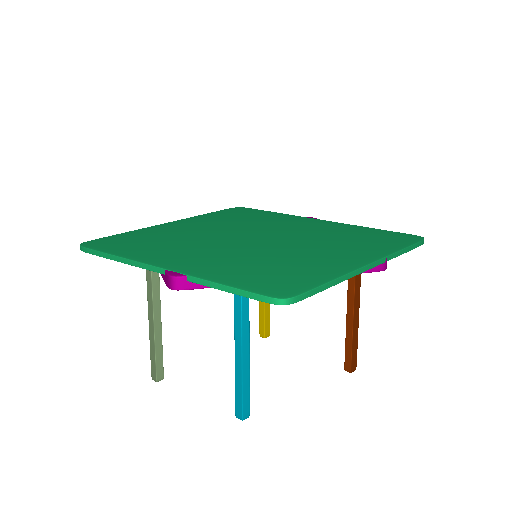}
    \end{subfigure}%
    \hfill%
    \begin{subfigure}[b]{0.11\linewidth}
	\centering
        \includegraphics[width=\linewidth]{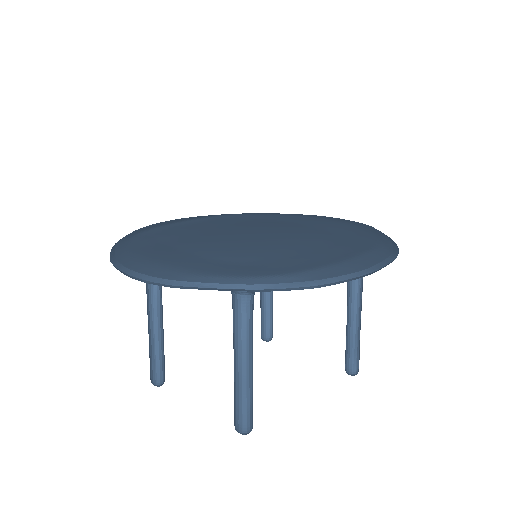}
    \end{subfigure}%
    \vskip\baselineskip%
    \vspace{-1.0em}
    \begin{subfigure}[b]{0.11\linewidth}
        \centering
	\includegraphics[width=\linewidth]{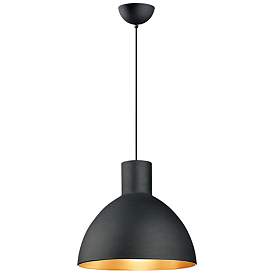}
    \end{subfigure}%
    \hfill%
    \begin{subfigure}[b]{0.11\linewidth}
        \centering
	\includegraphics[width=\linewidth]{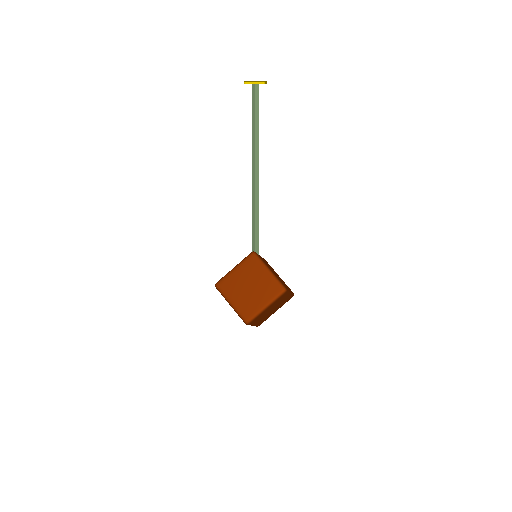}
    \end{subfigure}%
    \hfill%
    \begin{subfigure}[b]{0.11\linewidth}
	\centering
	\includegraphics[width=\linewidth]{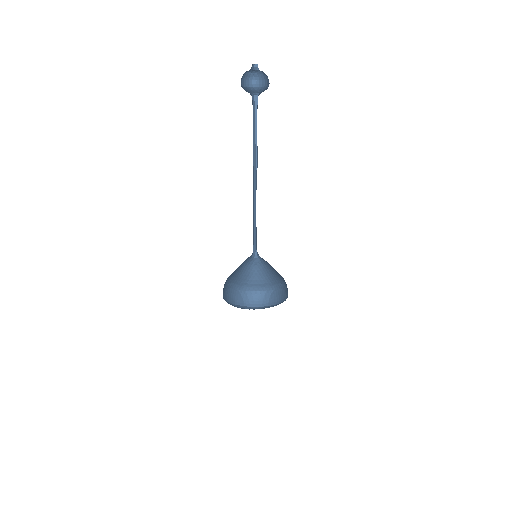}
    \end{subfigure}%
    \hfill%
    \begin{subfigure}[b]{0.11\linewidth}
	\centering
        \includegraphics[width=\linewidth]{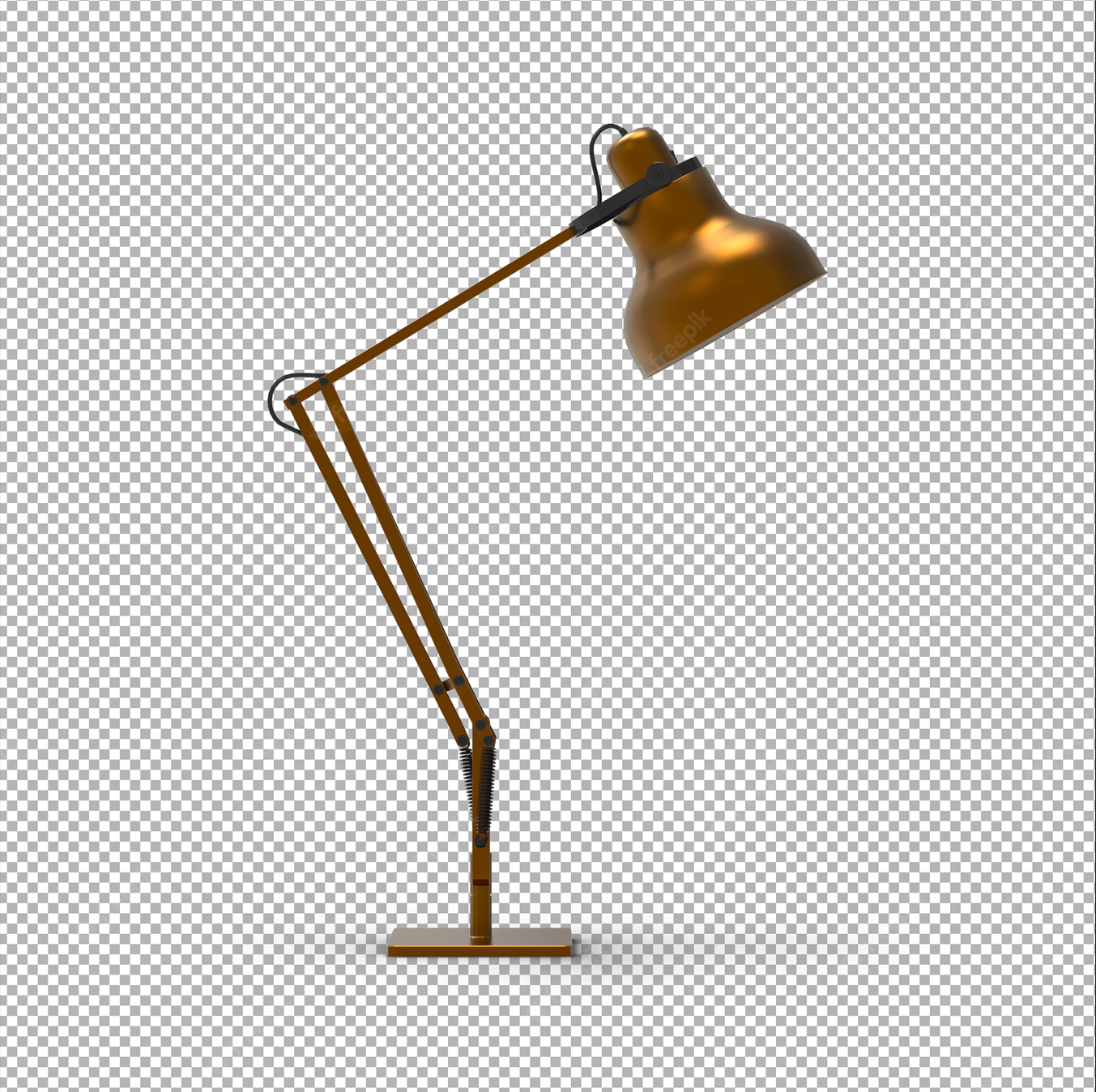}
    \end{subfigure}%
    \hfill%
    \begin{subfigure}[b]{0.11\linewidth}
        \centering
        \includegraphics[width=\linewidth]{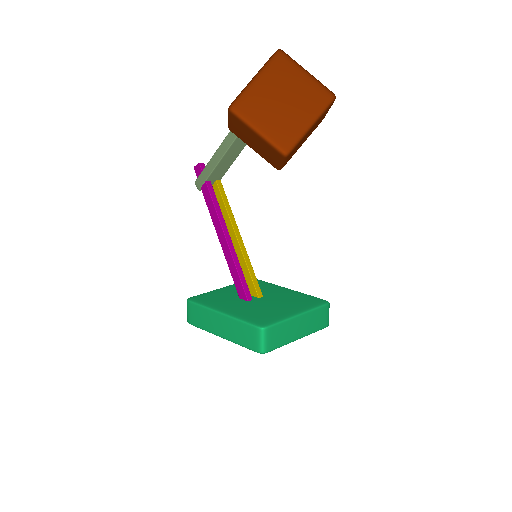}
    \end{subfigure}%
    \hfill%
    \begin{subfigure}[b]{0.11\linewidth}
	\centering
        \includegraphics[width=\linewidth]{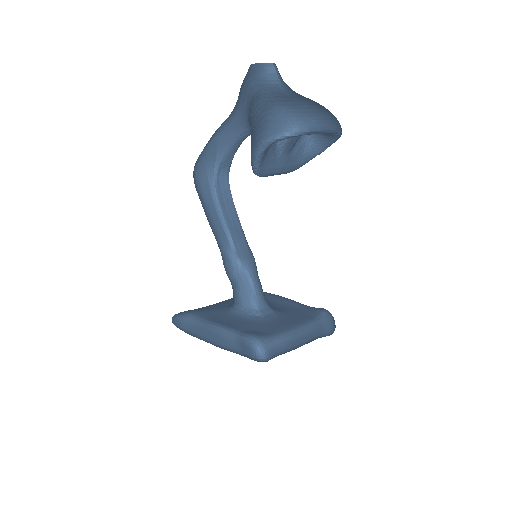}
    \end{subfigure}%
     \hfill%
    \begin{subfigure}[b]{0.11\linewidth}
	\centering
        \includegraphics[width=\linewidth]{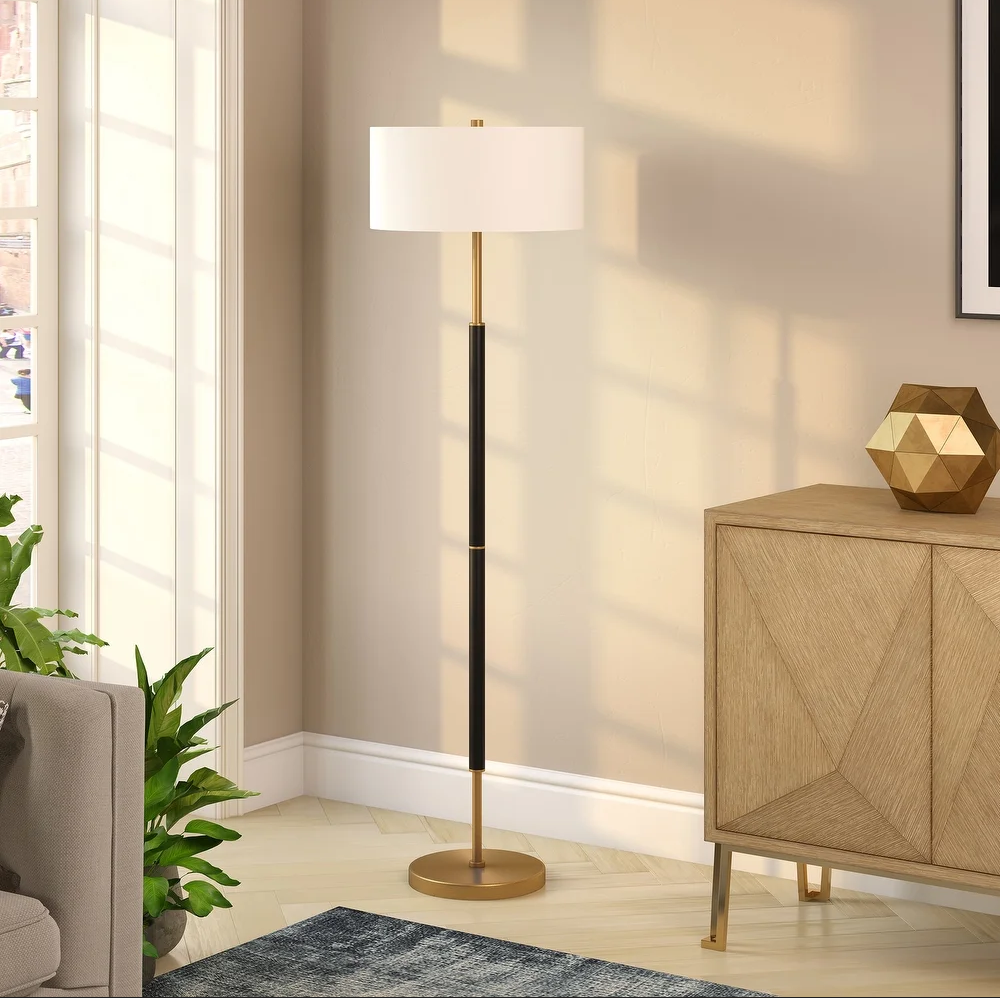}
    \end{subfigure}%
    \hfill%
    \begin{subfigure}[b]{0.11\linewidth}
        \centering
        \includegraphics[width=\linewidth]{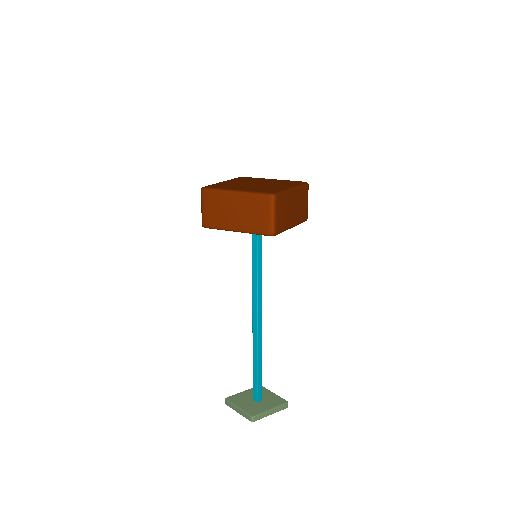}
    \end{subfigure}%
    \hfill%
    \begin{subfigure}[b]{0.11\linewidth}
	\centering
        \includegraphics[width=\linewidth]{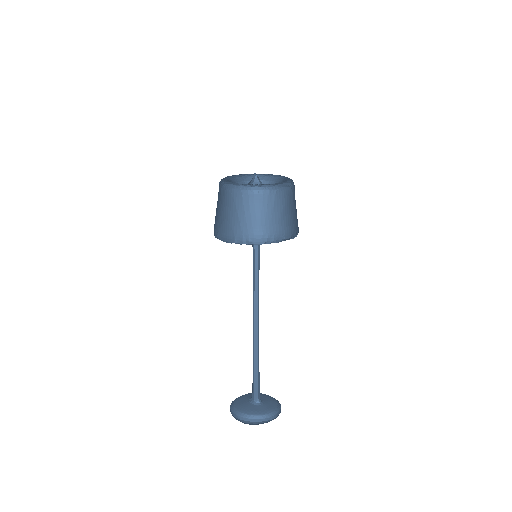}
    \end{subfigure}%
    \vskip\baselineskip%
    \vspace{-1.2em}
    \caption{{\bf Image-guided Shape Generation}. Conditioned on various images from
    different object categories, our model can generate plausible 3D shapes of
    chairs, tables, and lamps that match the input.}
    \label{fig:image_guided_generation}
\end{figure*}

%% file: fig/shape_generation_qualitative_lamps.tex
\begin{figure}
    \begin{subfigure}[t]{\linewidth}
    \centering
    \begin{subfigure}[b]{0.20\linewidth}
        \centering
	IM-NET
    \end{subfigure}%
    \hfill%
    \begin{subfigure}[b]{0.20\linewidth}
	\centering
        PQ-NET
    \end{subfigure}%
    \hfill%
    \begin{subfigure}[b]{0.20\linewidth}
	\centering
        ATISS
    \end{subfigure}%
    \hfill%
    \begin{subfigure}[b]{0.20\linewidth}
        \centering
        Ours-Parts
    \end{subfigure}%
    \hfill%
    \begin{subfigure}[b]{0.20\linewidth}
        \centering
        Ours
    \end{subfigure}
    \end{subfigure}
    \vskip\baselineskip%
    \vspace{-1.5em}
    \begin{subfigure}[b]{0.20\linewidth}
		\centering
		\includegraphics[width=\linewidth]{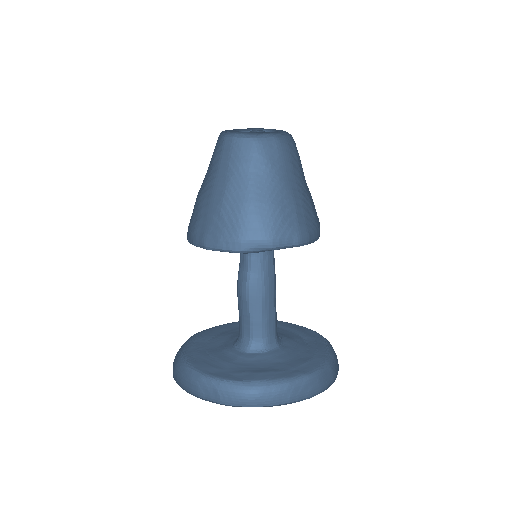}
    \end{subfigure}%
    \hfill%
    \begin{subfigure}[b]{0.20\linewidth}
		\centering
		\includegraphics[width=\linewidth]{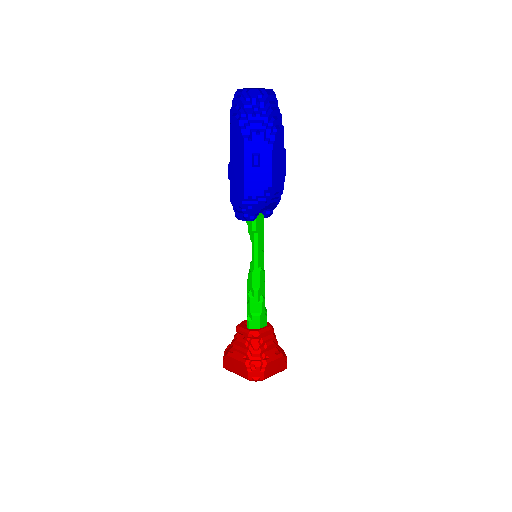}
    \end{subfigure}%
    \hfill%
    \begin{subfigure}[b]{0.20\linewidth}
		\centering
		\includegraphics[width=\linewidth]{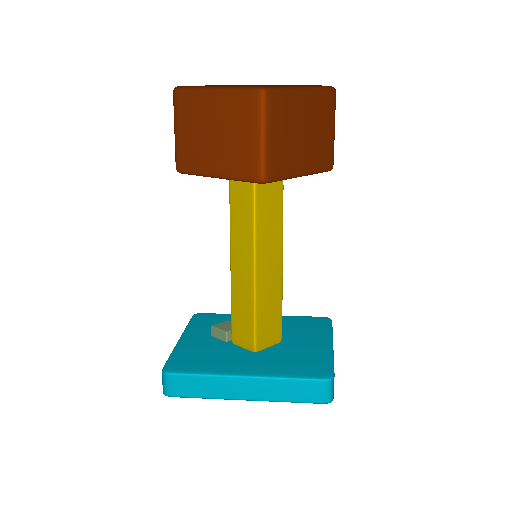}
    \end{subfigure}%
    \hfill%
    \begin{subfigure}[b]{0.20\linewidth}
		\centering
		\includegraphics[width=\linewidth]{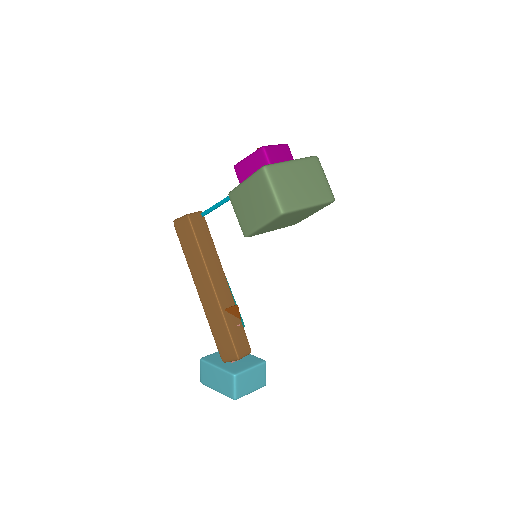}
    \end{subfigure}%
    \hfill%
    \begin{subfigure}[b]{0.20\linewidth}
		\centering
		\includegraphics[width=\linewidth]{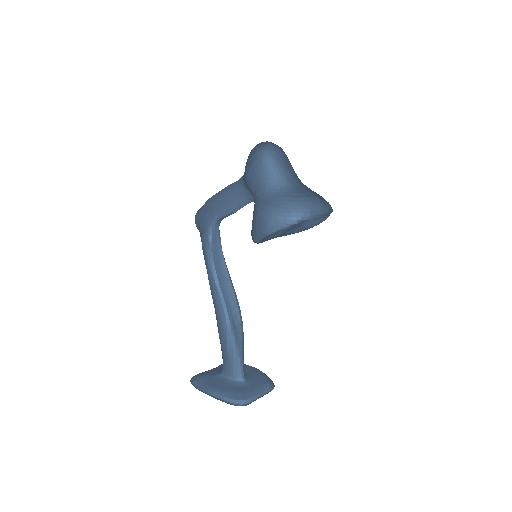}
    \end{subfigure}%
    \vskip\baselineskip%
    \vspace{-1.5em}
    \begin{subfigure}[b]{0.20\linewidth}
		\centering
		\includegraphics[width=\linewidth]{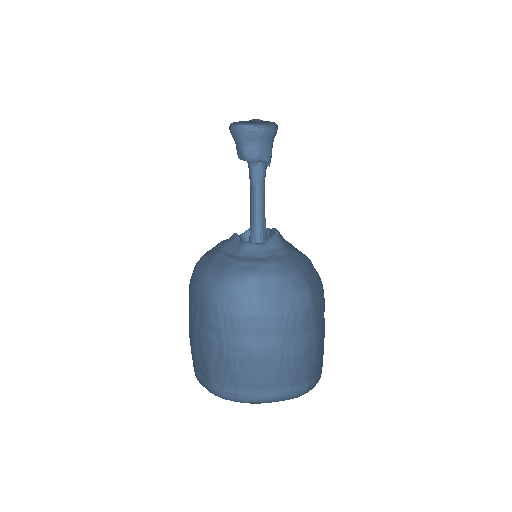}
    \end{subfigure}%
    \hfill%
    \begin{subfigure}[b]{0.20\linewidth}
		\centering
		\includegraphics[width=\linewidth]{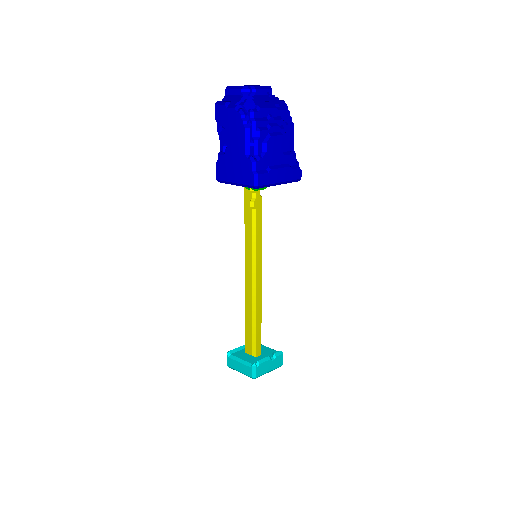}
    \end{subfigure}%
    \hfill%
    \begin{subfigure}[b]{0.20\linewidth}
		\centering
		\includegraphics[width=\linewidth]{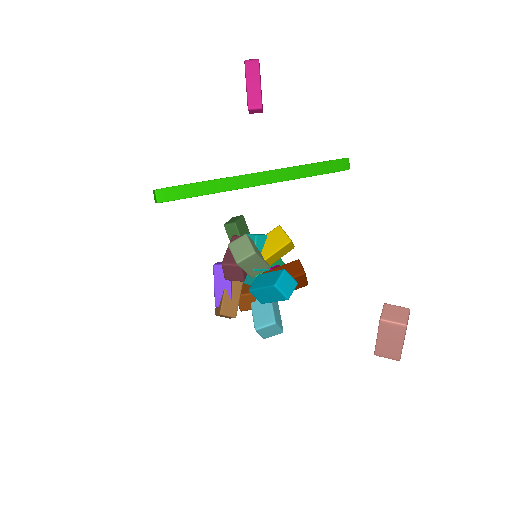}
    \end{subfigure}%
    \hfill%
    \begin{subfigure}[b]{0.20\linewidth}
		\centering
		\includegraphics[width=\linewidth]{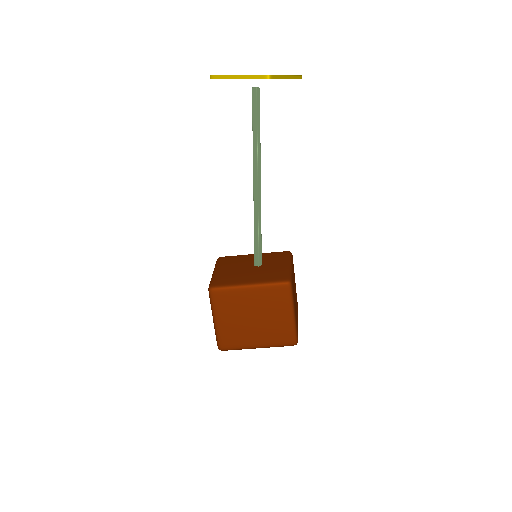}
    \end{subfigure}%
    \hfill%
    \begin{subfigure}[b]{0.20\linewidth}
		\centering
		\includegraphics[width=\linewidth]{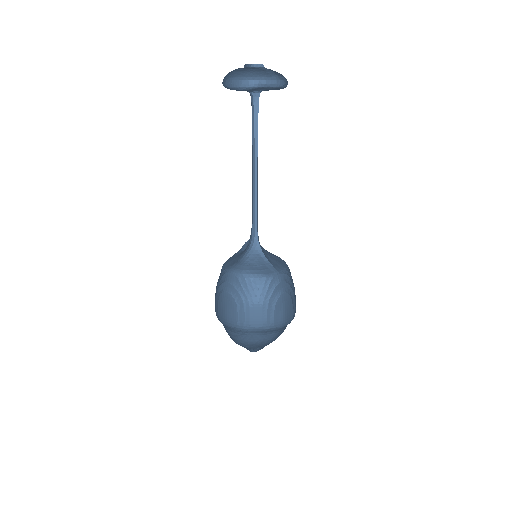}
    \end{subfigure}%
    \vskip\baselineskip%
    \vspace{-1.75em}
    \begin{subfigure}[b]{0.20\linewidth}
		\centering
		\includegraphics[width=\linewidth]{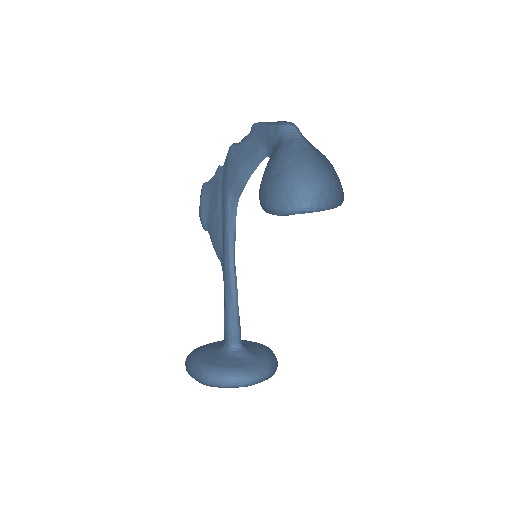}
    \end{subfigure}%
    \hfill%
    \begin{subfigure}[b]{0.20\linewidth}
		\centering
		\includegraphics[width=\linewidth]{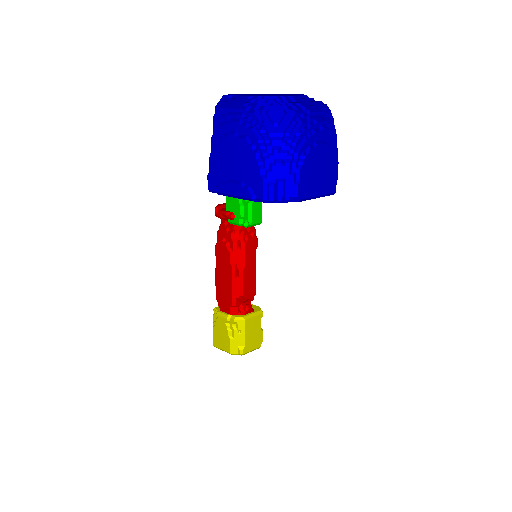}
    \end{subfigure}%
    \hfill%
    \begin{subfigure}[b]{0.20\linewidth}
		\centering
		\includegraphics[width=\linewidth]{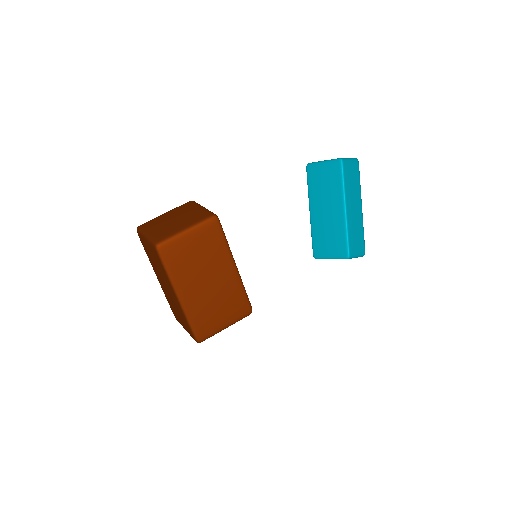}
    \end{subfigure}%
    \hfill%
    \begin{subfigure}[b]{0.20\linewidth}
		\centering
		\includegraphics[width=\linewidth]{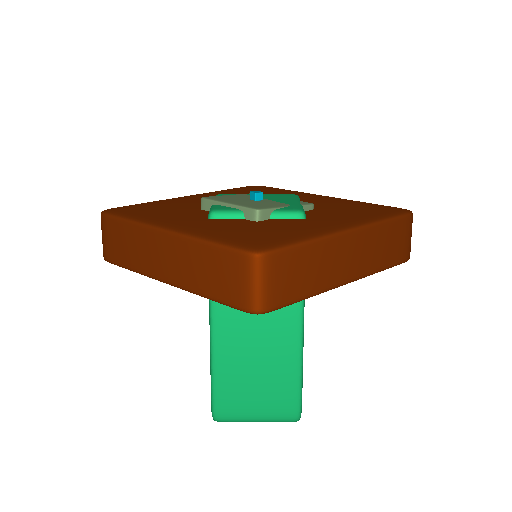}
    \end{subfigure}%
    \hfill%
    \begin{subfigure}[b]{0.20\linewidth}
		\centering
		\includegraphics[width=\linewidth]{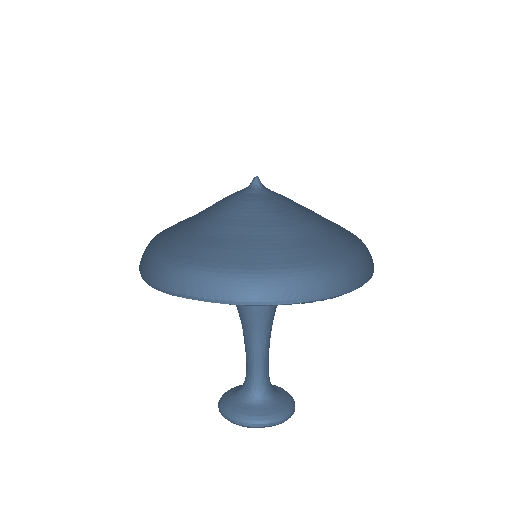}
    \end{subfigure}%
    \vskip\baselineskip%
    \vspace{-1.5em}
    \begin{subfigure}[b]{0.20\linewidth}
		\centering
		\includegraphics[width=\linewidth]{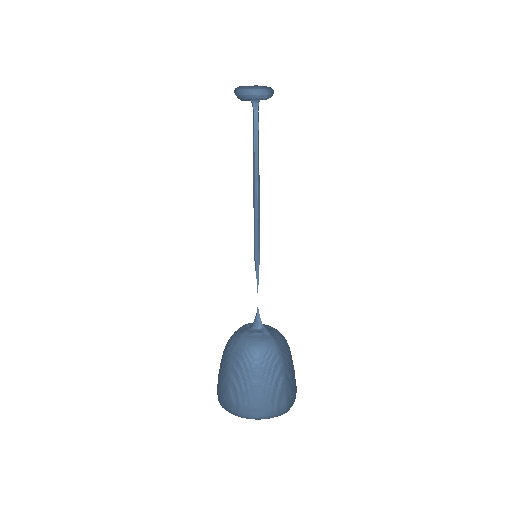}
    \end{subfigure}%
    \hfill%
    \begin{subfigure}[b]{0.20\linewidth}
		\centering
		\includegraphics[width=\linewidth]{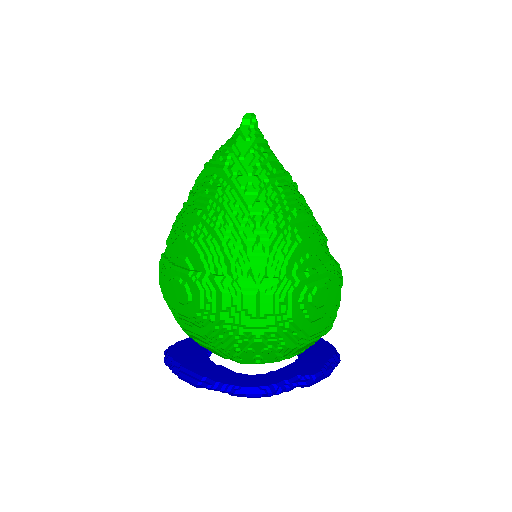}
    \end{subfigure}%
    \hfill%
    \begin{subfigure}[b]{0.20\linewidth}
		\centering
		\includegraphics[width=\linewidth]{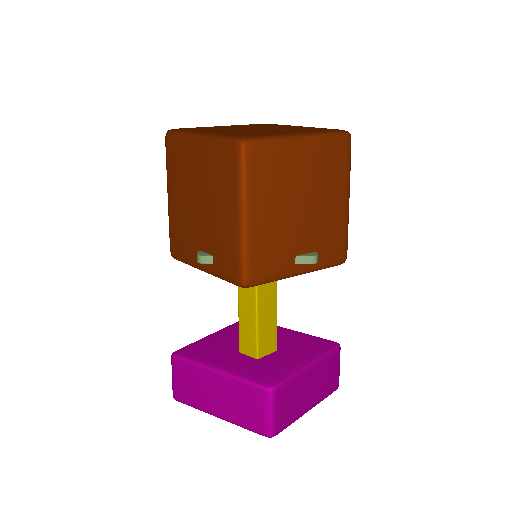}
    \end{subfigure}%
    \hfill%
    \begin{subfigure}[b]{0.20\linewidth}
		\centering
		\includegraphics[width=\linewidth]{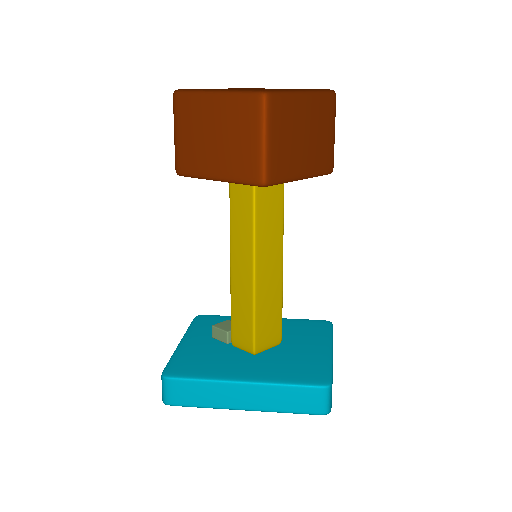}
    \end{subfigure}%
    \hfill%
    \begin{subfigure}[b]{0.20\linewidth}
		\centering
		\includegraphics[width=\linewidth]{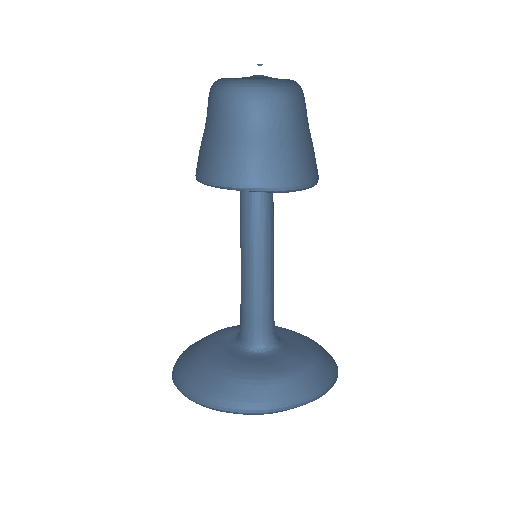}
    \end{subfigure}%
    \vskip\baselineskip%
    \vspace{-1.5em}
    \begin{subfigure}[b]{0.20\linewidth}
		\centering
		\includegraphics[width=\linewidth]{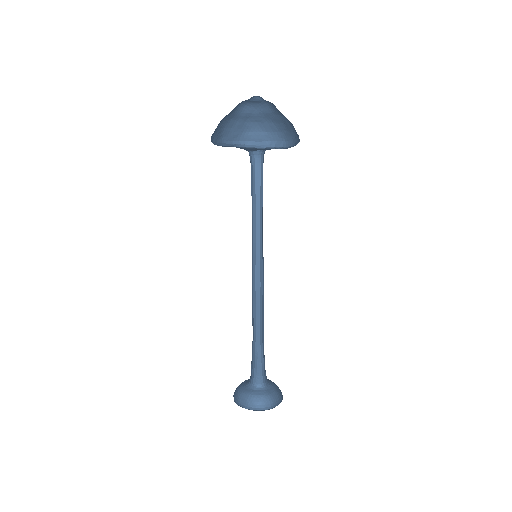}
    \end{subfigure}%
    \hfill%
    \begin{subfigure}[b]{0.20\linewidth}
		\centering
		\includegraphics[width=\linewidth]{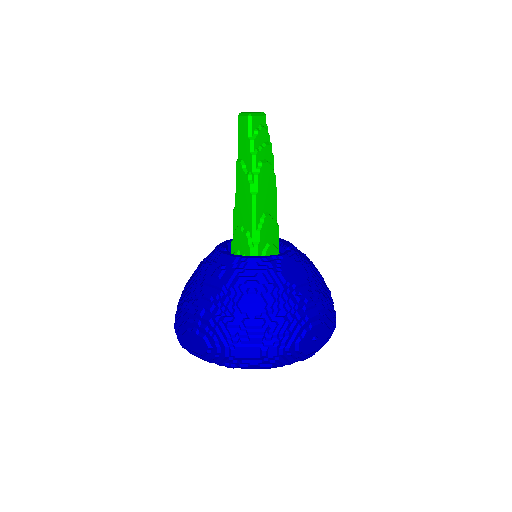}
    \end{subfigure}%
    \hfill%
    \begin{subfigure}[b]{0.20\linewidth}
		\centering
		\includegraphics[width=\linewidth]{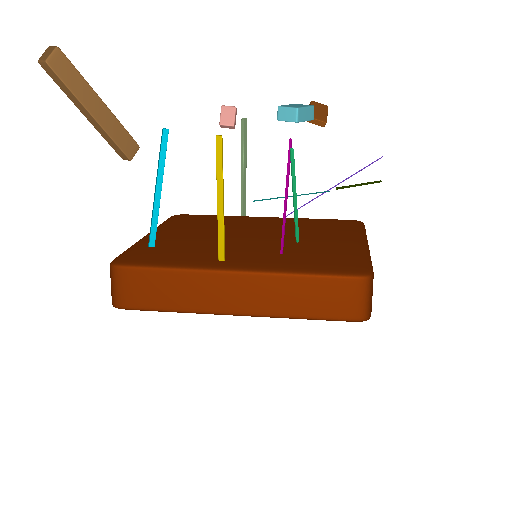}
    \end{subfigure}%
    \hfill%
    \begin{subfigure}[b]{0.20\linewidth}
		\centering
		\includegraphics[width=\linewidth]{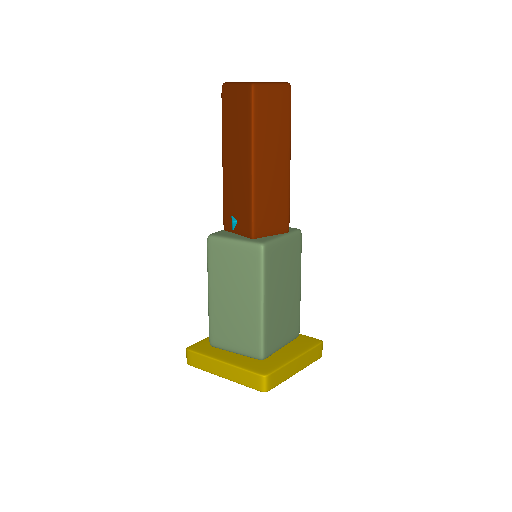}
    \end{subfigure}%
    \hfill%
    \begin{subfigure}[b]{0.20\linewidth}
		\centering
		\includegraphics[width=\linewidth]{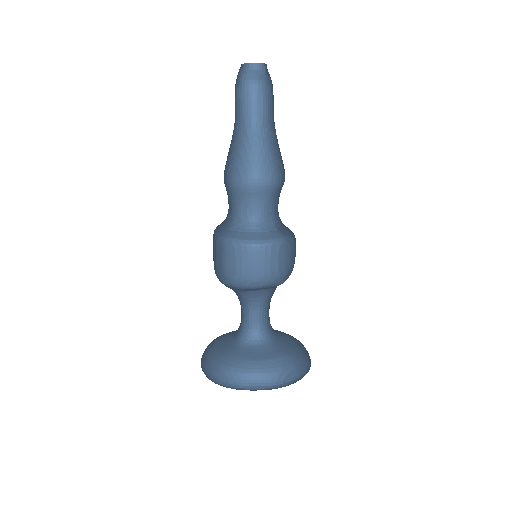}
    \end{subfigure}%
    \vskip\baselineskip%
    \vspace{-2.5em}
    \caption{{\bf Shape Generation Results on Lamps}. We showcase
    randomly generated lamps using our model, ATISS, PQ-NET and IM-NET.}
    \label{fig:shapenet_qualitative_comparison_lamps}
    \vspace{2.5em}
\end{figure}

%% file: fig/shape_generation_qualitative_all.tex
\begin{figure}
    \begin{subfigure}[t]{\linewidth}
    \centering
    \begin{subfigure}[b]{0.20\linewidth}
        \centering
	    IM-NET
    \end{subfigure}%
    \hfill%
    \begin{subfigure}[b]{0.20\linewidth}
	\centering
        PQ-NET
    \end{subfigure}%
    \hfill%
    \begin{subfigure}[b]{0.20\linewidth}
	\centering
        ATISS
    \end{subfigure}%
    \hfill%
    \begin{subfigure}[b]{0.20\linewidth}
        \centering
        Ours-Parts
    \end{subfigure}%
    \hfill%
    \begin{subfigure}[b]{0.20\linewidth}
        \centering
        Ours
    \end{subfigure}
    \end{subfigure}
    \vspace{-1.5em}
    \begin{subfigure}[b]{0.20\linewidth}
		\centering
		\includegraphics[width=\linewidth]{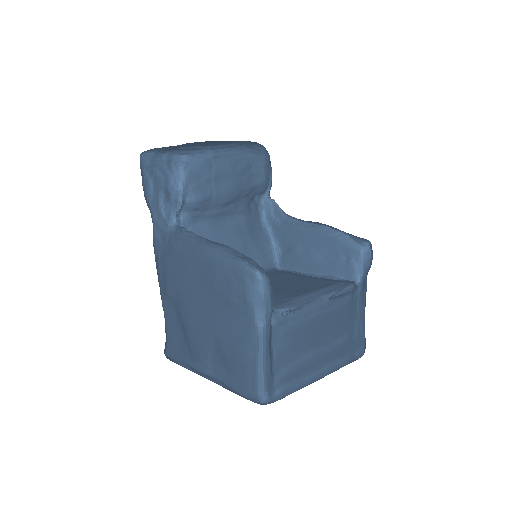}
    \end{subfigure}%
    \hfill%
    \begin{subfigure}[b]{0.20\linewidth}
		\centering
		\includegraphics[width=\linewidth]{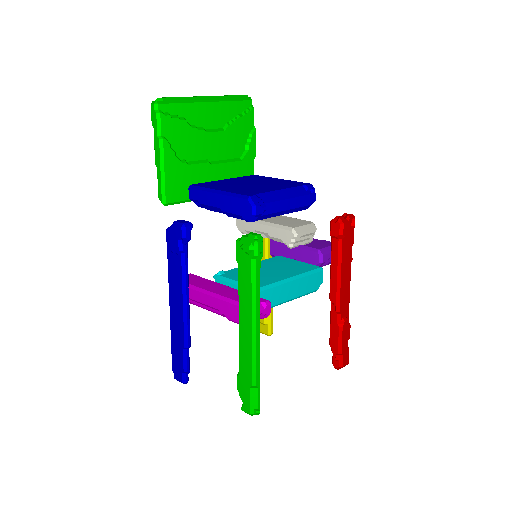}
    \end{subfigure}%
    \hfill%
    \begin{subfigure}[b]{0.20\linewidth}
		\centering
		\includegraphics[width=\linewidth]{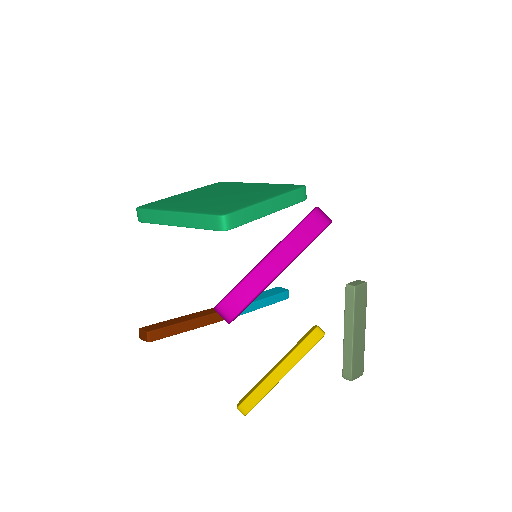}
    \end{subfigure}%
    \hfill%
    \begin{subfigure}[b]{0.20\linewidth}
		\centering
		\includegraphics[width=\linewidth]{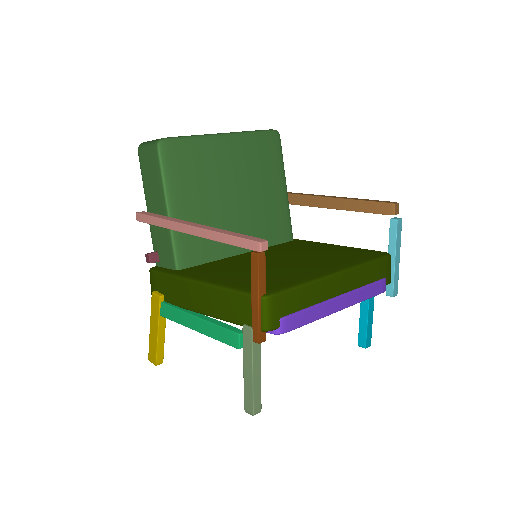}
    \end{subfigure}%
    \hfill%
    \begin{subfigure}[b]{0.20\linewidth}
		\centering
		\includegraphics[width=\linewidth]{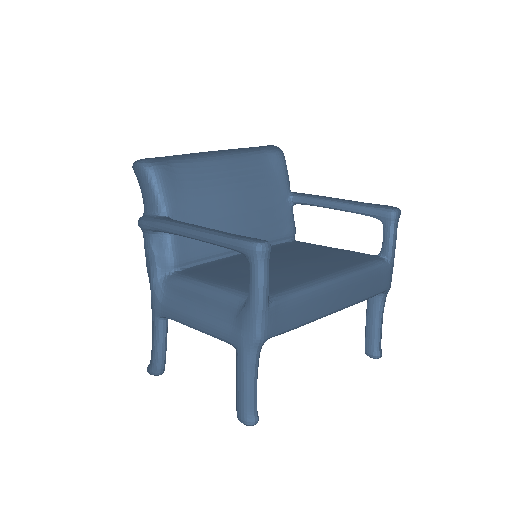}
    \end{subfigure}%
    \vspace{-0.75em}
    \vskip\baselineskip%
    \begin{subfigure}[b]{0.20\linewidth}
		\centering
		\includegraphics[width=\linewidth]{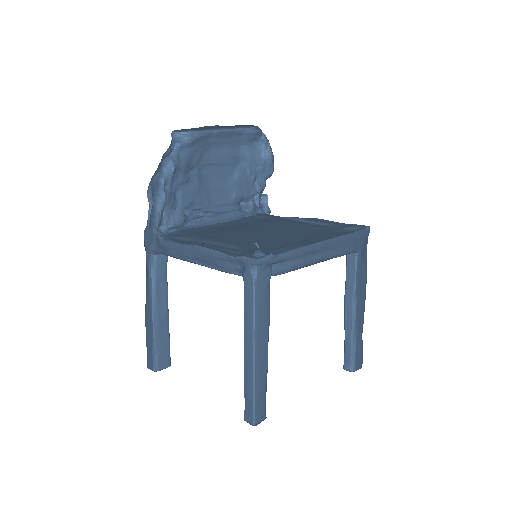}
    \end{subfigure}%
    \hfill%
    \begin{subfigure}[b]{0.20\linewidth}
		\centering
		\includegraphics[width=\linewidth]{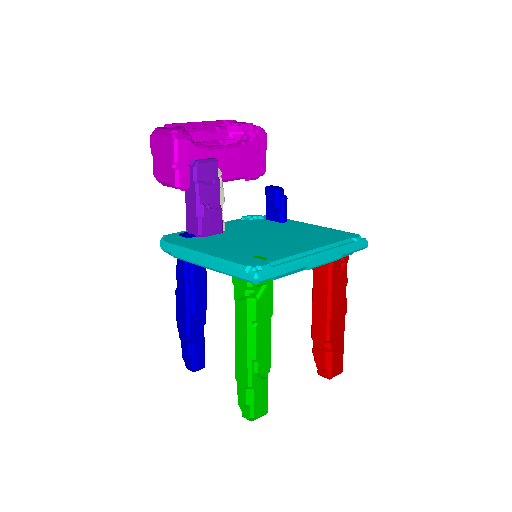}
    \end{subfigure}%
    \hfill%
    \begin{subfigure}[b]{0.20\linewidth}
		\centering
		\includegraphics[width=\linewidth]{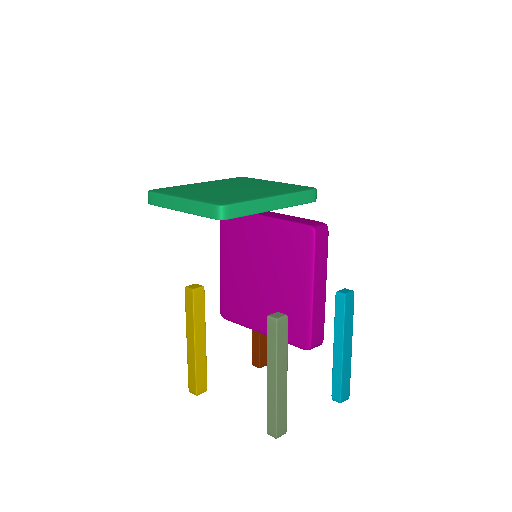}
    \end{subfigure}%
    \hfill%
    \begin{subfigure}[b]{0.20\linewidth}
		\centering
		\includegraphics[width=\linewidth]{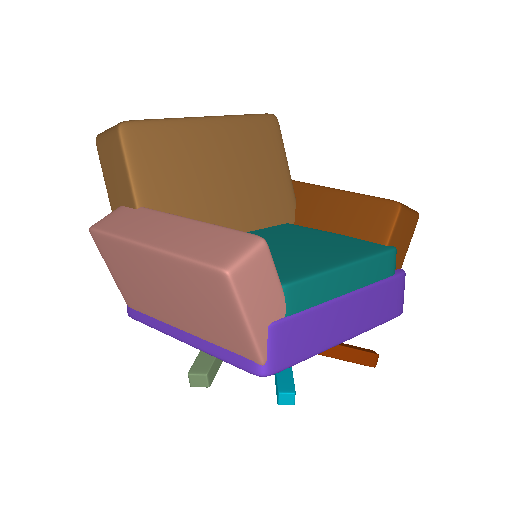}
    \end{subfigure}%
    \hfill%
    \begin{subfigure}[b]{0.20\linewidth}
		\centering
		\includegraphics[width=\linewidth]{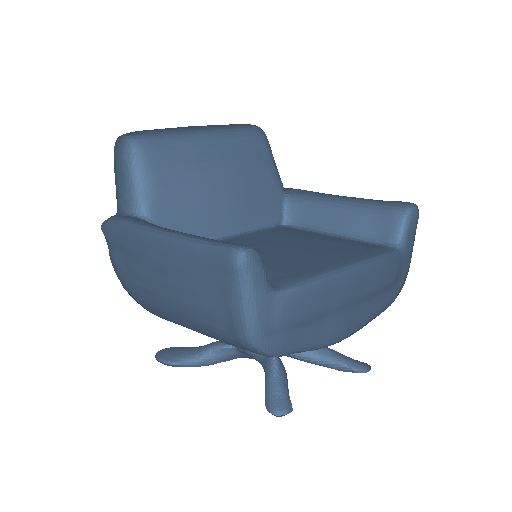}
    \end{subfigure}%
    \vskip\baselineskip%
   \vspace{-1.75em}
    \begin{subfigure}[b]{0.20\linewidth}
		\centering
		\includegraphics[width=\linewidth]{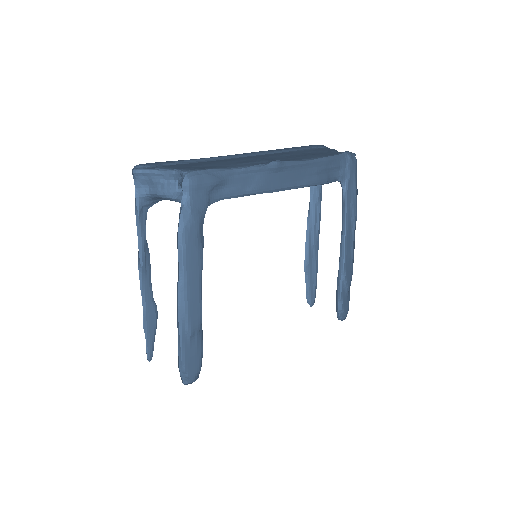}
    \end{subfigure}%
    \hfill%
    \begin{subfigure}[b]{0.20\linewidth}
		\centering
		\includegraphics[width=\linewidth]{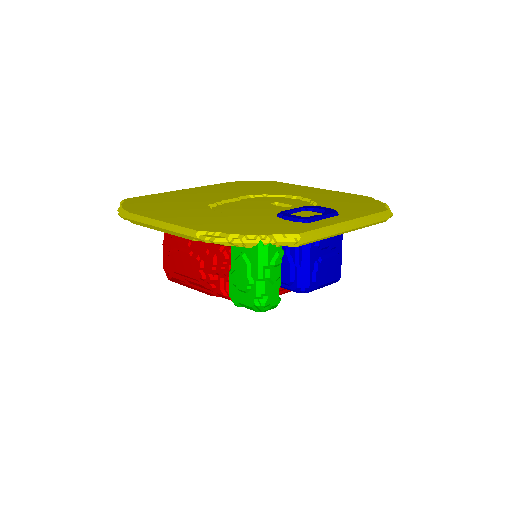}
    \end{subfigure}%
    \hfill%
    \begin{subfigure}[b]{0.20\linewidth}
		\centering
		\includegraphics[width=\linewidth]{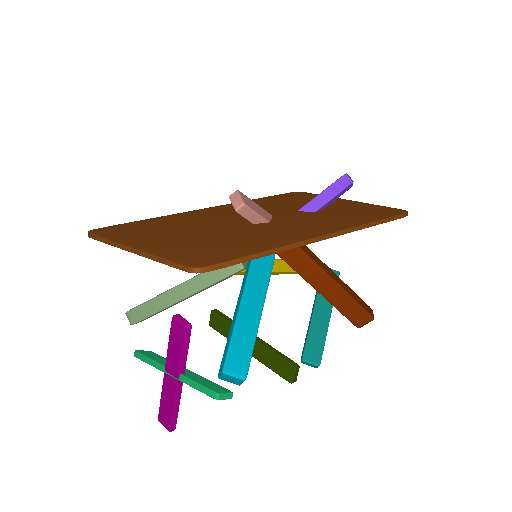}
    \end{subfigure}%
    \hfill%
    \begin{subfigure}[b]{0.20\linewidth}
		\centering
		\includegraphics[width=\linewidth]{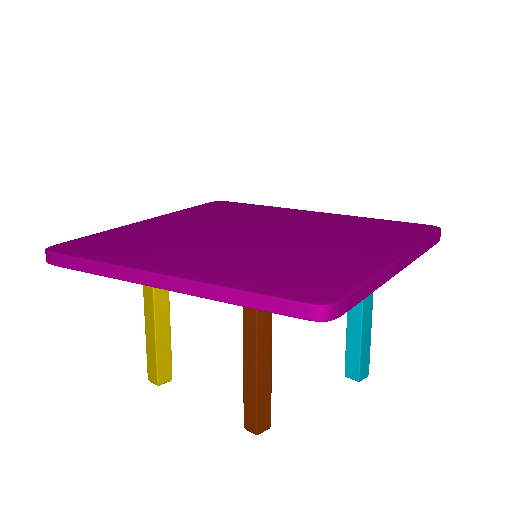}
    \end{subfigure}%
    \hfill%
    \begin{subfigure}[b]{0.20\linewidth}
		\centering
		\includegraphics[width=\linewidth]{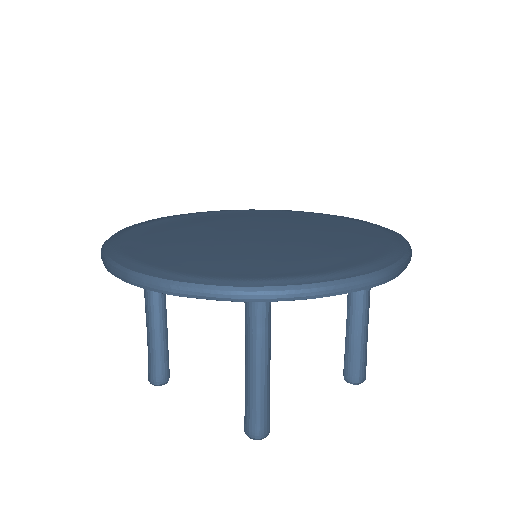}
    \end{subfigure}%
    \vskip\baselineskip%
   \vspace{-1.5em}
    \begin{subfigure}[b]{0.20\linewidth}
		\centering
		\includegraphics[width=\linewidth]{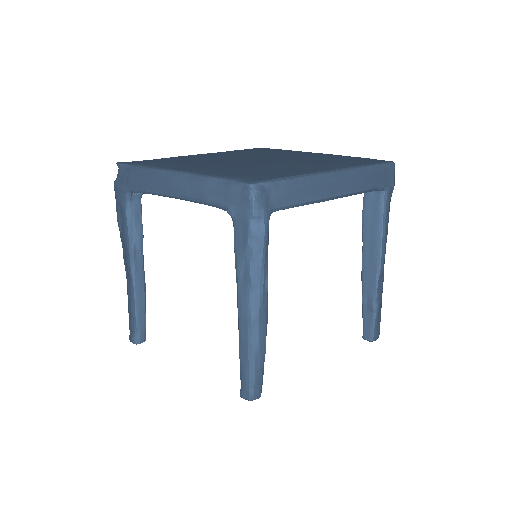}
    \end{subfigure}%
    \hfill%
    \begin{subfigure}[b]{0.20\linewidth}
		\centering
		\includegraphics[width=\linewidth]{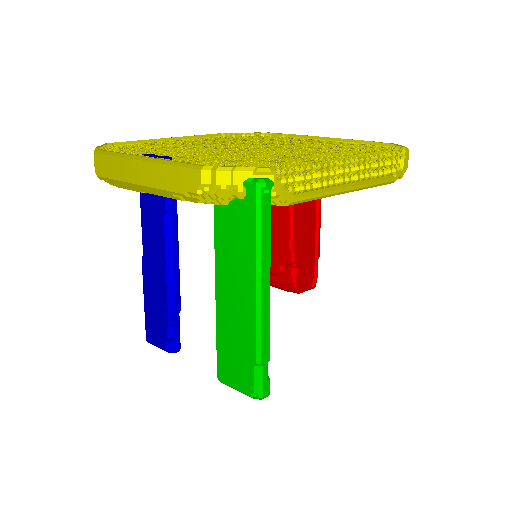}
    \end{subfigure}%
    \hfill%
    \begin{subfigure}[b]{0.20\linewidth}
		\centering
		\includegraphics[width=\linewidth]{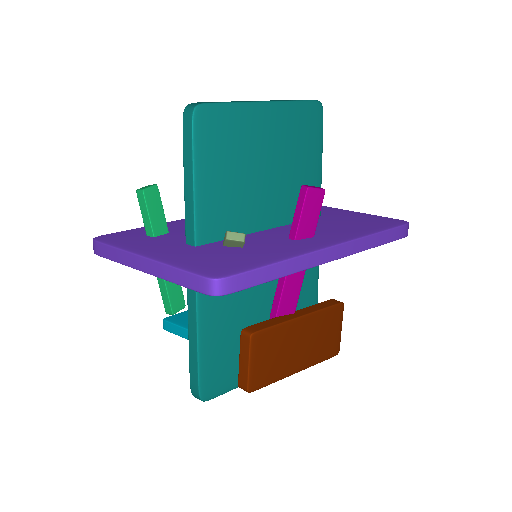}
    \end{subfigure}%
    \hfill%
    \begin{subfigure}[b]{0.20\linewidth}
		\centering
		\includegraphics[width=\linewidth]{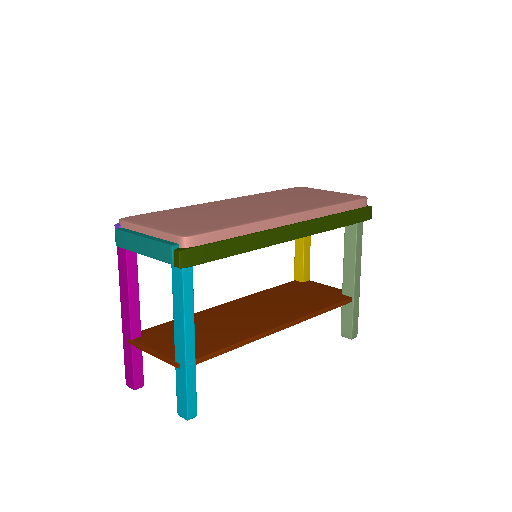}
    \end{subfigure}%
    \hfill%
    \begin{subfigure}[b]{0.20\linewidth}
		\centering
		\includegraphics[width=\linewidth]{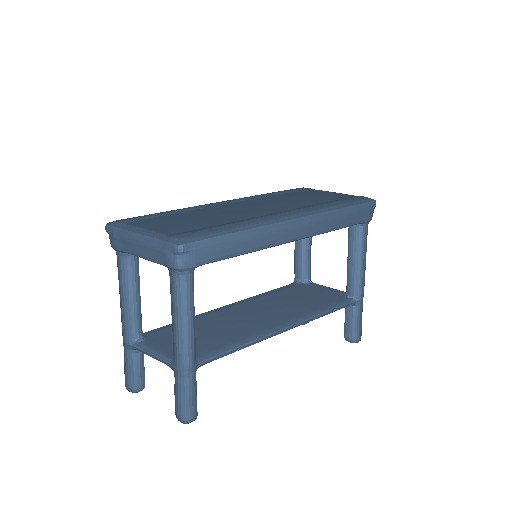}
    \end{subfigure}%
    \vskip\baselineskip%
   \vspace{-1.5em}
    \begin{subfigure}[b]{0.20\linewidth}
		\centering
		\includegraphics[width=\linewidth]{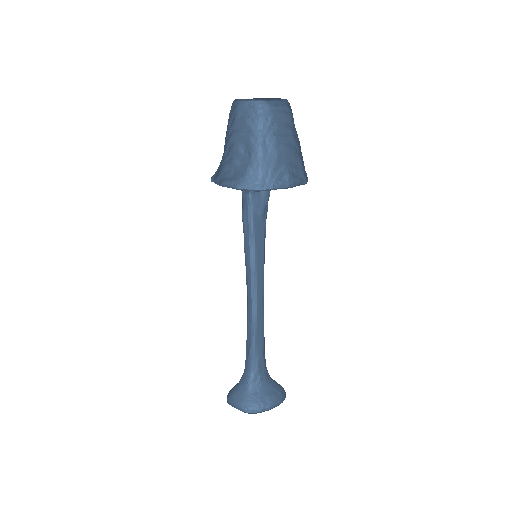}
    \end{subfigure}%
    \hfill%
    \begin{subfigure}[b]{0.20\linewidth}
		\centering
		\includegraphics[width=\linewidth]{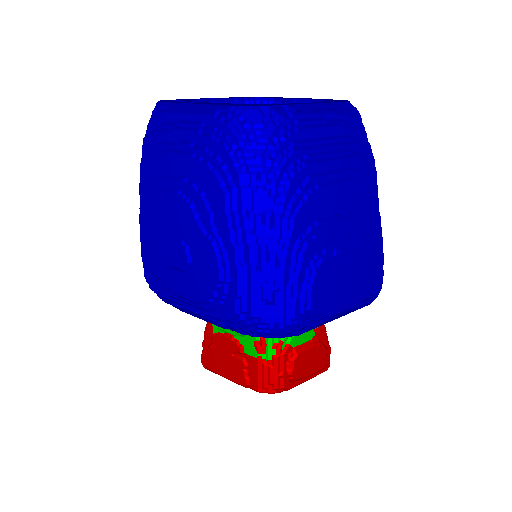}
    \end{subfigure}%
    \hfill%
    \begin{subfigure}[b]{0.20\linewidth}
		\centering
		\includegraphics[width=\linewidth]{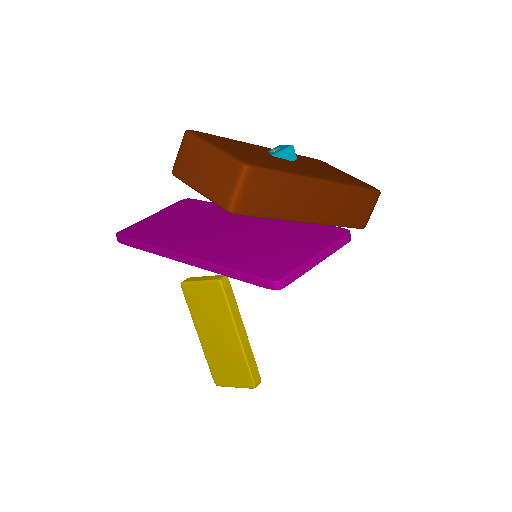}
    \end{subfigure}%
    \hfill%
    \begin{subfigure}[b]{0.20\linewidth}
		\centering
		\includegraphics[width=\linewidth]{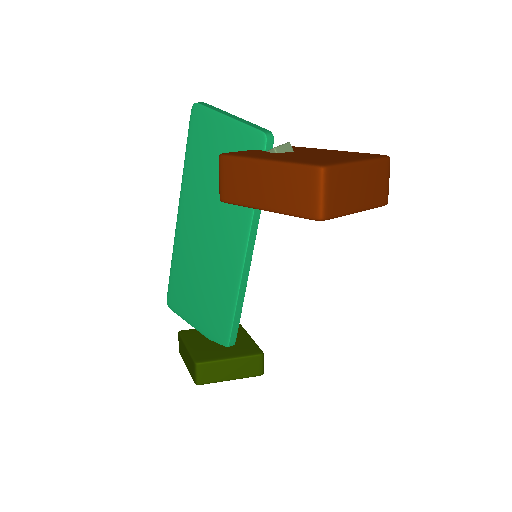}
    \end{subfigure}%
    \hfill%
    \begin{subfigure}[b]{0.20\linewidth}
		\centering
		\includegraphics[width=\linewidth]{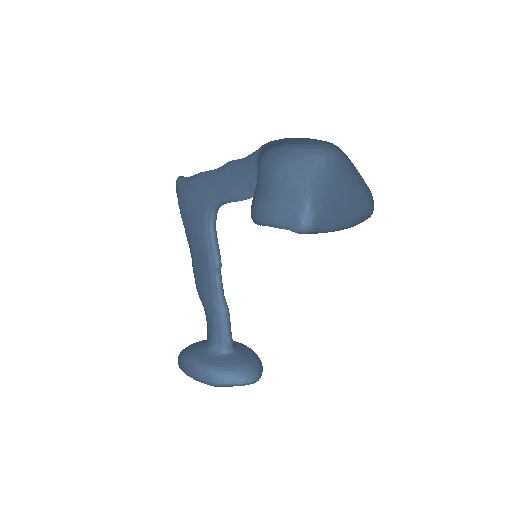}
    \end{subfigure}%
    \vskip\baselineskip%
    \vspace{-1.5em}
    \begin{subfigure}[b]{0.20\linewidth}
		\centering
		\includegraphics[width=\linewidth]{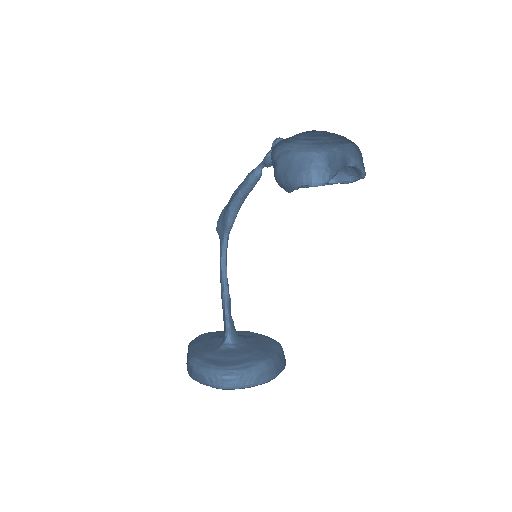}
    \end{subfigure}%
    \hfill%
    \begin{subfigure}[b]{0.20\linewidth}
		\centering
		\includegraphics[width=\linewidth]{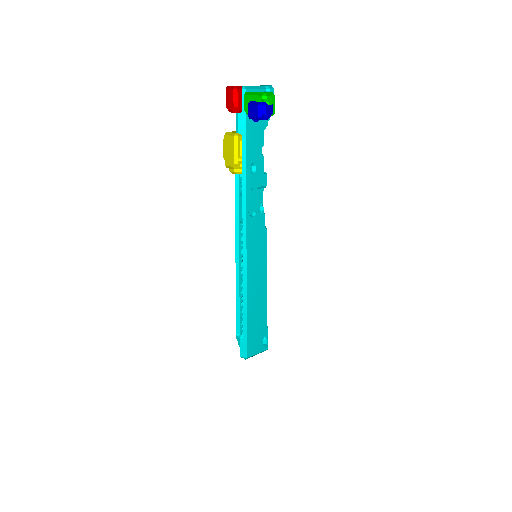}
    \end{subfigure}%
    \hfill%
    \begin{subfigure}[b]{0.20\linewidth}
		\centering
		\includegraphics[width=\linewidth]{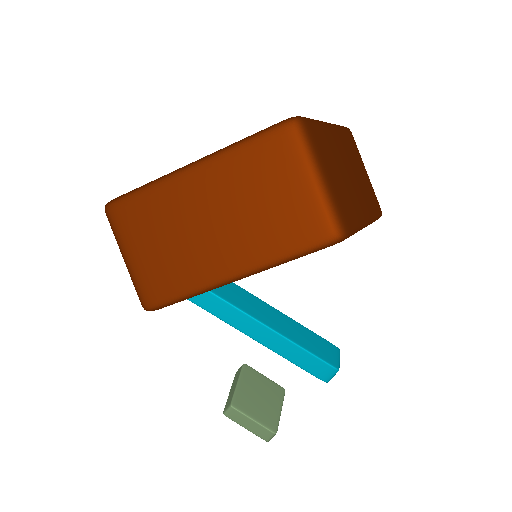}
    \end{subfigure}%
    \hfill%
    \begin{subfigure}[b]{0.20\linewidth}
		\centering
		\includegraphics[width=\linewidth]{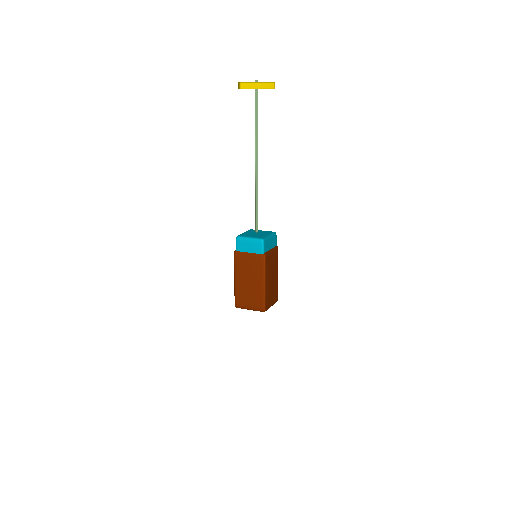}
    \end{subfigure}%
    \hfill%
    \begin{subfigure}[b]{0.20\linewidth}
		\centering
		\includegraphics[width=\linewidth]{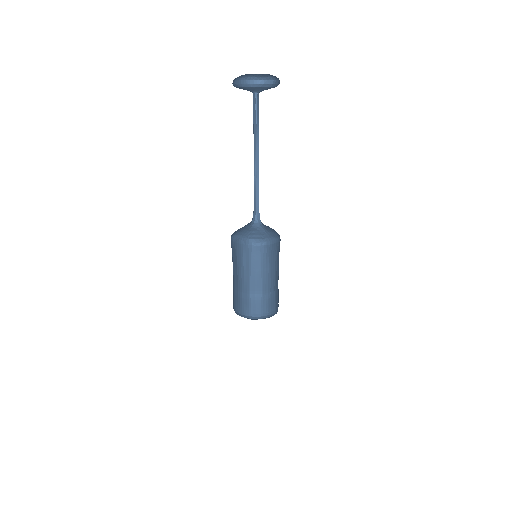}
    \end{subfigure}%
    \vskip\baselineskip%
    \vspace{-2.5em}
    \caption{{\bf Shape Generation Results}. We show randomly
    generated chairs, tables  and lamps using PASTA and our baselines, trained
    jointly on multiple objects categories.}
    \label{fig:shapenet_qualitative_comparison_all}
    \vspace{-2em}
\end{figure}

%% file: fig/shape_completion_qualitative_lamps.tex
\begin{figure}
    \begin{subfigure}[t]{\linewidth}
    \centering
    \begin{subfigure}[b]{0.20\linewidth}
		\centering
		Partial Input
    \end{subfigure}%
    \hfill%
    \begin{subfigure}[b]{0.20\linewidth}
        \centering
        PQ-NET
    \end{subfigure}%
    \hfill%
    \begin{subfigure}[b]{0.20\linewidth}
		\centering
       ATISS
    \end{subfigure}%
    \hfill%
    \begin{subfigure}[b]{0.20\linewidth}
        \centering
        Ours-Parts
    \end{subfigure}%
    \hfill%
    \begin{subfigure}[b]{0.20\linewidth}
        \centering
        Ours
    \end{subfigure}
    \end{subfigure}
    \vskip\baselineskip%
    \vspace{-1.0em}
    \begin{subfigure}[b]{0.20\linewidth}
        \centering
        \includegraphics[width=\linewidth]{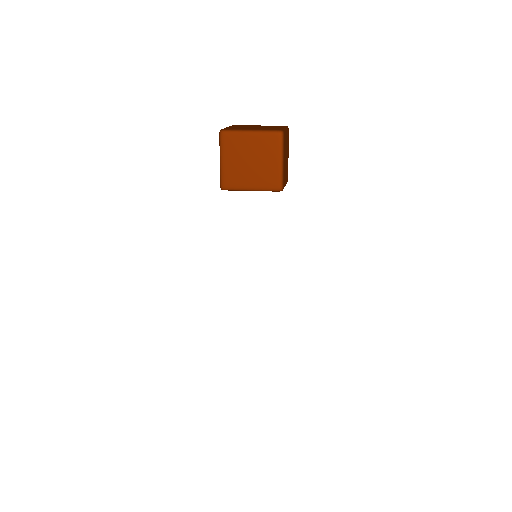}
    \end{subfigure}%
    \hfill%
    \begin{subfigure}[b]{0.20\linewidth}
		\centering
		\includegraphics[width=\linewidth]{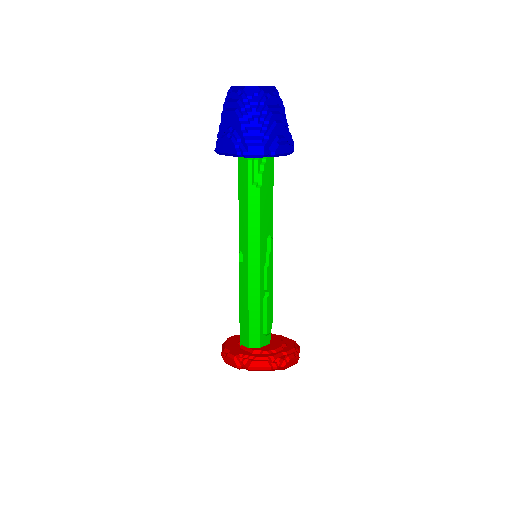}
    \end{subfigure}%
    \hfill%
    \begin{subfigure}[b]{0.20\linewidth}
        \centering
	\includegraphics[width=\linewidth]{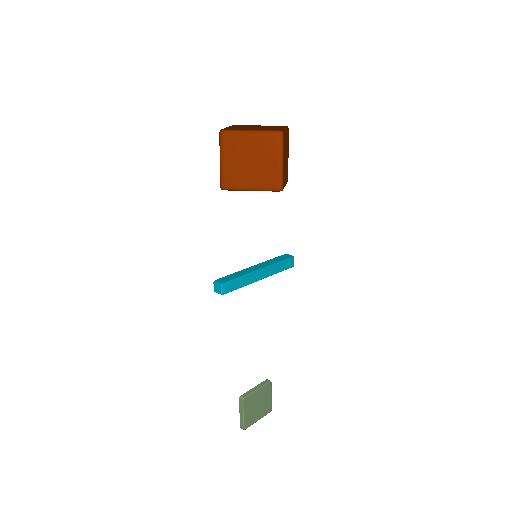}
    \end{subfigure}%
    \hfill%
    \begin{subfigure}[b]{0.20\linewidth}
		\centering
		\includegraphics[width=\linewidth]{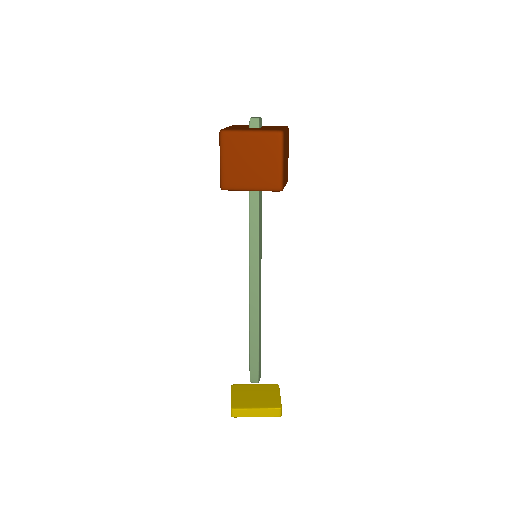}
    \end{subfigure}%
    \hfill%
    \begin{subfigure}[b]{0.20\linewidth}
		\centering
		\includegraphics[width=\linewidth]{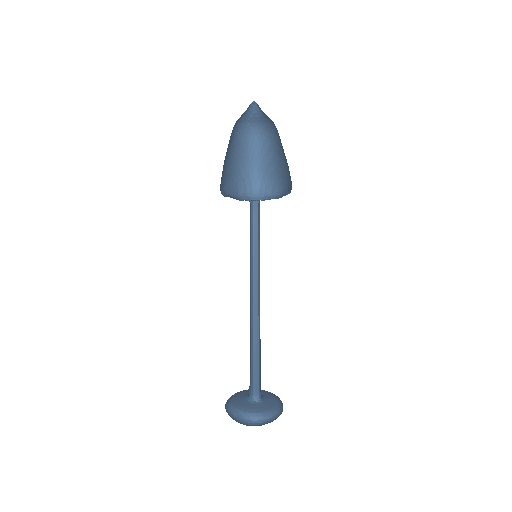}
    \end{subfigure}%
    \vskip\baselineskip%
    \vspace{-1.5em}
    \begin{subfigure}[b]{0.20\linewidth}
	\centering
        \includegraphics[width=\linewidth]{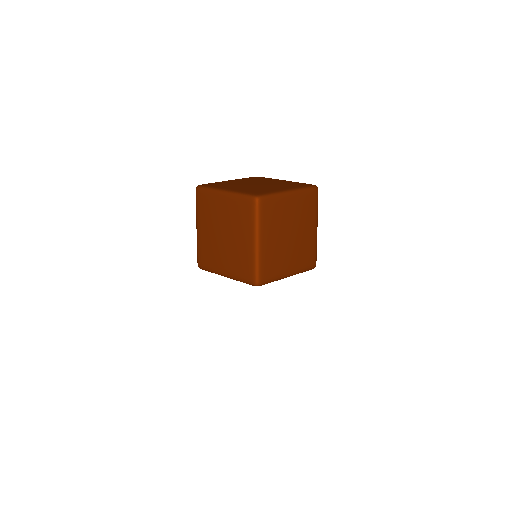}
    \end{subfigure}%
    \hfill%
    \begin{subfigure}[b]{0.20\linewidth}
		\centering
		\includegraphics[width=\linewidth]{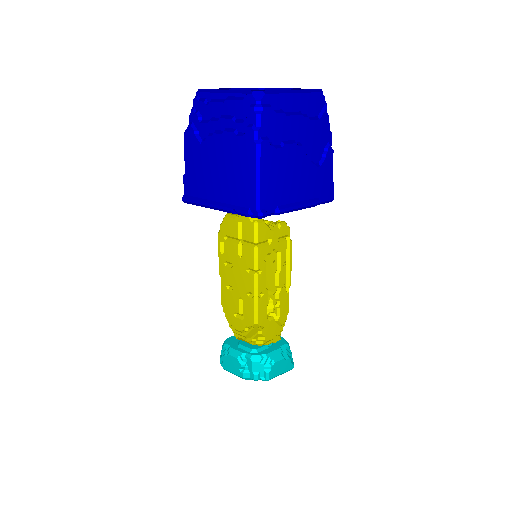}
    \end{subfigure}%
    \hfill%
    \begin{subfigure}[b]{0.20\linewidth}
	\centering
        \includegraphics[width=\linewidth]{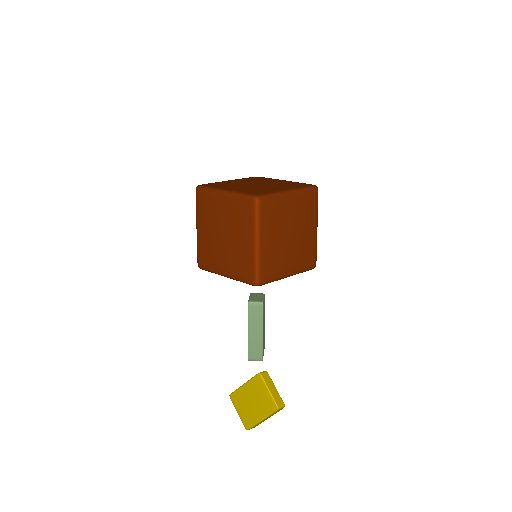}
    \end{subfigure}%
    \hfill%
    \begin{subfigure}[b]{0.20\linewidth}
		\centering
		\includegraphics[width=\linewidth]{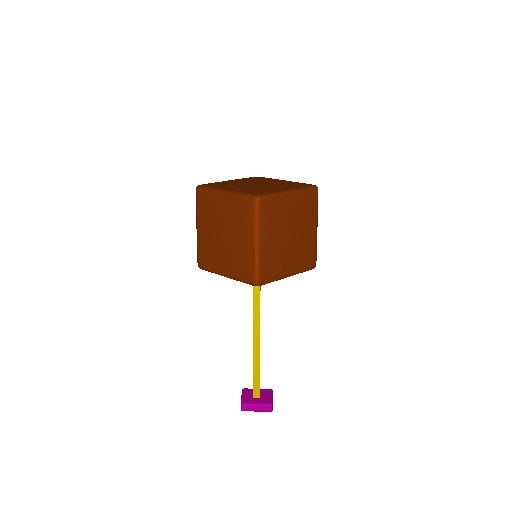}
    \end{subfigure}%
    \hfill%
    \begin{subfigure}[b]{0.20\linewidth}
		\centering
		\includegraphics[width=\linewidth]{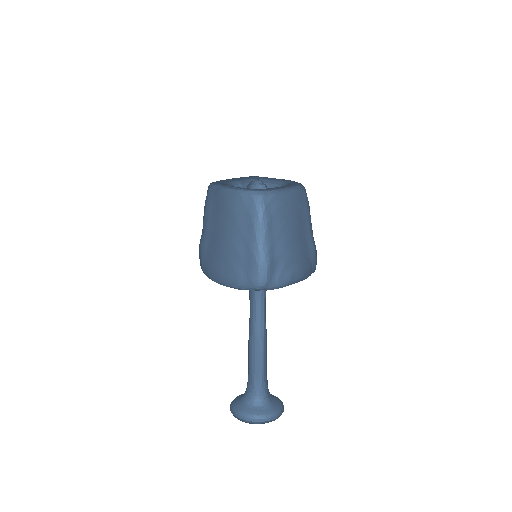}
    \end{subfigure}%
    \vskip\baselineskip%
    \vspace{-1.5em}
    \begin{subfigure}[b]{0.20\linewidth}
	\centering
        \includegraphics[width=\linewidth]{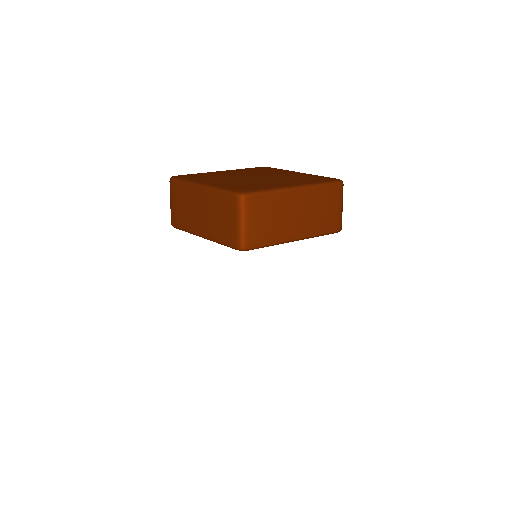}
    \end{subfigure}%
    \hfill%
    \begin{subfigure}[b]{0.20\linewidth}
		\centering
		\includegraphics[width=\linewidth]{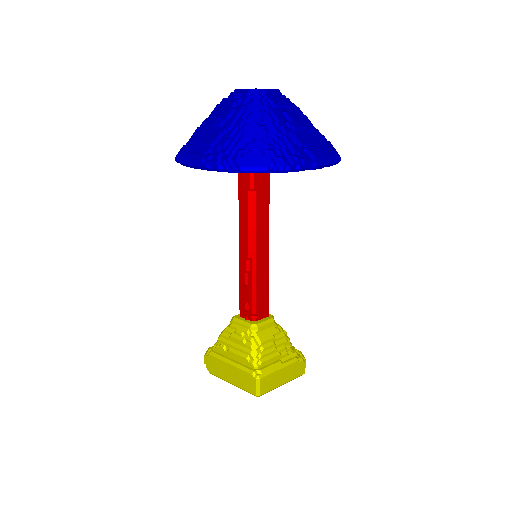}
    \end{subfigure}%
    \hfill%
    \begin{subfigure}[b]{0.20\linewidth}
	\centering
        \includegraphics[width=\linewidth]{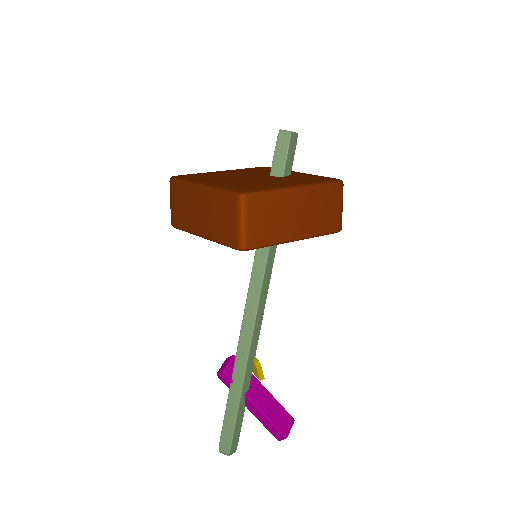}	
    \end{subfigure}%
    \hfill%
    \begin{subfigure}[b]{0.20\linewidth}
		\centering
		\includegraphics[width=\linewidth]{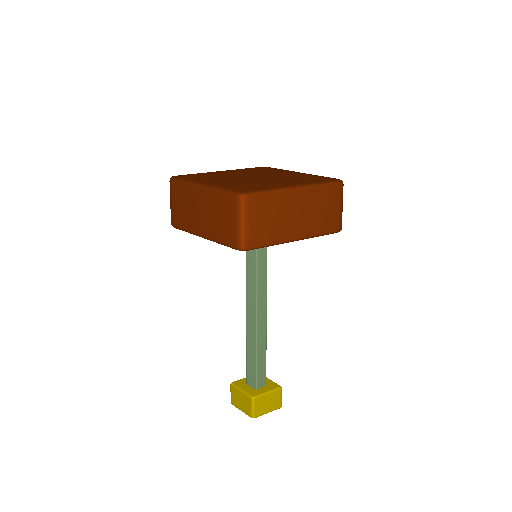}
    \end{subfigure}%
    \hfill%
    \begin{subfigure}[b]{0.20\linewidth}
		\centering
		\includegraphics[width=\linewidth]{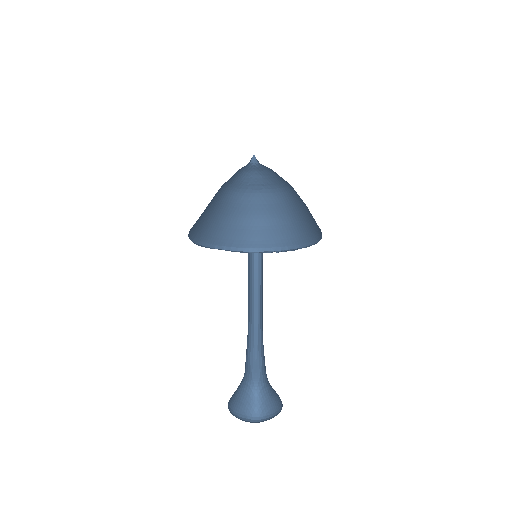}
    \end{subfigure}%
    \vskip\baselineskip%
    \vspace{-1.5em}
    \begin{subfigure}[b]{0.20\linewidth}
	\centering
        \includegraphics[width=\linewidth]{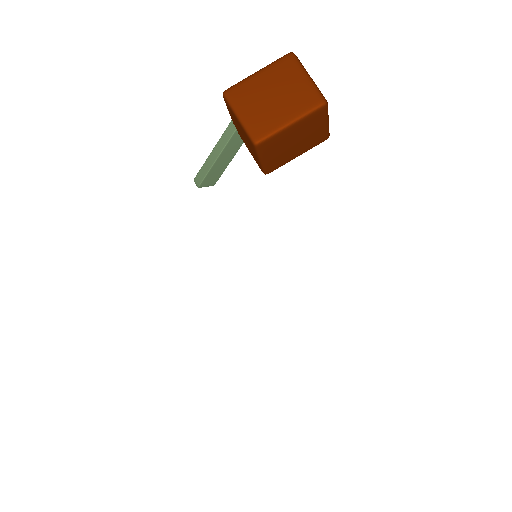}
    \end{subfigure}%
    \hfill%
    \begin{subfigure}[b]{0.20\linewidth}
		\centering
		\includegraphics[width=\linewidth]{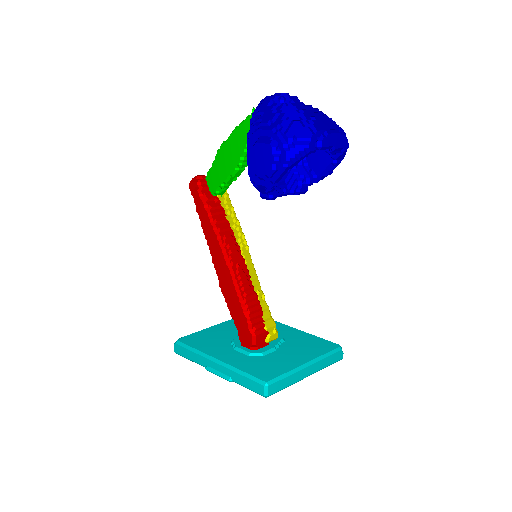}
    \end{subfigure}%
    \hfill%
    \begin{subfigure}[b]{0.20\linewidth}
		\centering
		\includegraphics[width=\linewidth]{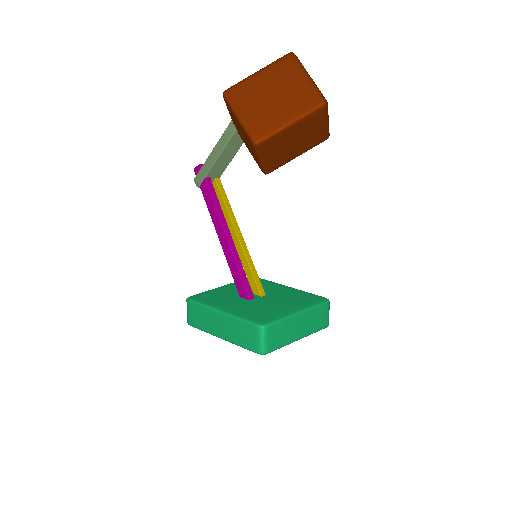}	
    \end{subfigure}%
    \hfill%
    \begin{subfigure}[b]{0.20\linewidth}
		\centering
		\includegraphics[width=\linewidth]{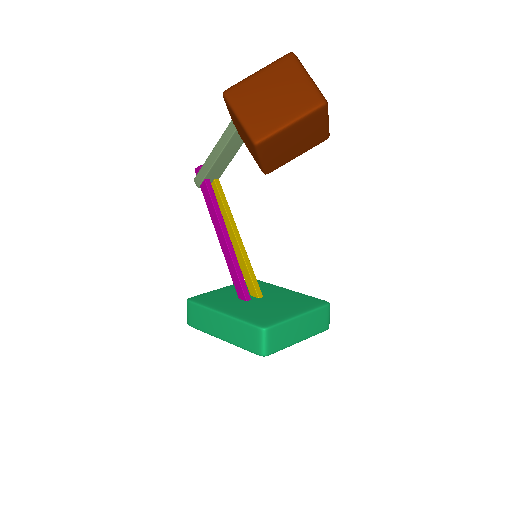}
    \end{subfigure}%
    \hfill%
    \begin{subfigure}[b]{0.20\linewidth}
		\centering
		\includegraphics[width=\linewidth]{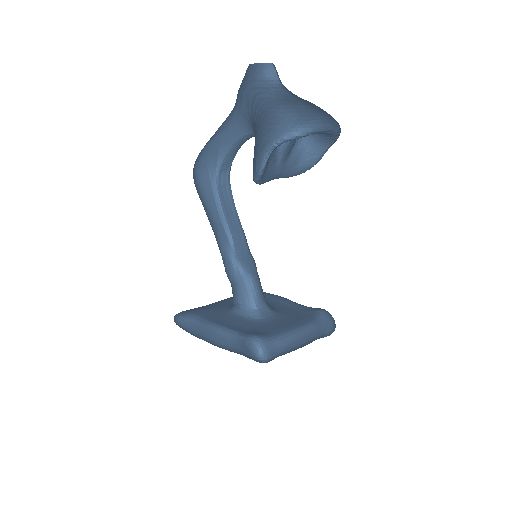}
    \end{subfigure}%
    \vskip\baselineskip%
    \vspace{-1.5em}
    \begin{subfigure}[b]{0.20\linewidth}
		\centering
        \includegraphics[width=\linewidth]{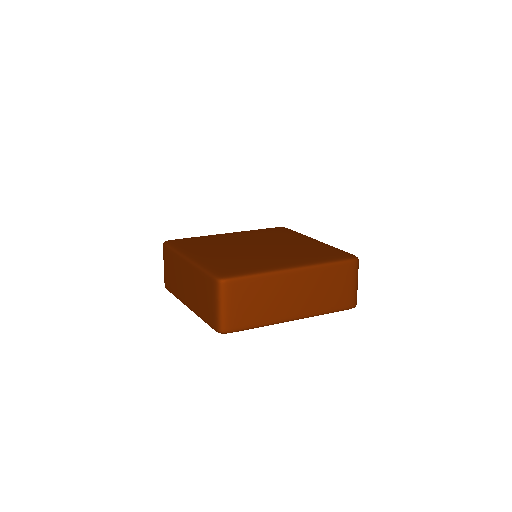}
    \end{subfigure}%
    \hfill%
    \begin{subfigure}[b]{0.20\linewidth}
		\centering
		\includegraphics[width=\linewidth]{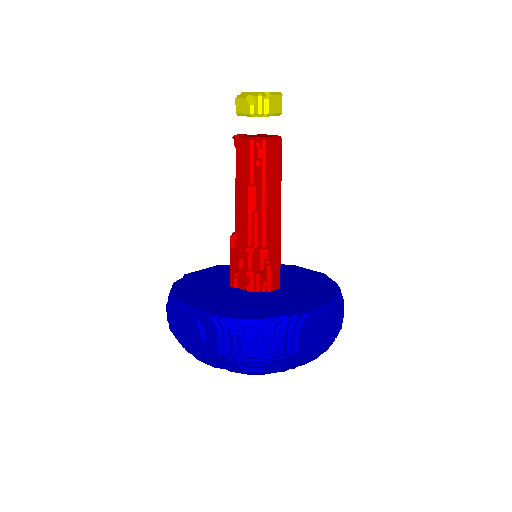}
    \end{subfigure}%
    \hfill%
    \begin{subfigure}[b]{0.20\linewidth}
	\centering
	\includegraphics[width=\linewidth]{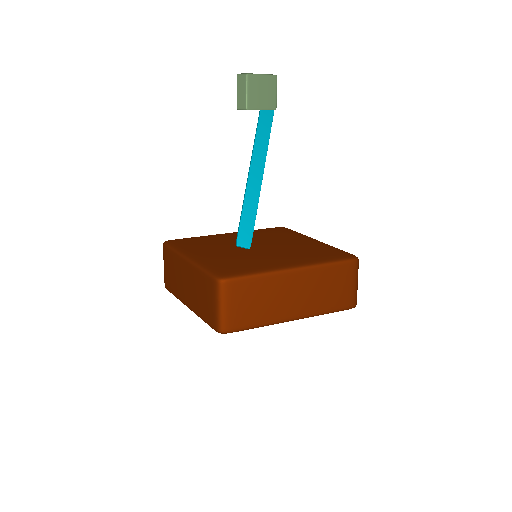}	
    \end{subfigure}%
    \hfill%
    \begin{subfigure}[b]{0.20\linewidth}
		\centering
		\includegraphics[width=\linewidth]{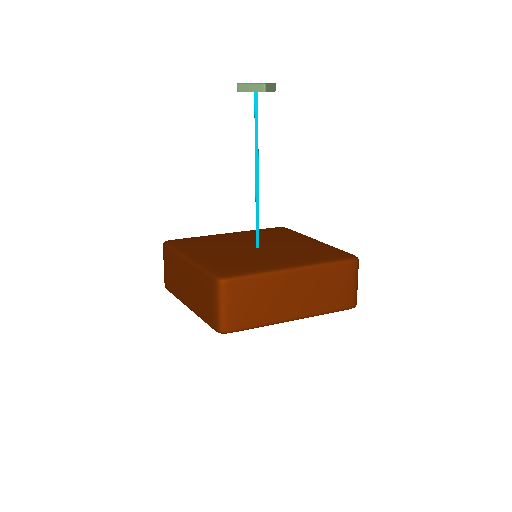}
    \end{subfigure}%
    \hfill%
    \begin{subfigure}[b]{0.20\linewidth}
		\centering
		\includegraphics[width=\linewidth]{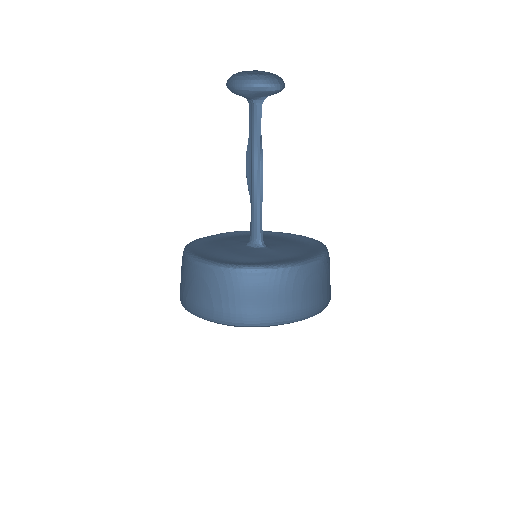}
    \end{subfigure}%
    \vskip\baselineskip%
    \vspace{-2.6em}
    \caption{{\bf Shape Completion Results on Lamps}. Starting from
    partial lamps, we show completions of our model,
    ATISS and PQ-NET.}
    \label{fig:shapenet_qualitative_completion_comparison_lamps}
    \vspace{2.5em}
\end{figure}

%% file: fig/shape_completion_qualitative_tables.tex
\begin{figure}
  \begin{subfigure}[t]{\linewidth}
  \centering
  \begin{subfigure}[b]{0.20\linewidth}
		\centering
		Partial Input
    \end{subfigure}%
    \hfill%
    \begin{subfigure}[b]{0.20\linewidth}
        \centering
        PQ-NET
    \end{subfigure}%
    \hfill%
    \begin{subfigure}[b]{0.20\linewidth}
		\centering
       ATISS
    \end{subfigure}%
    \hfill%
    \begin{subfigure}[b]{0.20\linewidth}
        \centering
        Ours-Parts
    \end{subfigure}%
    \hfill%
    \begin{subfigure}[b]{0.20\linewidth}
        \centering
        Ours
    \end{subfigure}
    \end{subfigure}
    \vspace{-1.5em}
    \begin{subfigure}[b]{0.20\linewidth}
	\centering
        \includegraphics[width=\linewidth]{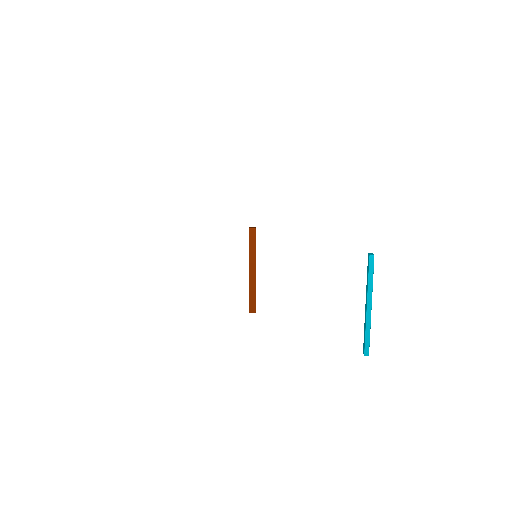}
    \end{subfigure}%
    \hfill%
    \begin{subfigure}[b]{0.20\linewidth}
		\centering
		\includegraphics[width=\linewidth]{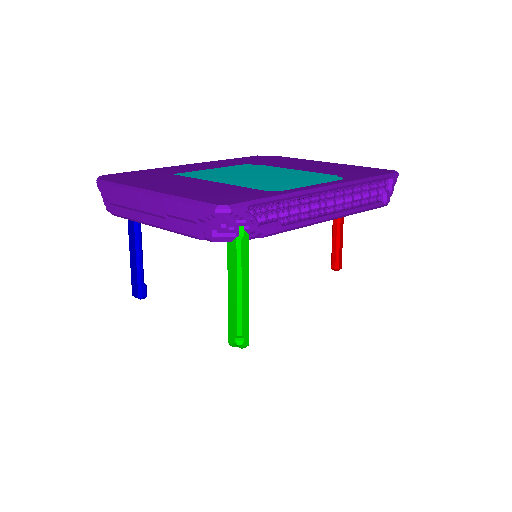}
    \end{subfigure}%
    \hfill%
    \begin{subfigure}[b]{0.20\linewidth}
	\centering
        \includegraphics[width=\linewidth]{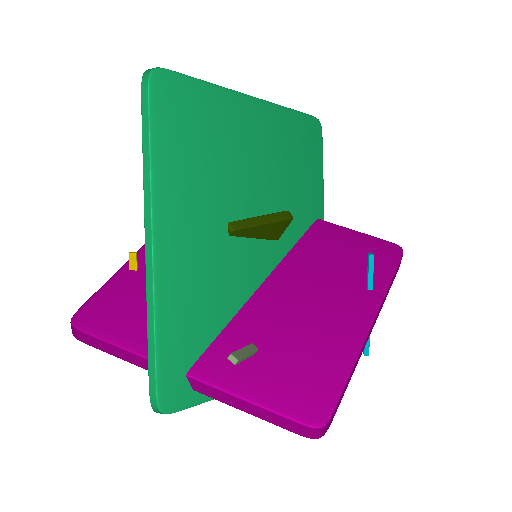}
    \end{subfigure}%
    \hfill%
    \begin{subfigure}[b]{0.20\linewidth}
		\centering
		\includegraphics[width=\linewidth]{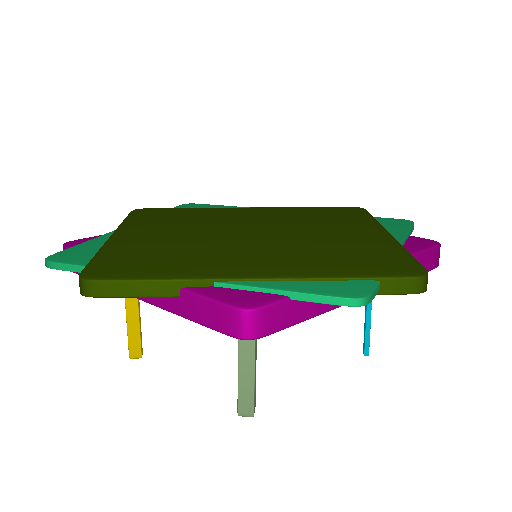}
    \end{subfigure}%
    \hfill%
    \begin{subfigure}[b]{0.20\linewidth}
		\centering
		\includegraphics[width=\linewidth]{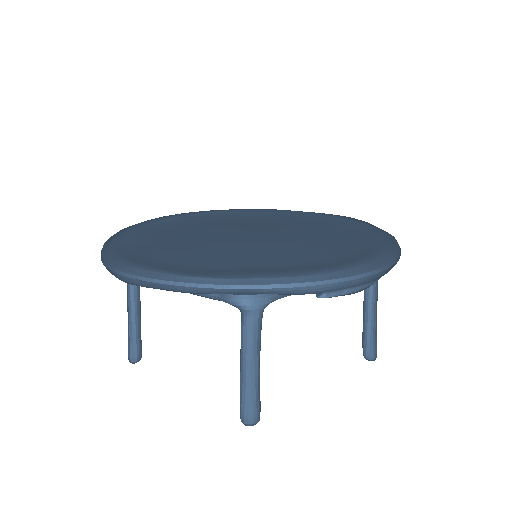}
    \end{subfigure}%
    \vskip\baselineskip%
    \vspace{-0.5em}
    \begin{subfigure}[b]{0.20\linewidth}
		\centering
	\includegraphics[width=\linewidth]{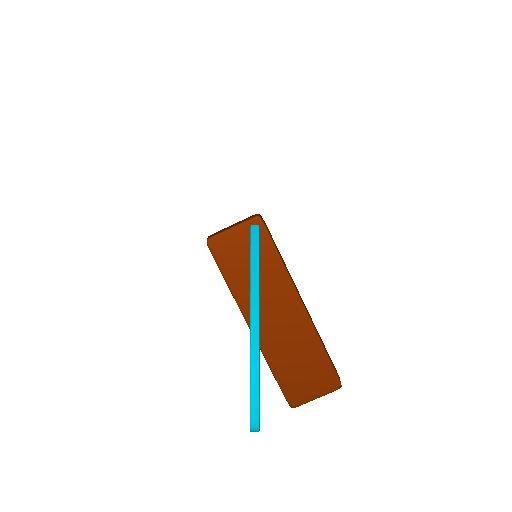}
    \end{subfigure}%
    \hfill%
    \begin{subfigure}[b]{0.20\linewidth}
		\centering
		\includegraphics[width=\linewidth]{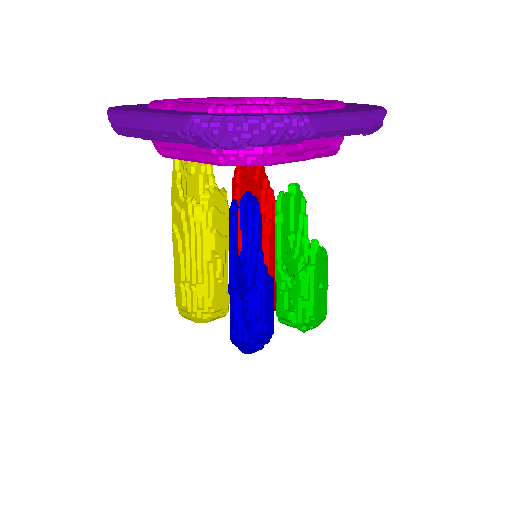}
    \end{subfigure}%
    \hfill%
    \begin{subfigure}[b]{0.20\linewidth}
	\centering
        \includegraphics[width=\linewidth]{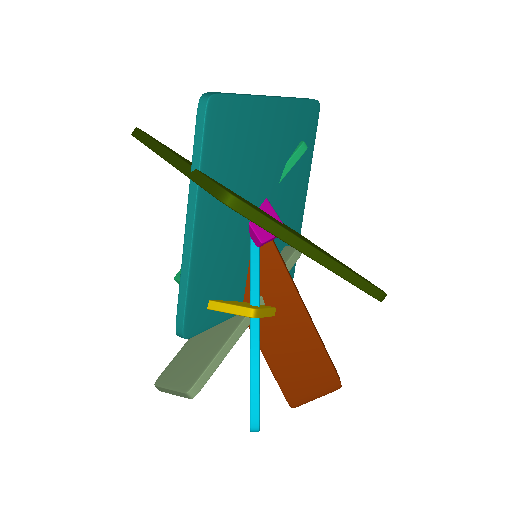}
    \end{subfigure}%
    \hfill%
    \begin{subfigure}[b]{0.20\linewidth}
		\centering
		\includegraphics[width=\linewidth]{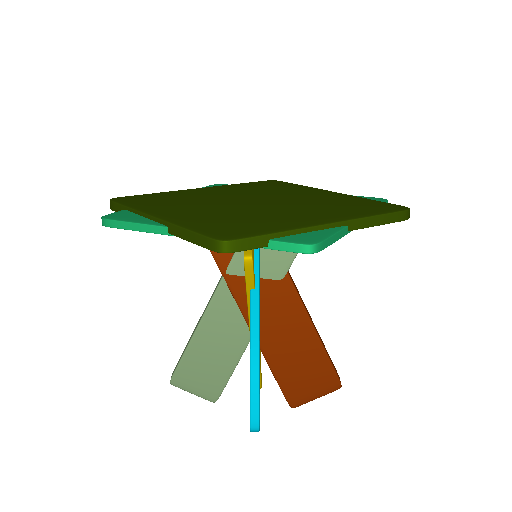}
    \end{subfigure}%
    \hfill%
    \begin{subfigure}[b]{0.20\linewidth}
		\centering
		\includegraphics[width=\linewidth]{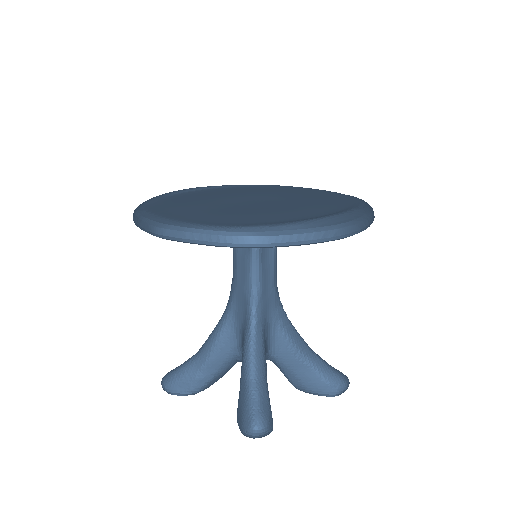}
    \end{subfigure}%
    \vskip\baselineskip%
    \vspace{-1.5em}
    \begin{subfigure}[b]{0.20\linewidth}
	\centering
        \includegraphics[width=\linewidth]{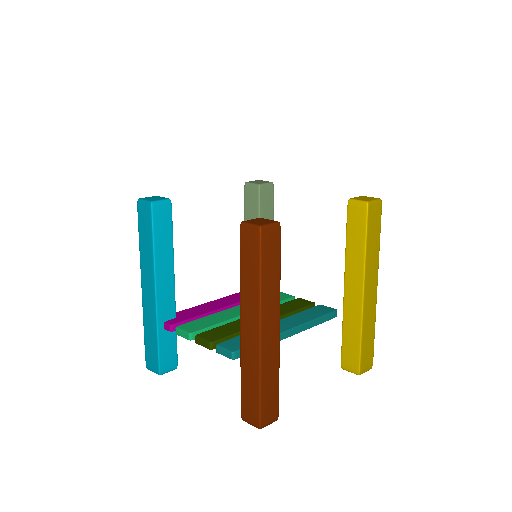}
    \end{subfigure}%
    \hfill%
    \begin{subfigure}[b]{0.20\linewidth}
		\centering
		\includegraphics[width=\linewidth]{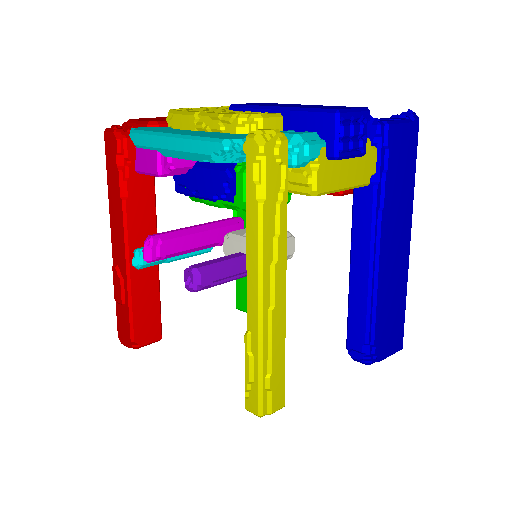}
    \end{subfigure}%
    \hfill%
    \begin{subfigure}[b]{0.20\linewidth}
	\centering
	\includegraphics[width=\linewidth]{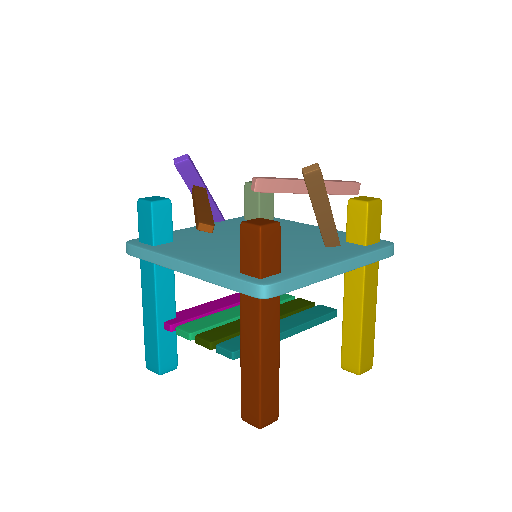}
    \end{subfigure}%
    \hfill%
    \begin{subfigure}[b]{0.20\linewidth}
		\centering
		\includegraphics[width=\linewidth]{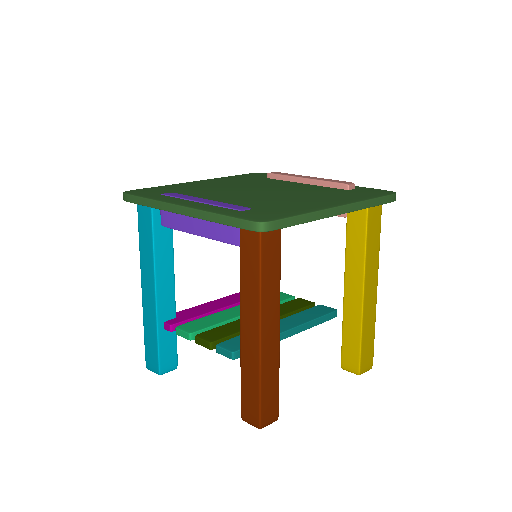}
    \end{subfigure}%
    \hfill%
    \begin{subfigure}[b]{0.20\linewidth}
		\centering
		\includegraphics[width=\linewidth]{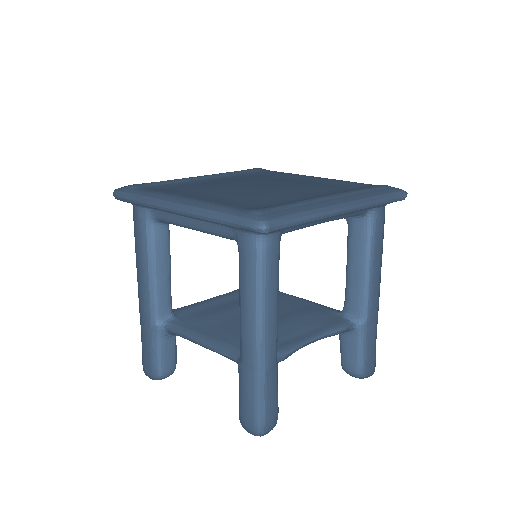}
    \end{subfigure}%
    \vskip\baselineskip%
    \vspace{-1.5em}
    \begin{subfigure}[b]{0.20\linewidth}
	\centering
        \includegraphics[width=\linewidth]{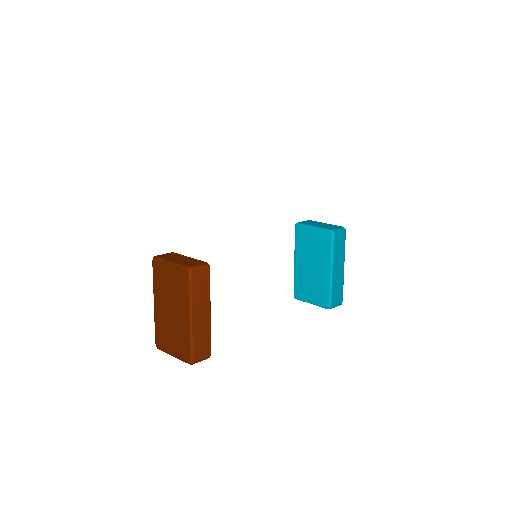}
    \end{subfigure}%
    \hfill%
    \begin{subfigure}[b]{0.20\linewidth}
		\centering
		\includegraphics[width=\linewidth]{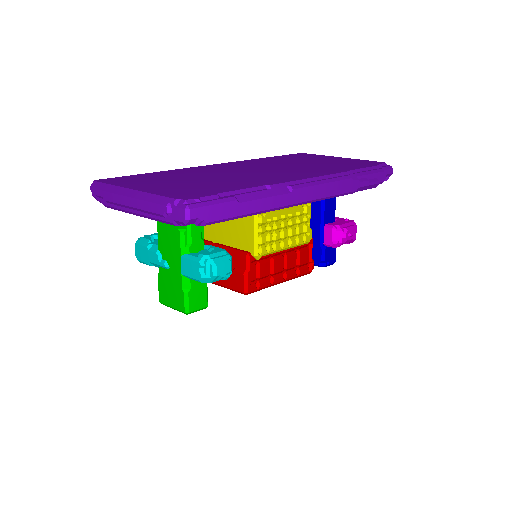}
    \end{subfigure}%
    \hfill%
    \begin{subfigure}[b]{0.20\linewidth}
        \centering
	\includegraphics[width=\linewidth]{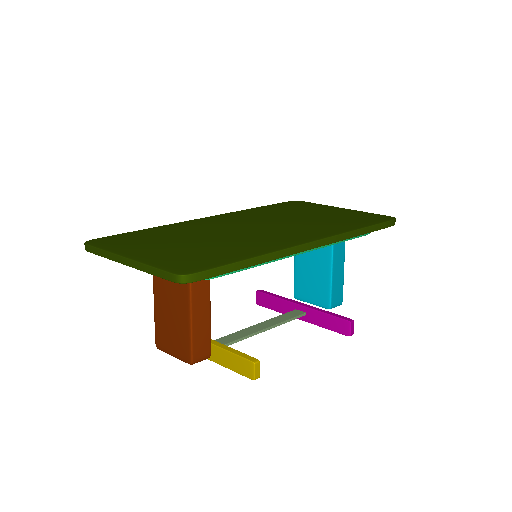}
    \end{subfigure}%
    \hfill%
    \begin{subfigure}[b]{0.20\linewidth}
		\centering
		\includegraphics[width=\linewidth]{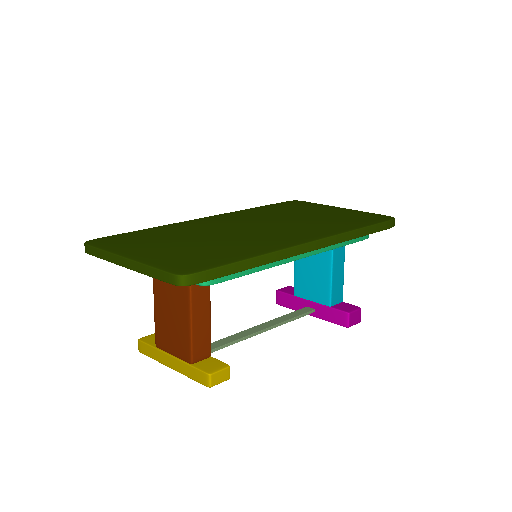}
    \end{subfigure}%
    \hfill%
    \begin{subfigure}[b]{0.20\linewidth}
		\centering
		\includegraphics[width=\linewidth]{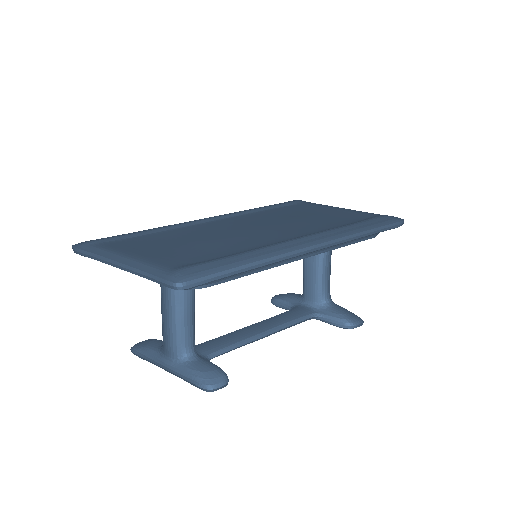}
    \end{subfigure}%
    \vskip\baselineskip%
    \vspace{-1.5em}
    \begin{subfigure}[b]{0.20\linewidth}
	\centering
        \includegraphics[width=\linewidth]{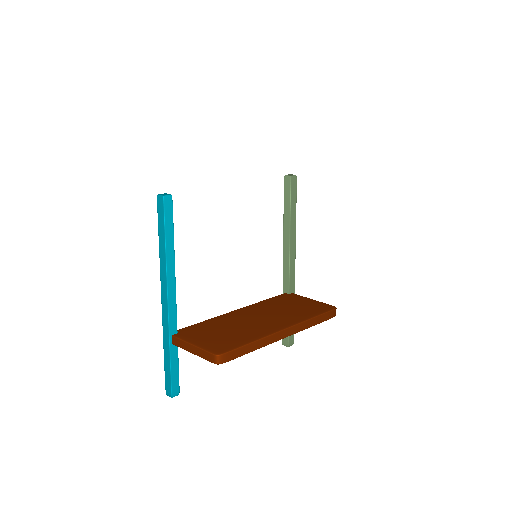}
    \end{subfigure}%
    \hfill%
    \begin{subfigure}[b]{0.20\linewidth}
		\centering
		\includegraphics[width=\linewidth]{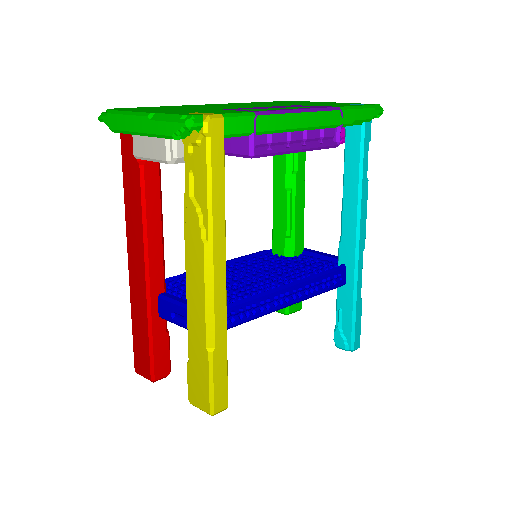}
    \end{subfigure}%
    \hfill%
    \begin{subfigure}[b]{0.20\linewidth}
	\centering
        \includegraphics[width=\linewidth]{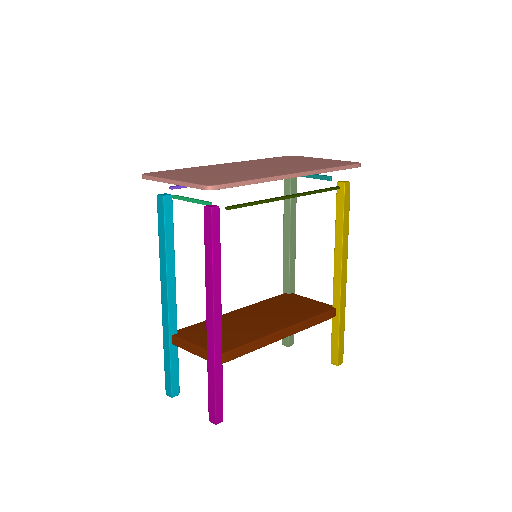}
    \end{subfigure}%
    \hfill%
    \begin{subfigure}[b]{0.20\linewidth}
		\centering
		\includegraphics[width=\linewidth]{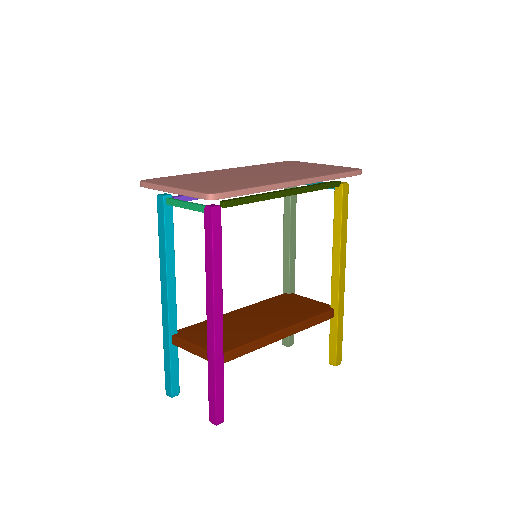}
    \end{subfigure}%
    \hfill%
    \begin{subfigure}[b]{0.20\linewidth}
		\centering
		\includegraphics[width=\linewidth]{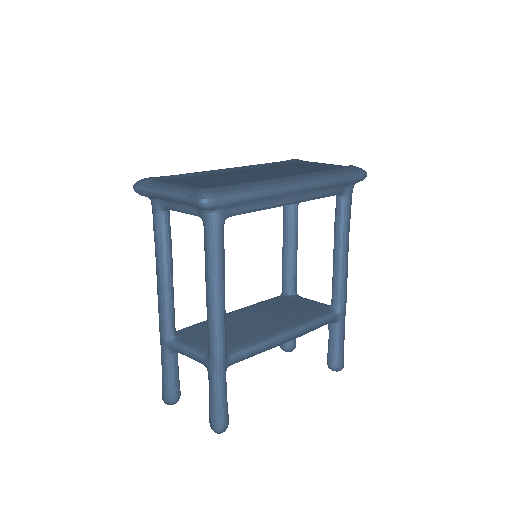}
    \end{subfigure}%
    \vskip\baselineskip%
    \vspace{-2.5em}
    \caption{{\bf Shape Completion Results on Tables}. Starting
    from partial tables, we show completions of our model,
    ATISS and PQ-NET.}
    \label{fig:shapenet_qualitative_completion_comparison_tables}
\end{figure}

%% file: supp_abstract.tex
\section*{Supplementary Material}
\subsection*{Abstract}
In this supplementary document, we provide a detailed overview of our
network architecture and the training procedure. Subsequently, we describe
the preprocessing steps that we followed to filter out problematic objects from
the PartNet dataset~\cite{Mo2019CVPR}. Next, we discuss how scheduled sampling
impacts the performance of our model on the
scene synthesis task. Finally, we provide additional qualitative and quantitative
results and analyze the limitations, future research directions, and potential
negative impact of our work on society.

%% file: supp_implementation_details.tex
\section{Implementation Details}
\label{sec:implementation_details_supp}

In this section, we provide a detailed description of the several components of
our network architecture (\secref{subsec:network}). Next, we describe our
training procedure (\secref{subsec:training}) and the generation protocol
(\secref{subsec:generation}). Finally, we detail our metrics computation
(\secref{subsec:metrics}) and discuss our baselines (\secref{subsec:baselines}).

\subsection{Network Architecture}
\label{subsec:network}

Here we describe the individual components of our network architecture and
provide additional implementation details.  Our model comprises two main
components: the \emph{object generator} that sequentially generates objects as
unordered sequences of labelled parts, where each part is parametrized using a
3D cuboidal primitive, and a \emph{blending network} that composes part
sequences in a meaningful way and synthesizes high quality implicit shapes. 

\emph{Object Generator }
We implement our object generator using an autoregressive transformer architecture
similar to ATISS~\cite{Paschalidou2021NEURIPS}. In particular, it comprises three main
components: (i) the \emph{part encoder} that takes the per-part attributes and
maps them to an embedding vector, (ii) the transformer encoder that takes the
embeddings for each part in the sequence, the embedding of the bounding box and
a learnable query embedding $\bq$ and predicts the feature vector $\bF$ and (iii)
the part decoder that takes $\bF$ and predicts the attributes of the next
object to be added in the scene.

The part encoder simply takes the attributes of each part $p_j=\{\bc_j,
\bs_j, \bt_j, \bo_j\}$ and maps them to an embedding vector $\bz_j$ as follows:
\begin{equation}
    \bz_j = s_\theta \big( \left[ \lambda(\bc_j); \gamma(\bs_j); \gamma(\bt_j); \gamma(\bo_j) \right] \big).
    \label{eq:part_encoder_supp}
\end{equation}
For the part label $\bc_j$, we use a learnable
embedding, denoted as $\lambda(\cdot)$, which is simply a matrix of size $C \times 64$, where $C$ is the
total number of part labels in the dataset.
The positional encoding layer, denoted as $\gamma(\cdot)$, that is applied on the remaining attributes
can be expressed as follows:
\begin{equation}
    \gamma(p) = (\sin(2^0\pi p), \cos(2^0\pi p), \dots, \sin(2^{L-1}\pi p), \cos(2^{L-1}\pi p))
    \label{eq:positional_encoding}
\end{equation}
where $p$ can be any of the size, translation or rotation. We follow \cite{Paschalidou2021NEURIPS} and set
$L=32$. Once the attributes are embedded to a higher dimensional space either using $\lambda(\cdot)$
or $\gamma(\cdot)$, we concatenate them to a $512$-dimensional feature vector, which is then passed
to another linear layer and generates the final embedding vector $\bz_j \in \mathbb{R}^{64}$

Similar to \cite{Vaswani2017NIPS, Paschalidou2021NEURIPS} we implement our transformer 
encoder as a multi-head attention transformer without any positional
encoding. Our transformer consists of $4$ layers with $8$ heads. The
queries, keys and values have $72$ dimensions and the intermediate
representations for the MLPs have $1024$ dimensions. We implement
the transformer architecture using the transformer library provided by
Katharopoulos \etal
\cite{Katharopoulos2020ICML}\footnote{\href{https://github.com/idiap/fast-transformers}{https://github.com/idiap/fast-transformers}}.
The learnable embedding vector $\bq$ and the predicted feature vector $\bF$
have both $64$ dimensions.

The part decoder takes the feature vector $\bF$ as input and
autoregressively predicts the attributes of the next part to be generated. The function
$c_\theta(\cdot)$ that is used for the part labels is a linear layer with $64$ hidden dimensions that
predicts C per-label probabilities. The functions, 
$t_\theta^{\text{coarse}}(\cdot)$, $t_\theta^{\text{fine}}(\cdot)$, $o_\theta^{\text{coarse}}(\cdot)$,
$o_\theta^{\text{fine}}(\cdot)$, $s_\theta^{\text{coarse}}(\cdot)$ and
$s_\theta^{\text{fine}}(\cdot)$ are implemented using a 2-layer MLP
with RELU non-linearities with hidden size $128$ and output size $64$. Using the
$t_\theta^{\text{fine}}(\cdot)$,
$o_\theta^{\text{fine}}(\cdot)$ and $s_\theta^{\text{fine}}(\cdot)$ we predict
the mean, variance and mixing coefficients for the $K$ logistic distributions for each attribute.
In our experiments, we set $K=10$.

\emph{Blending Network }
Our blending network consists of two main components: (i) a \emph{part-encoder}
that maps the part attributes into an embedding vector, which is implemented as
discussed before and in Sec. 3.2 of our main submission and (ii) a transformer
decoder without self-attention that takes the part embeddings and a set of
query points and predicts their occupancy probabilities, namely whether they
are inside or outside the surface boundary. We implement our transformer decoder
as a multi-head cross-attention transformer without self-attention.
Our transformer consists of $42$ layers with $8$ attention heads. The
queries, keys and values have $72$ dimensions and the intermediate
representations for the MLPs have $1024$ dimensions. To implement the
transformer decoder architecture we use the transformer library provided by Katharopoulos \etal
\cite{Katharopoulos2020ICML}\footnote{\href{https://github.com/idiap/fast-transformers}{https://github.com/idiap/fast-transformers}}.

\subsection{Training Protocol}
\label{subsec:training}

As already discussed in our main submission, we train the two components of our
model independently. To train the autoregressive transformer encoder, we use
the Adam optimizer \cite{Kingma2015ICLR} with learning rate $\eta=10^{-4}$ and
weight decay $10^{-3}$. For the other hyperparameters of Adam we use the
PyTorch defaults: $\beta_1=0.9$, $\beta_2 = 0.999$ and $\epsilon= 10^{-8}$. For
both category-specific training as well as joint training experiments, we train
the autoregressive transformer encoder with a batch size of $128$ for $700$k
iterations. During training, we do not perform any type of rotation
augmentation. To determine when to stop training, we follow
\cite{Paschalidou2021NEURIPS} and evaluate the validation metric every 1000
iterations and use the model that performed best as our final model. Likewise,
to train our blending network that is implemented using a transformer decoder
we use the Adam optimizer \cite{Kingma2015ICLR} with learning rate
$\eta=10^{-4}$ with no weight decay. For the other hyperparameters of Adam we
use the PyTorch defaults: $\beta_1=0.9$, $\beta_2 = 0.999$ and $\epsilon=
10^{-8}$. For the case of category-specific training, we train the transformer
on each object category with a batch size of $32$ for $150k$ iterations. For
the case of the joint-training on multiple object categories, we train the
blending network with a batch size of $32$ for 300k iterations. To train
the blending network, we need to generate occupancy pairs, namely points
accompanied by a label indicating whether this point lies inside or outside the
target mesh. To this end, we sample
$180,000$ points uniformly in a cube ranging from -1 to 1 centered at $(0, 0,
0)$ plus $20,000$ points from the surface for each mesh and compute which of
these points lie inside or outside the
mesh. During training, we sample $2048$ occupancy pairs from an unbalanced
distribution that, in expectation, results in an equal number of points with
positive and negative labels. Note that we follow common practice  and compute importance
sampling weights in order to reweigh our loss and create an unbiased
estimator of the loss with uniform sampling similar to \cite{Paschalidou2019CVPR}.
All our experiments were conducted on a single NVIDIA 2080 Ti GPU, with $11$GB
of memory, and training of both components of our architecture takes approximately 2 days.

\subsection{Generation Protocol}
\label{subsec:generation}

In this section, we discuss the sampling process for generating a novel part
arrangement. When we perform generation from scratch, we condition our
generation on a bounding box that specifies the object boundaries. Note that
when we want to generate novel shapes from a model that was trained jointly on
multiple object categories, we do not have to explicit condition on a specific
category. For the case of shape completion from an incomplete sequence of parts,
we condition our generation on the cuboidal primitives as well as the bounding box
that specifies the object boundaries. A similar concept is adapted for the case
of the language- and image-guided generation. As soon as a sequence of cuboidal
parts is produced, we pass it to the transformer decoder that composes the
cuboids and synthesizes an implicit 3D shape of high quality.

\subsection{Metrics}
\label{subsec:metrics}

As mentioned in the main submission, to evaluate the plausibility and the
diversity of the generated shapes using our model and our baselines, we report
the Coverage Score (COV) and the Minimum Matching Distance (MMD)
\cite{Achlioptas2018ICML} using the Chamfer-$L_2$ distance between points sampled
on the surface of the real and the generated shapes.
In particular, 
MMD measures the \emph{quality of the generated shapes} by computing how likely it
is that a generated shape looks like a shape from the reference set of shapes.
On the other hand, COV measures how many shape variations are covered by the
generated shapes, by computing the percentage of reference shapes that are closest
to at least one generated shape.

Let us denote $G$ the set of generated shapes and $R$ the set of reference shapes
from the test set. To estimate the similarity between
two shapes from the two sets, we use the Chamfer Distance (CD), which is simply
the distance between a set of points sampled on the surface of the reference
and the generated mesh. Namely, given a set of $N$ sampled points on the
surface of the reference $\cX=\{\bx_i\}_{i=1}^N$ and the generated shape
$\cY=\{\by_i\}_{i=1}^N$ the Chamfer Distance (CD) becomes
\begin{equation}
    {\text{CD}}(\cX, \cY) = \frac{1}{N} \sum_{\bx \in \cX}
    \min_{\by \in \cY}\mid \mid{\bx - \by \mid \mid}_2^2 +
    \frac{1}{M} \sum_{\by \in \cY}
    \min_{\bx \in \cX}\mid\mid{\by - \bx \mid \mid}_2^2.
    \label{eq:chamfer}
\end{equation}
In all our evaluations, we set $N=2048$. Note that to compute the Chamfer distance
between our generated part-based representations and real objects from the test
set, we sample points on the surface of the union of the generated cuboids and compute
their distance to points sampled from the reference shapes.

The Minimum Matching Distance (MMD) is the average distance
between each shape from the generated set $G$ to its closest shape in the reference
set $R$ and can be defined as:
\begin{equation}
    {\text{MMD}}(G, R) = \frac{1}{\lvert R \rvert} \sum_{\cX \in R} \min_{\cY \in G} {\text{CD}}(\cX, \cY).
    \label{eq:mmd}
\end{equation}
Intuitively, MMD measures how likely it is that a generated shape is similar to
a reference shape in terms of Chamfer Distance and is a metric of the plausibility
of the generated shapes. Namely, a high MMD score indicates that the shapes in
the generated set $G$ faithfully represent the shapes in the reference set $R$.

The Coverage score (COV) measures the percentage of shapes in the reference set
that are closest to each shape from the generated set. In particular, for each
shape in the generated set $G$, we assign its closest shape from the reference set 
$R$. In our measurement, we only consider shapes from $R$ that are closest to at
least one shape in $G$. Formally, COV is defined as
\begin{equation}
    {\text{COV}}(G, R) = \frac{\lvert \{\argmin_{\cX \in R}{\text{CD}}(\cX, \cY) \mid \cY \in G \}\rvert}{\lvert G \rvert}
    \label{eq:coverage}
\end{equation}
Intuitively, COV measures the diversity of the generated shapes in comparison
to the reference set. In other words, a high Coverage indicates that most of
the shapes in the reference set $R$ are roughly represented by the set of
generated shapes $G$. To ensure a fair comparison with our baselines, we
generate $2000$ shapes per object category for each baseline and compare it
with $225, 1216, 1477$ test shapes from the lamps, chairs and tables categories
respectively. Note that both the generated and the target shapes from the test set
are scaled within the unit cube before the metrics computations.

\subsection{Baselines}
\label{subsec:baselines}

In this section, we provide additional details regarding our baselines. We
compare our model with IM-NET~\cite{Chen2019CVPR}, PQ-NET~\cite{Wu2020CVPR}
and ATISS~\cite{Paschalidou2021NEURIPS}. For all our experiments, we retrain
all baselines, using the released code provided by the authors. For the case of
ATISS, which is an autoregressive transformer originally introduced for indoor
scene synthesis, we adapt the original
code\footnote{\href{https://github.com/nv-tlabs/ATISS}{https://github.com/nv-tlabs/ATISS}}
for the task of part-based object generation and train ATISS using
the per-object part sequences from PartNet~\cite{Mo2019CVPR}. Note that
IM-NET~\cite{Chen2019CVPR} is not directly comparable to our model as it does not
consider any parts. However, we include it in our evaluation as a powerful
implicit-based generative model for 3D shapes.

\boldparagraph{IM-NET}%
IM-NET~\cite{Chen2019CVPR} was among the first methods that proposed to implicitly
represent 3D object geometries in the weights of a neural network. In particular,
given an input feature representation and a 3D point, their model predicts whether
the query 3D point lies inside or outside the object's surface boundaries. IM-NET can be
combined with several generative frameworks such as Variational Autoencoders
(VAEs)~\cite{Kingma2014ICLR} or Generative Adversarial Networks
(GANs)~\cite{Goodfellow2014NIPS}. In our experiments, we consider the latter,
referred to as IM-GAN in the original paper, that utilizes a latent
GAN~\cite{Achlioptas2018ICML} directly trained on the latent feature space of a voxel-based
autoencoder. Note that IM-GAN relies on two-stage training, namely, first train
the autoencoder and then train the GAN on the autoencoder's latent space. In
our experiments, we train IM-NET using their Tensorflow
implementation~\cite{Abadi2016SIGPLAN}\footnote{
\href{https://github.com/czq142857/IM-NET}{https://github.com/czq142857/IM-NET}}
with the default parameters until convergence, on
the preprocessed data released by the authors. Training
of both the autoencoder and the latent GAN took approximately 2 days on a single
NVIDIA 2080 Ti GPU.

\boldparagraph{PQ-NET}%
In PQ-NET~\cite{Wu2020CVPR}, the authors introduce a generative model that
synthesizes 3D shapes sequentially using a set of parts parametrized as 
volumetric Signed Distance Fields (SDFs). In detail, PQ-NET comprises three core
components that need to be trained sequentially. First, the \emph{part
autoencoder} learns a mapping between the voxelized part-based representation
of a 3D shape to a volumetric SDF. Next, a sequence-to-sequence
autoencoder, implemented with two RNNs~\cite{Schuster1997ITSP} is employed, which
takes as input a sequence of per-part features and maps them to a latent
feature representation that describes the assembled 3D shape. This representation
is then passed to a sequential decoder that predicts a sequence of meaningful
parts. Lastly, to generate novel 3D shapes, a
GAN~\cite{Achlioptas2018ICML} is trained on the latent space of the sequential autoencoder. 
We train
PQ-NET using the provided PyTorch \cite{Paszke2016ARXIV}\footnote{\href{https://github.com/ChrisWu1997/PQ-NET}{https://github.com/ChrisWu1997/PQ-NET}} implementation with the
default parameters until convergence. Note that in our experiments, we do not
exclude shapes with more than $10$ shapes from our training data, as in the
original paper. Instead, we consider chairs
and tables with at most $50$, and lamps with no more than $30$ parts. Unlike our model,
considering shapes with a larger number of parts is not possible, because it
results in excessive memory usage that
prevents training PQ-NET's seq2seq module on a single GPU. Specifically, PQ-NET's part auto-encoder requires 3
NVIDIA 2080 Ti GPU. Part seq2seq module and GAN require 1 NVIDIA 2080 Ti. The training of all three components took approximately 4-5 days.

\boldparagraph{ATISS}%
In ATISS~\cite{Paschalidou2021NEURIPS}, the authors introduce an autoregressive
transformer architecture for indoor scene synthesis. In our experiments, we
repurpose the original
PyTorch~\cite{Paszke2016ARXIV}\footnote{\href{https://github.com/nv-tlabs/ATISS}{https://github.com/nv-tlabs/ATISS}} implementation in order to be able to utilize it for the
object generation task. In particular, each object is represented as a collection
of labelled cuboidal primitives and we train ATISS to maximize the log-likelihood
of all possible permutations of part arrangements in a collection of training data.
We train ATISS using the default parameters for $1200$ epochs on an NVIDIA 2080 Ti GPU
for approximately 1-2 days, without any data augmentation.
Unlike PQ-NET, to train ATISS, we do not filter out
shapes with a larger number of parts. Specifically, we consider the same set of
shapes used to train our model, namely chairs, tables
and lamps with a maximum number of $144,164,191$ parts respectively.

\subsection{Mesh Extraction}

To extract meshes from the predicted occupancy field, we
employ the Marching Cubes algorithm \cite{Lorensen1987SIGGRAPH}.
In particular, we start from a voxel grid of $128^3$
initial resolution for which we predict occupancy values. Next, we follow the
process proposed in \cite{Mescheder2019CVPR} and extract the approximate
isosurface with Marching Cubes using the code
provided by Mescheder \etal \cite{Mescheder2019CVPR}. Note that the same process
is followed to extract meshes from the implicit representations learned
both with PQ-NET~\cite{Wu2020CVPR} and IM-NET~\cite{Chen2019CVPR}.

%% file: supp_data_processing.tex
\section{Data Processing}
\label{sec:data_processing}
We use PartNet \cite{Mo2019SIGGRAPH} as the dataset to evaluate our model and our baselines. We remove 123 objects in total across all categories due to their invalid part hierarchical structures (\ie missing nodes \etc). 
To train our blending network, we assume explicit 3D supervision in the form of a wateright mesh.
To acquire this, we align ShapeNet~\cite{Chang2015ARXIV} objects with PartNet objects using the scripts provided in the official PartNet repository\footnote{\href{https://partnet.cs.stanford.edu}{https://partnet.cs.stanford.edu/}}. We then convert aligned objects into watertight meshes using the code provided
by Stutz \etal \cite{Stutz2018CVPR}\footnote{\href{https://github.com/paschalidoud/mesh\_fusion\_simple}{https://github.com/paschalidoud/mesh\_fusion\_simple}}.
To train the variant of our model for language-guided generation, we remove samples, whose descriptions are too long
and cannot be handled by CLIP \cite{Radford2021ICML}.
After this processing step, we have 4000 training samples for chairs, 5439 for tables, and 1424 for lamps. 

\figref{fig:shapenet_statistics}
visualizes the part sequences for all object categories. We note that
chairs contain more samples with longer sequences, while lamps tend to have objects with fewer components,
hence making them easier to generate.
\input{fig/sequence_lengths_distribution.tex} 

%% file: fig/sequence_lengths_distribution.tex
\begin{figure}[h!]
  \centering
  \begin{subfigure}[b]{\linewidth}
    \includegraphics[width=\linewidth]{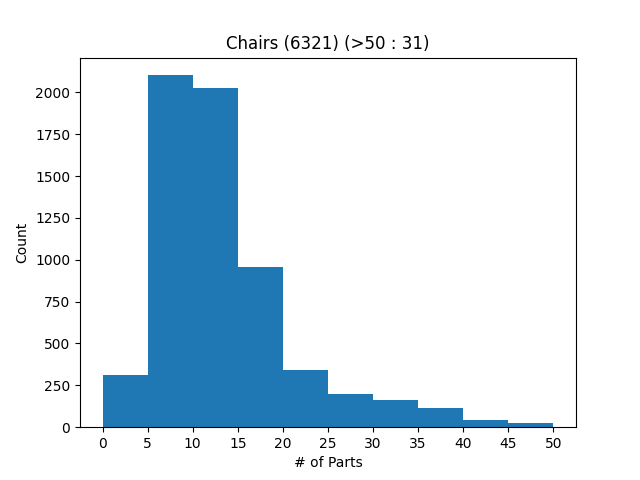}
  \end{subfigure}
  \begin{subfigure}[b]{\linewidth}
    \includegraphics[width=\linewidth]{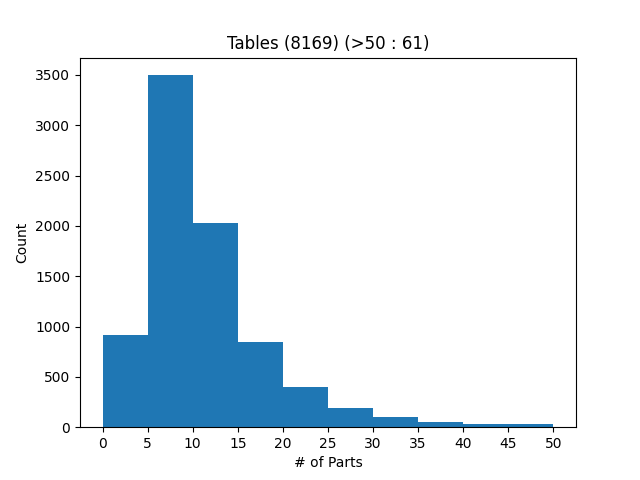}
  \end{subfigure}
    \begin{subfigure}[b]{\linewidth}
    \includegraphics[width=\linewidth]{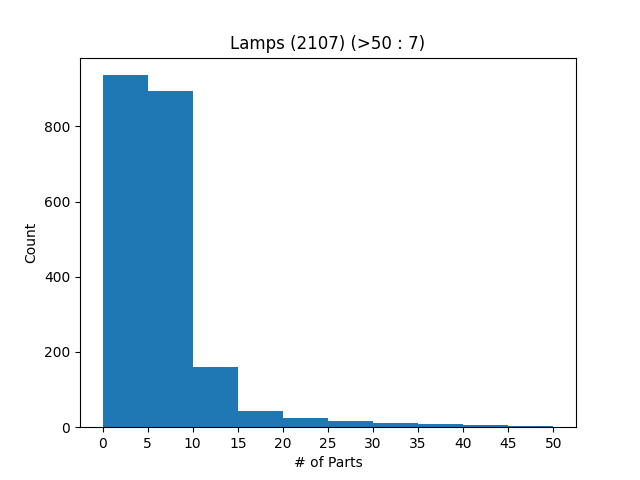}
  \end{subfigure}
  \caption{\bf{Sequence Lengths of all Object Categories.}}
  \label{fig:shapenet_statistics}
\end{figure}

%% file: supp_ablations.tex
\section{Impact of Schedule Sampling}

\begin{figure}
    \begin{subfigure}[t]{0.2\linewidth}
	    \centering
	    \includegraphics[width=\linewidth]{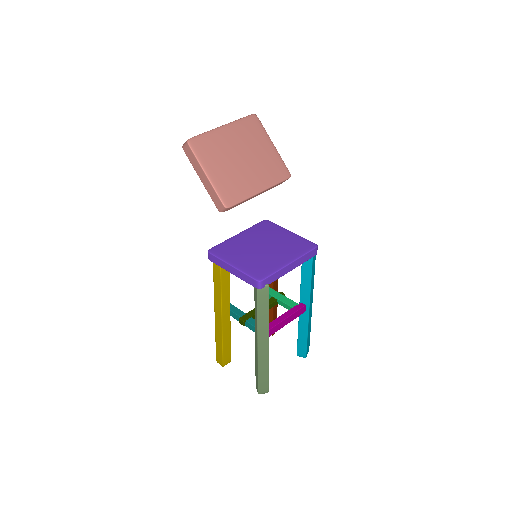}
    \end{subfigure}%
    \hfill%
    \begin{subfigure}[t]{0.2\linewidth}
	    \centering
	    \includegraphics[width=\linewidth]{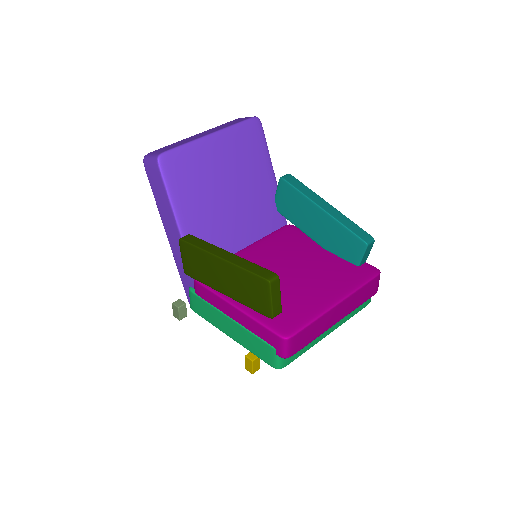}
    \end{subfigure}%
    \hfill%
    \begin{subfigure}[t]{0.2\linewidth}
	    \centering
	    \includegraphics[width=\linewidth]{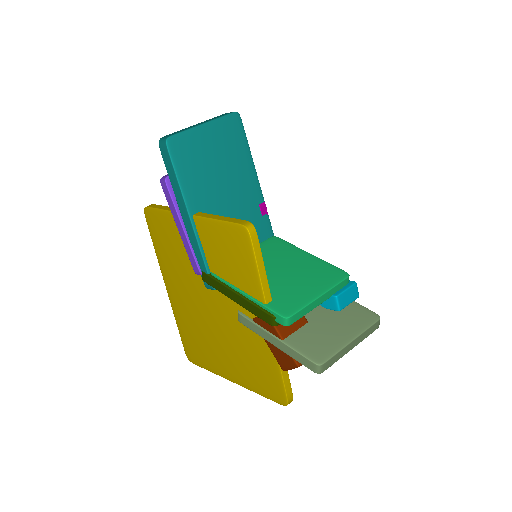}
    \end{subfigure}%
    \hfill%
    \begin{subfigure}[t]{0.2\linewidth}
	    \centering
	    \includegraphics[width=\linewidth]{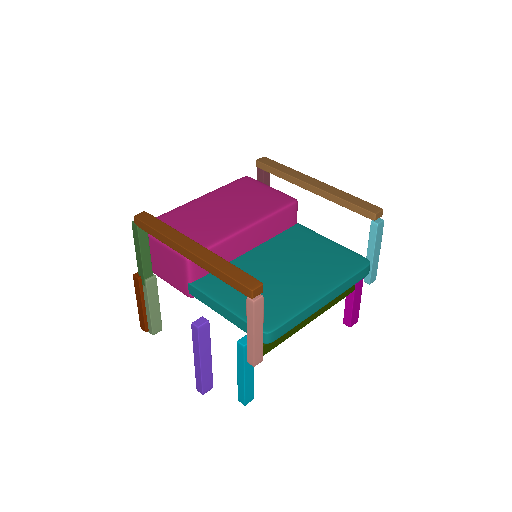}
    \end{subfigure}%
    \hfill%
    \begin{subfigure}[t]{0.2\linewidth}
	    \centering
	    \includegraphics[width=\linewidth]{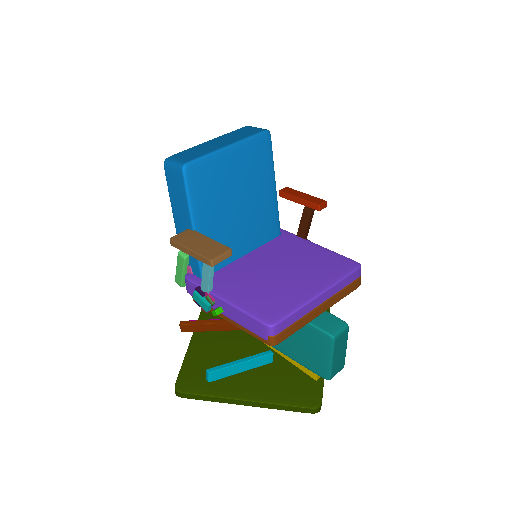}
    \end{subfigure}%
    \vskip\baselineskip%
    \vspace{-2em}
    \begin{subfigure}[t]{\linewidth}
        \centering
        Ours w/o Schedule Sampling
    \end{subfigure}
    \vskip\baselineskip%
    \vspace{-1em}
    \begin{subfigure}[t]{0.2\linewidth}
	    \centering
	    \includegraphics[width=\linewidth]{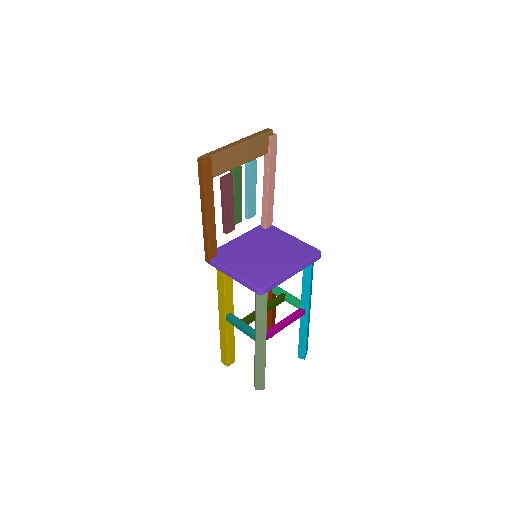}
    \end{subfigure}%
    \hfill%
    \begin{subfigure}[t]{0.2\linewidth}
	    \centering
	    \includegraphics[width=\linewidth]{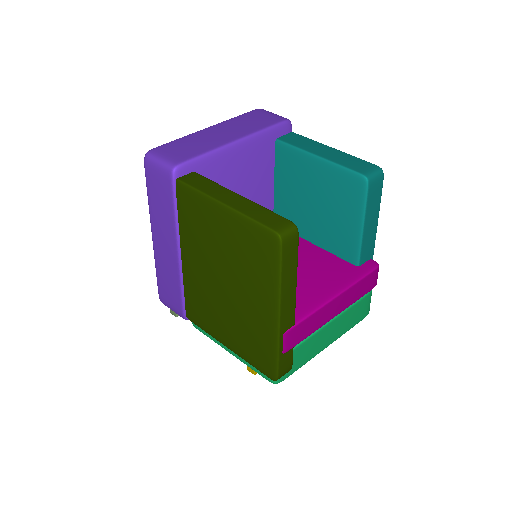}
    \end{subfigure}%
    \hfill%
    \begin{subfigure}[t]{0.2\linewidth}
	    \centering
	    \includegraphics[width=\linewidth]{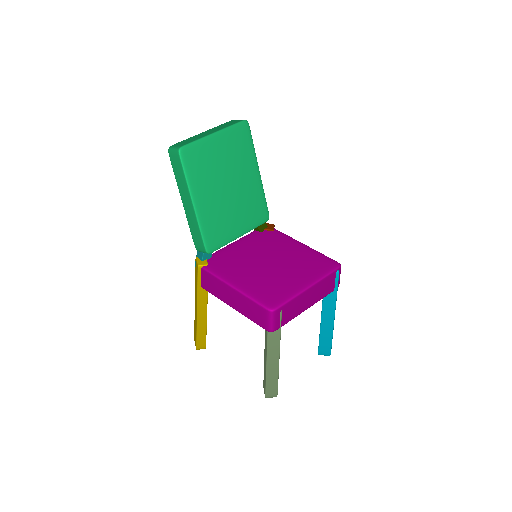}
    \end{subfigure}%
    \hfill%
    \begin{subfigure}[t]{0.2\linewidth}
	    \centering
	    \includegraphics[width=\linewidth]{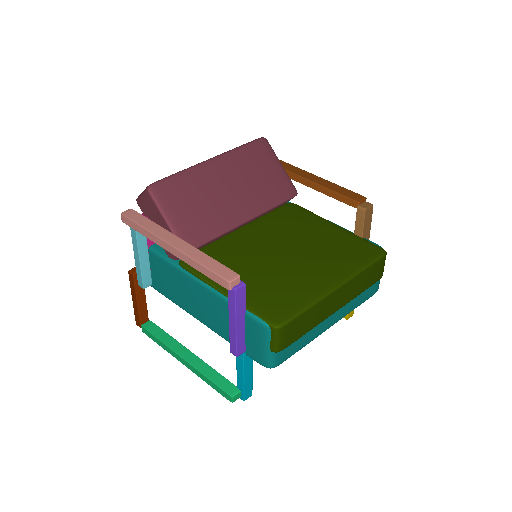}
    \end{subfigure}%
    \hfill%
    \begin{subfigure}[t]{0.2\linewidth}
	    \centering
	    \includegraphics[width=\linewidth]{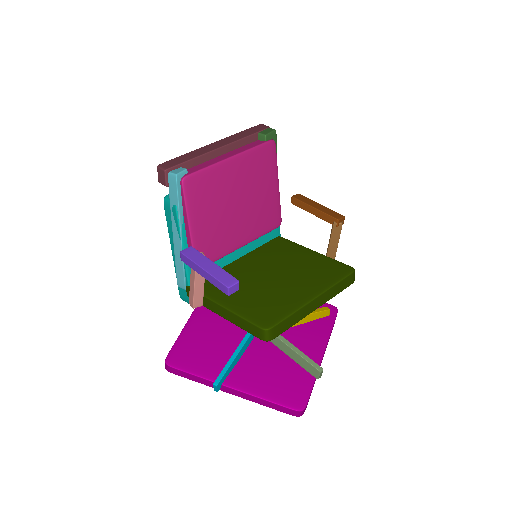}
    \end{subfigure}%
    \vskip\baselineskip%
    \vspace{-2em}
    \begin{subfigure}[t]{\linewidth}
        \centering
        Ours
    \end{subfigure}
    \vskip\baselineskip%
    \vspace{-2.2em}
    \caption{{\bf Impact of Schedule Sampling}. We show randomly generated
    samples of our model trained with teacher forcing(first row) and compare
    with randomly generated samples of our model trained with our schedule
    sampling strategy (second row).}
    \label{fig:schedule_sampling_impact}
    \vspace{-1.2em}
\end{figure}

In this section, we investigate how schedule sampling affects the generation
capabilities of our model. For this experiment, we perform category specific
training on the Chair category. In particular, we train two variants of our
model, one with teacher forcing and one with schedule sampling until
convergence. We compare the two model variants \wrt their generation
performance in \tabref{tab:schedule_sampling_quantitative}.  We note that
training our network only with teacher forcing, significantly deteriorates
performance and results in generations of lower quality.
\begin{table}[!h]
\resizebox{\columnwidth}{!}{%
    \centering
    \begin{tabular}{c|c|c}
        \toprule
        & MMD-CD ($\downarrow$) & COV-CD ($ \%, \uparrow$) \\
        \midrule
        Ours w/o Schedule Sampling & 3.74 & 56.99 \\
        Ours & \bf{3.21} & \bf{57.73} \\
        \bottomrule
    \end{tabular}
    }
    \caption{{\bf Ablation Study on Schedule Sampling.} This table shows a quantitative
    comparison of our approach trained with teacher forcing and our proposed 
    schedule sampling strategy. We compare the two variants of our model \wrt
    the MMD-CD ($\downarrow$) and the COV-CD ($\uparrow$) between generated
    and real shapes from the test set.}
    \label{tab:schedule_sampling_quantitative}
    \vspace{-2.2em}
\end{table}
This is also validated
from our qualitative comparison in \figref{fig:schedule_sampling_impact}, where
we visualize randomly generated chairs using both models. We observe that the variant
of our model trained only with teacher forcing tends to produce part arrangements, 
with parts placed in unnatural positions. On the contrary, when training our
model with our proposed schedule sampling strategy, we observe that our model
consistently generate plausible part sequences. This is expected as schedule sampling
allows training our model with imperfect data, which makes it more robust to imperfect
generations. Namely, even if one of the generated parts is problematic, our
network can produce plausible parts in the next generation steps.

%% file: supp_additional_results.tex
\section{Additional Experimental Results}

In this section, we provide additional information regarding our experiments on
PartNet~\cite{Mo2019CVPR}. In particular, we consider three categories: \emph{Chair,
Table} and \emph{Lamp}, which contain 4489, 5705, and 1554 shapes
respectively.  For the \emph{Chair} category, there are 47 different 
classes (\eg back surface horizontal bar, arm holistic frame \etc) in total,
while for the \emph{Lamp} and \emph{Table}, we have 32 and 43 part categories
respectively. Note that unlike prior works such as PQ-NET~\cite{Wu2020CVPR}
that only consider shapes that have less than $10$ parts, we consider shapes
with a significantly larger number of components. In particular, chairs can have up to
$144$ parts, tables $164$ parts and lamps up to $191$ parts.
For more details, regarding our training data, we refer reader to
\secref{sec:data_processing}. In this section, we provide additional
qualitative results for all experiments discussed in our main submission. 

\begin{figure}
    \begin{subfigure}[t]{0.2\linewidth}
	    \centering
	    \includegraphics[width=\linewidth]{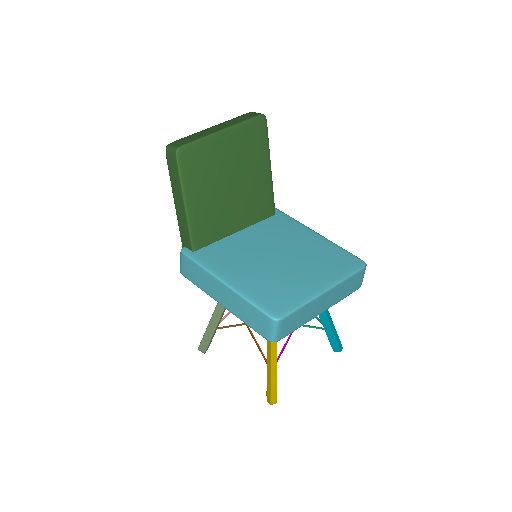}
    \end{subfigure}%
    \hfill%
    \begin{subfigure}[t]{0.2\linewidth}
	    \centering
	    \includegraphics[width=\linewidth]{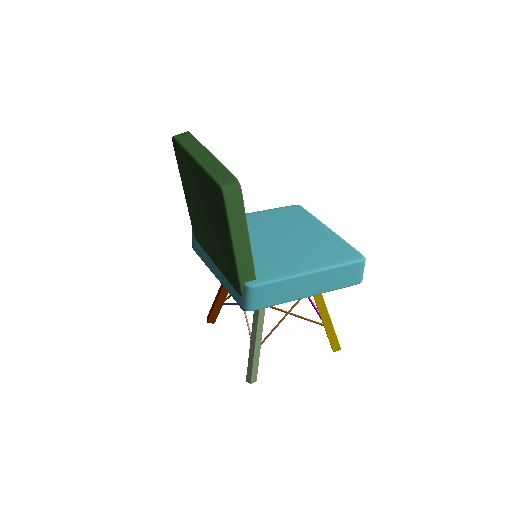}
    \end{subfigure}%
    \hfill%
    \begin{subfigure}[t]{0.2\linewidth}
	    \centering
	    \includegraphics[width=\linewidth]{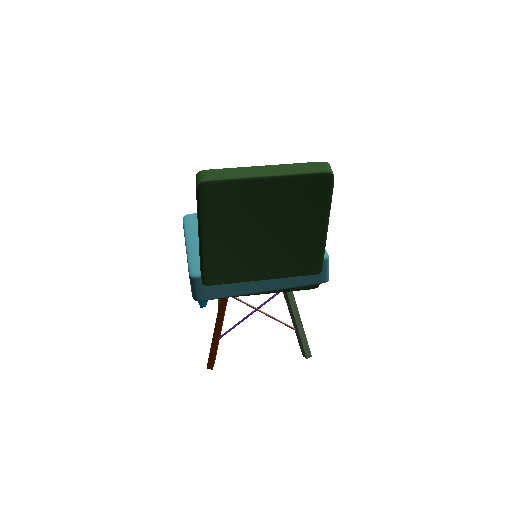}
    \end{subfigure}%
    \hfill%
    \begin{subfigure}[t]{0.2\linewidth}
	    \centering
	    \includegraphics[width=\linewidth]{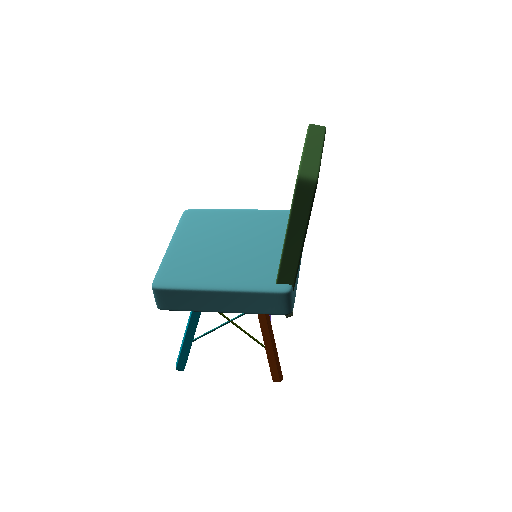}
    \end{subfigure}%
    \hfill%
    \begin{subfigure}[t]{0.2\linewidth}
	    \centering
	    \includegraphics[width=\linewidth]{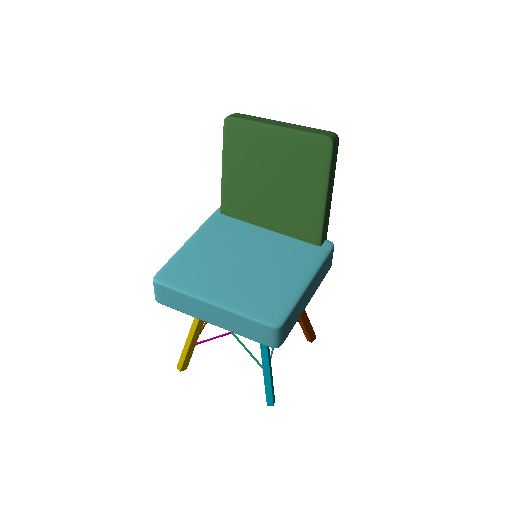}
    \end{subfigure}%
    \vskip\baselineskip%
    \vspace{-3em}
    \begin{subfigure}[t]{0.2\linewidth}
	    \centering
	    \includegraphics[width=\linewidth]{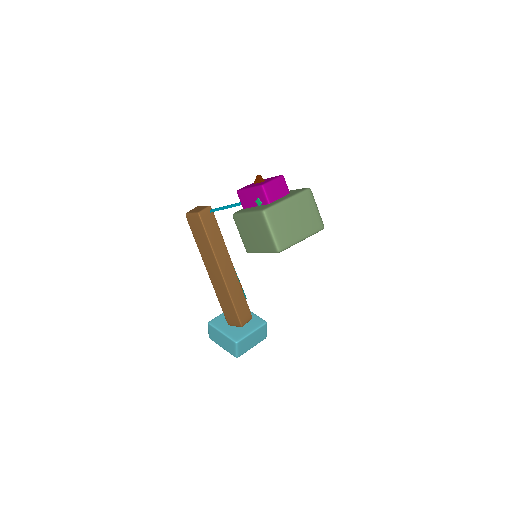}
    \end{subfigure}%
    \hfill%
    \begin{subfigure}[t]{0.2\linewidth}
	    \centering
	    \includegraphics[width=\linewidth]{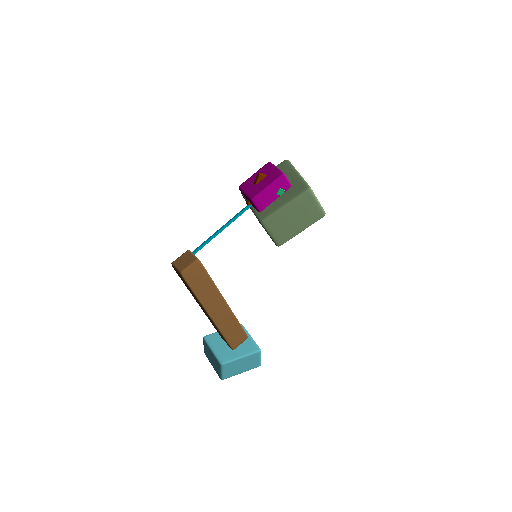}
    \end{subfigure}%
    \hfill%
    \begin{subfigure}[t]{0.2\linewidth}
	    \centering
	    \includegraphics[width=\linewidth]{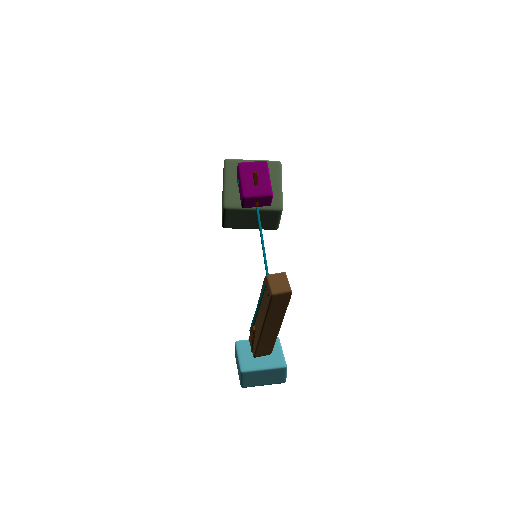}
    \end{subfigure}%
    \hfill%
    \begin{subfigure}[t]{0.2\linewidth}
	    \centering
	    \includegraphics[width=\linewidth]{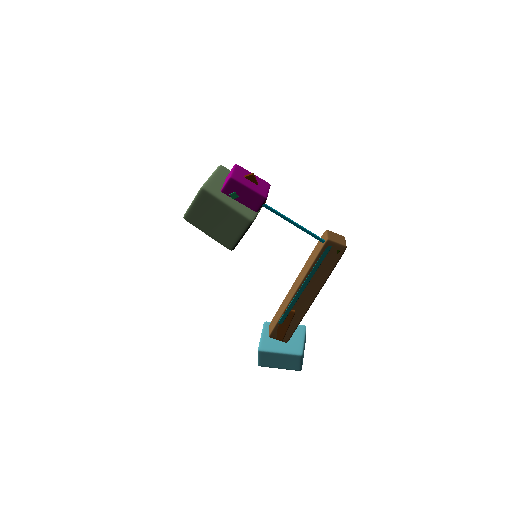}
    \end{subfigure}%
    \hfill%
    \begin{subfigure}[t]{0.2\linewidth}
	    \centering
	    \includegraphics[width=\linewidth]{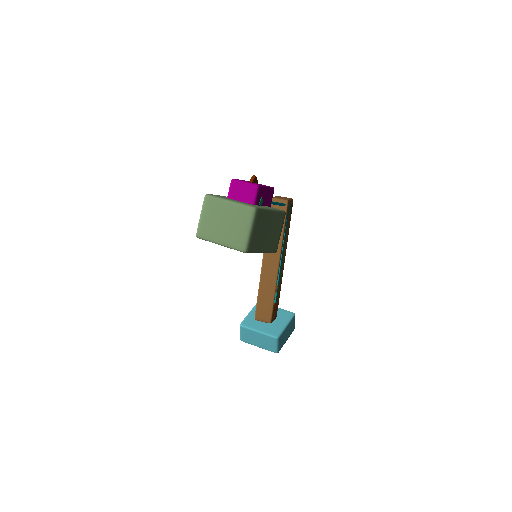}
    \end{subfigure}%
    \vskip\baselineskip%
    \vspace{-2.2em}
    \caption{{\bf Example Rendering of Target Images for FID Computation}.}
    \label{fig:images_for_fid_computation}
    \vspace{-1.2em}
\end{figure}

\input{tab/shapenet_completion_quantitative_supp}
\subsection{Shape Generation}

In this experiment, we investigate the ability of our model to generate
plausible part-aware 3D geometries, conditioned on various bounding boxes that
specify the object's boundaries.
\figref{fig:shapenet_qualitative_comparison_chairs_supp} shows seven randomly
generated chairs using our model, ATISS~\cite{Paschalidou2021NEURIPS},
PQ-NET~\cite{Wu2020CVPR} and IM-NET~\cite{Chen2019CVPR}. Note that for this
experiment, we perform category-specific training, namely we train a different
model for each object type.
For our model, we visualize both the synthesized part arrangements (Ours-Parts)
and the output of our blending network that composes the sequence of the
generated cuboids into a single high-quality implicit shape. We observe that our
synthesized part sequences are consistently meaningful. For the case of ATISS,
we note that the generated cuboids are placed in unnatural positions, hence
producing non-functional objects. We hypothesize that this is mainly due to the
large number of parts that typically compose chairs. On the contrary, we
observe that both PQ-NET and IM-NET produce plausible chairs. For the case of tables
(see \figref{fig:shapenet_qualitative_comparison_tables_supp}), we observe that our model
consistently generates meaningful part sequences, whereas ATISS again produces
part arrangements, where parts are placed in unnatural positions. For PQ-NET, we observe that
in some cases, it tends to generate non-functional tables with less legs (see
5th row in \figref{fig:shapenet_qualitative_comparison_tables_supp}). Finally, we also provide
seven generated lamps of our model and our baselines in \figref{fig:shapenet_qualitative_comparison_lamps_supp}.
For the case of lamps, both our model and ATISS seem to be able to generate realistic sequences of cuboids.
For all the experiments in this section, as well as the results in Sec. 4.1 of our main submission
we condition the generation on randomly sampled bounding boxes from the test set.

For all object categories, we observe that our blending network composes the
generated part sequences in a meaningful way and synthesizes novel 3D shapes
that respect the provided part-based structure.  The output representation of
our blending network is of higher quality than IM-NET~\cite{Chen2019CVPR} that
also generates implicit 3D shapes, as indicated by our quantitative evaluation
from Table 1 in our main submission.
\input{fig/shape_generation_qualitative_chairs_supp}
\input{fig/shape_generation_qualitative_tables_supp}
\input{fig/shape_generation_qualitative_lamps_supp}

\subsection{Shape Completion}

Starting from a partial object, parameterized with a set of cuboidal parts, we
want to evaluate whether our model and our baselines can generate plausible
part arrangements. Since IM-NET cannot be used to complete 3D shapes from a
partial sequence of parts, we exclude it from our evaluation.  To measure the
quality of the generated parts, in this experiment, we also report the
classification accuracy of a classifier trained to discriminate real from
synthetic objects. In particular, as we represent objects using a collection of
cuboidal primitives, we implement our classifier using a transformer encoder
\cite{Vaswani2017NIPS} trained to discriminate real from generated 3D labelled
cuboids. Furthermore, in our evaluation, we also the FID
score~\cite{Heusel2017NIPS}.  For the FID score computation, we generate the
same amount of objects as in the test set and render them at $512\times512$
resolution using $5$ random camera views.  To evaluate the realism of the
generated parts, we render objects using their part-based representation, as
show in \figref{fig:images_for_fid_computation}, and we compare with
corresponding part-based renderings from the ground-truth objects from the test
set, rendered using the same camera distribution. We follow common practice and
repeat the metric computation for FID $10$ times and report the mean. The
quantitative results for this experiment are summarized in
\tabref{tab:shape_completion_quantitative_supp}. Note that we compare our model
only two ATISS \wrt FID and classification accuracy, which also generates
objects as a sequence of cuboidal primitives.

To generate the partial objects, we randomly sample objects from the test set
and generate the partial input by removing an arbitrary set of parts.
As mentioned also in our main submission, to ensure a fair comparison to
PQ-NET, for this task instead of conditioning on the partial set of cuboids, we
utilize the corresponding 3D parts that were used during the PQ-NET's training.
The 1st column in
\figref{fig:shapenet_qualitative_completion_comparison_chairs_supp},
\figref{fig:shapenet_qualitative_completion_comparison_tables_supp} and
\figref{fig:shapenet_qualitative_completion_comparison_lamps_supp} shows the
partial input in the form of 3D cuboids that is used in the case of PASTA and ATISS.
For PQ-NET, we utilize the actual 3D parts that correspond to each cuboid. Both
from the quantitative analysis in
\tabref{tab:shape_completion_quantitative_supp}, as well as the qualitative
results in \figref{fig:shapenet_qualitative_completion_comparison_chairs_supp}
and \figref{fig:shapenet_qualitative_completion_comparison_tables_supp}, we
observe that our model completes the partial input chairs and table in a more
plausible way, than both PQ-NET and ATISS. Note that, since PQ-NET employs a
sequence-to-sequence autoencoder to generate part sequences, there is no
guarantee that the original part sequence, will also appear in the completed
sequence. For the case of ATISS, we observe that it struggles completing the
partial input in a meaningful way. For example, the added parts are placed in
non realistic positions (see last row in
\figref{fig:shapenet_qualitative_completion_comparison_chairs_supp} and 3rd row
in \figref{fig:shapenet_qualitative_completion_comparison_tables_supp}. For the case of lamps,
we note that all three models can successfully complete the partial sequence.

\subsection{Size-guided Generation}%
Now we examine whether our model can generate objects
of different sizes. Note that as we condition the generation of parts on a
bounding box that defines the object boundaries, our model can generate shapes
of arbitrary sizes. In this experiment, we generate several bounding boxes,
with different size parameters and demonstrate the ability of our model to generate short and
tall lamps (see 1st and 3rd lamp in 3rd row in
\figref{fig:shapenet_generations_variable_sizes}, respectively), or smaller and
bigger tables (see 1st and 2nd tables in 2nd row in
\figref{fig:shapenet_generations_variable_sizes}). We believe that this is an
important application of our model that allows users to precisely specify
the size of the generated object. For all experiments presented in
\figref{fig:shapenet_generations_variable_sizes}, we perform category-specific training per object type.
\input{fig/shape_generation_variable_sizes}

\subsection{Language-guided Generation}
Starting from a text prompt and a bounding box that defines the object's
boundaries, we want to examine the ability of our model to generate plausible
part arrangements that match the input text descriptions.  To this end, for
this task, we utilize the part labels provided in PartNet~\cite{Mo2019CVPR} and
generate text descriptions that describe the part-based structure for each
object. Some examples of the produced text descriptions for various object
categories are summarized below:
\begin{itemize}
    \item \textit{A chair with four leg, one bar stretcher, three runners, one
    seat single surface, one arm horizontal bar, two arm near vertical bars,
    two arm horizontal bars, two arm near vertical bars, one arm horizontal
    bar, and one back single surface.}
    \item \textit{A chair with four leg with two runners, one seat single surface, and one back single
    surface.}
    \item \textit{A table with one drawer front, one handle, one drawer front,
    one handle, two vertical side panels, one bottom panel, four leg, one back
    panel, one vertical front panel, and one board.}
    \item \textit{A table with one central support, one pedestal, one tabletop
    connector, one other, one board, and one tabletop frame.}
    \item \textit{A lamp with one lamp shade, one light bulb, one other, one
    chain, and one lamp base part.}
    \item \textit{A lamp with one lamp arm straight bar, one lamp shade, one
    light bulb, one other, one lamp base part, and one lamp body.}
\end{itemize}
Using these text descriptions, we utilize a pre-trained
CLIP~\cite{Radford2021ICML}\footnote{\href{https://github.com/OpenAI/CLIP}{https://github.com/OpenAI/CLIP}}
text encoder to extract embeddings for the shape's textual descriptions. During
training, we condition our generation both on the CLIP text embeddings and the
embedding produced from the bounding box, containing the object. Note that
during training the CLIP text-encoder is not optimized with the rest of our
architecture. While our model was not trained with free-text descriptions, we
showcase that by exploiting CLIP's powerful latent space our model can generate
plausible part arrangements and 3D objects that match the input text prompt
(see \figref{fig:text_guided_generation_supp}). To be able to control the shape
of the generated shape \eg. generate a small chair or a narrow table, it simply
suffices to condition the generation on a bounding box that fits these
criteria. For now, we do this manually, namely we generate a bounding box that
fits the text input.
\input{fig/shape_completion_qualitative_chairs_supp}
\input{fig/shape_completion_qualitative_tables_supp}
\input{fig/shape_completion_qualitative_lamps_supp}
\input{fig/text_guided_generation_free_text_supp}

\subsection{Image-guided Generation}
For this task, we utilize the variant of our model that was trained for
language-guided generation without any re-training. CLIP~\cite{Radford2021ICML}
learns a common latent space between images and sentences that describe them.
Therefore, we take advantage of CLIP's joint latent space, and use the variant of
our model trained for language-guided generation, to synthesize plausible 3D
shapes from images, simply by replacing the CLIP's text encoder, with the
corresponding CLIP image encoder. While our model was only trained with text
embeddings, we showcase that it can successfully generate part sequences that
match the input image (see \figref{fig:image_guided_generation_supp}). Note that
the recovered parts capture fine geometric details such as the circular base of the 2nd chair,
in the second row in the \figref{fig:image_guided_generation_supp}, and our model is able to
generate realistic shapes conditioned on images with and without backgrounds.
\input{fig/shape_generation_image_guided_supp}

\subsection{Generating Part Variations}
Finally, we also demonstrate that our model can produce plausible part
variations for a specific part, selected by the user. For example in
\figref{fig:part_variation}, we select the back of the chair, highlighted with
red and show that our model can generate variations of the selected part with
different sizes.
\input{fig/part_variations}

%% file: tab/shapenet_completion_quantitative_supp.tex
\begin{table*}
\resizebox{\linewidth}{!}{%
    \centering
    \begin{tabular}{l|ccc|ccc|ccc|cccc}
        \toprule
        \multicolumn{1}{c}{\,} &
        \multicolumn{3}{c}{MMD-CD ($\downarrow$)} & \multicolumn{3}{c}{COV-CD ($ \%, \uparrow$)} &
        \multicolumn{3}{c}{FID ($\downarrow$)} & \multicolumn{3}{c}{Classification Accuracy} \\
        & Chair & Table & Lamp & Chair & Table & Lamp & Chair & Table & Lamp & Chair & Table
        & Lamp\\
        \toprule
        PQ-Net & 4.69 & 3.64 & \bf{4.55} & 35.77 & 42.31 & 53.33 & - & - & - & - & - & -\\
        ATISS & 4.33 & 3.40 & 5.90 & 44.57 & 43.33 & \bf{60.88} & 13.63 & 30.87 & 25.86 & 76.85 $\pm$ 1.8 & 72.16 $\pm$ 1.2 &  45.7 $\pm$ 9.84 \\
        \midrule
        Ours-Parts & 3.43 & 2.66 & 5.72 & \bf{50.49} & 56.13  & 57.78 &  \bf{3.92} & \bf{3.90} & \bf{15.2} & \bf{60.83 $\pm$ 0.59} & \bf{65.9 $\pm$ 0.01} & \bf{48.55 $\pm$ 10.16} \\ 
        Ours & \bf{3.11} & \bf{2.33} & 5.58 & \bf{49.00} & \bf{58.56} & 51.00 & - & - & - & - & - & -\\
        \bottomrule
    \end{tabular}
    }
    \caption{{\bf Shape Completion.} We measure
    the MMD-CD ($\downarrow$), the COV-CD ($\uparrow$), the
    FID score ($\downarrow$) and the Parts Classification Accuracy between the
    part-based representations of completed and real shapes from the test set.
    Classificiation accuracy closer to 0.5 is better.}
    \label{tab:shape_completion_quantitative_supp}
    \vspace{-1.2em}
\end{table*}

%% file: fig/shape_generation_qualitative_chairs_supp.tex
\begin{figure*}
    \begin{subfigure}[t]{\linewidth}
    \centering
    \begin{subfigure}[b]{0.20\linewidth}
        \centering
	    IM-NET
    \end{subfigure}%
    \hfill%
    \begin{subfigure}[b]{0.20\linewidth}
	\centering
        PQ-NET
    \end{subfigure}%
    \hfill%
    \begin{subfigure}[b]{0.20\linewidth}
	\centering
        ATISS
    \end{subfigure}%
    \hfill%
    \begin{subfigure}[b]{0.20\linewidth}
        \centering
        Ours-Parts
    \end{subfigure}%
    \hfill%
    \begin{subfigure}[b]{0.20\linewidth}
        \centering
        Ours
    \end{subfigure}
    \end{subfigure}
    \vspace{-1.5em}
    \begin{subfigure}[b]{0.20\linewidth}
		\centering
		\includegraphics[width=0.8\linewidth]{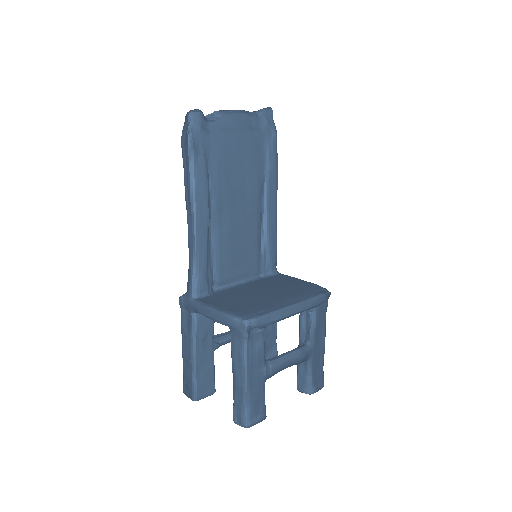}
    \end{subfigure}%
    \hfill%
    \begin{subfigure}[b]{0.20\linewidth}
		\centering
		\includegraphics[width=0.8\linewidth]{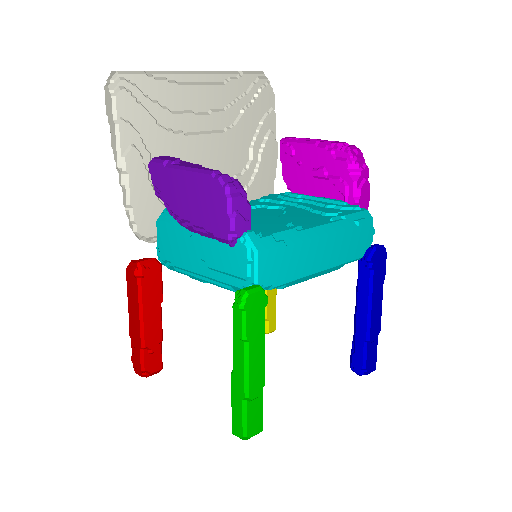}
    \end{subfigure}%
    \hfill%
    \begin{subfigure}[b]{0.20\linewidth}
		\centering
		\includegraphics[width=0.8\linewidth]{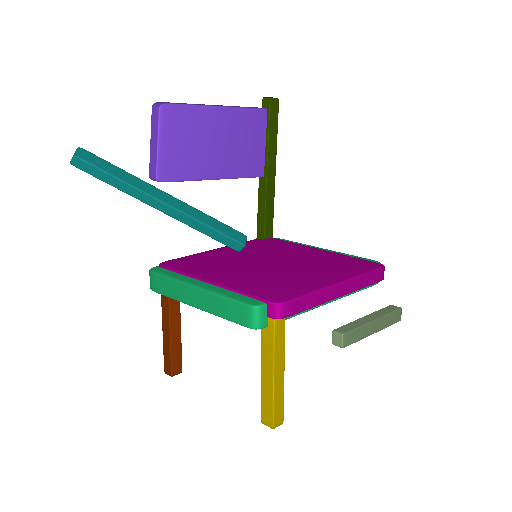}
    \end{subfigure}%
    \hfill%
    \begin{subfigure}[b]{0.20\linewidth}
		\centering
		\includegraphics[width=0.8\linewidth]{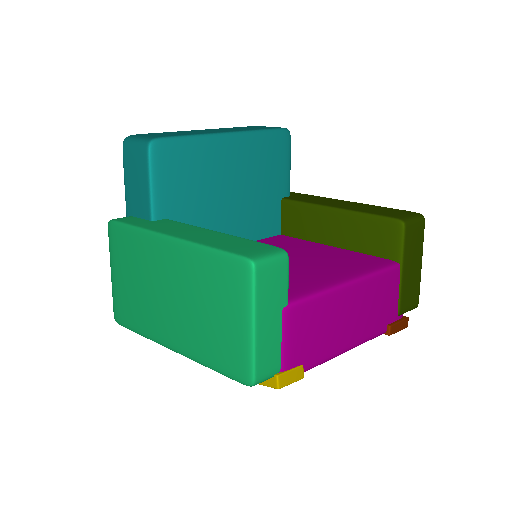}
    \end{subfigure}%
    \hfill%
    \begin{subfigure}[b]{0.20\linewidth}
		\centering
		\includegraphics[width=0.8\linewidth]{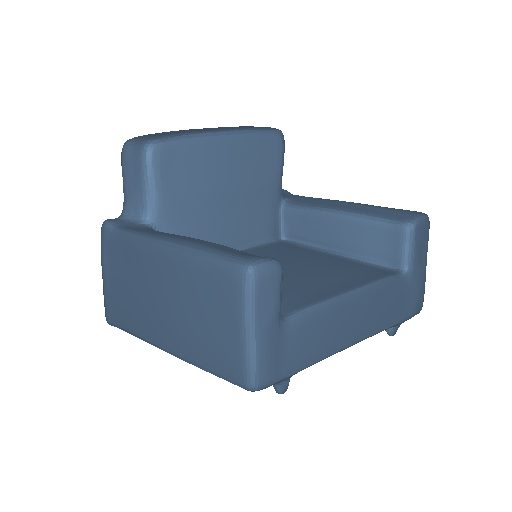}
    \end{subfigure}%
    \vskip\baselineskip%
    \vspace{-1.25em}
    \begin{subfigure}[b]{0.20\linewidth}
		\centering
		\includegraphics[width=0.8\linewidth]{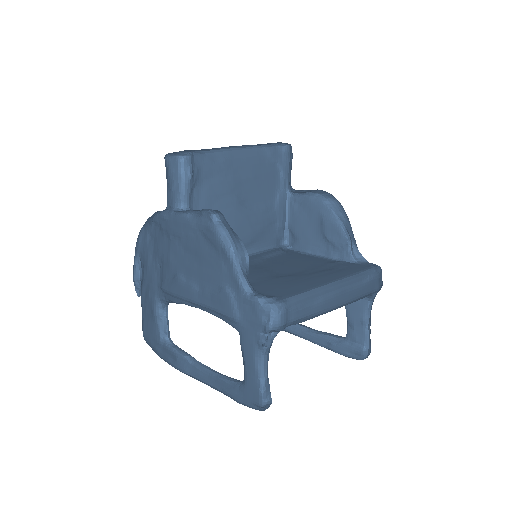}
    \end{subfigure}%
    \hfill%
    \begin{subfigure}[b]{0.20\linewidth}
		\centering
		\includegraphics[width=0.8\linewidth]{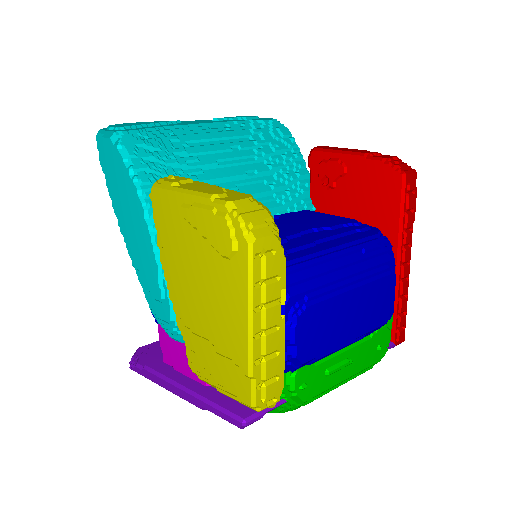}
    \end{subfigure}%
    \hfill%
    \begin{subfigure}[b]{0.20\linewidth}
		\centering
		\includegraphics[width=0.8\linewidth]{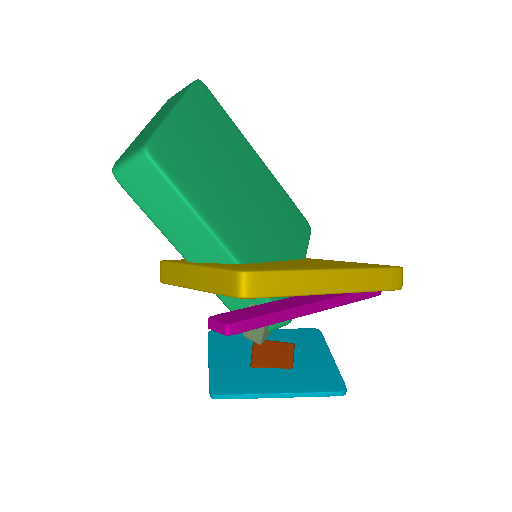}
    \end{subfigure}%
    \hfill%
    \begin{subfigure}[b]{0.20\linewidth}
		\centering
		\includegraphics[width=0.8\linewidth]{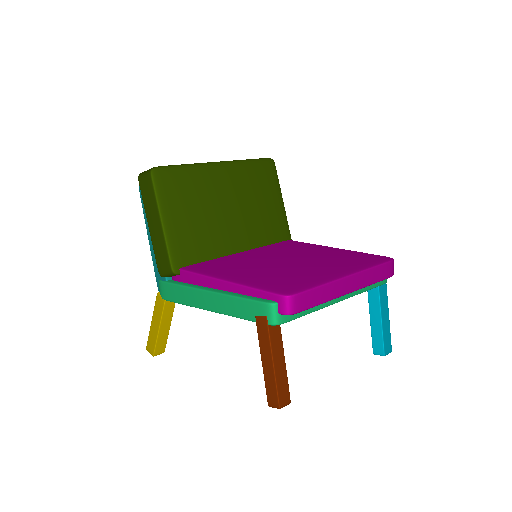}
    \end{subfigure}%
    \hfill%
    \begin{subfigure}[b]{0.20\linewidth}
		\centering
		\includegraphics[width=0.8\linewidth]{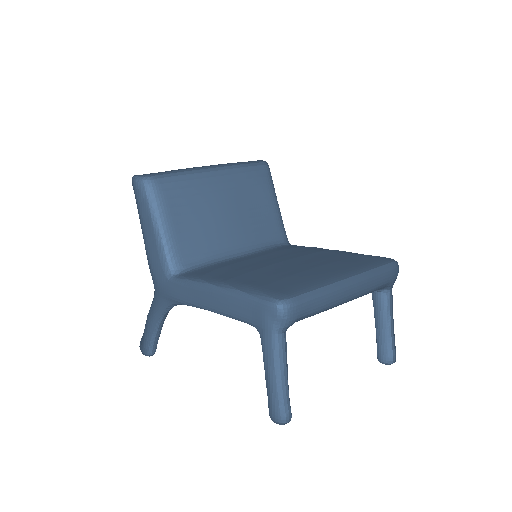}
    \end{subfigure}%
    \vskip\baselineskip%
    \vspace{-1.25em}
    \begin{subfigure}[b]{0.20\linewidth}
		\centering
		\includegraphics[width=0.8\linewidth]{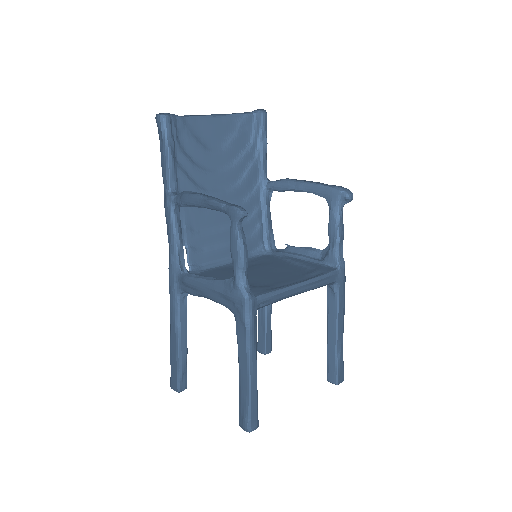}
    \end{subfigure}%
    \hfill%
    \begin{subfigure}[b]{0.20\linewidth}
		\centering
		\includegraphics[width=0.8\linewidth]{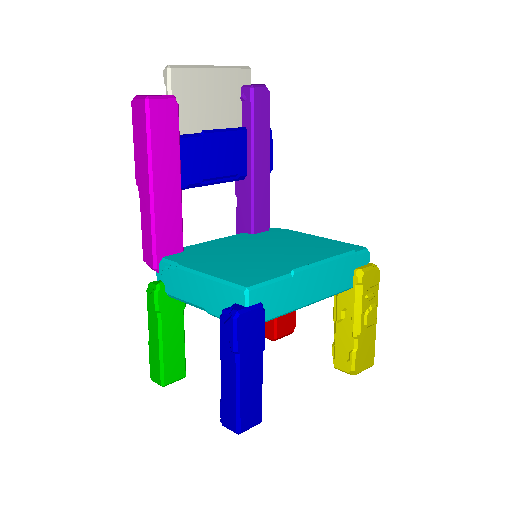}
    \end{subfigure}%
    \hfill%
    \begin{subfigure}[b]{0.20\linewidth}
		\centering
		\includegraphics[width=0.8\linewidth]{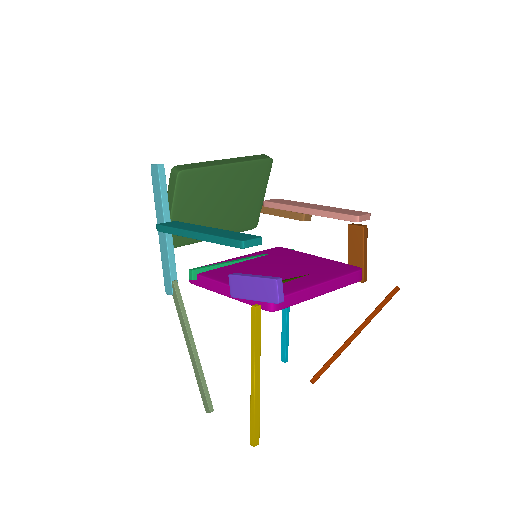}
    \end{subfigure}%
    \hfill%
    \begin{subfigure}[b]{0.20\linewidth}
		\centering
		\includegraphics[width=0.8\linewidth]{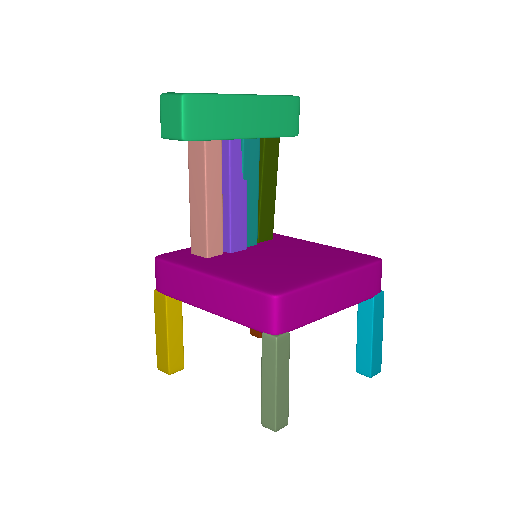}
    \end{subfigure}%
    \hfill%
    \begin{subfigure}[b]{0.20\linewidth}
		\centering
		\includegraphics[width=0.8\linewidth]{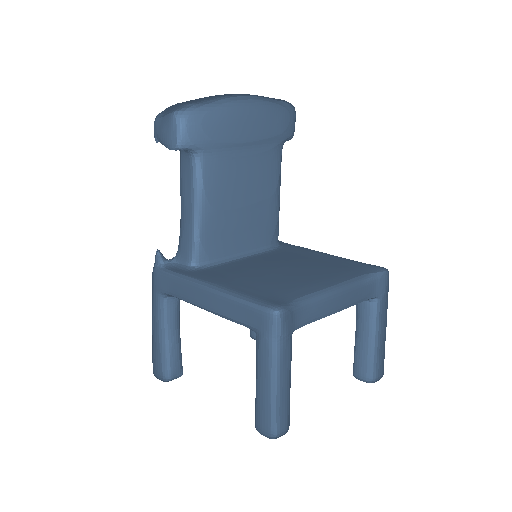}
    \end{subfigure}%
    \vskip\baselineskip%
   \vspace{-1.55em}
    \begin{subfigure}[b]{0.20\linewidth}
		\centering
		\includegraphics[width=0.8\linewidth]{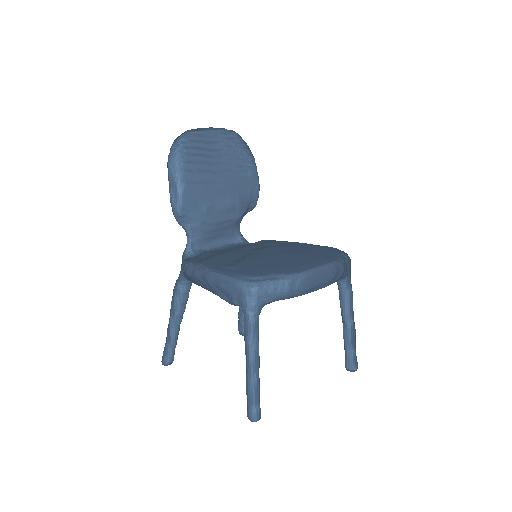}
    \end{subfigure}%
    \hfill%
    \begin{subfigure}[b]{0.20\linewidth}
		\centering
		\includegraphics[width=0.8\linewidth]{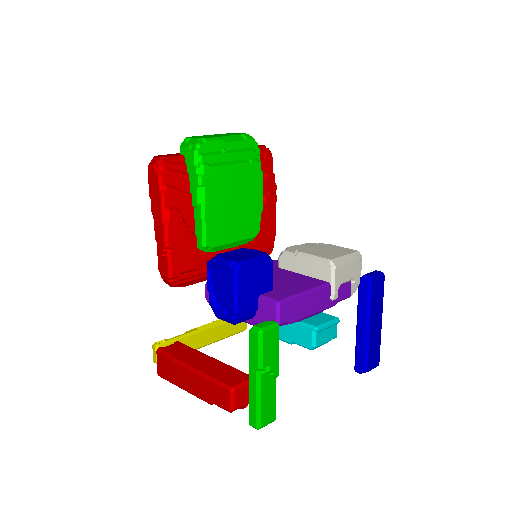}
    \end{subfigure}%
    \hfill%
    \begin{subfigure}[b]{0.20\linewidth}
		\centering
		\includegraphics[width=0.8\linewidth]{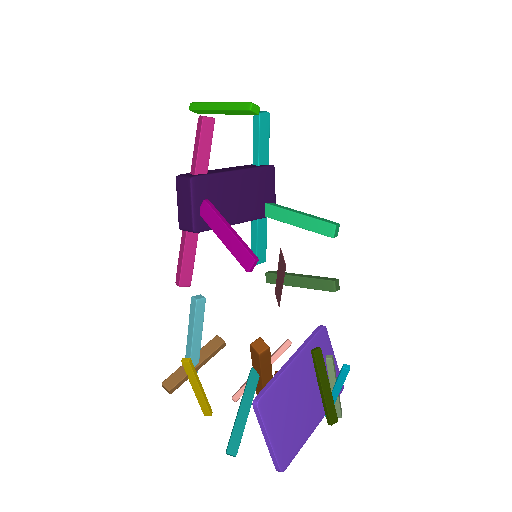}
    \end{subfigure}%
    \hfill%
    \begin{subfigure}[b]{0.20\linewidth}
		\centering
		\includegraphics[width=0.8\linewidth]{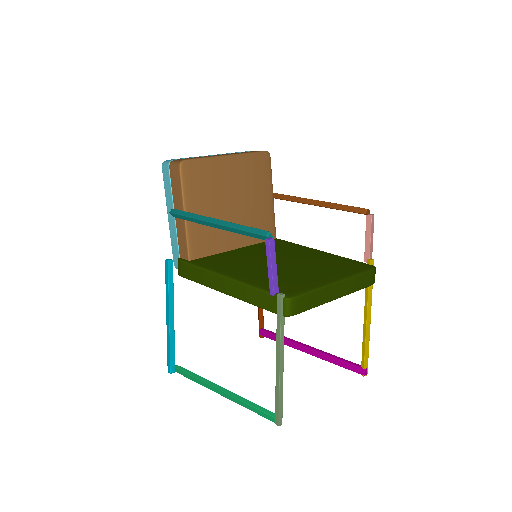}
    \end{subfigure}%
    \hfill%
    \begin{subfigure}[b]{0.20\linewidth}
		\centering
		\includegraphics[width=0.8\linewidth]{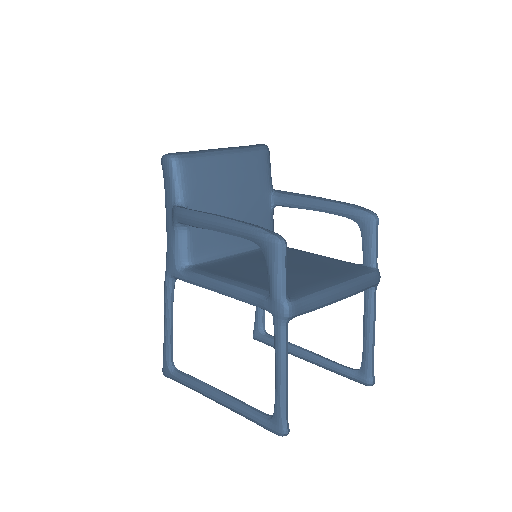}
    \end{subfigure}%
   \vskip\baselineskip%
   \vspace{-1.55em}
    \begin{subfigure}[b]{0.20\linewidth}
		\centering
		\includegraphics[width=0.8\linewidth]{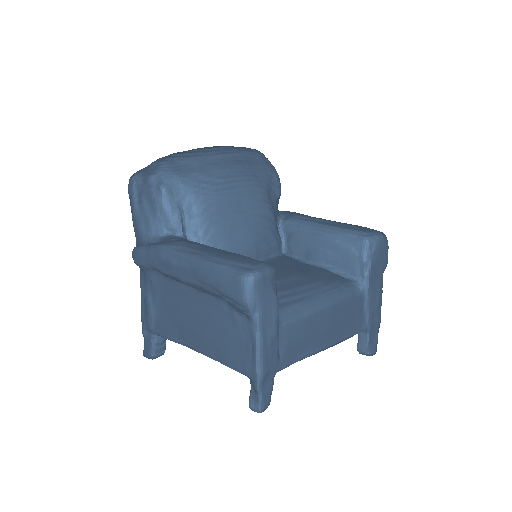}
    \end{subfigure}%
    \hfill%
    \begin{subfigure}[b]{0.20\linewidth}
		\centering
		\includegraphics[width=0.8\linewidth]{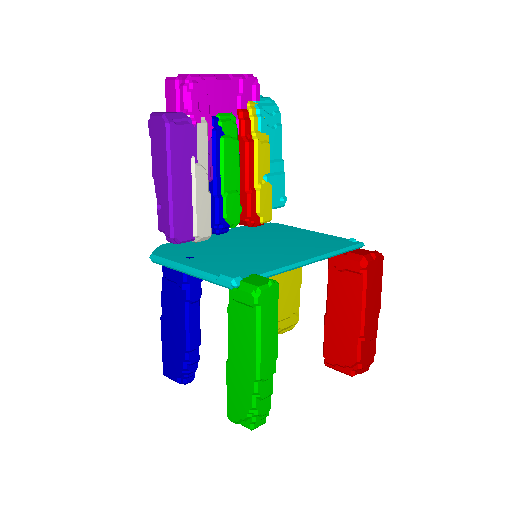}
    \end{subfigure}%
    \hfill%
    \begin{subfigure}[b]{0.20\linewidth}
		\centering
		\includegraphics[width=0.8\linewidth]{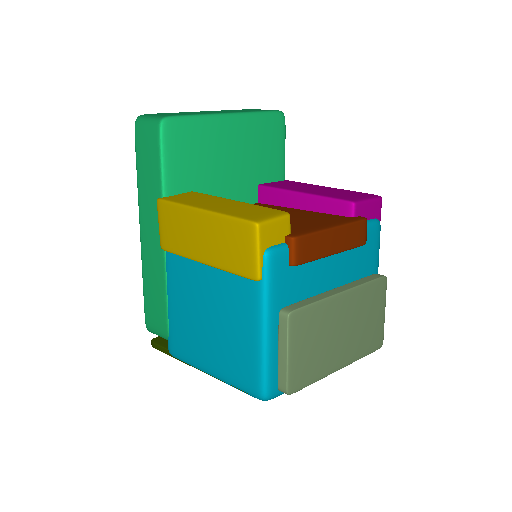}
    \end{subfigure}%
    \hfill%
    \begin{subfigure}[b]{0.20\linewidth}
		\centering
		\includegraphics[width=0.8\linewidth]{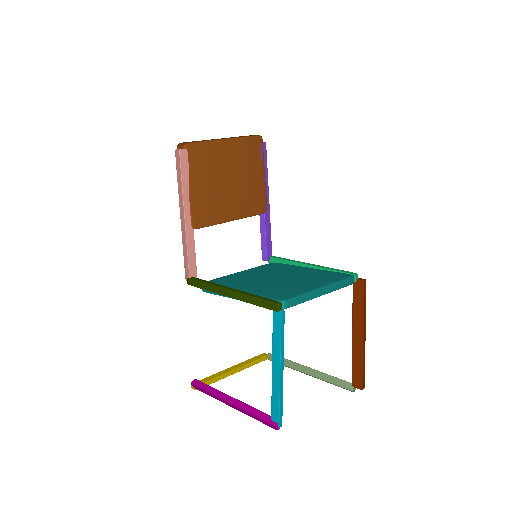}
    \end{subfigure}%
    \hfill%
    \begin{subfigure}[b]{0.20\linewidth}
		\centering
		\includegraphics[width=0.8\linewidth]{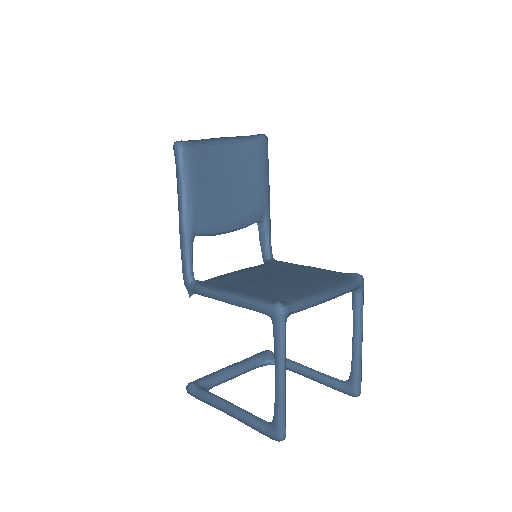}
    \end{subfigure}%
    \vskip\baselineskip%
    \vspace{-0.75em}
    \begin{subfigure}[b]{0.20\linewidth}
		\centering
		\includegraphics[width=0.8\linewidth]{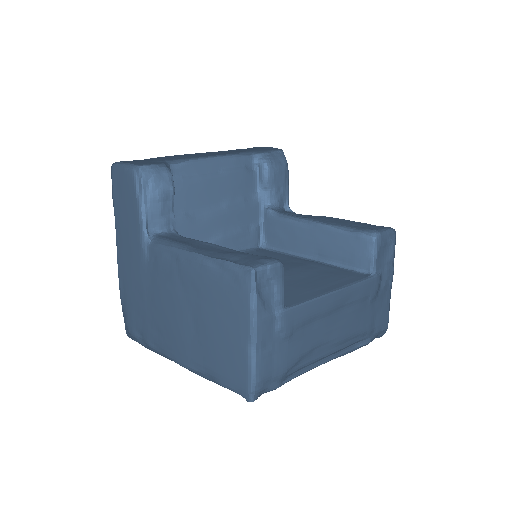}
    \end{subfigure}%
    \hfill%
    \begin{subfigure}[b]{0.20\linewidth}
		\centering
		\includegraphics[width=0.8\linewidth]{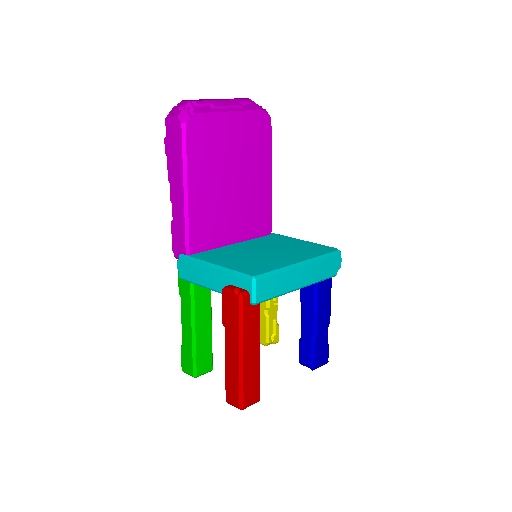}
    \end{subfigure}%
    \hfill%
    \begin{subfigure}[b]{0.20\linewidth}
		\centering
		\includegraphics[width=0.8\linewidth]{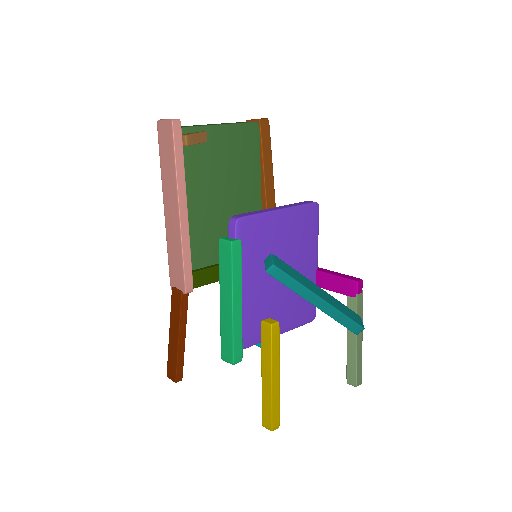}
    \end{subfigure}%
    \hfill%
    \begin{subfigure}[b]{0.20\linewidth}
		\centering
		\includegraphics[width=0.8\linewidth]{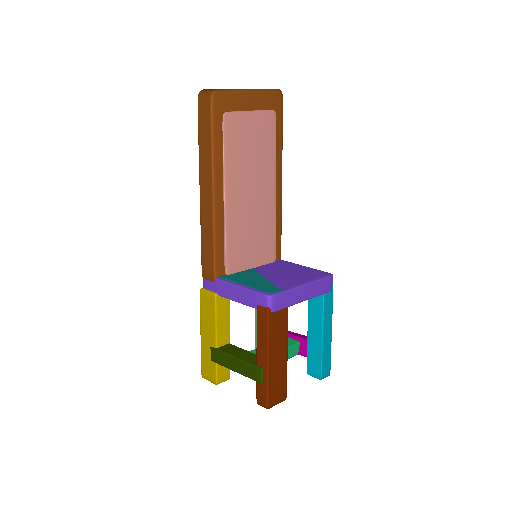}
    \end{subfigure}%
    \hfill%
    \begin{subfigure}[b]{0.20\linewidth}
		\centering
		\includegraphics[width=0.8\linewidth]{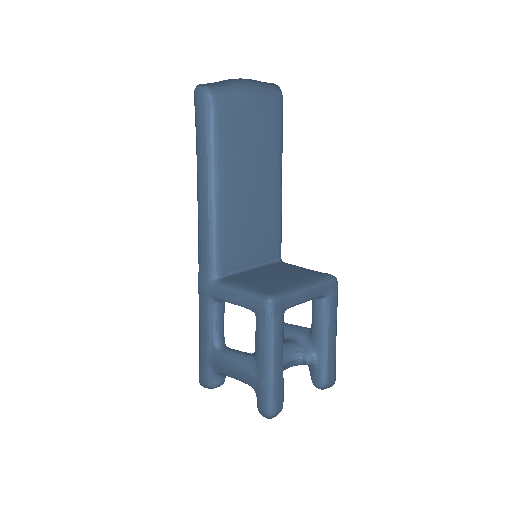}
    \end{subfigure}%
    \vskip\baselineskip%
    \vspace{-1.25em}
    \begin{subfigure}[b]{0.20\linewidth}
		\centering
		\includegraphics[width=0.8\linewidth]{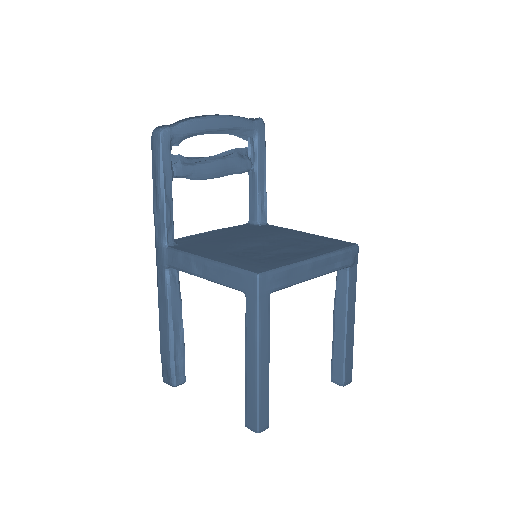}
    \end{subfigure}%
    \hfill%
    \begin{subfigure}[b]{0.20\linewidth}
		\centering
		\includegraphics[width=0.8\linewidth]{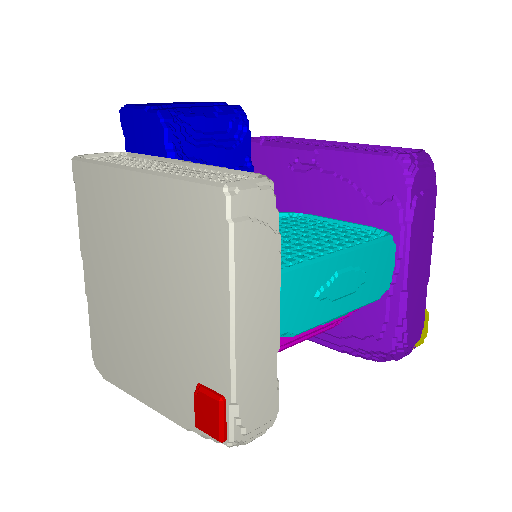}
    \end{subfigure}%
    \hfill%
    \begin{subfigure}[b]{0.20\linewidth}
		\centering
		\includegraphics[width=0.8\linewidth]{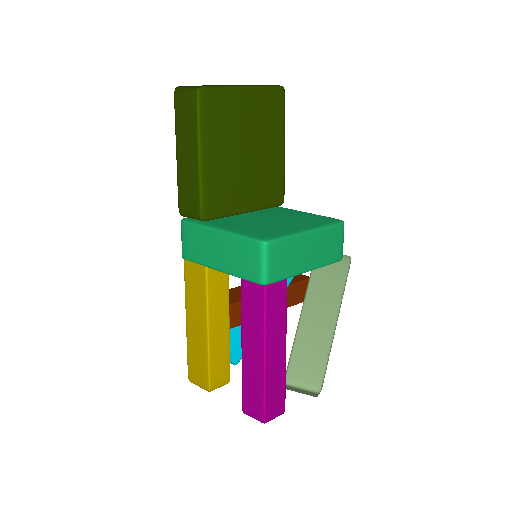}
    \end{subfigure}%
    \hfill%
    \begin{subfigure}[b]{0.20\linewidth}
		\centering
		\includegraphics[width=0.8\linewidth]{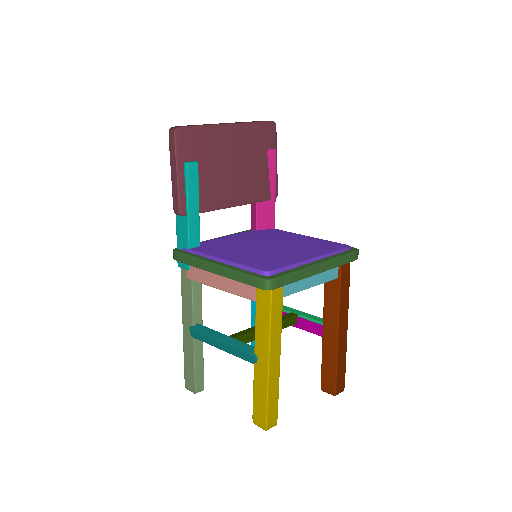}
    \end{subfigure}%
    \hfill%
    \begin{subfigure}[b]{0.20\linewidth}
		\centering
		\includegraphics[width=0.8\linewidth]{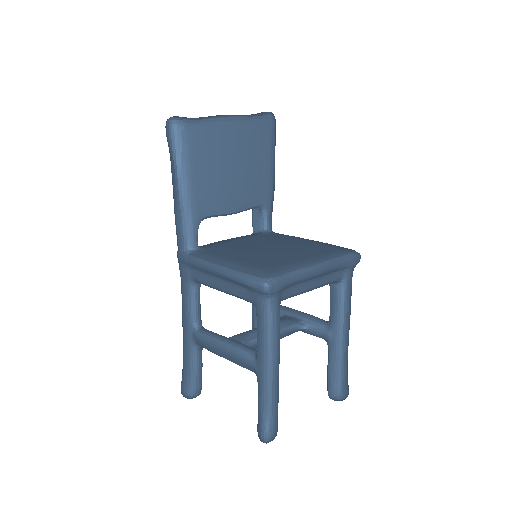}
    \end{subfigure}%
    \vskip\baselineskip%
    \vspace{-1.5em}
    \caption{{\bf Shape Generation Results on Chairs}. We show
    randomly generated chairs using our model,
    ATISS~\cite{Paschalidou2021NEURIPS}, PQ-NET~\cite{Wu2020CVPR} and
    IM-NET~\cite{Chen2019CVPR}.}
    \label{fig:shapenet_qualitative_comparison_chairs_supp}
\end{figure*}

%% file: fig/shape_generation_qualitative_tables_supp.tex
\begin{figure*}
    \begin{subfigure}[t]{\linewidth}
    \centering
    \begin{subfigure}[b]{0.20\linewidth}
        \centering
	    IM-NET
    \end{subfigure}%
    \hfill%
    \begin{subfigure}[b]{0.20\linewidth}
	\centering
        PQ-NET
    \end{subfigure}%
    \hfill%
    \begin{subfigure}[b]{0.20\linewidth}
	\centering
        ATISS
    \end{subfigure}%
    \hfill%
    \begin{subfigure}[b]{0.20\linewidth}
        \centering
        Ours-Parts
    \end{subfigure}%
    \hfill%
    \begin{subfigure}[b]{0.20\linewidth}
        \centering
        Ours
    \end{subfigure}
    \end{subfigure}
    \vspace{-1.5em}
    \begin{subfigure}[b]{0.20\linewidth}
		\centering
		\includegraphics[width=0.8\linewidth]{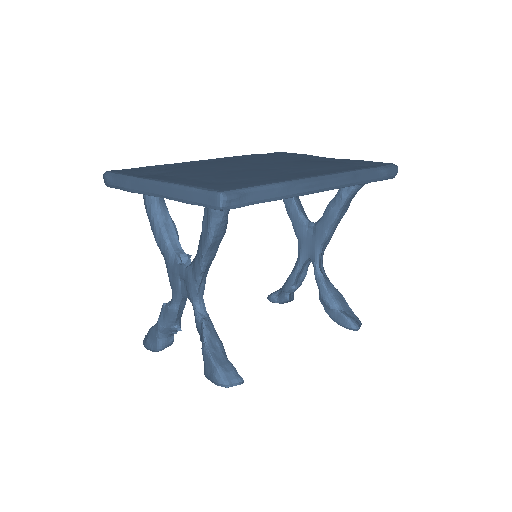}
    \end{subfigure}%
    \hfill%
    \begin{subfigure}[b]{0.20\linewidth}
		\centering
		\includegraphics[width=0.8\linewidth]{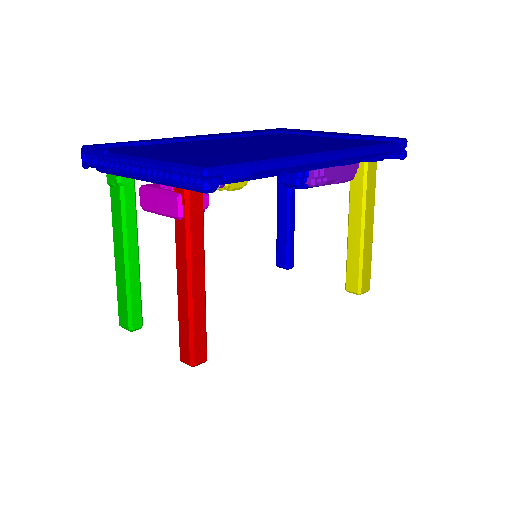}
    \end{subfigure}%
    \hfill%
    \begin{subfigure}[b]{0.20\linewidth}
		\centering
		\includegraphics[width=0.8\linewidth]{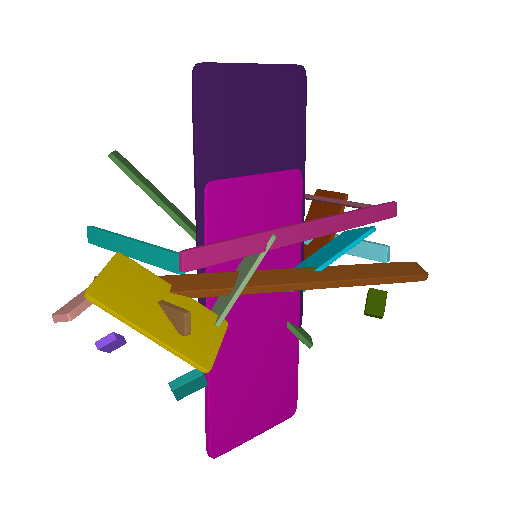}
    \end{subfigure}%
    \hfill%
    \begin{subfigure}[b]{0.20\linewidth}
		\centering
		\includegraphics[width=0.8\linewidth]{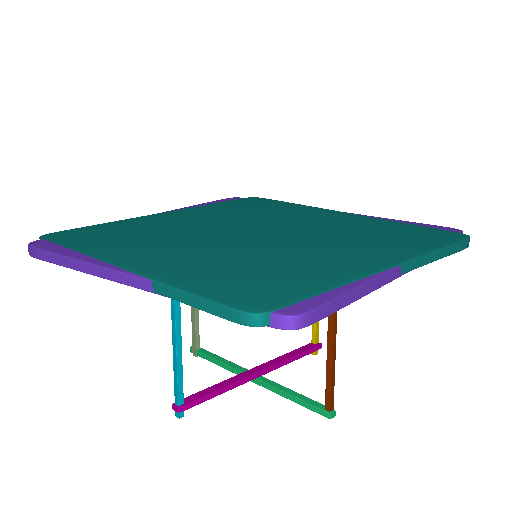}
    \end{subfigure}%
    \hfill%
    \begin{subfigure}[b]{0.20\linewidth}
		\centering
		\includegraphics[width=0.8\linewidth]{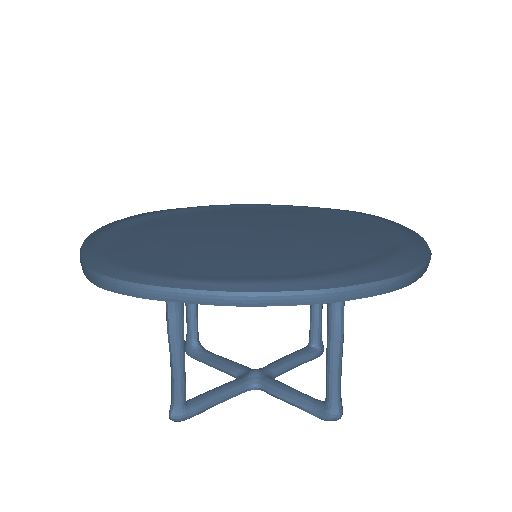}
    \end{subfigure}%
    \vskip\baselineskip%
    \vspace{-1.25em}
    \begin{subfigure}[b]{0.20\linewidth}
		\centering
		\includegraphics[width=0.8\linewidth]{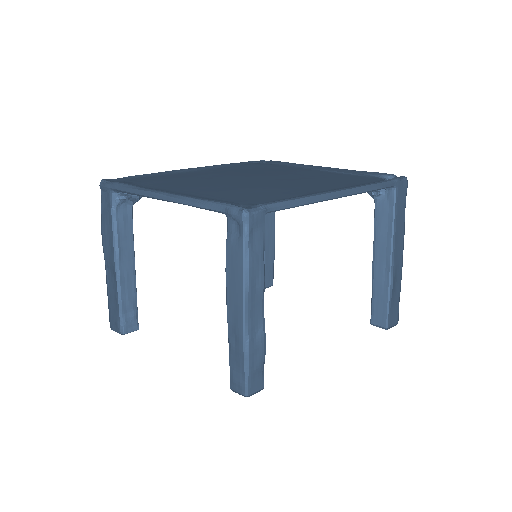}
    \end{subfigure}%
    \hfill%
    \begin{subfigure}[b]{0.20\linewidth}
		\centering
		\includegraphics[width=0.8\linewidth]{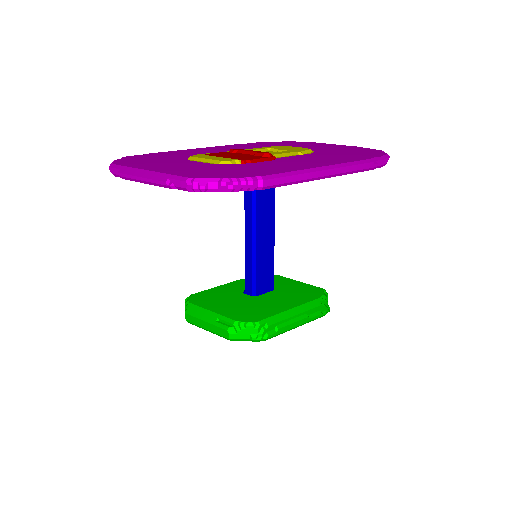}
    \end{subfigure}%
    \hfill%
    \begin{subfigure}[b]{0.20\linewidth}
		\centering
		\includegraphics[width=0.8\linewidth]{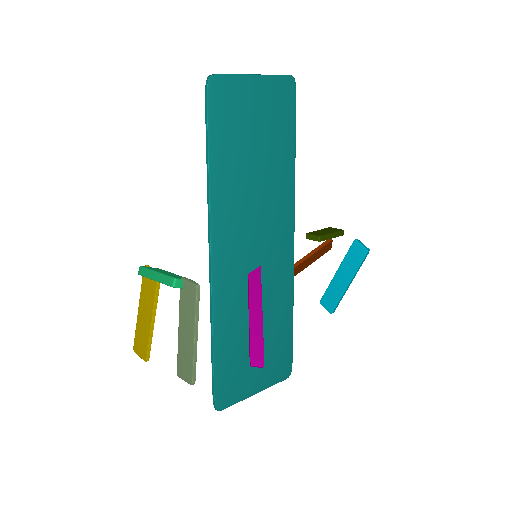}
    \end{subfigure}%
    \hfill%
    \begin{subfigure}[b]{0.20\linewidth}
		\centering
		\includegraphics[width=0.8\linewidth]{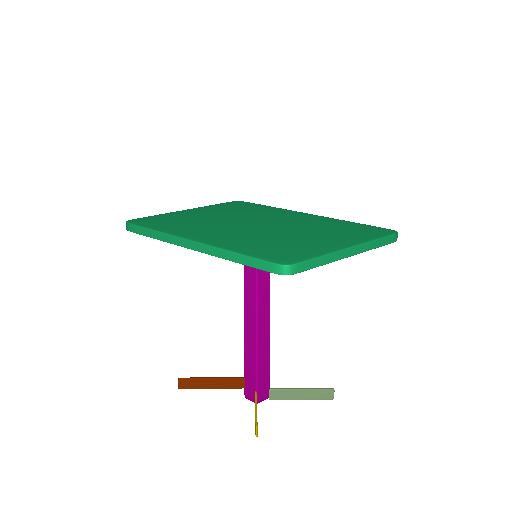}
    \end{subfigure}%
    \hfill%
    \begin{subfigure}[b]{0.20\linewidth}
		\centering
		\includegraphics[width=0.8\linewidth]{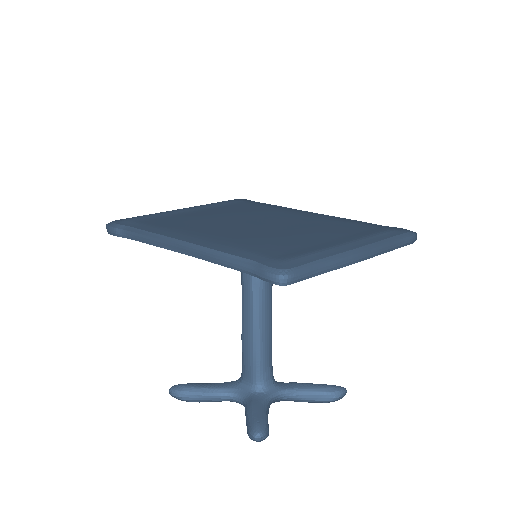}
    \end{subfigure}%
    \vskip\baselineskip%
    \vspace{-1.25em}
    \begin{subfigure}[b]{0.20\linewidth}
		\centering
		\includegraphics[width=0.8\linewidth]{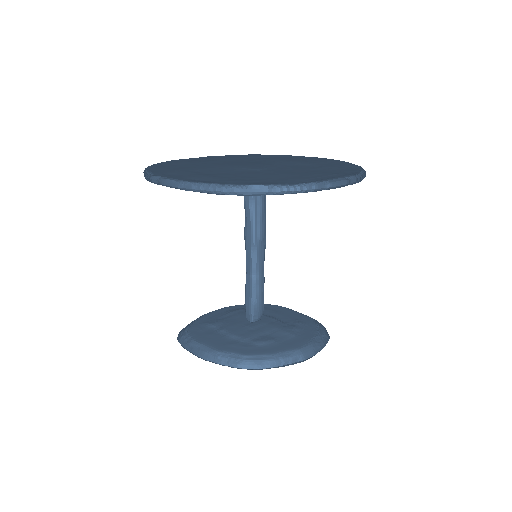}
    \end{subfigure}%
    \hfill%
    \begin{subfigure}[b]{0.20\linewidth}
		\centering
		\includegraphics[width=0.8\linewidth]{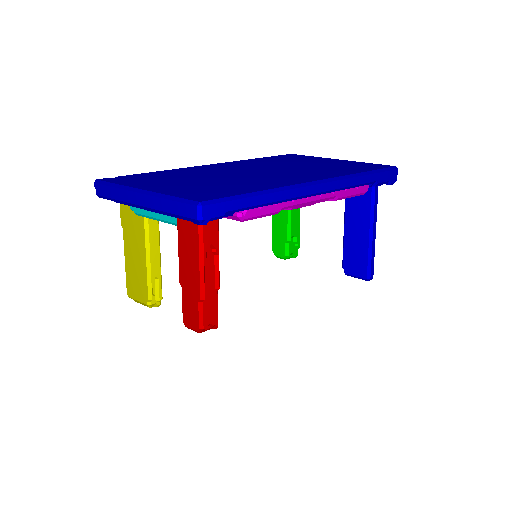}
    \end{subfigure}%
    \hfill%
    \begin{subfigure}[b]{0.20\linewidth}
		\centering
		\includegraphics[width=0.8\linewidth]{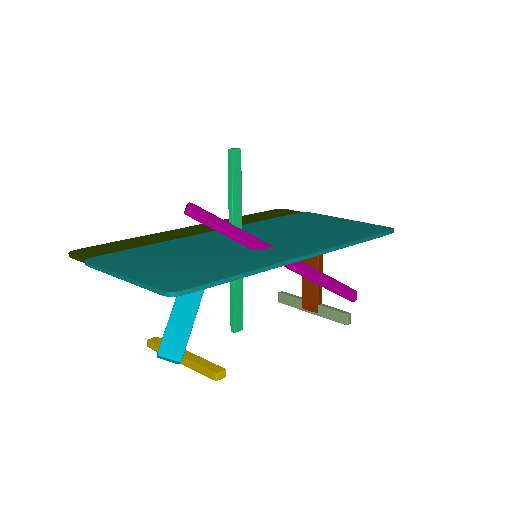}
    \end{subfigure}%
    \hfill%
    \begin{subfigure}[b]{0.20\linewidth}
		\centering
		\includegraphics[width=0.8\linewidth]{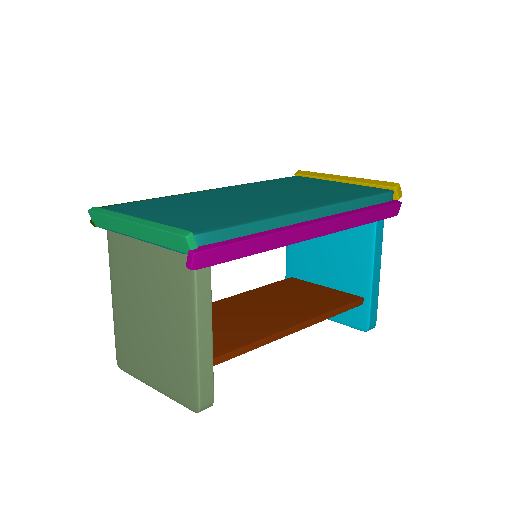}
    \end{subfigure}%
    \hfill%
    \begin{subfigure}[b]{0.20\linewidth}
		\centering
		\includegraphics[width=0.8\linewidth]{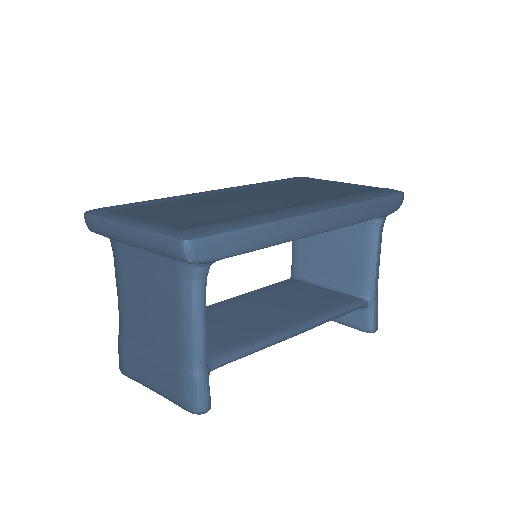}
    \end{subfigure}%
    \vskip\baselineskip%
   \vspace{-1.25em}
    \begin{subfigure}[b]{0.20\linewidth}
		\centering
		\includegraphics[width=0.8\linewidth]{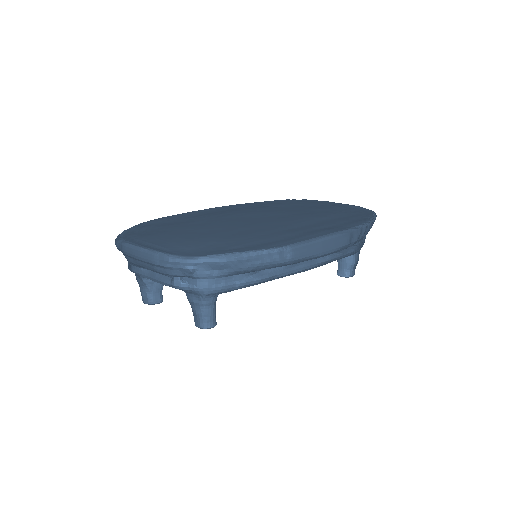}
    \end{subfigure}%
    \hfill%
    \begin{subfigure}[b]{0.20\linewidth}
		\centering
		\includegraphics[width=0.8\linewidth]{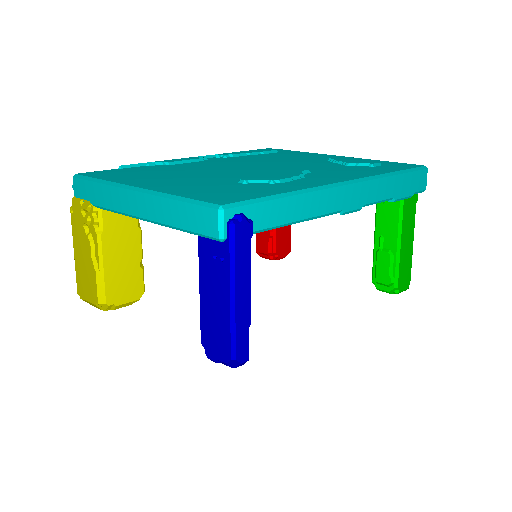}
    \end{subfigure}%
    \hfill%
    \begin{subfigure}[b]{0.20\linewidth}
		\centering
		\includegraphics[width=0.8\linewidth]{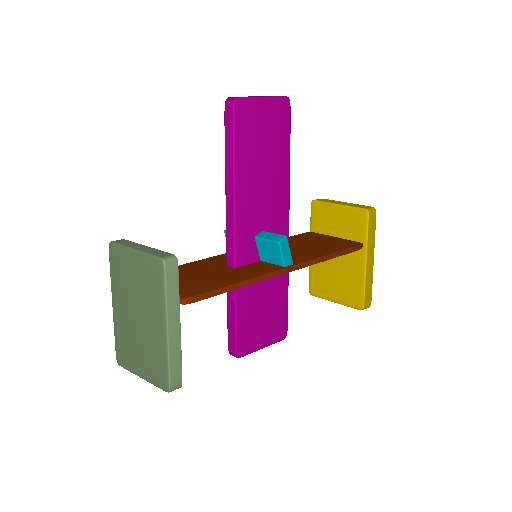}
    \end{subfigure}%
    \hfill%
    \begin{subfigure}[b]{0.20\linewidth}
		\centering
		\includegraphics[width=0.8\linewidth]{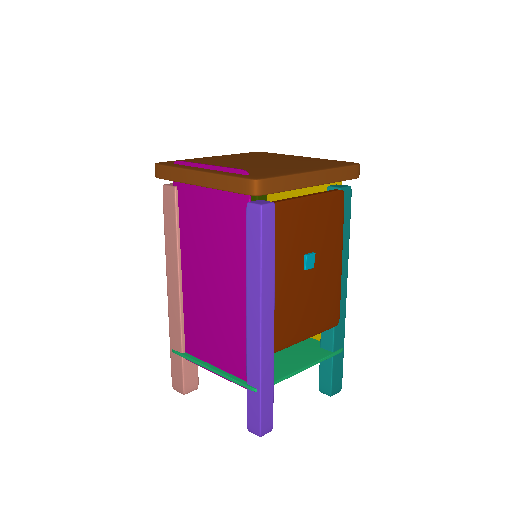}
    \end{subfigure}%
    \hfill%
    \begin{subfigure}[b]{0.20\linewidth}
		\centering
		\includegraphics[width=0.8\linewidth]{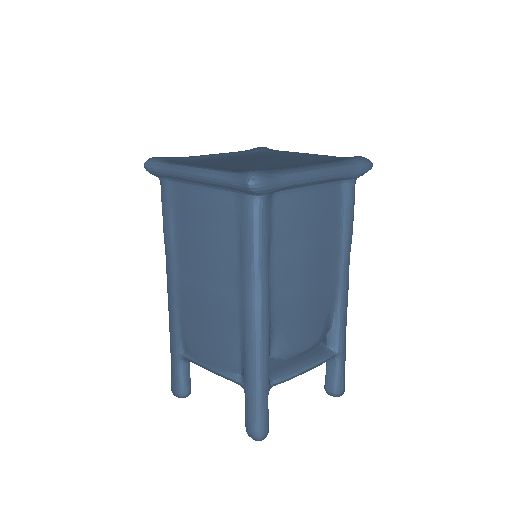}
    \end{subfigure}%
    \vskip\baselineskip%
   \vspace{-1.25em}
    \begin{subfigure}[b]{0.20\linewidth}
		\centering
		\includegraphics[width=0.8\linewidth]{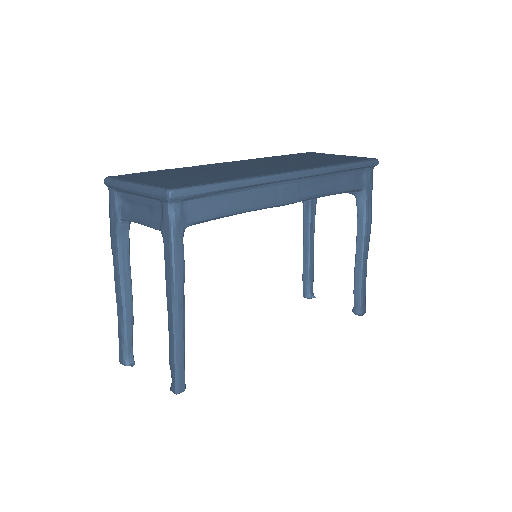}
    \end{subfigure}%
    \hfill%
    \begin{subfigure}[b]{0.20\linewidth}
		\centering
		\includegraphics[width=0.8\linewidth]{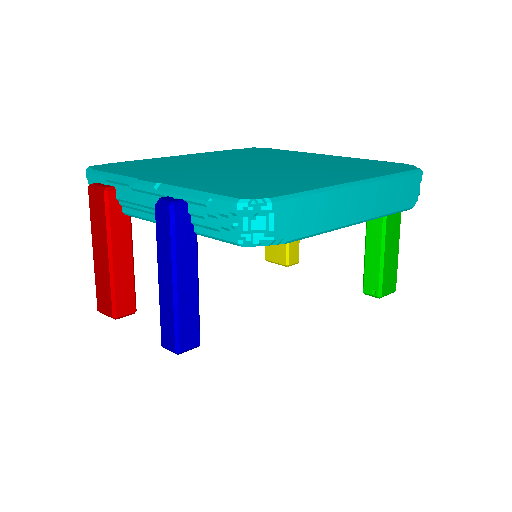}
    \end{subfigure}%
    \hfill%
    \begin{subfigure}[b]{0.20\linewidth}
		\centering
		\includegraphics[width=0.8\linewidth]{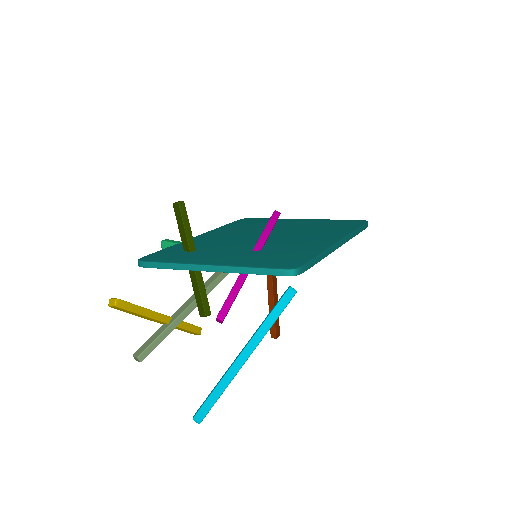}
    \end{subfigure}%
    \hfill%
    \begin{subfigure}[b]{0.20\linewidth}
		\centering
		\includegraphics[width=0.8\linewidth]{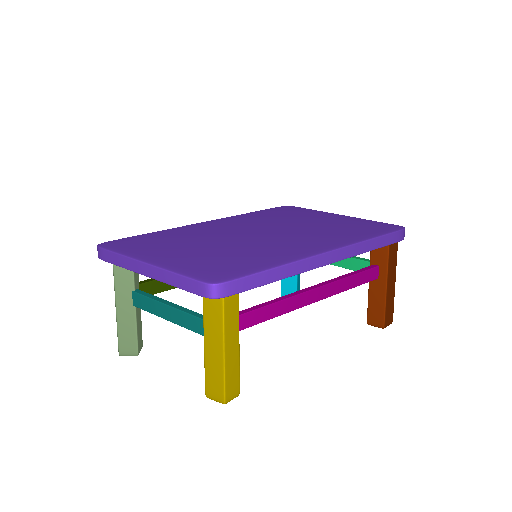}
    \end{subfigure}%
    \hfill%
    \begin{subfigure}[b]{0.20\linewidth}
		\centering
		\includegraphics[width=0.8\linewidth]{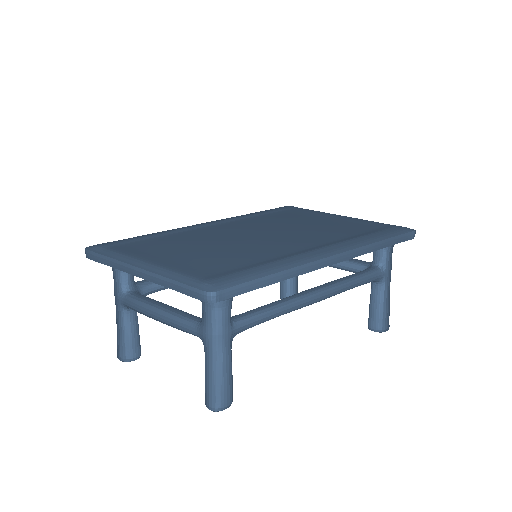}
    \end{subfigure}%
    \vskip\baselineskip%
    \vspace{-1.25em}
    \begin{subfigure}[b]{0.20\linewidth}
		\centering
		\includegraphics[width=0.8\linewidth]{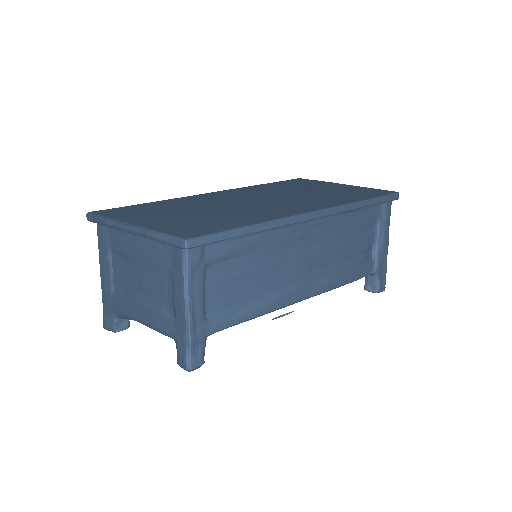}
    \end{subfigure}%
    \hfill%
    \begin{subfigure}[b]{0.20\linewidth}
		\centering
		\includegraphics[width=0.8\linewidth]{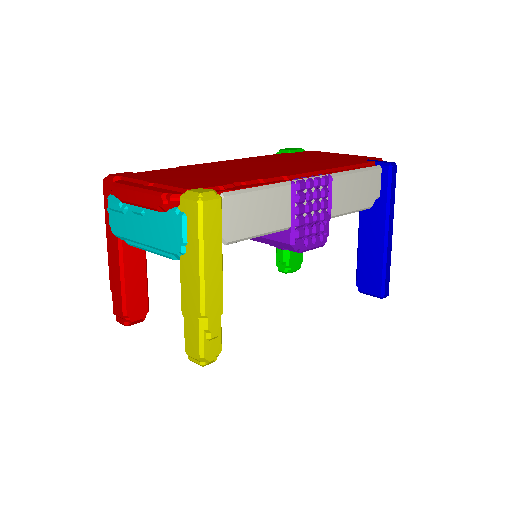}
    \end{subfigure}%
    \hfill%
    \begin{subfigure}[b]{0.20\linewidth}
		\centering
		\includegraphics[width=0.8\linewidth]{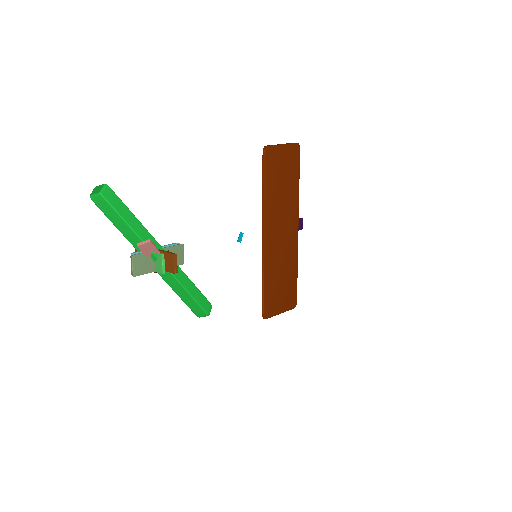}
    \end{subfigure}%
    \hfill%
    \begin{subfigure}[b]{0.20\linewidth}
		\centering
		\includegraphics[width=0.8\linewidth]{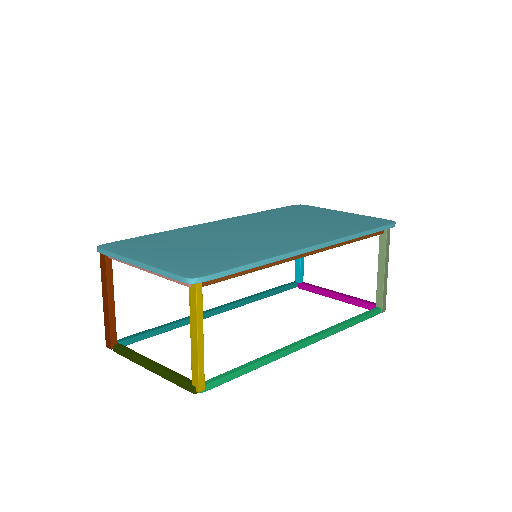}
    \end{subfigure}%
    \hfill%
    \begin{subfigure}[b]{0.20\linewidth}
		\centering
		\includegraphics[width=0.8\linewidth]{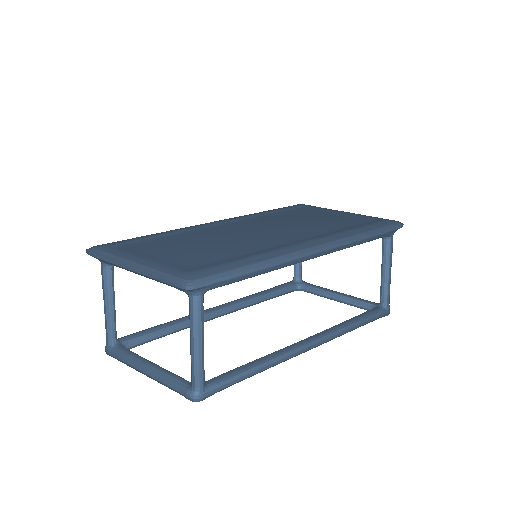}
    \end{subfigure}%
    \vskip\baselineskip%
    \vspace{-1.25em}
    \begin{subfigure}[b]{0.20\linewidth}
		\centering
		\includegraphics[width=0.8\linewidth]{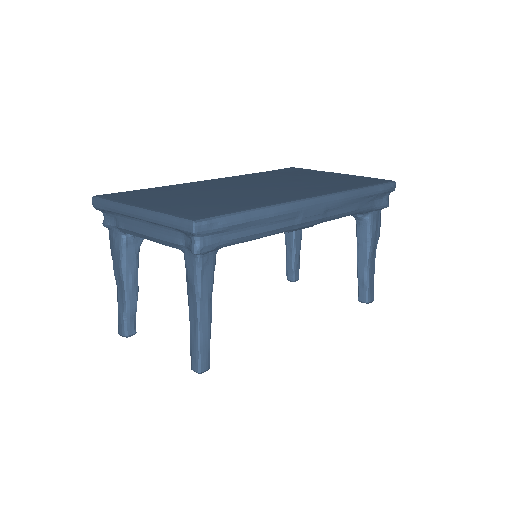}
    \end{subfigure}%
    \hfill%
    \begin{subfigure}[b]{0.20\linewidth}
		\centering
		\includegraphics[width=0.8\linewidth]{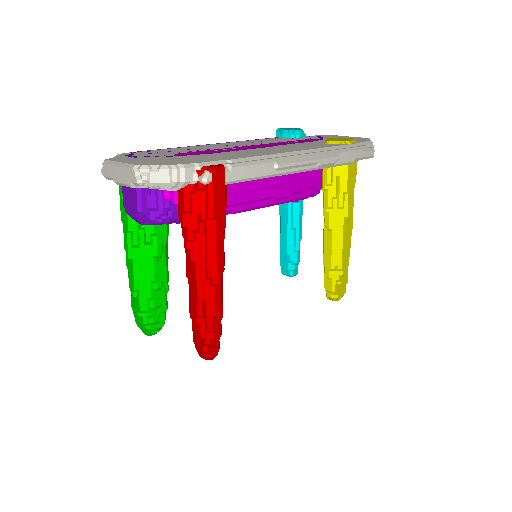}
    \end{subfigure}%
    \hfill%
    \begin{subfigure}[b]{0.20\linewidth}
		\centering
		\includegraphics[width=0.8\linewidth]{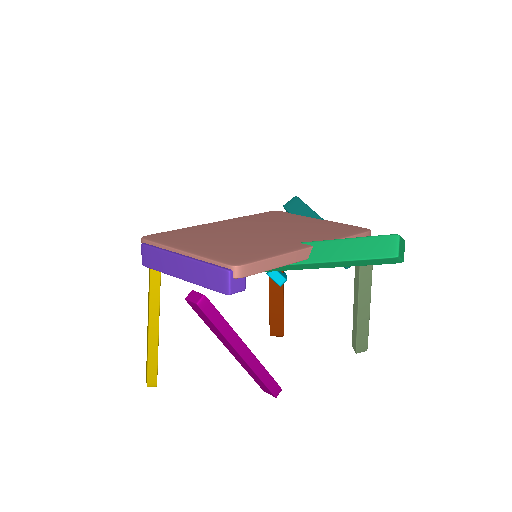}
    \end{subfigure}%
    \hfill%
    \begin{subfigure}[b]{0.20\linewidth}
		\centering
		\includegraphics[width=0.8\linewidth]{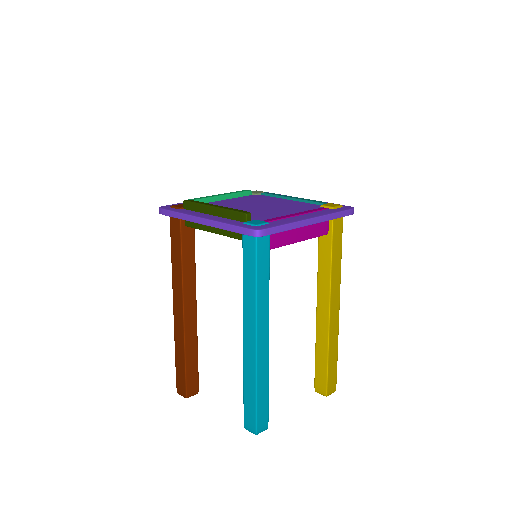}
    \end{subfigure}%
    \hfill%
    \begin{subfigure}[b]{0.20\linewidth}
		\centering
		\includegraphics[width=0.8\linewidth]{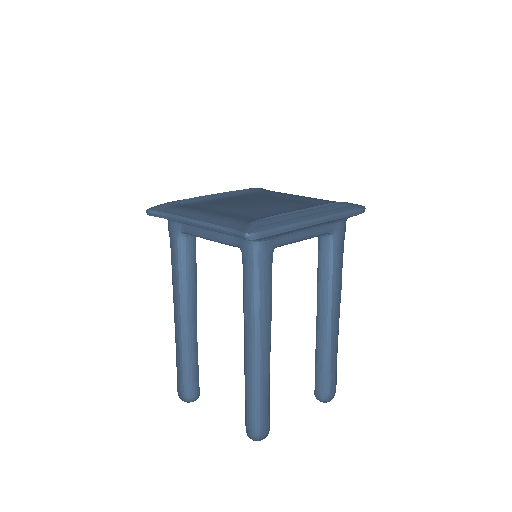}
    \end{subfigure}%
    \vskip\baselineskip%
    \vspace{-1.5em}
    \caption{{\bf Shape Generation Results on Tables}. We showcase randomly generated tables using our model, ATISS~\cite{Paschalidou2021NEURIPS}, PQ-NET~\cite{Wu2020CVPR} and IM-NET~\cite{Chen2019CVPR}.}
    \label{fig:shapenet_qualitative_comparison_tables_supp}
\end{figure*}

%% file: fig/shape_generation_qualitative_lamps_supp.tex
\begin{figure*}
    \begin{subfigure}[t]{\linewidth}
    \centering
    \begin{subfigure}[b]{0.20\linewidth}
        \centering
	    IM-NET
    \end{subfigure}%
    \hfill%
    \begin{subfigure}[b]{0.20\linewidth}
	\centering
        PQ-NET
    \end{subfigure}%
    \hfill%
    \begin{subfigure}[b]{0.20\linewidth}
	\centering
        ATISS
    \end{subfigure}%
    \hfill%
    \begin{subfigure}[b]{0.20\linewidth}
        \centering
        Ours-Parts
    \end{subfigure}%
    \hfill%
    \begin{subfigure}[b]{0.20\linewidth}
        \centering
        Ours
    \end{subfigure}
    \end{subfigure}
    \vspace{-1.5em}
    \begin{subfigure}[b]{0.20\linewidth}
		\centering
		\includegraphics[width=0.8\linewidth]{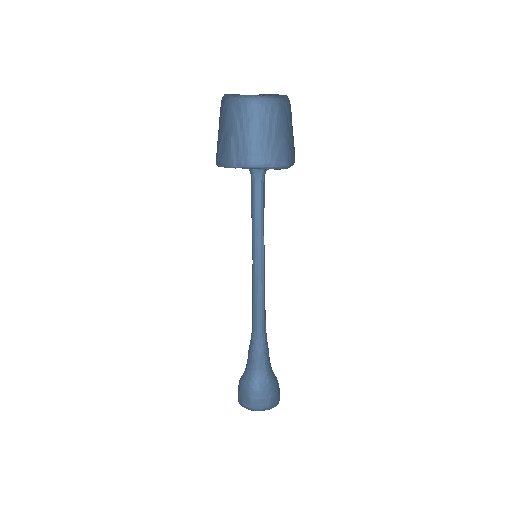}
    \end{subfigure}%
    \hfill%
    \begin{subfigure}[b]{0.20\linewidth}
		\centering
		\includegraphics[width=0.8\linewidth]{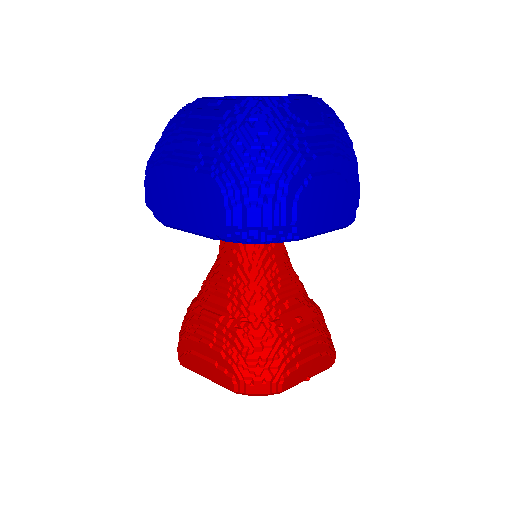}
    \end{subfigure}%
    \hfill%
    \begin{subfigure}[b]{0.20\linewidth}
		\centering
		\includegraphics[width=0.8\linewidth]{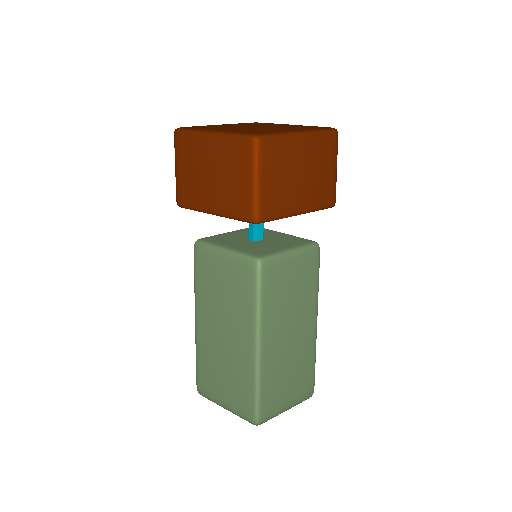}
    \end{subfigure}%
    \hfill%
    \begin{subfigure}[b]{0.20\linewidth}
		\centering
		\includegraphics[width=0.8\linewidth]{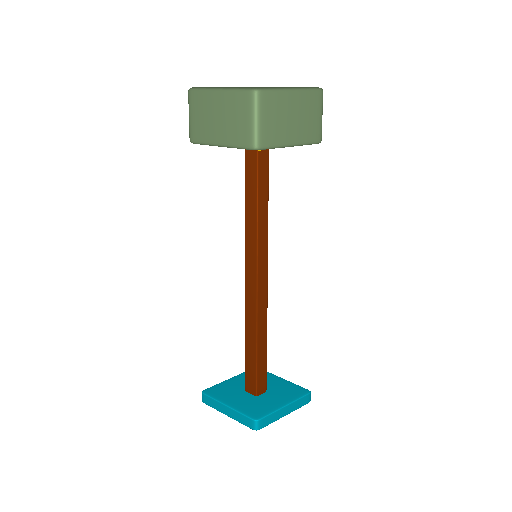}
    \end{subfigure}%
    \hfill%
    \begin{subfigure}[b]{0.20\linewidth}
		\centering
		\includegraphics[width=0.8\linewidth]{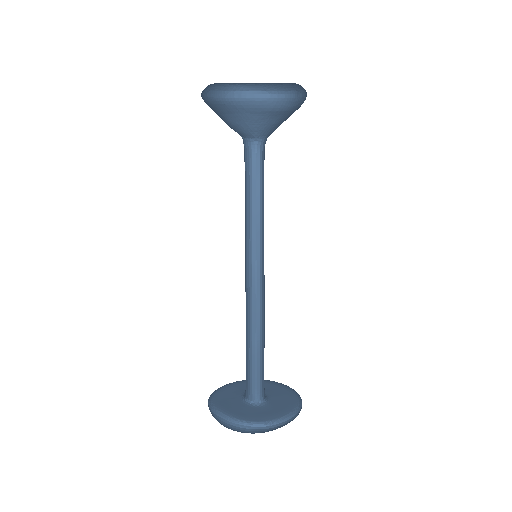}
    \end{subfigure}%
    \vskip\baselineskip%
    \vspace{-1.25em}
    \begin{subfigure}[b]{0.20\linewidth}
		\centering
		\includegraphics[width=0.8\linewidth]{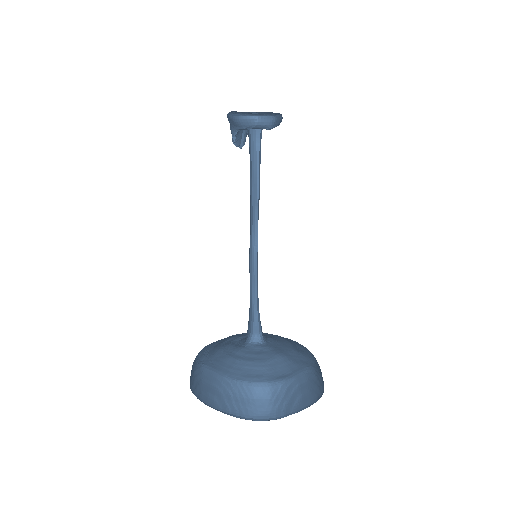}
    \end{subfigure}%
    \hfill%
    \begin{subfigure}[b]{0.20\linewidth}
		\centering
		\includegraphics[width=0.8\linewidth]{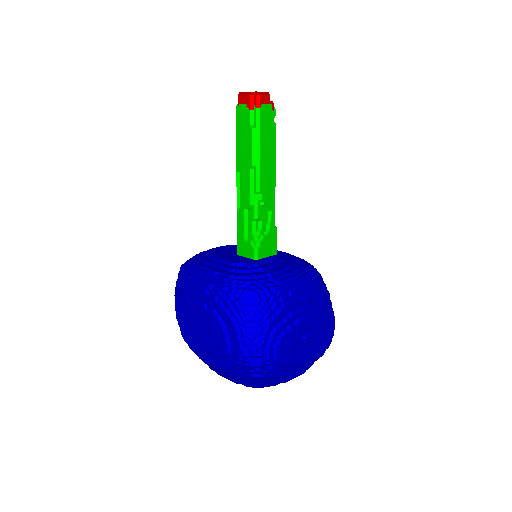}
    \end{subfigure}%
    \hfill%
    \begin{subfigure}[b]{0.20\linewidth}
		\centering
		\includegraphics[width=0.8\linewidth]{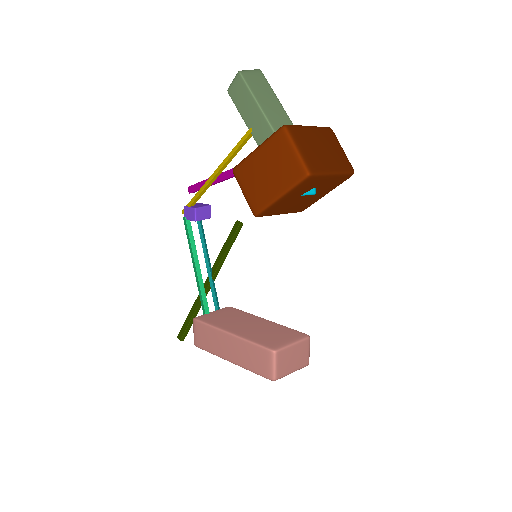}
    \end{subfigure}%
    \hfill%
    \begin{subfigure}[b]{0.20\linewidth}
		\centering
		\includegraphics[width=0.8\linewidth]{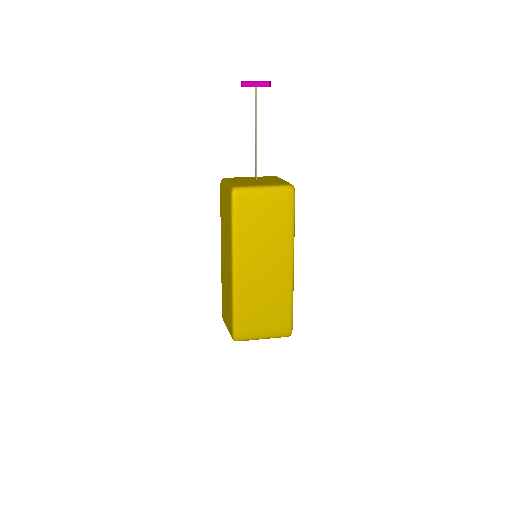}
    \end{subfigure}%
    \hfill%
    \begin{subfigure}[b]{0.20\linewidth}
		\centering
		\includegraphics[width=0.8\linewidth]{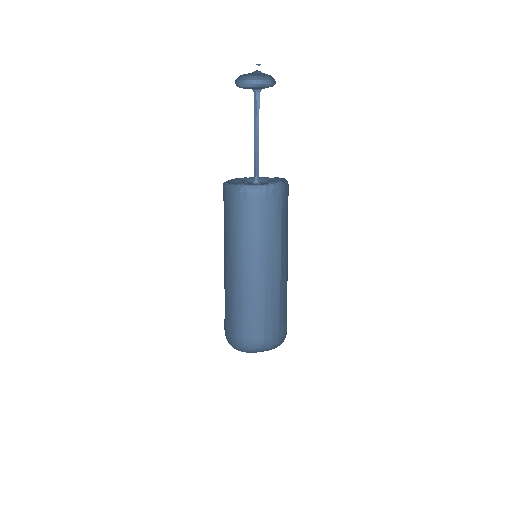}
    \end{subfigure}%
    \vskip\baselineskip%
    \vspace{-0.5em}
    \begin{subfigure}[b]{0.20\linewidth}
		\centering
		\includegraphics[width=0.8\linewidth]{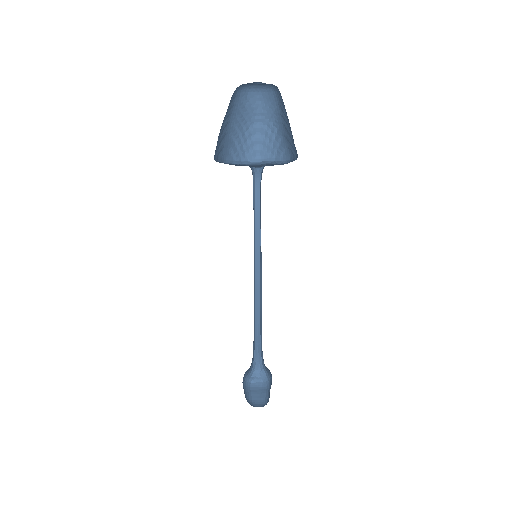}
    \end{subfigure}%
    \hfill%
    \begin{subfigure}[b]{0.20\linewidth}
		\centering
		\includegraphics[width=0.8\linewidth]{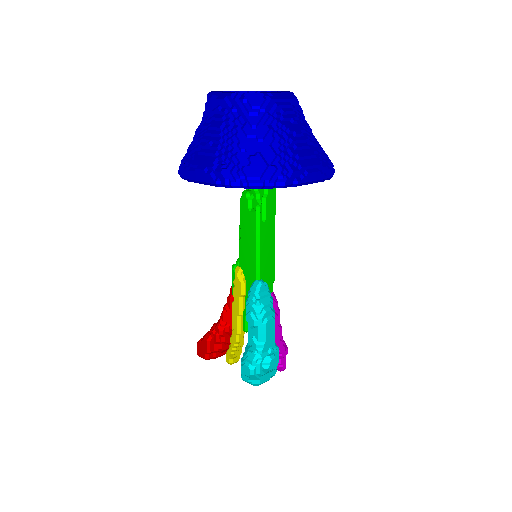}
    \end{subfigure}%
    \hfill%
    \begin{subfigure}[b]{0.20\linewidth}
		\centering
		\includegraphics[width=0.8\linewidth]{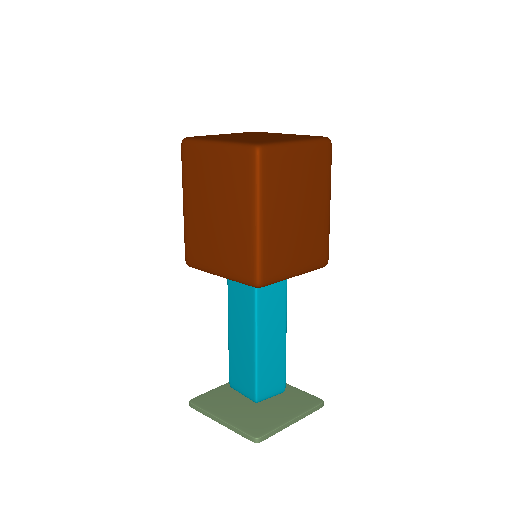}
    \end{subfigure}%
    \hfill%
    \begin{subfigure}[b]{0.20\linewidth}
		\centering
		\includegraphics[width=0.8\linewidth]{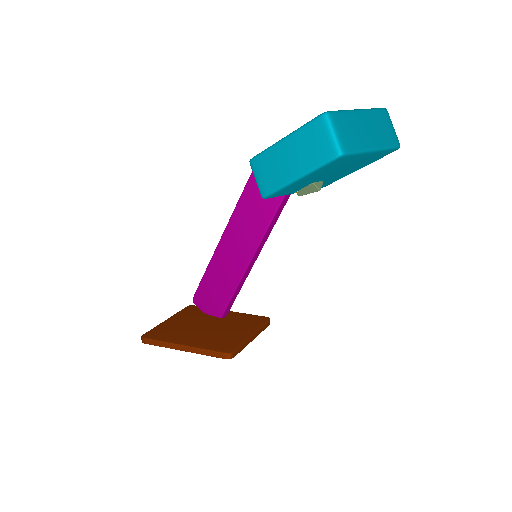}
    \end{subfigure}%
    \hfill%
    \begin{subfigure}[b]{0.20\linewidth}
		\centering
		\includegraphics[width=0.8\linewidth]{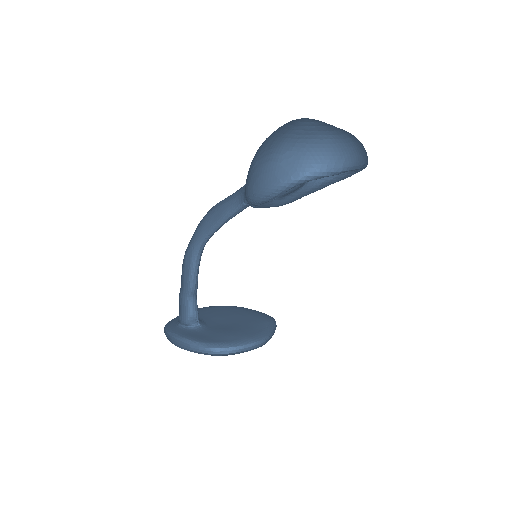}
    \end{subfigure}%
    \vskip\baselineskip%
   \vspace{-1.25em}
    \begin{subfigure}[b]{0.20\linewidth}
		\centering
		\includegraphics[width=0.8\linewidth]{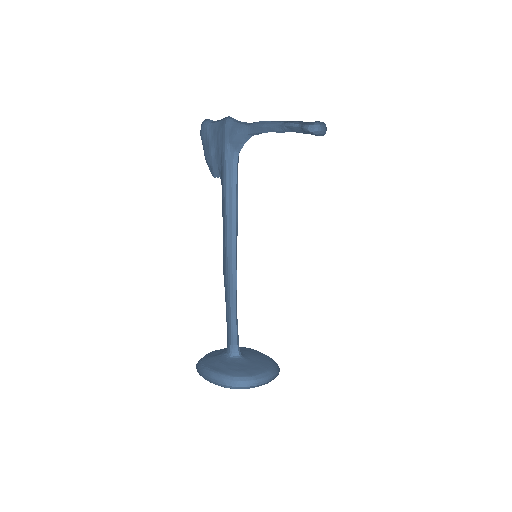}
    \end{subfigure}%
    \hfill%
    \begin{subfigure}[b]{0.20\linewidth}
		\centering
		\includegraphics[width=0.8\linewidth]{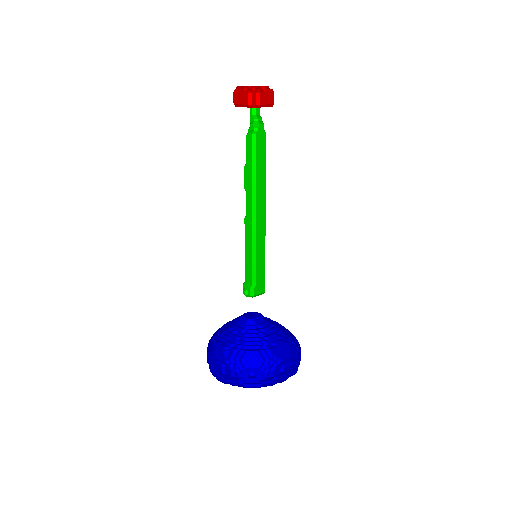}
    \end{subfigure}%
    \hfill%
    \begin{subfigure}[b]{0.20\linewidth}
		\centering
		\includegraphics[width=0.8\linewidth]{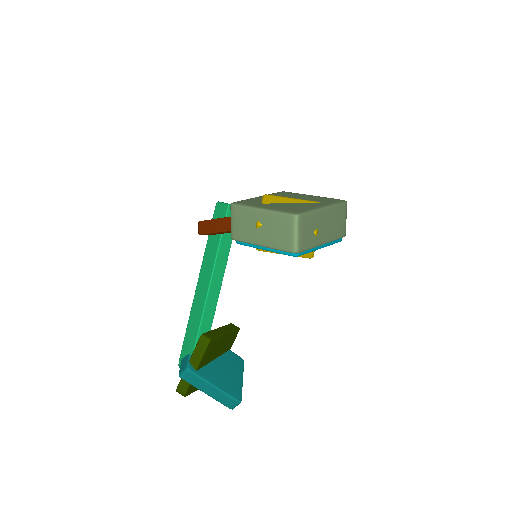}
    \end{subfigure}%
    \hfill%
    \begin{subfigure}[b]{0.20\linewidth}
		\centering
		\includegraphics[width=0.8\linewidth]{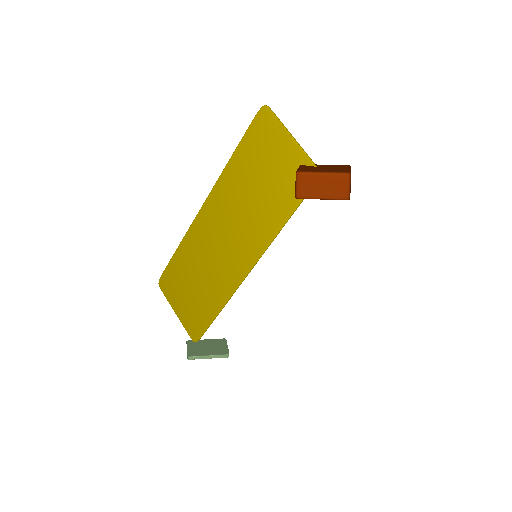}
    \end{subfigure}%
    \hfill%
    \begin{subfigure}[b]{0.20\linewidth}
		\centering
		\includegraphics[width=0.8\linewidth]{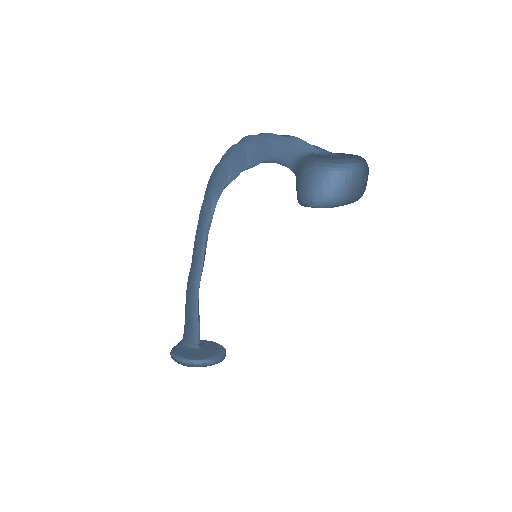}
    \end{subfigure}%
    \vskip\baselineskip%
   \vspace{-0.5em}
    \begin{subfigure}[b]{0.20\linewidth}
		\centering
		\includegraphics[width=0.8\linewidth]{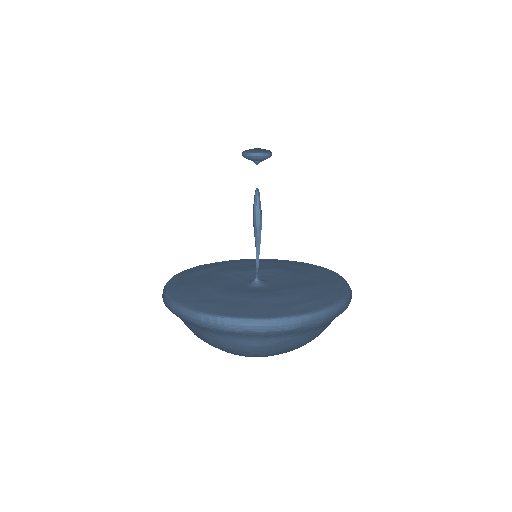}
    \end{subfigure}%
    \hfill%
    \begin{subfigure}[b]{0.20\linewidth}
		\centering
		\includegraphics[width=0.8\linewidth]{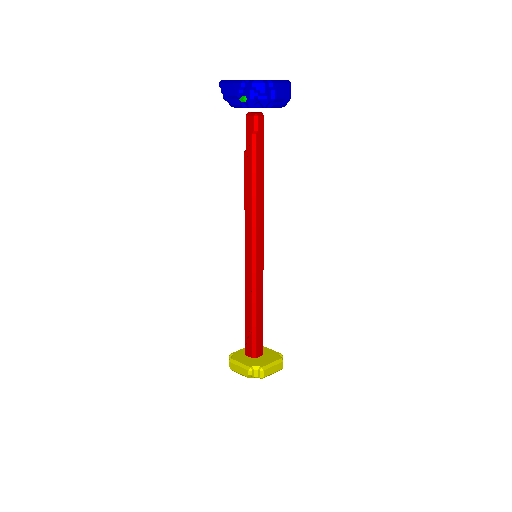}
    \end{subfigure}%
    \hfill%
    \begin{subfigure}[b]{0.20\linewidth}
		\centering
		\includegraphics[width=0.8\linewidth]{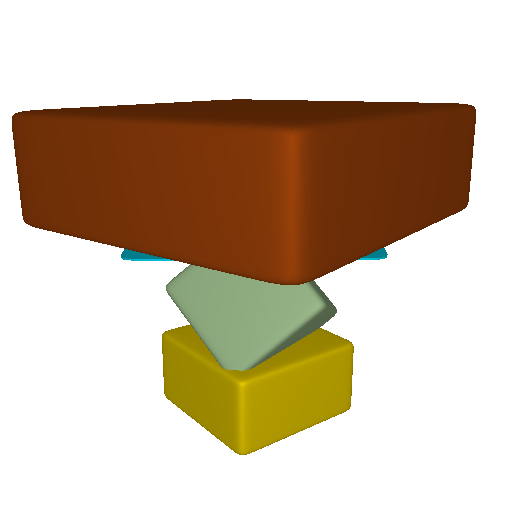}
    \end{subfigure}%
    \hfill%
    \begin{subfigure}[b]{0.20\linewidth}
		\centering
		\includegraphics[width=0.8\linewidth]{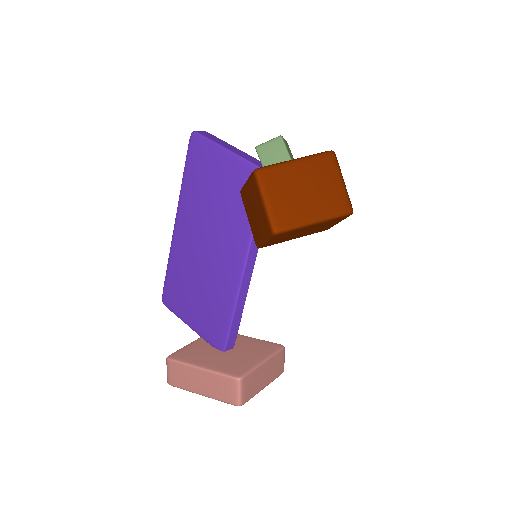}
    \end{subfigure}%
    \hfill%
    \begin{subfigure}[b]{0.20\linewidth}
		\centering
		\includegraphics[width=0.8\linewidth]{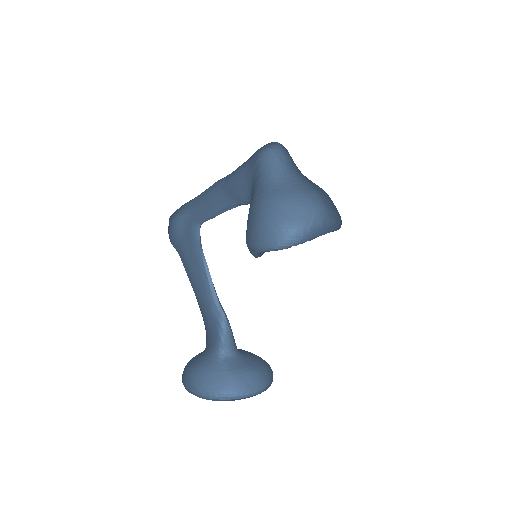}
    \end{subfigure}%
    \vskip\baselineskip%
    \vspace{-1.25em}
    \begin{subfigure}[b]{0.20\linewidth}
		\centering
		\includegraphics[width=0.8\linewidth]{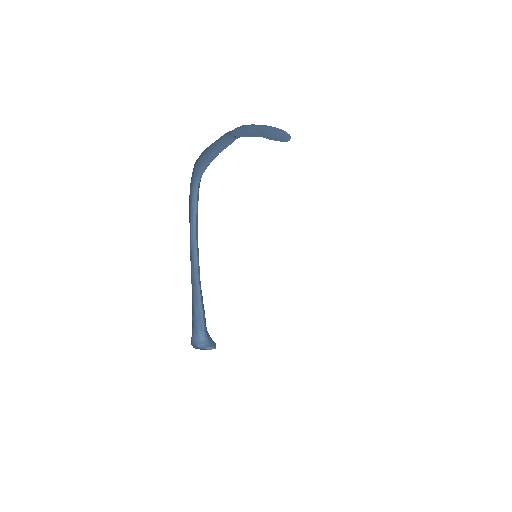}
    \end{subfigure}%
    \hfill%
    \begin{subfigure}[b]{0.20\linewidth}
		\centering
		\includegraphics[width=0.8\linewidth]{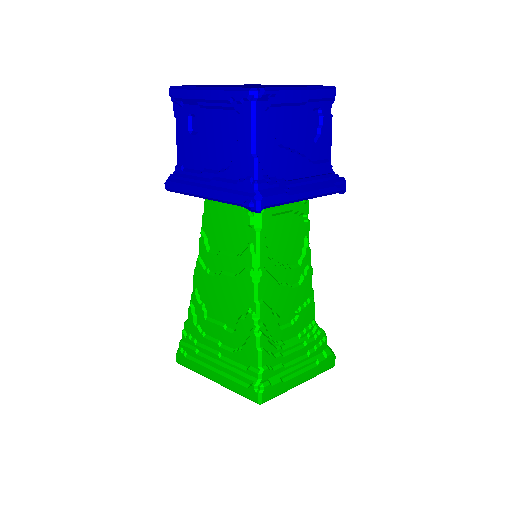}
    \end{subfigure}%
    \hfill%
    \begin{subfigure}[b]{0.20\linewidth}
		\centering
		\includegraphics[width=0.8\linewidth]{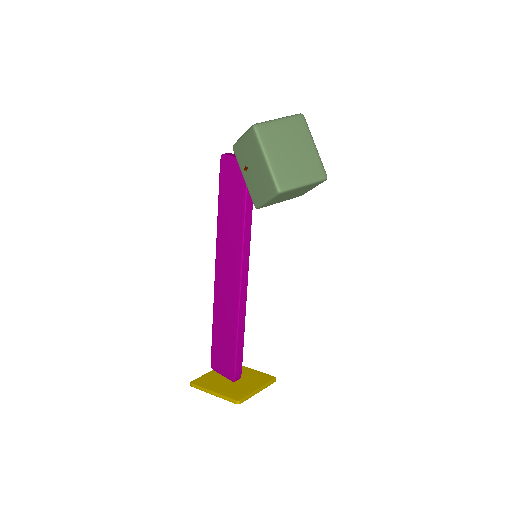}
    \end{subfigure}%
    \hfill%
    \begin{subfigure}[b]{0.20\linewidth}
		\centering
		\includegraphics[width=0.8\linewidth]{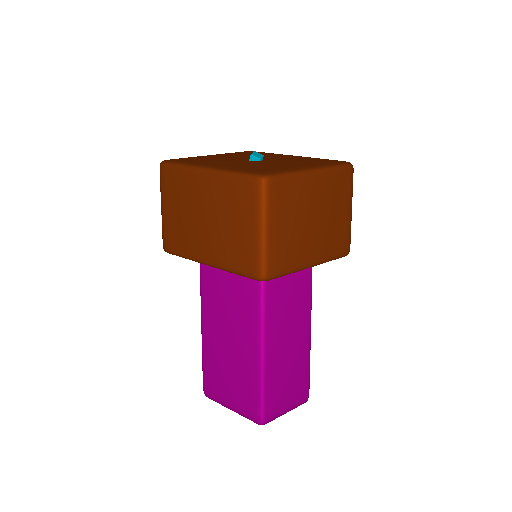}
    \end{subfigure}%
    \hfill%
    \begin{subfigure}[b]{0.20\linewidth}
		\centering
		\includegraphics[width=0.8\linewidth]{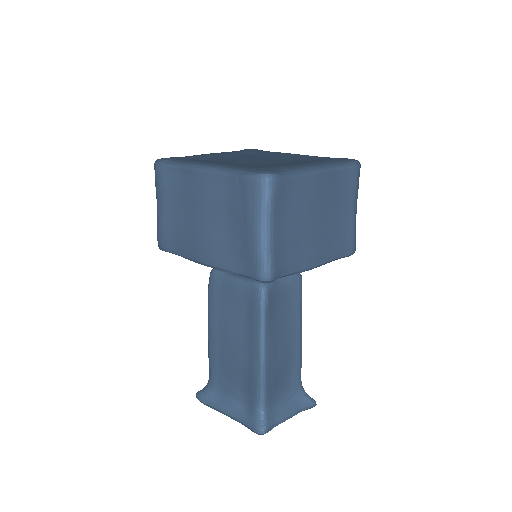}
    \end{subfigure}%
    \vskip\baselineskip%
    \vspace{-1.25em}
    \begin{subfigure}[b]{0.20\linewidth}
		\centering
		\includegraphics[width=0.8\linewidth]{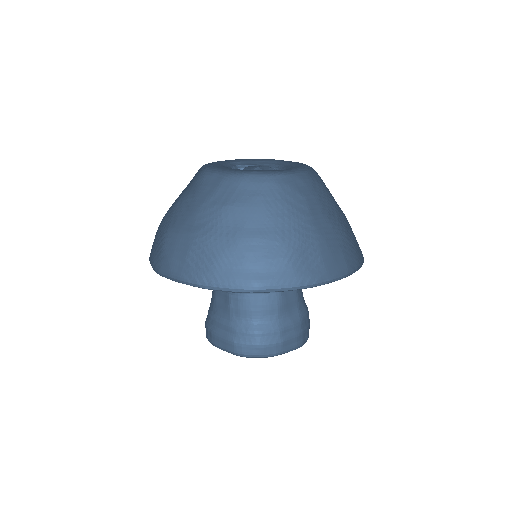}
    \end{subfigure}%
    \hfill%
    \begin{subfigure}[b]{0.20\linewidth}
		\centering
		\includegraphics[width=0.8\linewidth]{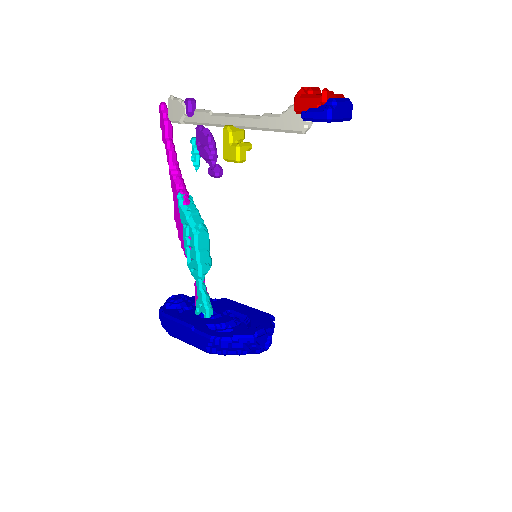}
    \end{subfigure}%
    \hfill%
    \begin{subfigure}[b]{0.20\linewidth}
		\centering
		\includegraphics[width=0.8\linewidth]{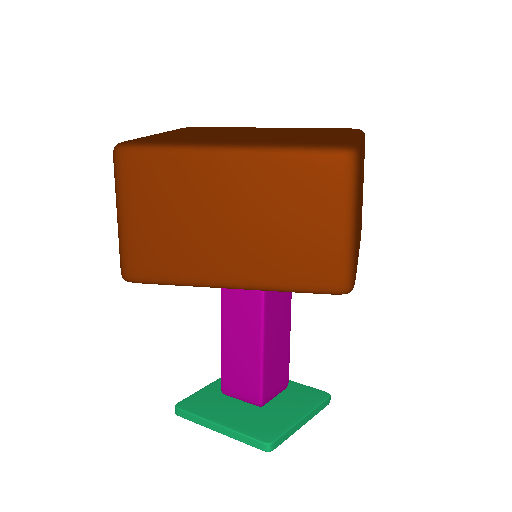}
    \end{subfigure}%
    \hfill%
    \begin{subfigure}[b]{0.20\linewidth}
		\centering
		\includegraphics[width=0.8\linewidth]{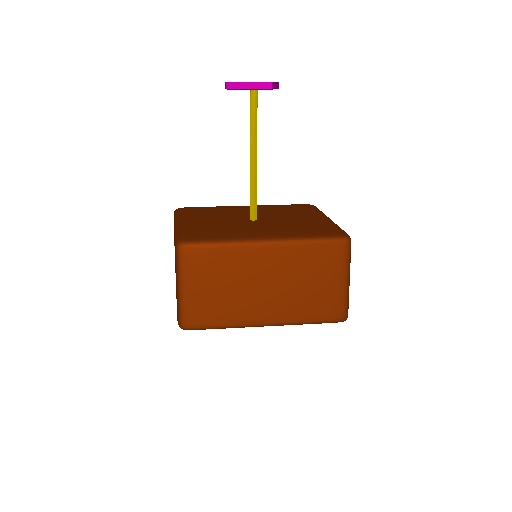}
    \end{subfigure}%
    \hfill%
    \begin{subfigure}[b]{0.20\linewidth}
		\centering
		\includegraphics[width=0.8\linewidth]{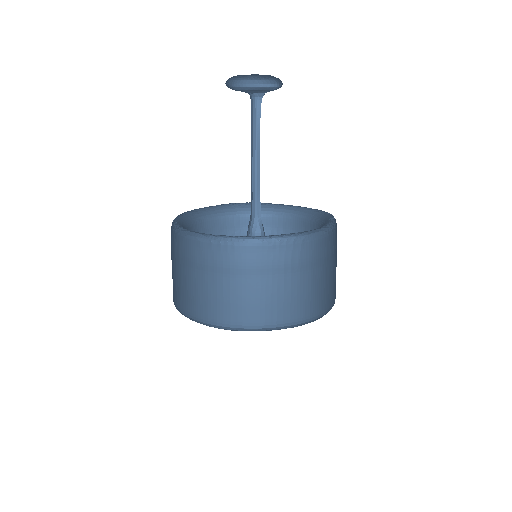}
    \end{subfigure}%
    \vskip\baselineskip%
    \vspace{-1.5em}
    \caption{{\bf Shape Generation Results on Lamps}. We showcase randomly generated lamps using our model, ATISS~\cite{Paschalidou2021NEURIPS}, PQ-NET~\cite{Wu2020CVPR} and IM-NET~\cite{Chen2019CVPR}.}
    \label{fig:shapenet_qualitative_comparison_lamps_supp}
\end{figure*}

%% file: fig/shape_generation_variable_sizes.tex
\begin{figure}
    \centering
    \vskip\baselineskip%
    \vspace{-0.5em}
    \begin{subfigure}[b]{0.15\linewidth}
		\centering
		\includegraphics[width=\linewidth]{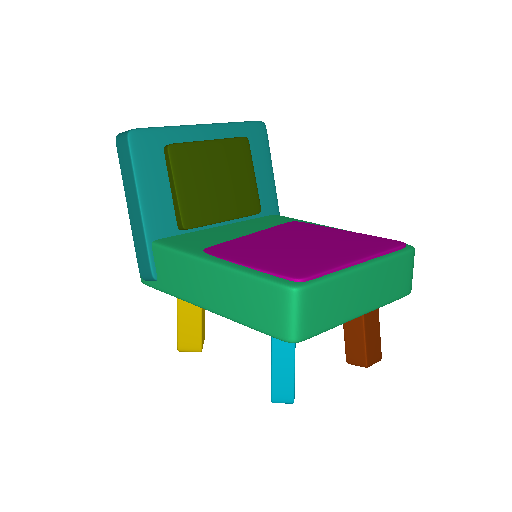}
    \end{subfigure}%
    \hfill%
    \begin{subfigure}[b]{0.15\linewidth}
		\centering
		\includegraphics[width=\linewidth]{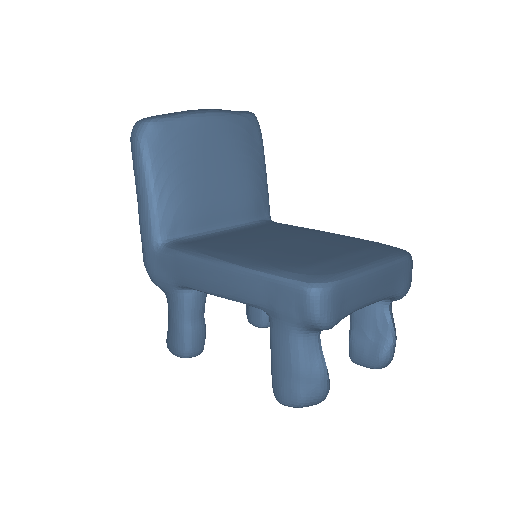}
    \end{subfigure}%
    \hfill%
    \begin{subfigure}[b]{0.15\linewidth}
		\centering
		\includegraphics[width=\linewidth]{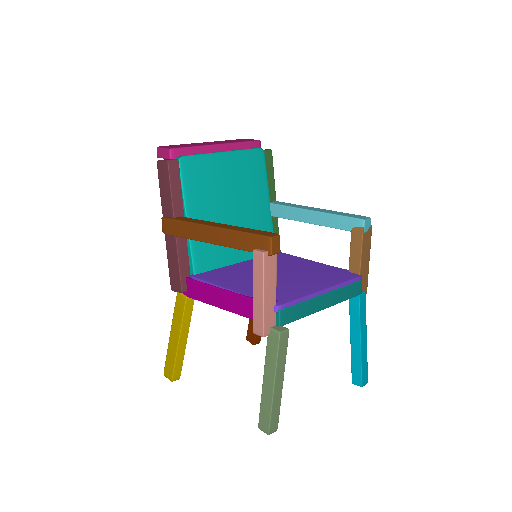}
    \end{subfigure}%
    \hfill%
    \begin{subfigure}[b]{0.15\linewidth}
		\centering
		\includegraphics[width=\linewidth]{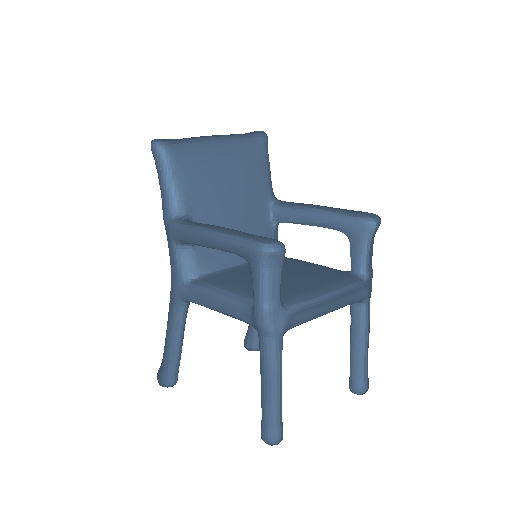}
    \end{subfigure}%
    \hfill%
    \begin{subfigure}[b]{0.15\linewidth}
		\centering
		\includegraphics[width=\linewidth]{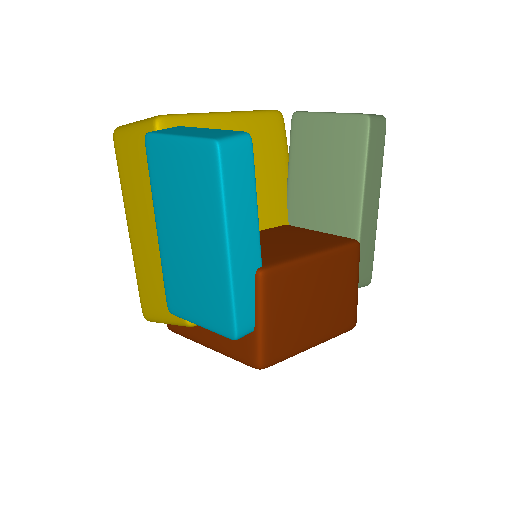}
    \end{subfigure}%
    \hfill%
    \begin{subfigure}[b]{0.15\linewidth}
		\centering
		\includegraphics[width=\linewidth]{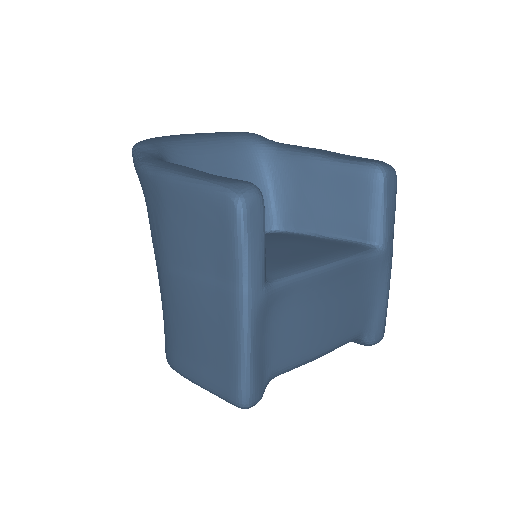}
    \end{subfigure}%
    \vskip\baselineskip%
    \begin{subfigure}[b]{0.15\linewidth}
		\centering
		\includegraphics[width=\linewidth]{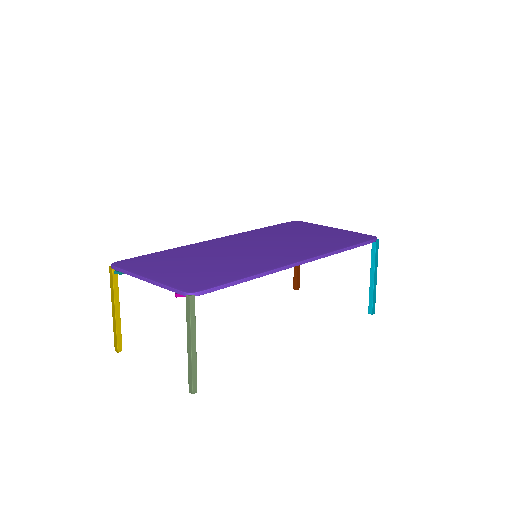}
    \end{subfigure}%
    \hfill%
    \begin{subfigure}[b]{0.15\linewidth}
		\centering
		\includegraphics[width=\linewidth]{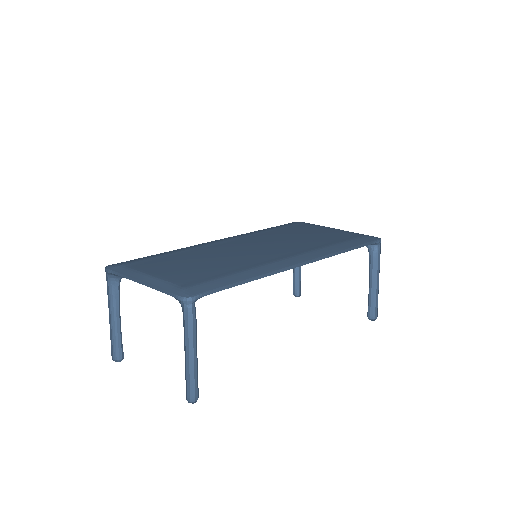}
    \end{subfigure}%
    \hfill%
    \begin{subfigure}[b]{0.15\linewidth}
		\centering
		\includegraphics[width=\linewidth]{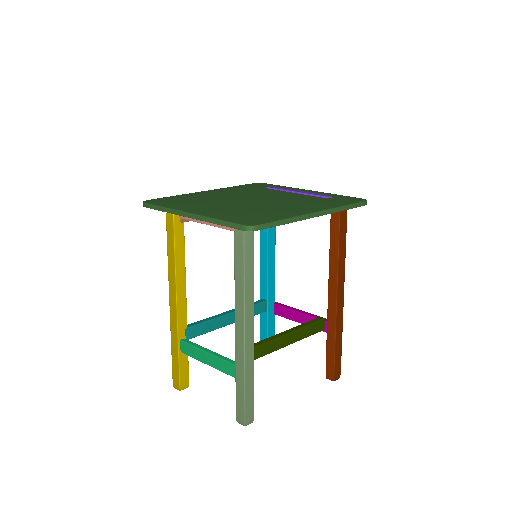}
    \end{subfigure}%
    \hfill%
    \begin{subfigure}[b]{0.15\linewidth}
		\centering
		\includegraphics[width=\linewidth]{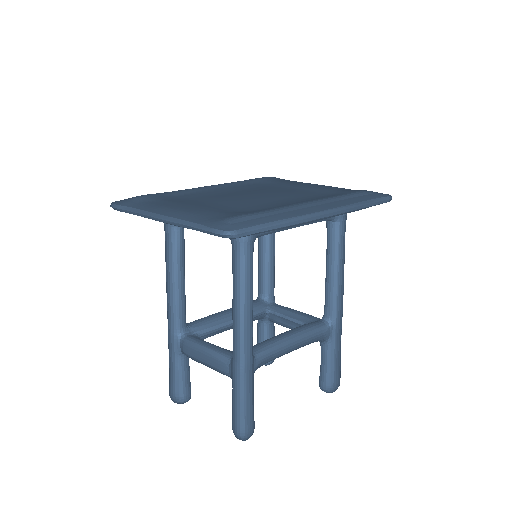}
    \end{subfigure}%
    \hfill%
    \begin{subfigure}[b]{0.15\linewidth}
		\centering
		\includegraphics[width=\linewidth]{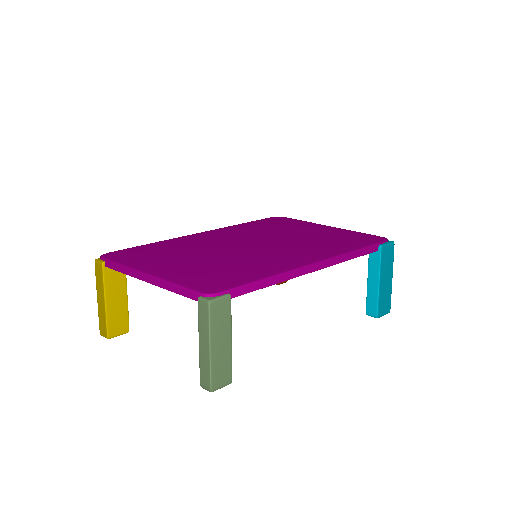}
    \end{subfigure}%
    \hfill%
    \begin{subfigure}[b]{0.15\linewidth}
		\centering
		\includegraphics[width=\linewidth]{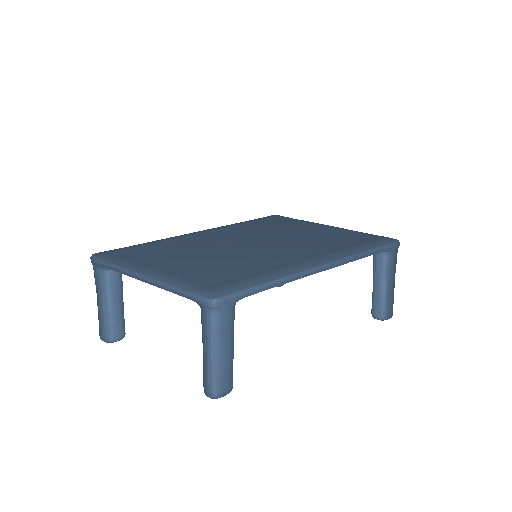}
    \end{subfigure}%
    \vskip\baselineskip%
    \begin{subfigure}[b]{0.15\linewidth}
		\centering
		\includegraphics[width=0.8\linewidth]{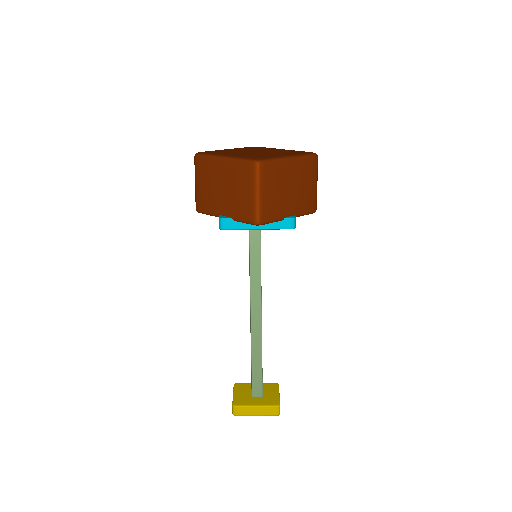}
    \end{subfigure}%
    \begin{subfigure}[b]{0.15\linewidth}
		\centering
		\includegraphics[width=0.8\linewidth]{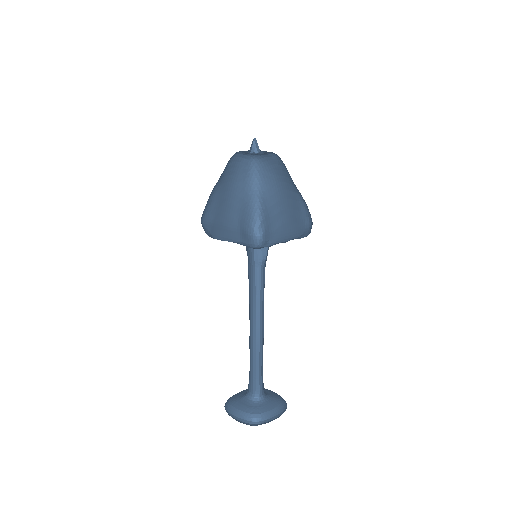}
    \end{subfigure}%
    \hfill%
    \begin{subfigure}[b]{0.15\linewidth}
		\centering
		\includegraphics[width=\linewidth]{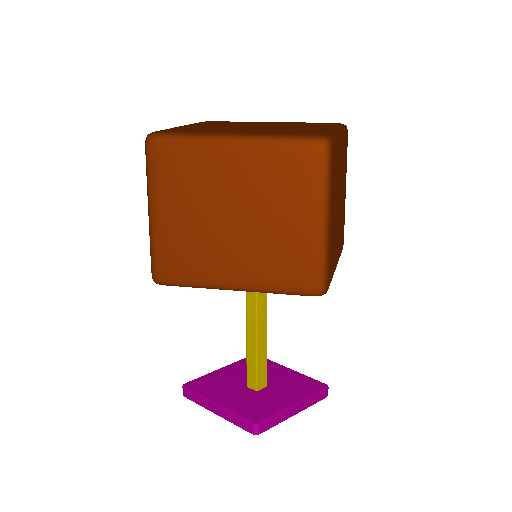}
    \end{subfigure}%
    \hfill%
    \begin{subfigure}[b]{0.15\linewidth}
		\centering
		\includegraphics[clip,width=\linewidth]{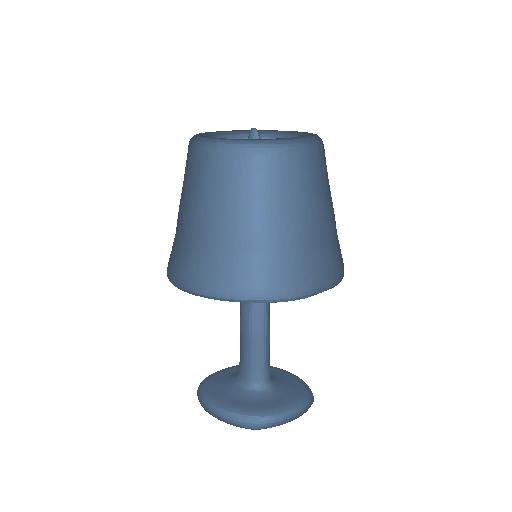}
    \end{subfigure}%
    \hfill%
    \begin{subfigure}[b]{0.15\linewidth}
		\centering
		\includegraphics[clip,width=\linewidth]{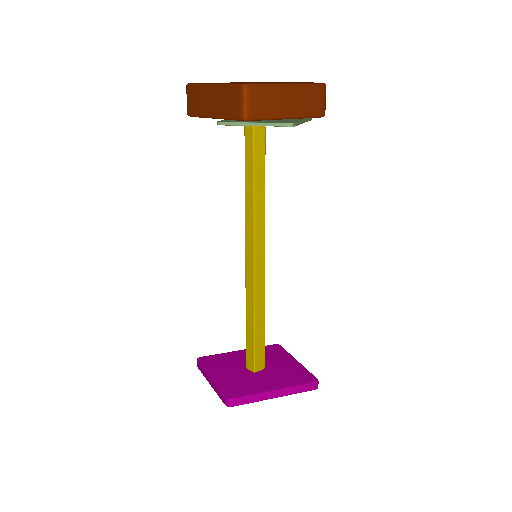}
    \end{subfigure}%
    \hfill%
    \begin{subfigure}[b]{0.15\linewidth}
		\centering
		\includegraphics[width=\linewidth]{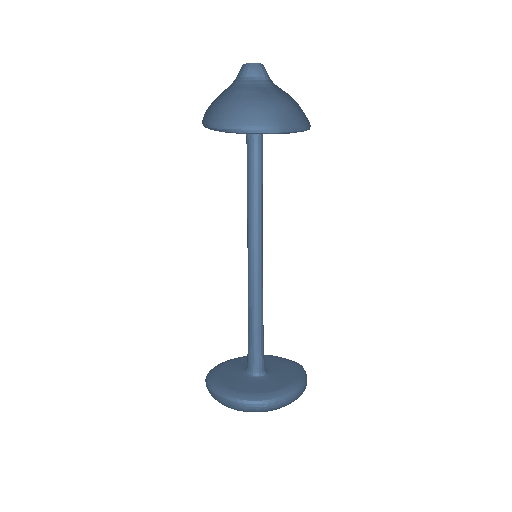}
    \end{subfigure}%
    \vskip\baselineskip%
    \vspace{-1.2em}
    \caption{{\bf Size-guided Shape Generation}. Conditioned on bounding boxes of different sizes, our model can generate shapes that
    match the conditioning..}
    \label{fig:shapenet_generations_variable_sizes}
    \vspace{-1.2em}
\end{figure}

%% file: fig/shape_completion_qualitative_chairs_supp.tex
\begin{figure*}
    \begin{subfigure}[t]{\linewidth}
    \centering
    \begin{subfigure}[b]{0.20\linewidth}
		\centering
		Partial Input
    \end{subfigure}%
    \hfill%
    \begin{subfigure}[b]{0.20\linewidth}
        \centering
        PQ-NET
    \end{subfigure}%
    \hfill%
    \begin{subfigure}[b]{0.20\linewidth}
		\centering
       ATISS 
    \end{subfigure}%
    \hfill%
    \begin{subfigure}[b]{0.20\linewidth}
        \centering
        Ours-Parts
    \end{subfigure}%
    \hfill%
    \begin{subfigure}[b]{0.20\linewidth}
        \centering
        Ours
    \end{subfigure}
    \end{subfigure}
    \vspace{-1.5em}
    \begin{subfigure}[b]{0.20\linewidth}
        \centering
        \includegraphics[width=0.8\linewidth]{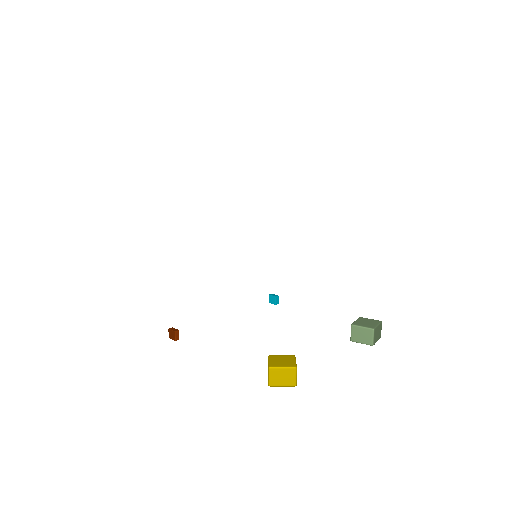}
    \end{subfigure}%
    \hfill%
    \begin{subfigure}[b]{0.20\linewidth}
	\centering
        \includegraphics[width=0.8\linewidth]{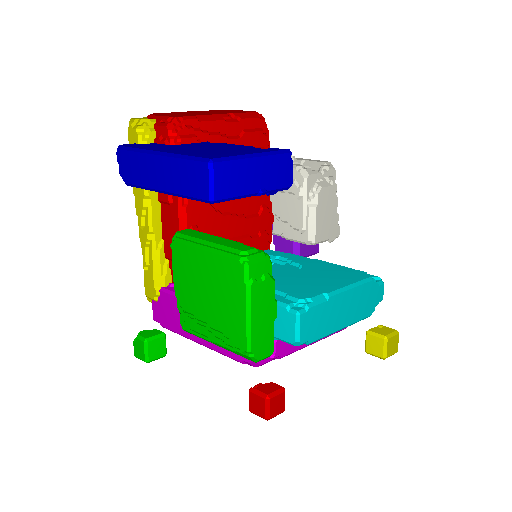}
    \end{subfigure}%
    \hfill%
    \begin{subfigure}[b]{0.20\linewidth}
        \centering
        \includegraphics[width=0.8\linewidth]{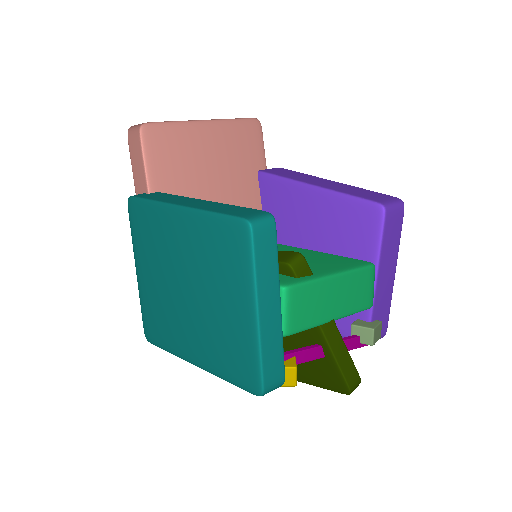}
    \end{subfigure}%
    \hfill%
    \begin{subfigure}[b]{0.20\linewidth}
	\centering
        \includegraphics[width=0.8\linewidth]{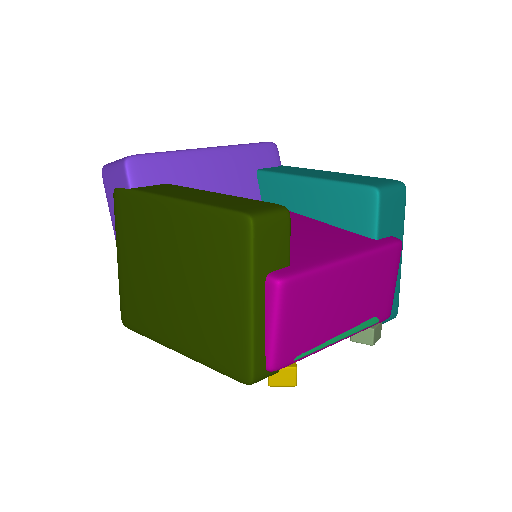}
    \end{subfigure}%
    \hfill%
    \begin{subfigure}[b]{0.20\linewidth}
	\centering
        \includegraphics[width=0.8\linewidth]{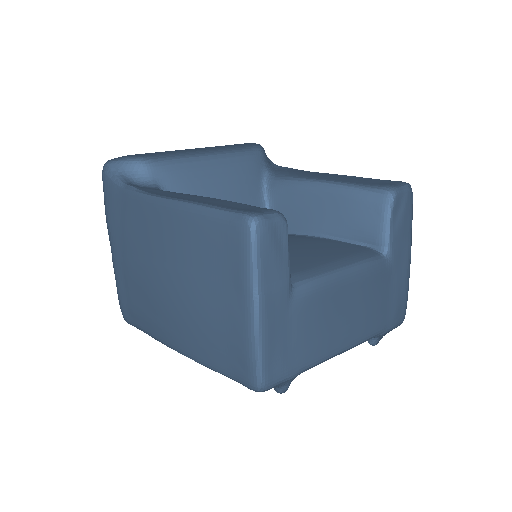}
    \end{subfigure}%
    \vskip\baselineskip%
    \vspace{-0.5em}
    \begin{subfigure}[b]{0.20\linewidth}
        \centering
        \includegraphics[width=0.8\linewidth]{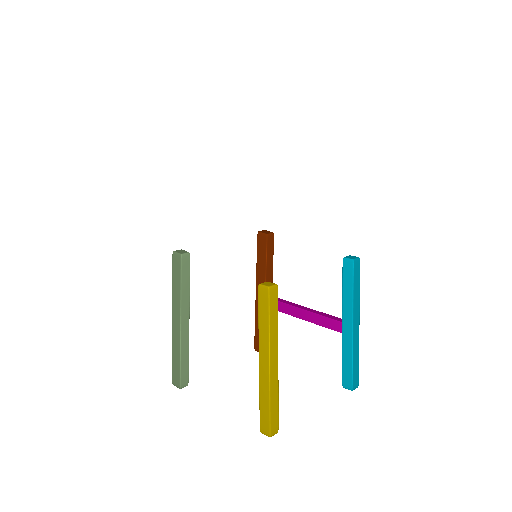}
    \end{subfigure}%
    \hfill%
    \begin{subfigure}[b]{0.20\linewidth}
	\centering
        \includegraphics[width=0.8\linewidth]{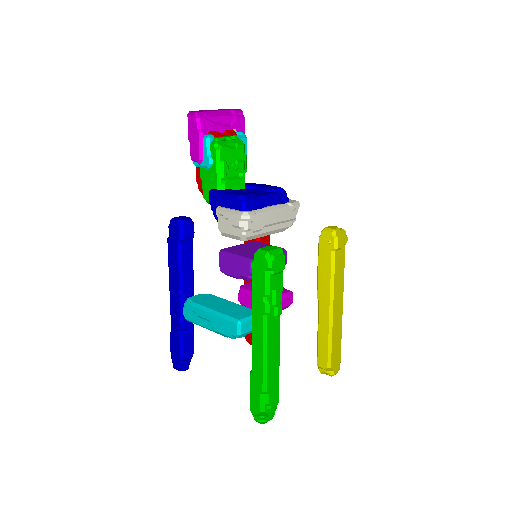}
    \end{subfigure}%
    \hfill%
    \begin{subfigure}[b]{0.20\linewidth}
        \centering
        \includegraphics[width=0.8\linewidth]{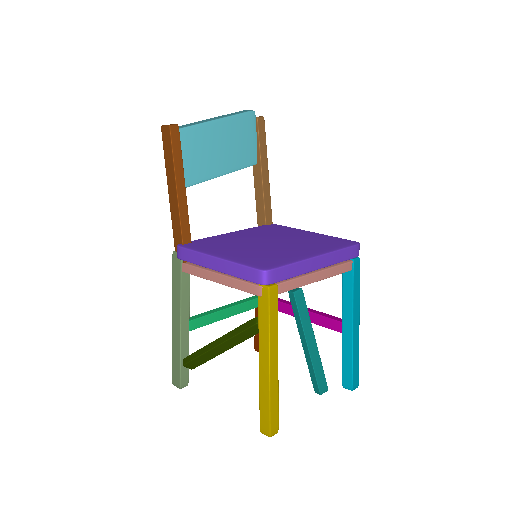}
    \end{subfigure}%
    \hfill%
    \begin{subfigure}[b]{0.20\linewidth}
	\centering
        \includegraphics[width=0.8\linewidth]{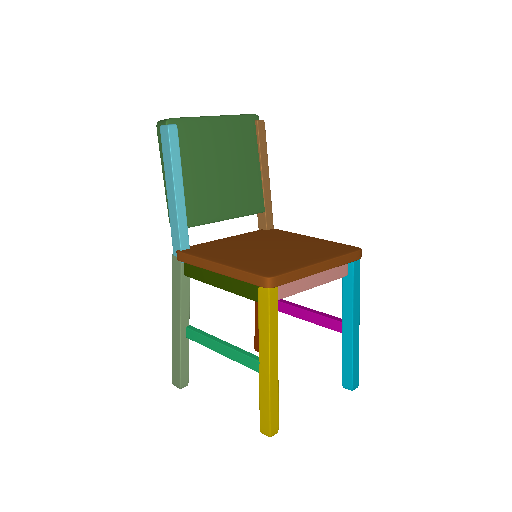}
    \end{subfigure}%
    \hfill%
    \begin{subfigure}[b]{0.20\linewidth}
	\centering
        \includegraphics[width=0.8\linewidth]{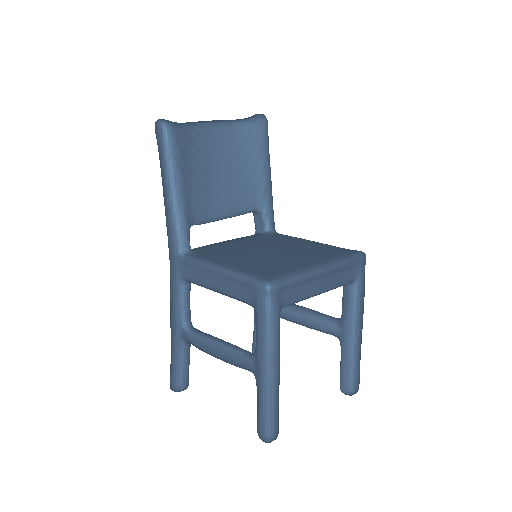}
    \end{subfigure}%
    \vskip\baselineskip%
    \vspace{-1.25em}
    \begin{subfigure}[b]{0.20\linewidth}
        \centering
        \includegraphics[width=0.8\linewidth]{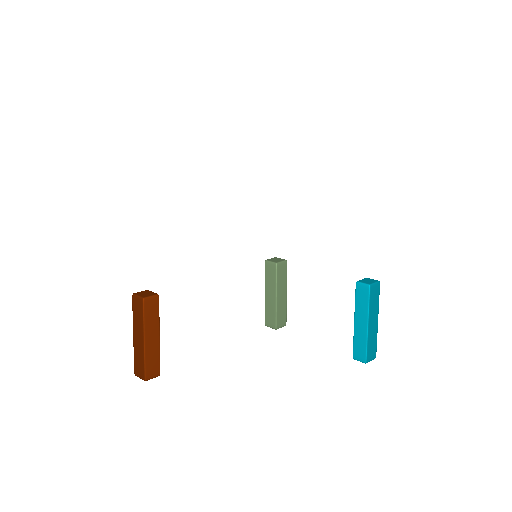}
    \end{subfigure}%
    \hfill%
    \begin{subfigure}[b]{0.20\linewidth}
	\centering
        \includegraphics[width=0.8\linewidth]{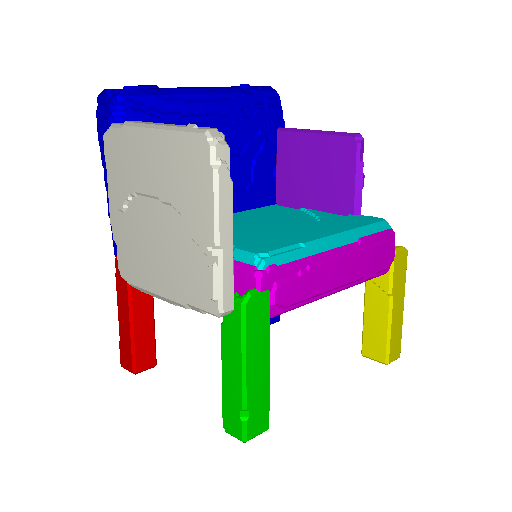}
    \end{subfigure}%
    \hfill%
    \begin{subfigure}[b]{0.20\linewidth}
        \centering
        \includegraphics[width=0.8\linewidth]{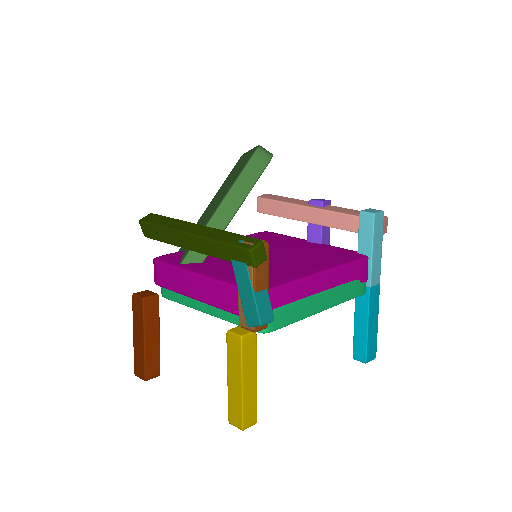}
    \end{subfigure}%
    \hfill%
    \begin{subfigure}[b]{0.20\linewidth}
	\centering
        \includegraphics[width=0.8\linewidth]{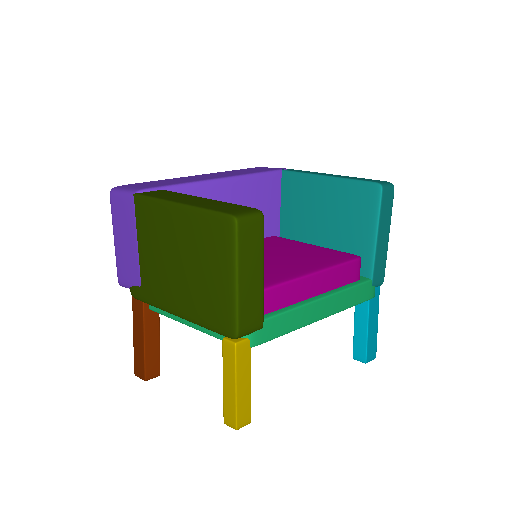}
    \end{subfigure}%
    \hfill%
    \begin{subfigure}[b]{0.20\linewidth}
	\centering
        \includegraphics[width=0.8\linewidth]{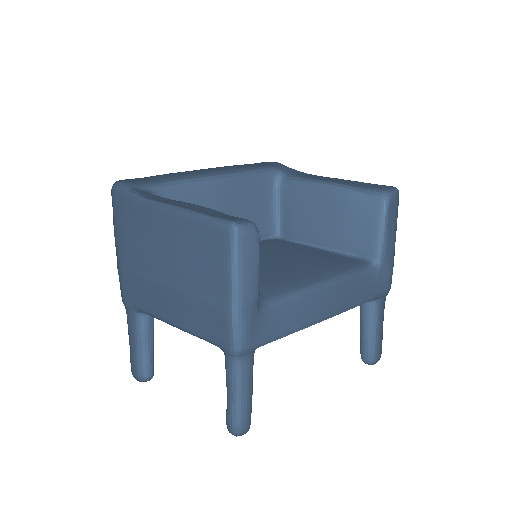}
    \end{subfigure}%
    \vskip\baselineskip%
    \vspace{-1.25em}
    \begin{subfigure}[b]{0.20\linewidth}
        \centering
        \includegraphics[width=0.8\linewidth]{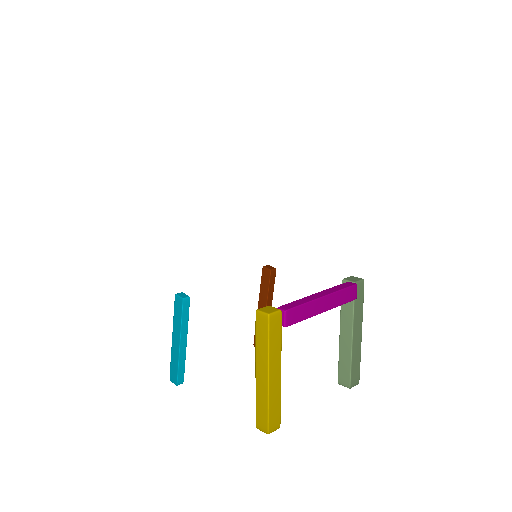}
    \end{subfigure}%
    \hfill%
    \begin{subfigure}[b]{0.20\linewidth}
	\centering
        \includegraphics[width=0.8\linewidth]{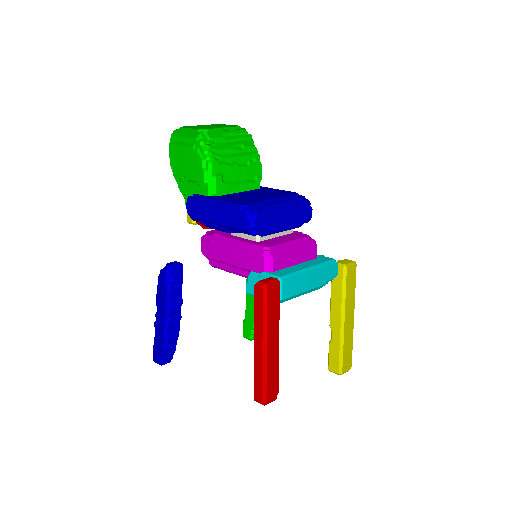}
    \end{subfigure}%
    \hfill%
    \begin{subfigure}[b]{0.20\linewidth}
        \centering
        \includegraphics[width=0.8\linewidth]{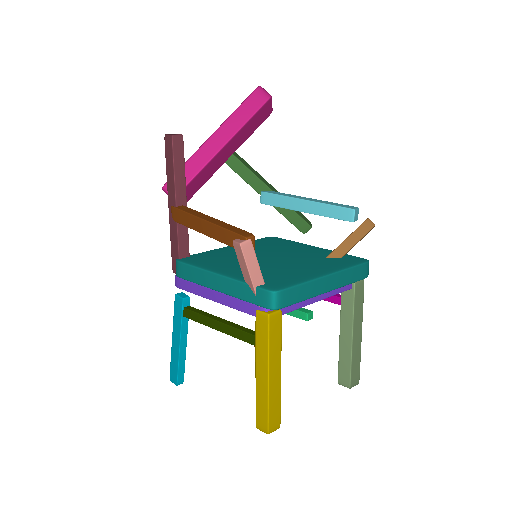}
    \end{subfigure}%
    \hfill%
    \begin{subfigure}[b]{0.20\linewidth}
	\centering
        \includegraphics[width=0.8\linewidth]{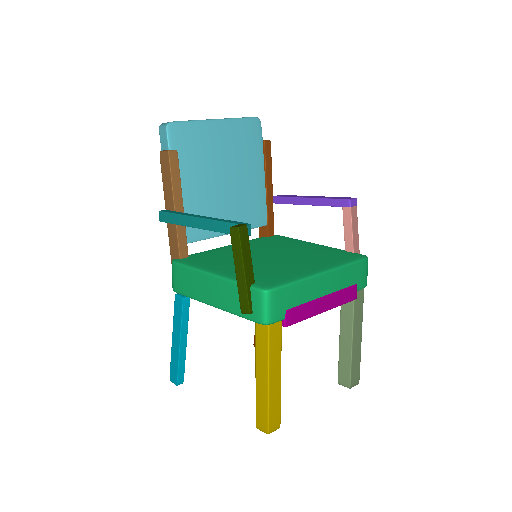}
    \end{subfigure}%
    \hfill%
    \begin{subfigure}[b]{0.20\linewidth}
	\centering
        \includegraphics[width=0.8\linewidth]{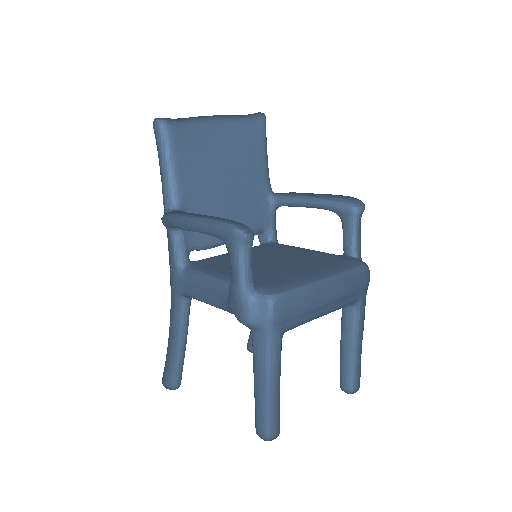}
    \end{subfigure}%
    \vskip\baselineskip%
    \vspace{-1.25em}
     \begin{subfigure}[b]{0.20\linewidth}
        \centering
        \includegraphics[width=0.8\linewidth]{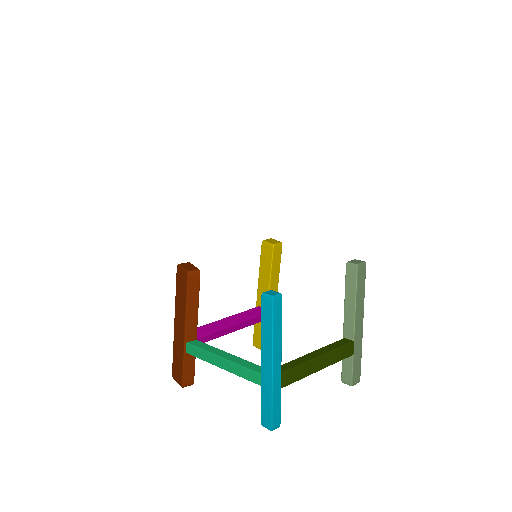}
    \end{subfigure}%
    \hfill%
    \begin{subfigure}[b]{0.20\linewidth}
	\centering
        \includegraphics[width=0.8\linewidth]{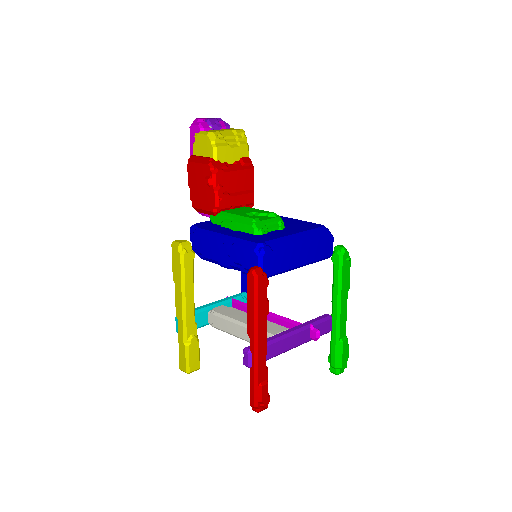}
    \end{subfigure}%
    \hfill%
    \begin{subfigure}[b]{0.20\linewidth}
        \centering
        \includegraphics[width=0.8\linewidth]{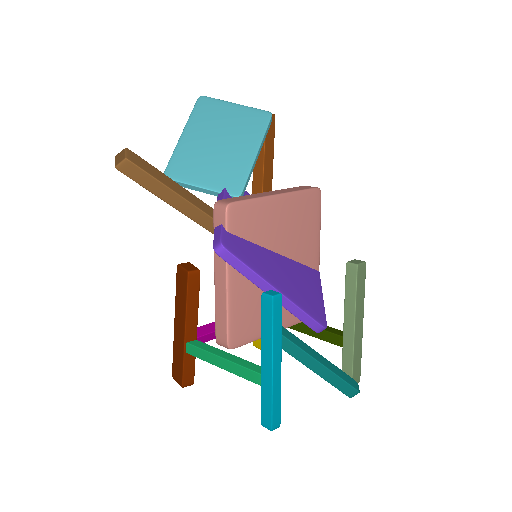}
    \end{subfigure}%
    \hfill%
    \begin{subfigure}[b]{0.20\linewidth}
	\centering
        \includegraphics[width=0.8\linewidth]{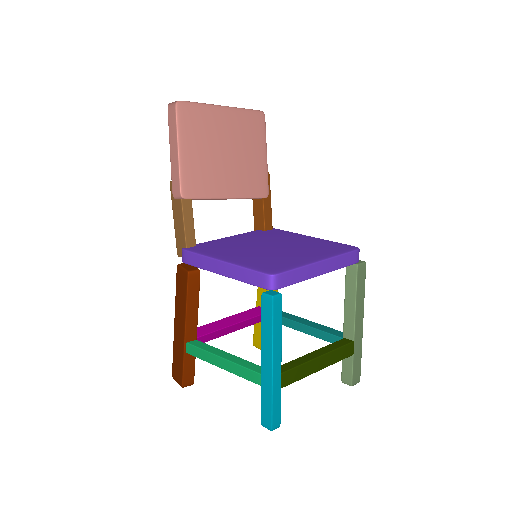}
    \end{subfigure}%
    \hfill%
    \begin{subfigure}[b]{0.20\linewidth}
	\centering
        \includegraphics[width=0.8\linewidth]{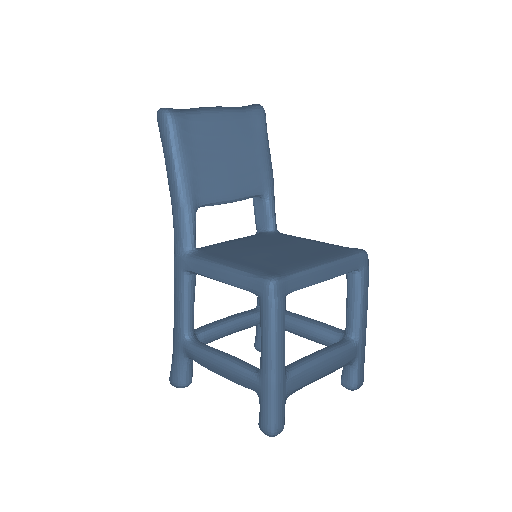}
    \end{subfigure}%
       \vskip\baselineskip%
    \vspace{-1.25em}
     \begin{subfigure}[b]{0.20\linewidth}
        \centering
        \includegraphics[width=0.8\linewidth]{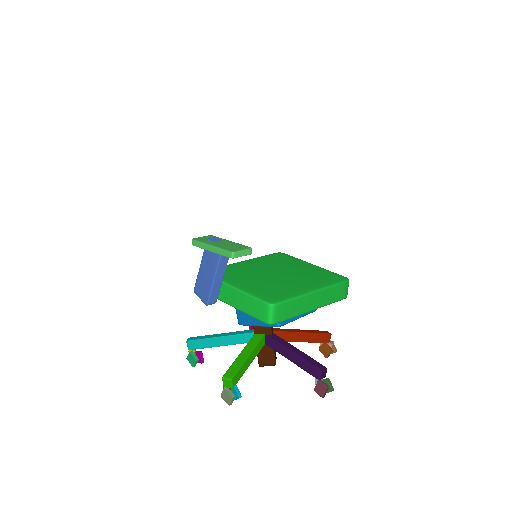}
    \end{subfigure}%
    \hfill%
    \begin{subfigure}[b]{0.20\linewidth}
	\centering
        \includegraphics[width=0.8\linewidth]{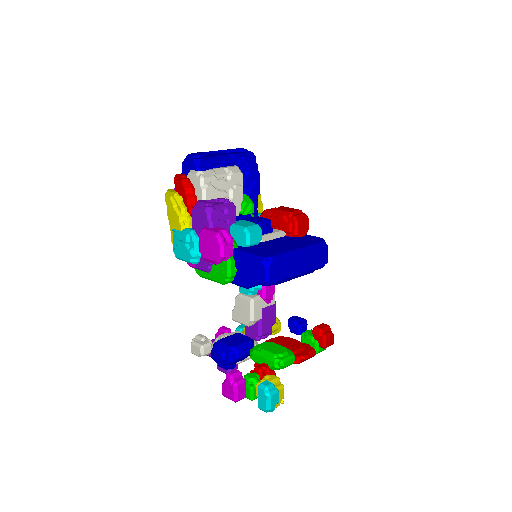}
    \end{subfigure}%
    \hfill%
    \begin{subfigure}[b]{0.20\linewidth}
        \centering
        \includegraphics[width=0.8\linewidth]{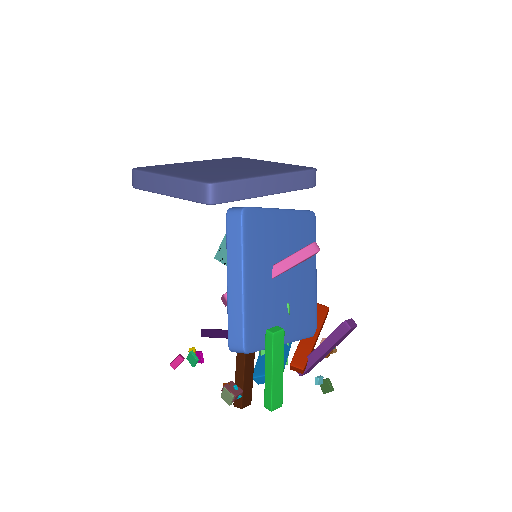}
    \end{subfigure}%
    \hfill%
    \begin{subfigure}[b]{0.20\linewidth}
	\centering
        \includegraphics[width=0.8\linewidth]{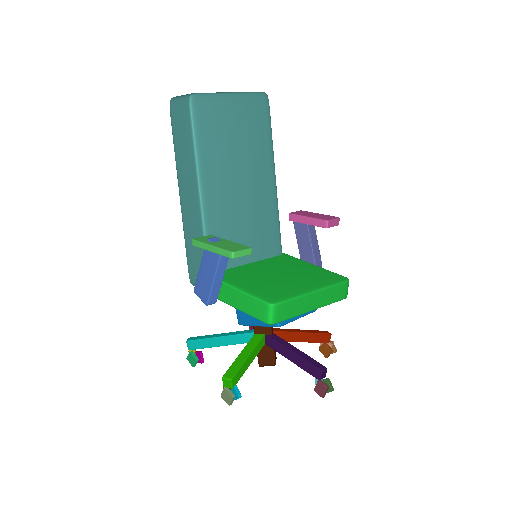}
    \end{subfigure}%
    \hfill%
    \begin{subfigure}[b]{0.20\linewidth}
	\centering
        \includegraphics[width=0.8\linewidth]{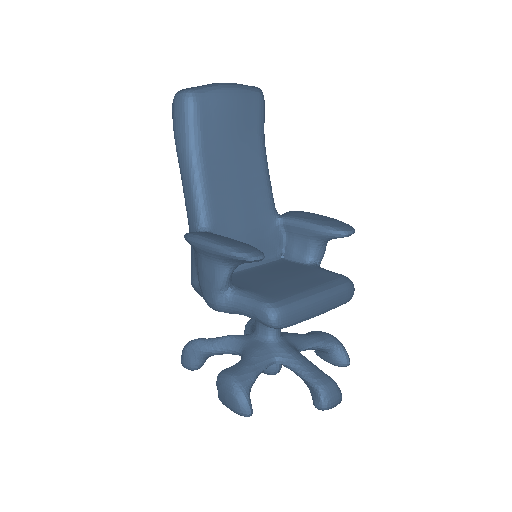}
    \end{subfigure}%
    \vskip\baselineskip%
    \vspace{-1.5em}
    \caption{{\bf Shape Completion Results on Chairs}. Starting
    from partial chairs, we show completions of our model,
    ATISS~\cite{Paschalidou2021NEURIPS} and PQ-NET~\cite{Wu2020CVPR}.}
    \label{fig:shapenet_qualitative_completion_comparison_chairs_supp}
\end{figure*}

%% file: fig/shape_completion_qualitative_tables_supp.tex
\begin{figure*}
    \begin{subfigure}[t]{\linewidth}
    \centering
  \begin{subfigure}[b]{0.20\linewidth}
		\centering
		Partial Input
    \end{subfigure}%
    \hfill%
    \begin{subfigure}[b]{0.20\linewidth}
        \centering
        PQ-NET
    \end{subfigure}%
    \hfill%
    \begin{subfigure}[b]{0.20\linewidth}
		\centering
       ATISS
    \end{subfigure}%
    \hfill%
    \begin{subfigure}[b]{0.20\linewidth}
        \centering
        Ours-Parts
    \end{subfigure}%
    \hfill%
    \begin{subfigure}[b]{0.20\linewidth}
        \centering
        Ours
    \end{subfigure}
    \end{subfigure}
    \vspace{-1.5em}
    \begin{subfigure}[b]{0.20\linewidth}
        \centering
        \includegraphics[width=0.8\linewidth]{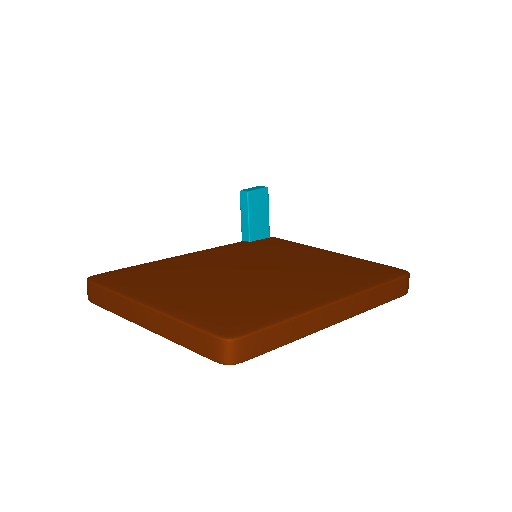}
    \end{subfigure}%
    \hfill%
    \begin{subfigure}[b]{0.20\linewidth}
	\centering
        \includegraphics[width=0.8\linewidth]{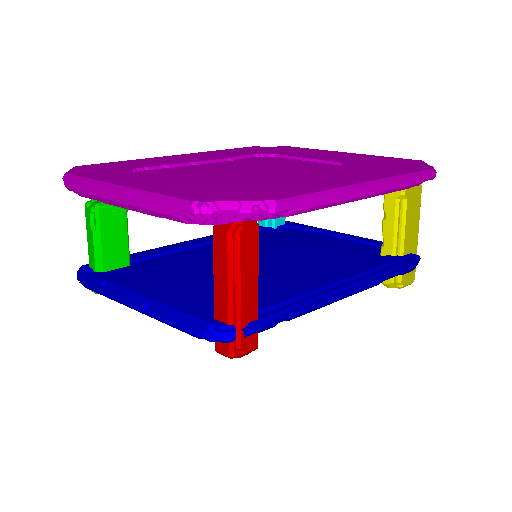}
    \end{subfigure}%
    \hfill%
    \begin{subfigure}[b]{0.20\linewidth}
        \centering
        \includegraphics[width=0.8\linewidth]{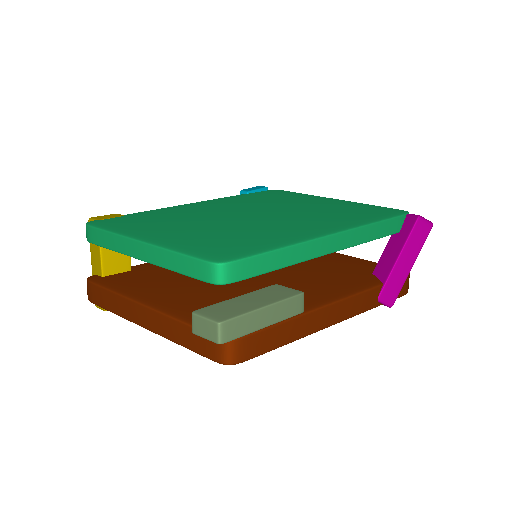}
    \end{subfigure}%
    \hfill%
    \begin{subfigure}[b]{0.20\linewidth}
	\centering
        \includegraphics[width=0.8\linewidth]{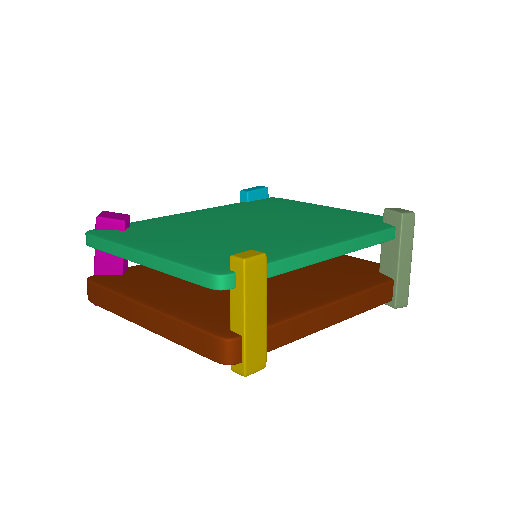}
    \end{subfigure}%
    \hfill%
    \begin{subfigure}[b]{0.20\linewidth}
	\centering
        \includegraphics[width=0.8\linewidth]{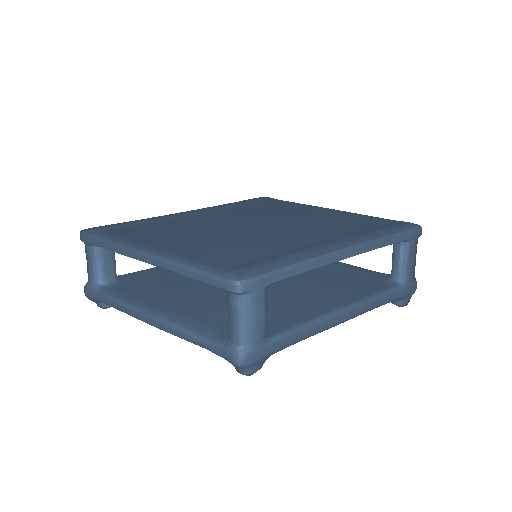}
    \end{subfigure}%
    \vskip\baselineskip%
    \vspace{-1.25em}
    \begin{subfigure}[b]{0.20\linewidth}
        \centering
        \includegraphics[width=0.8\linewidth]{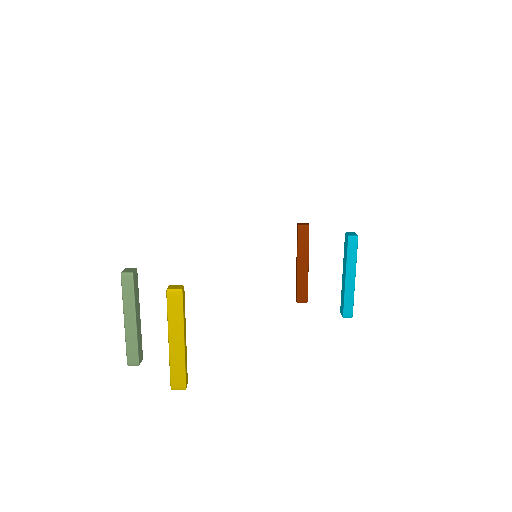}
    \end{subfigure}%
    \hfill%
    \begin{subfigure}[b]{0.20\linewidth}
	\centering
        \includegraphics[width=0.8\linewidth]{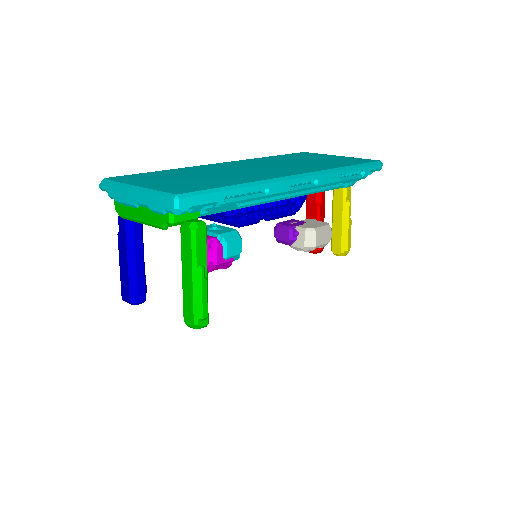}
    \end{subfigure}%
    \hfill%
    \begin{subfigure}[b]{0.20\linewidth}
        \centering
        \includegraphics[width=0.8\linewidth]{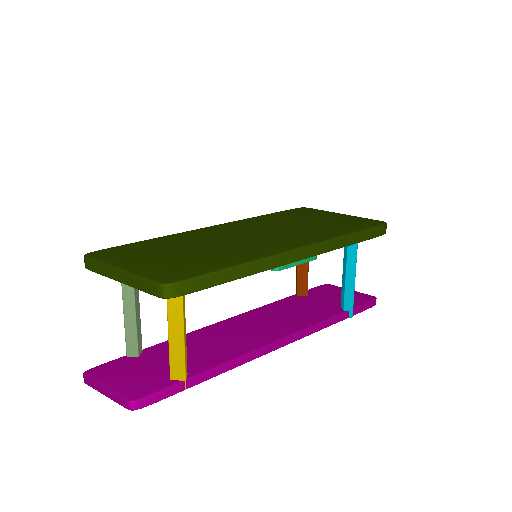}
    \end{subfigure}%
    \hfill%
    \begin{subfigure}[b]{0.20\linewidth}
	\centering
        \includegraphics[width=0.8\linewidth]{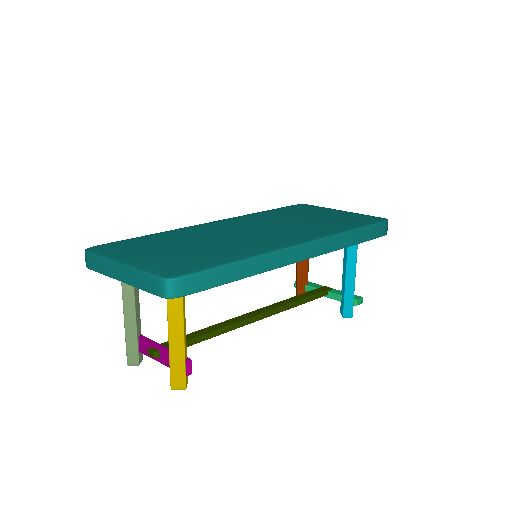}
    \end{subfigure}%
    \hfill%
    \begin{subfigure}[b]{0.20\linewidth}
	\centering
        \includegraphics[width=0.8\linewidth]{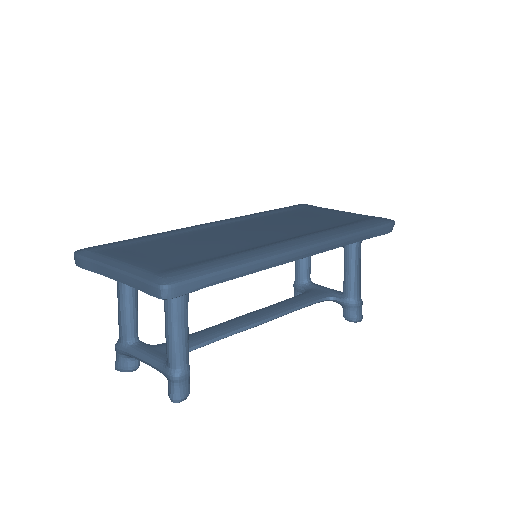}
    \end{subfigure}%
    \vskip\baselineskip%
    \vspace{-1.25em}
    \begin{subfigure}[b]{0.20\linewidth}
        \centering
        \includegraphics[width=0.8\linewidth]{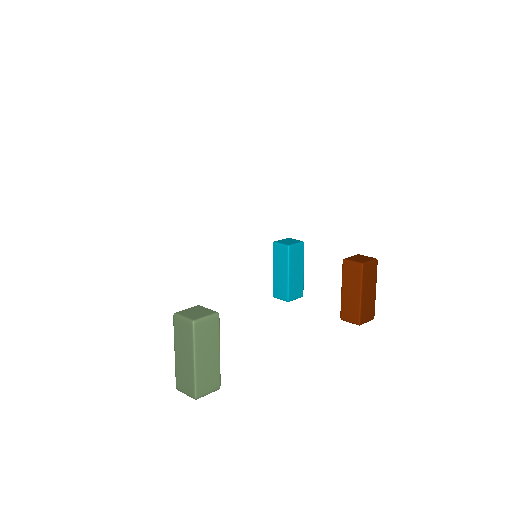}
    \end{subfigure}%
    \hfill%
    \begin{subfigure}[b]{0.20\linewidth}
	\centering
        \includegraphics[width=0.8\linewidth]{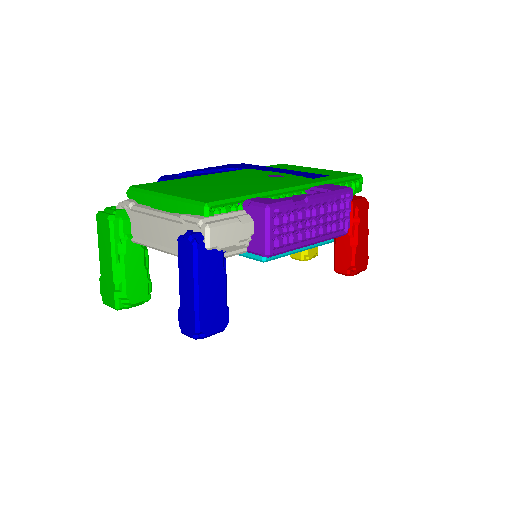}
    \end{subfigure}%
    \hfill%
    \begin{subfigure}[b]{0.20\linewidth}
        \centering
        \includegraphics[width=0.8\linewidth]{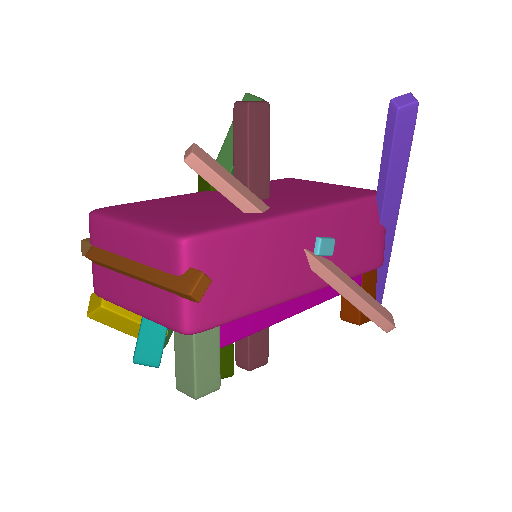}
    \end{subfigure}%
    \hfill%
    \begin{subfigure}[b]{0.20\linewidth}
	\centering
        \includegraphics[width=0.8\linewidth]{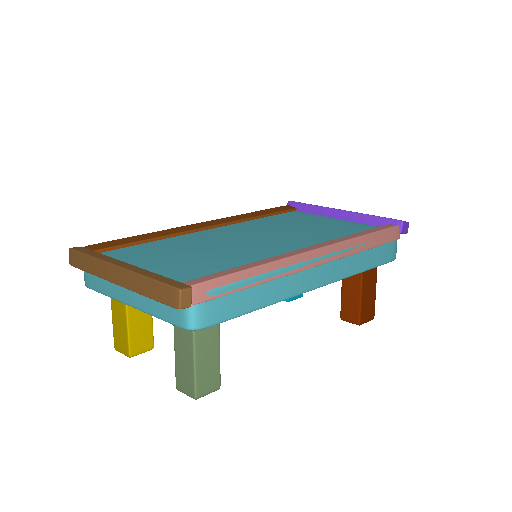}
    \end{subfigure}%
    \hfill%
    \begin{subfigure}[b]{0.20\linewidth}
	\centering
        \includegraphics[width=0.8\linewidth]{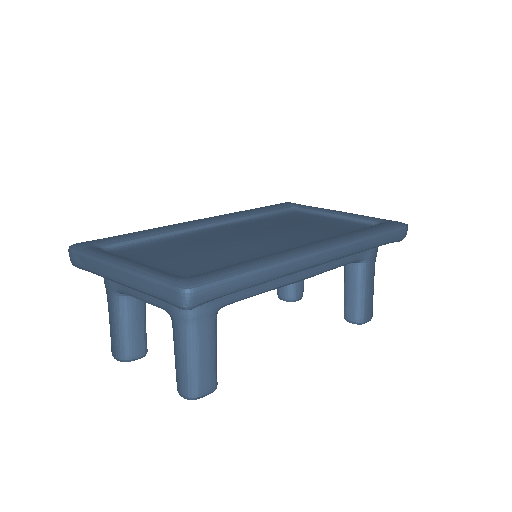}
    \end{subfigure}%
    \vskip\baselineskip%
    \vspace{-1.25em}
    \begin{subfigure}[b]{0.20\linewidth}
        \centering
        \includegraphics[width=0.8\linewidth]{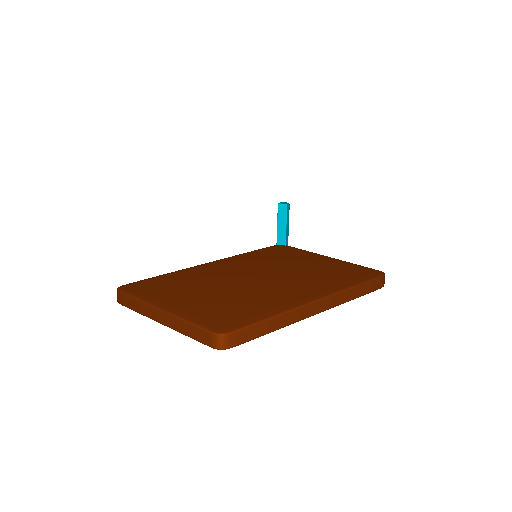}
    \end{subfigure}%
    \hfill%
    \begin{subfigure}[b]{0.20\linewidth}
	\centering
        \includegraphics[width=0.8\linewidth]{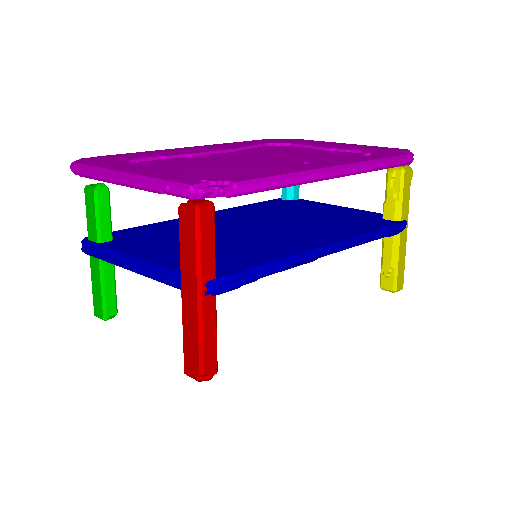}
    \end{subfigure}%
    \hfill%
    \begin{subfigure}[b]{0.20\linewidth}
        \centering
        \includegraphics[width=0.8\linewidth]{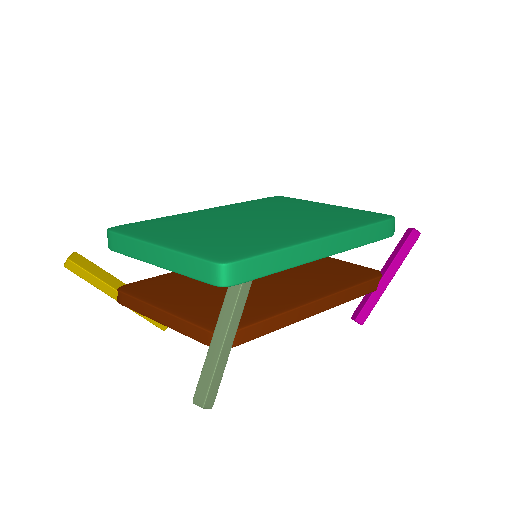}
    \end{subfigure}%
    \hfill%
    \begin{subfigure}[b]{0.20\linewidth}
	\centering
        \includegraphics[width=0.8\linewidth]{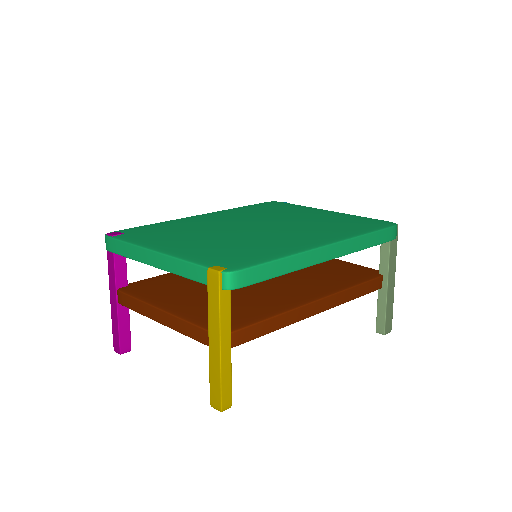}
    \end{subfigure}%
    \hfill%
    \begin{subfigure}[b]{0.20\linewidth}
	\centering
        \includegraphics[width=0.8\linewidth]{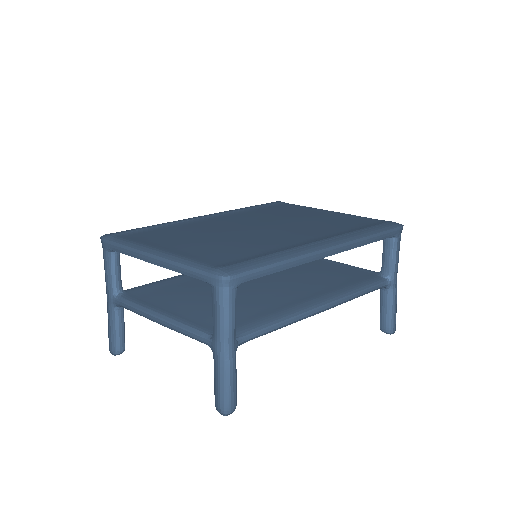}
    \end{subfigure}%
    \vskip\baselineskip%
    \vspace{-1.25em}
     \begin{subfigure}[b]{0.20\linewidth}
        \centering
        \includegraphics[width=0.8\linewidth]{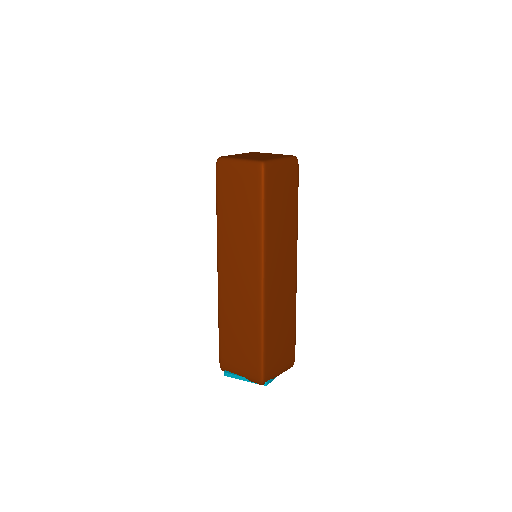}
    \end{subfigure}%
    \hfill%
    \begin{subfigure}[b]{0.20\linewidth}
	\centering
        \includegraphics[width=0.8\linewidth]{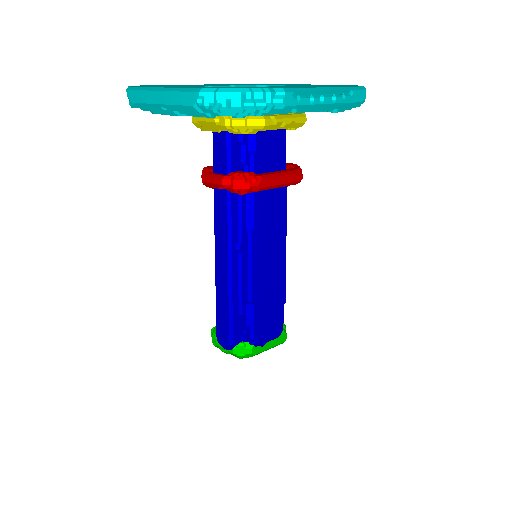}
    \end{subfigure}%
    \hfill%
    \begin{subfigure}[b]{0.20\linewidth}
        \centering
        \includegraphics[width=0.8\linewidth]{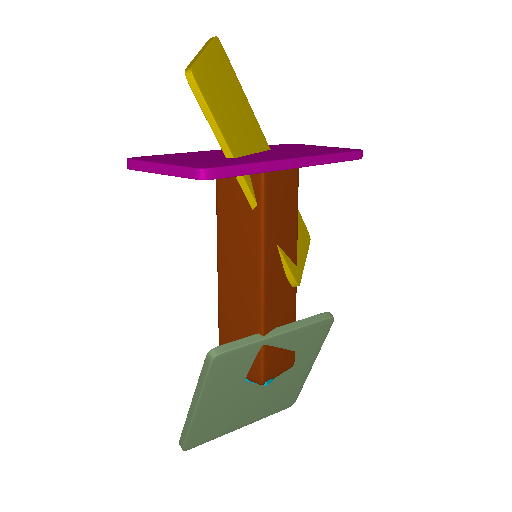}
    \end{subfigure}%
    \hfill%
    \begin{subfigure}[b]{0.20\linewidth}
	\centering
        \includegraphics[width=0.8\linewidth]{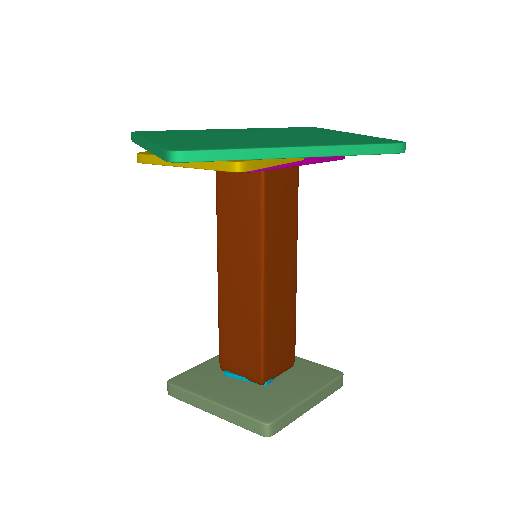}
    \end{subfigure}%
    \hfill%
    \begin{subfigure}[b]{0.20\linewidth}
	\centering
        \includegraphics[width=0.8\linewidth]{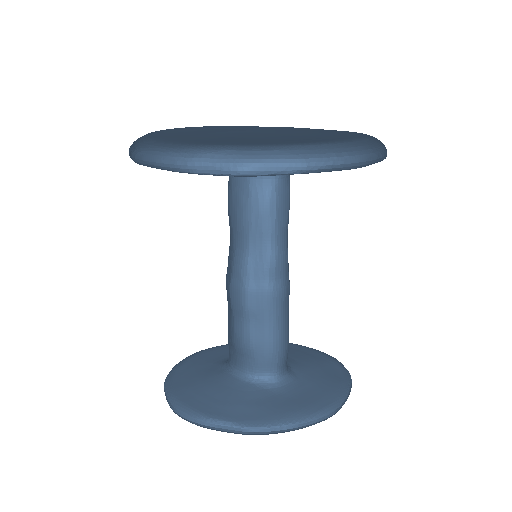}
    \end{subfigure}%
       \vskip\baselineskip%
    \vspace{-0.5em}
     \begin{subfigure}[b]{0.20\linewidth}
        \centering
        \includegraphics[width=0.8\linewidth]{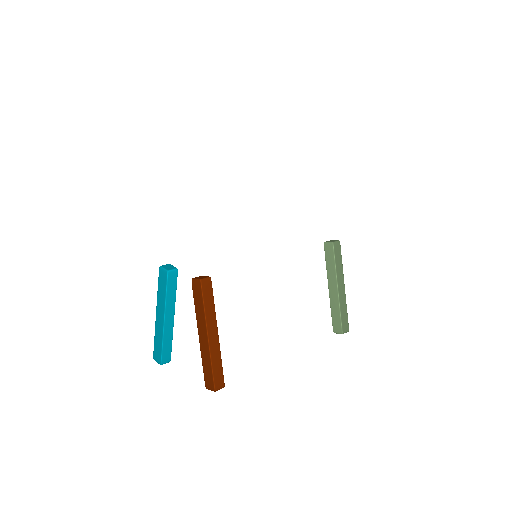}
    \end{subfigure}%
    \hfill%
    \begin{subfigure}[b]{0.20\linewidth}
	\centering
        \includegraphics[width=0.8\linewidth]{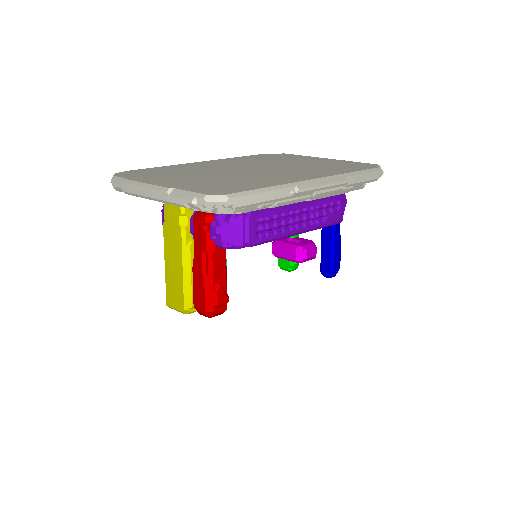}
    \end{subfigure}%
    \hfill%
    \begin{subfigure}[b]{0.20\linewidth}
        \centering
        \includegraphics[width=0.8\linewidth]{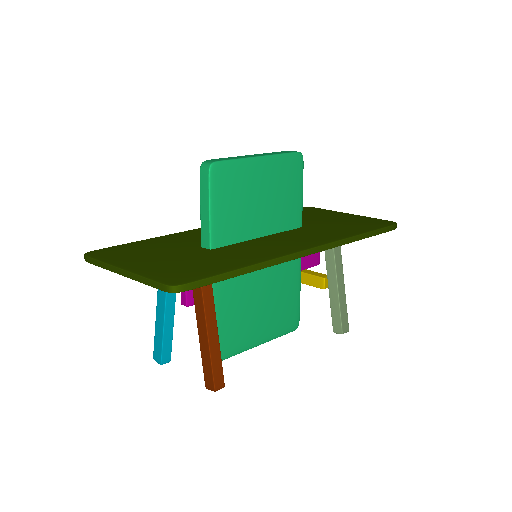}
    \end{subfigure}%
    \hfill%
    \begin{subfigure}[b]{0.20\linewidth}
	\centering
        \includegraphics[width=0.8\linewidth]{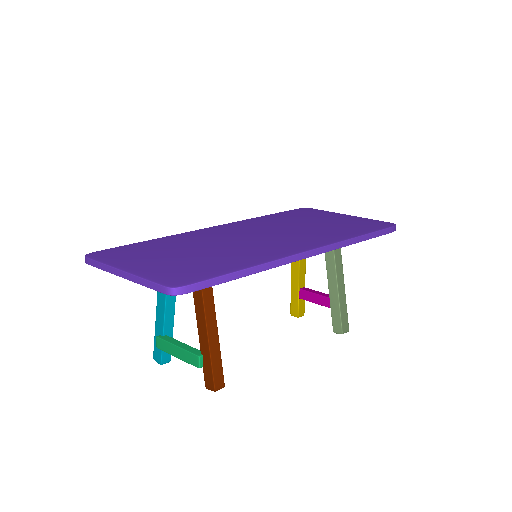}
    \end{subfigure}%
    \hfill%
    \begin{subfigure}[b]{0.20\linewidth}
	\centering
        \includegraphics[width=0.8\linewidth]{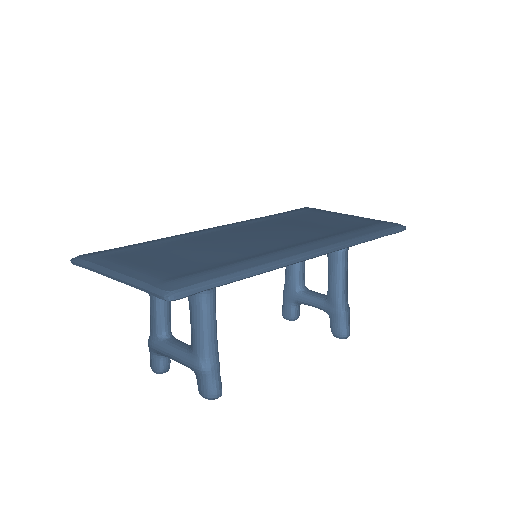}
    \end{subfigure}%
    \vskip\baselineskip%
    \vspace{-1.5em}
    \caption{{\bf Shape Completion Results on Tables}. Starting from partial tables, we show completions of our model, ATISS~\cite{Paschalidou2021NEURIPS} and PQ-Net~\cite{Wu2020CVPR}.}
    \label{fig:shapenet_qualitative_completion_comparison_tables_supp}
\end{figure*}

%% file: fig/shape_completion_qualitative_lamps_supp.tex
\begin{figure*}
    \begin{subfigure}[t]{\linewidth}
    \centering
    \begin{subfigure}[b]{0.20\linewidth}
	\centering
	Partial Input
    \end{subfigure}%
    \hfill%
    \begin{subfigure}[b]{0.20\linewidth}
        \centering
        PQ-NET
    \end{subfigure}%
    \hfill%
    \begin{subfigure}[b]{0.20\linewidth}
		\centering
       ATISS
    \end{subfigure}%
    \hfill%
    \begin{subfigure}[b]{0.20\linewidth}
        \centering
        Ours-Parts
    \end{subfigure}%
    \hfill%
    \begin{subfigure}[b]{0.20\linewidth}
        \centering
        Ours
    \end{subfigure}
    \end{subfigure}
    \vskip\baselineskip%
    \vspace{-1.5em}
    \begin{subfigure}[b]{0.20\linewidth}
        \centering
        \includegraphics[width=0.8\linewidth]{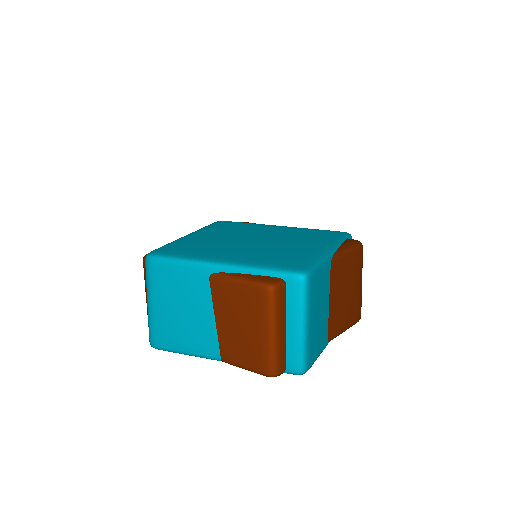}
    \end{subfigure}%
    \hfill%
    \begin{subfigure}[b]{0.20\linewidth}
	\centering
        \includegraphics[width=0.8\linewidth]{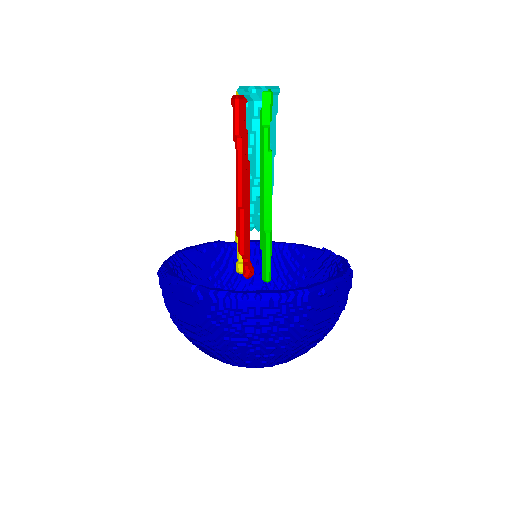}
    \end{subfigure}%
    \hfill%
    \begin{subfigure}[b]{0.20\linewidth}
        \centering
        \includegraphics[width=0.8\linewidth]{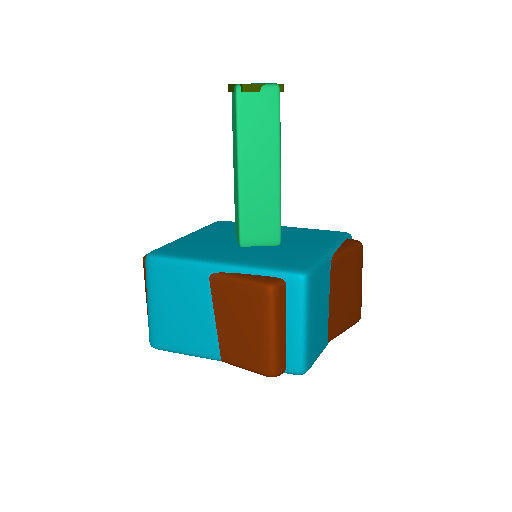}
    \end{subfigure}%
    \hfill%
    \begin{subfigure}[b]{0.20\linewidth}
	\centering
        \includegraphics[width=0.8\linewidth]{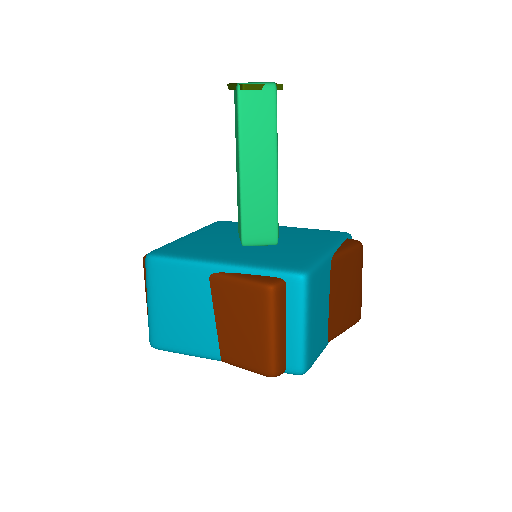}
    \end{subfigure}%
    \hfill%
    \begin{subfigure}[b]{0.20\linewidth}
	\centering
        \includegraphics[width=0.8\linewidth]{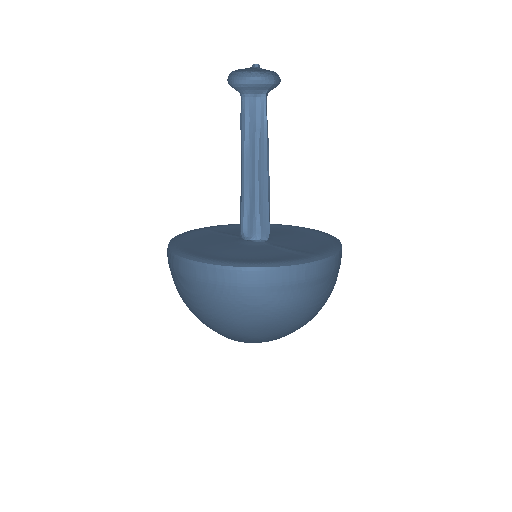}
    \end{subfigure}%
    \vskip\baselineskip%
    \vspace{-1.25em}
     \begin{subfigure}[b]{0.20\linewidth}
        \centering
        \includegraphics[width=0.8\linewidth]{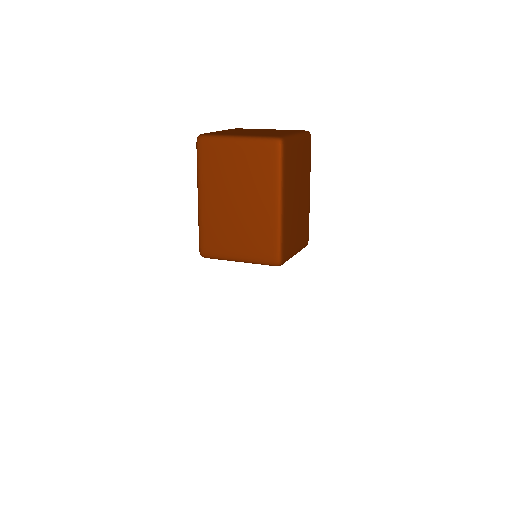}
    \end{subfigure}%
    \hfill%
    \begin{subfigure}[b]{0.20\linewidth}
	\centering
        \includegraphics[width=0.8\linewidth]{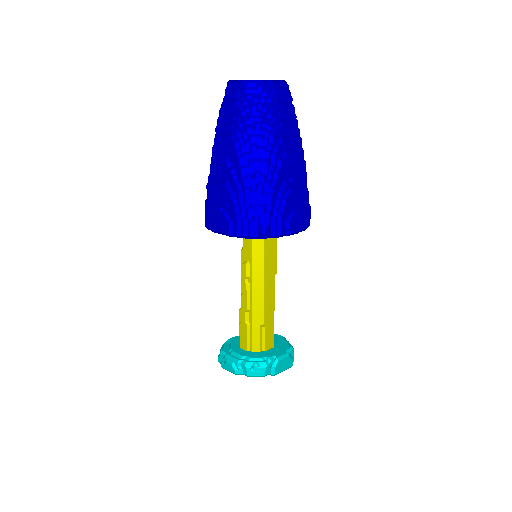}
    \end{subfigure}%
    \hfill%
    \begin{subfigure}[b]{0.20\linewidth}
        \centering
        \includegraphics[width=0.8\linewidth]{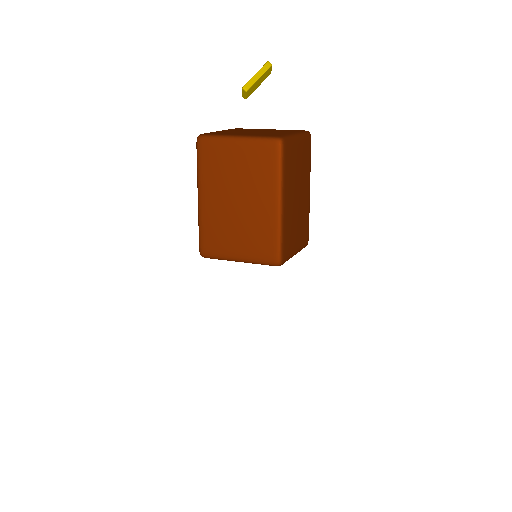}
    \end{subfigure}%
    \hfill%
    \begin{subfigure}[b]{0.20\linewidth}
	\centering
        \includegraphics[width=0.8\linewidth]{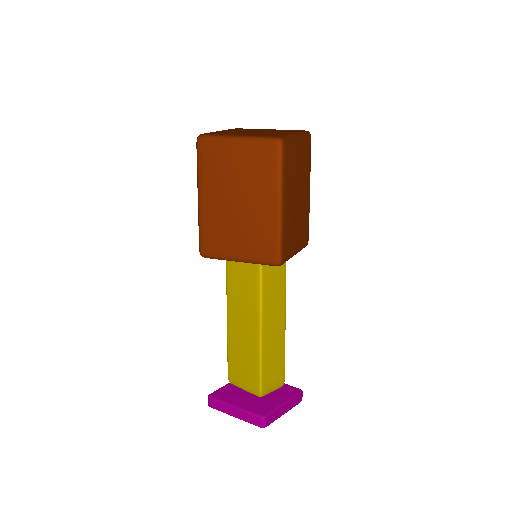}
    \end{subfigure}%
    \hfill%
    \begin{subfigure}[b]{0.20\linewidth}
	\centering
        \includegraphics[width=0.8\linewidth]{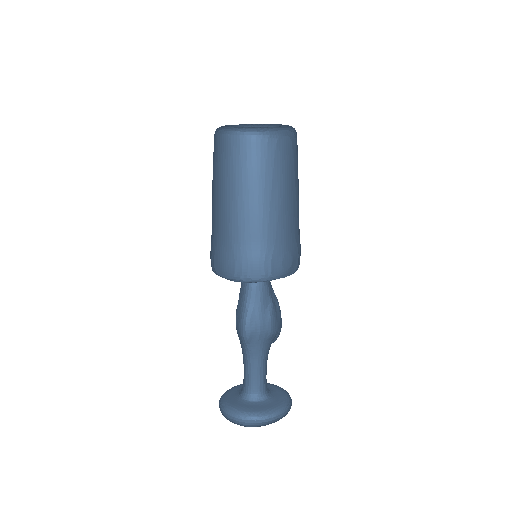}
    \end{subfigure}%
    \vskip\baselineskip%
    \vspace{-1.25em}
    \begin{subfigure}[b]{0.20\linewidth}
        \centering
        \includegraphics[width=0.8\linewidth]{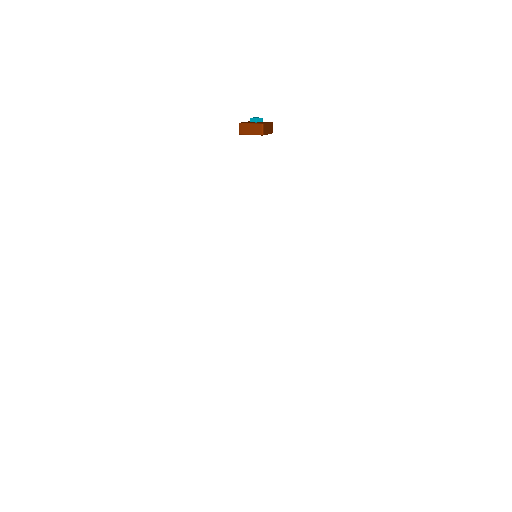}
    \end{subfigure}%
    \hfill%
    \begin{subfigure}[b]{0.20\linewidth}
	\centering
        \includegraphics[width=0.8\linewidth]{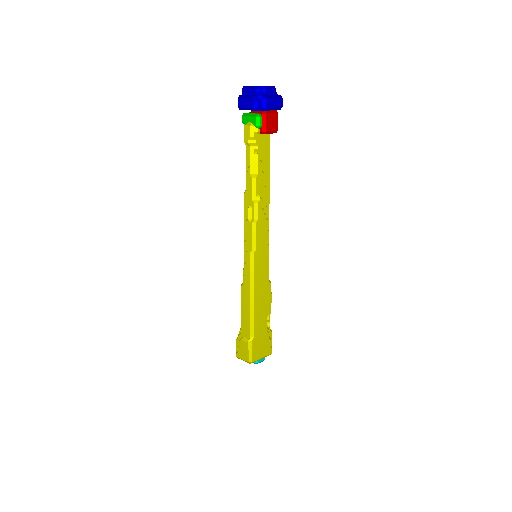}
    \end{subfigure}%
    \hfill%
    \begin{subfigure}[b]{0.20\linewidth}
        \centering
        \includegraphics[width=0.8\linewidth]{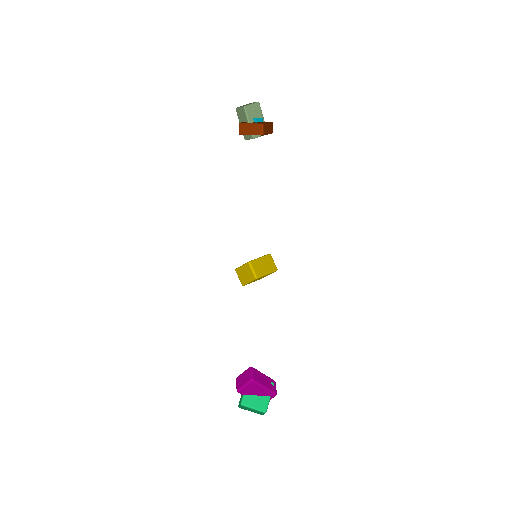}
    \end{subfigure}%
    \hfill%
    \begin{subfigure}[b]{0.20\linewidth}
	\centering
        \includegraphics[width=0.8\linewidth]{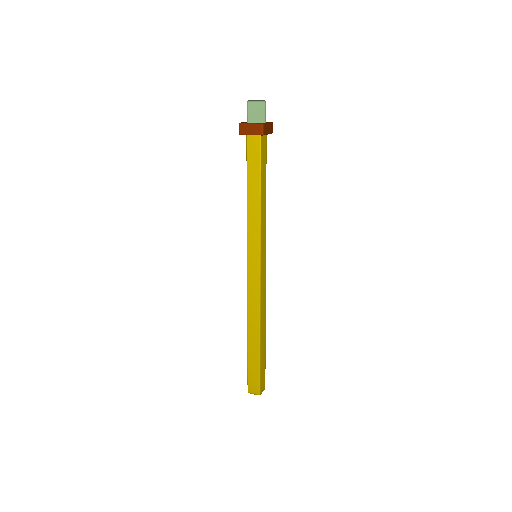}
    \end{subfigure}%
    \hfill%
    \begin{subfigure}[b]{0.20\linewidth}
	\centering
        \includegraphics[width=0.8\linewidth]{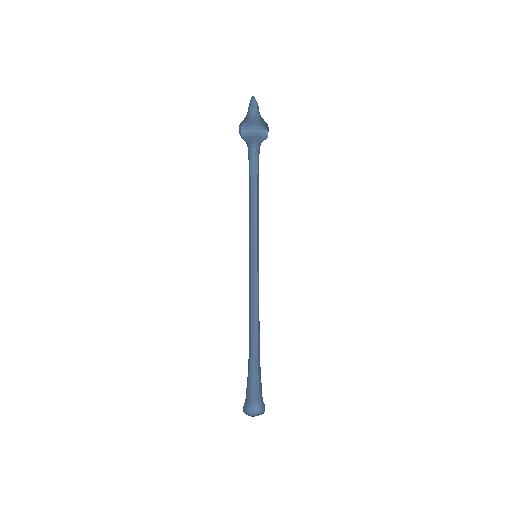}
    \end{subfigure}%
    \vskip\baselineskip%
    \vspace{-1.25em}
    \begin{subfigure}[b]{0.20\linewidth}
        \centering
        \includegraphics[width=0.8\linewidth]{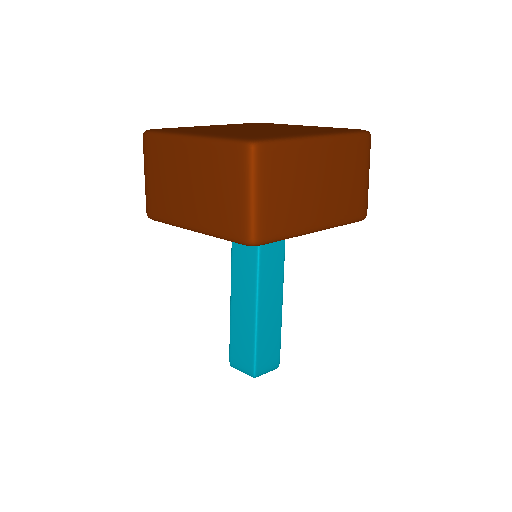}
    \end{subfigure}%
    \hfill%
    \begin{subfigure}[b]{0.20\linewidth}
	\centering
        \includegraphics[width=0.8\linewidth]{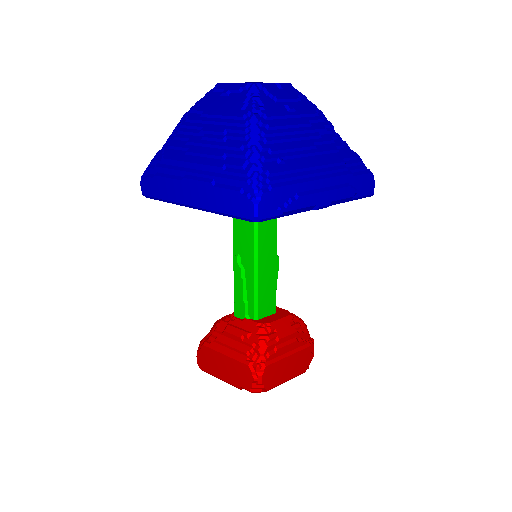}
    \end{subfigure}%
    \hfill%
    \begin{subfigure}[b]{0.20\linewidth}
        \centering
        \includegraphics[width=0.8\linewidth]{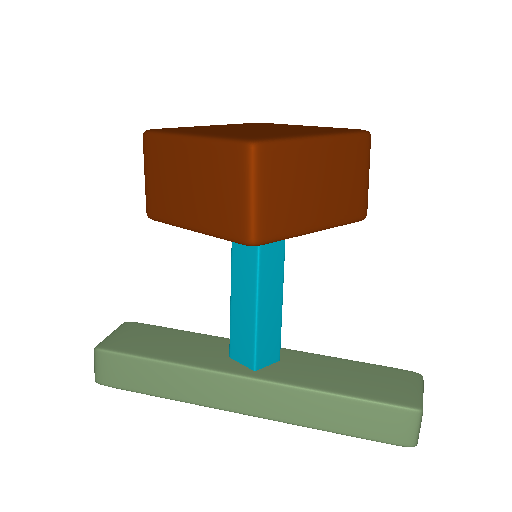}
    \end{subfigure}%
    \hfill%
    \begin{subfigure}[b]{0.20\linewidth}
	\centering
        \includegraphics[width=0.8\linewidth]{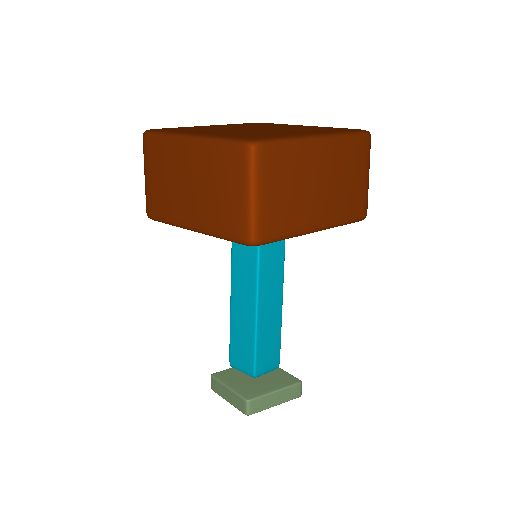}
    \end{subfigure}%
    \hfill%
    \begin{subfigure}[b]{0.20\linewidth}
	\centering
        \includegraphics[width=0.8\linewidth]{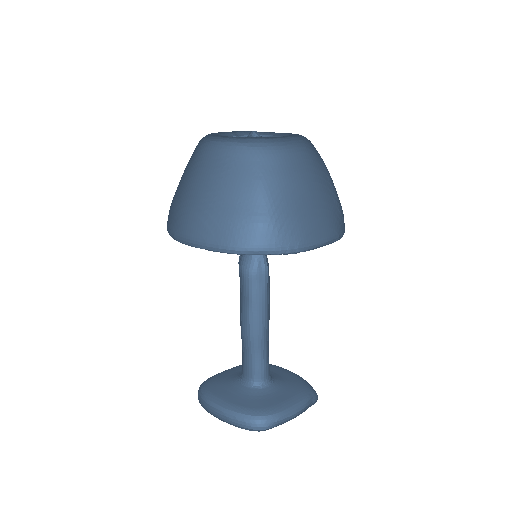}
    \end{subfigure}%
    \vskip\baselineskip%
    \vspace{-1.25em}
     \begin{subfigure}[b]{0.20\linewidth}
        \centering
        \includegraphics[width=0.8\linewidth]{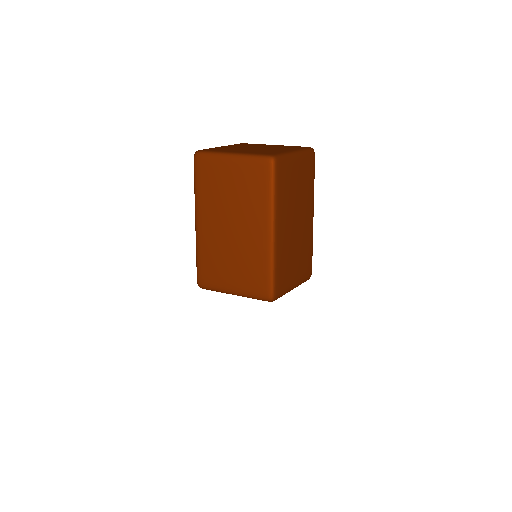}
    \end{subfigure}%
    \hfill%
    \begin{subfigure}[b]{0.20\linewidth}
	\centering
        \includegraphics[width=0.8\linewidth]{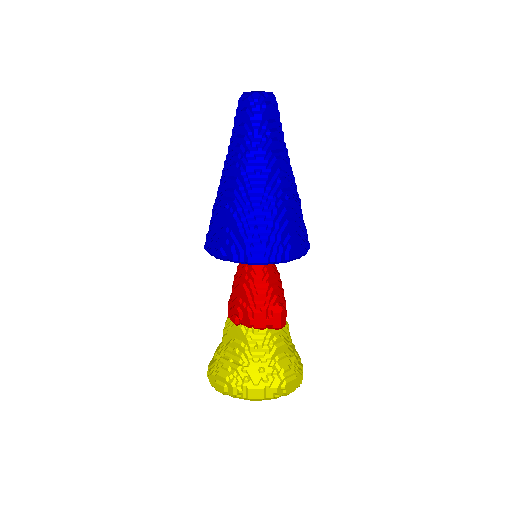}
    \end{subfigure}%
    \hfill%
    \begin{subfigure}[b]{0.20\linewidth}
        \centering
        \includegraphics[width=0.8\linewidth]{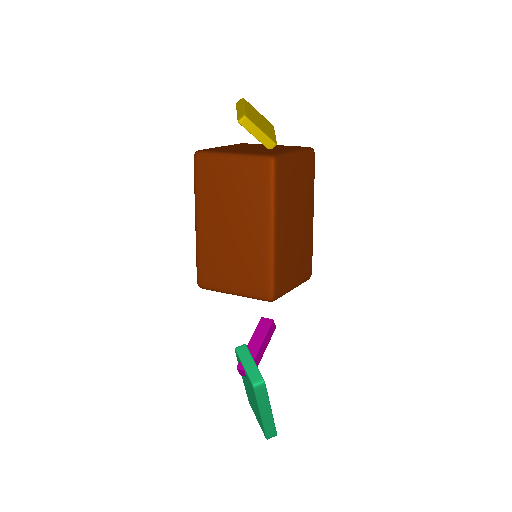}
    \end{subfigure}%
    \hfill%
    \begin{subfigure}[b]{0.20\linewidth}
	\centering
        \includegraphics[width=0.8\linewidth]{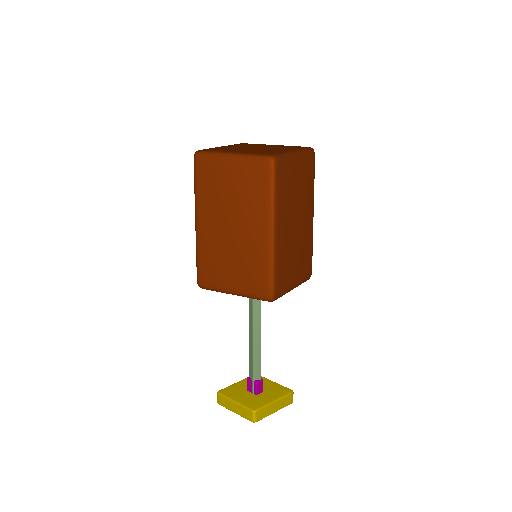}
    \end{subfigure}%
    \hfill%
    \begin{subfigure}[b]{0.20\linewidth}
	\centering
        \includegraphics[width=0.8\linewidth]{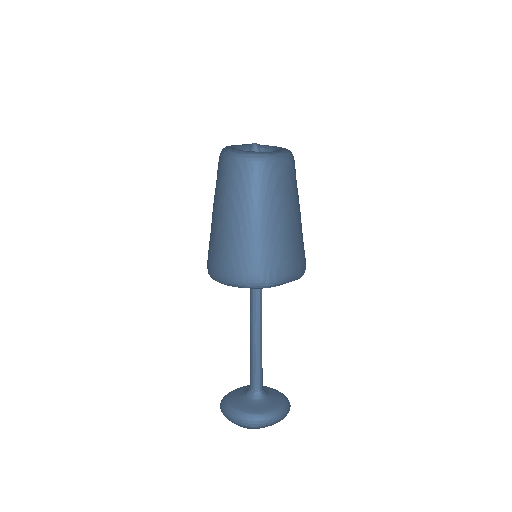}
    \end{subfigure}%
     \vskip\baselineskip%
    \vspace{-1.25em}
     \begin{subfigure}[b]{0.20\linewidth}
        \centering
        \includegraphics[width=0.8\linewidth]{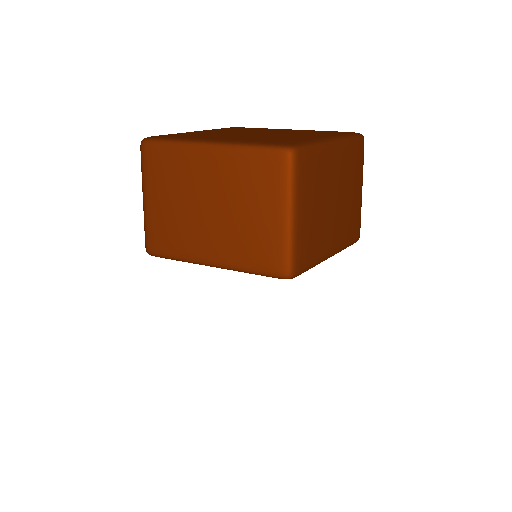}
    \end{subfigure}%
    \hfill%
    \begin{subfigure}[b]{0.20\linewidth}
	\centering
        \includegraphics[width=0.8\linewidth]{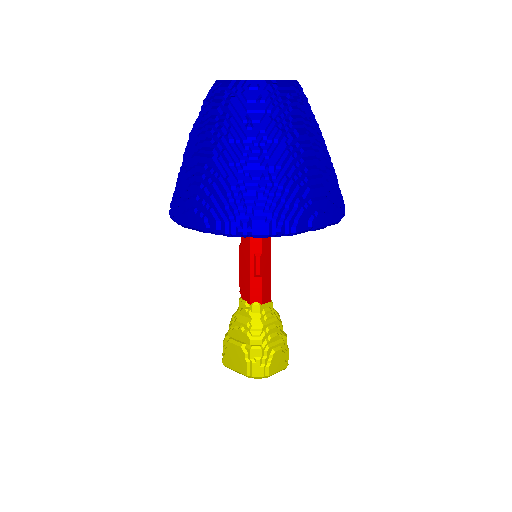}
    \end{subfigure}%
    \hfill%
    \begin{subfigure}[b]{0.20\linewidth}
        \centering
        \includegraphics[width=0.8\linewidth]{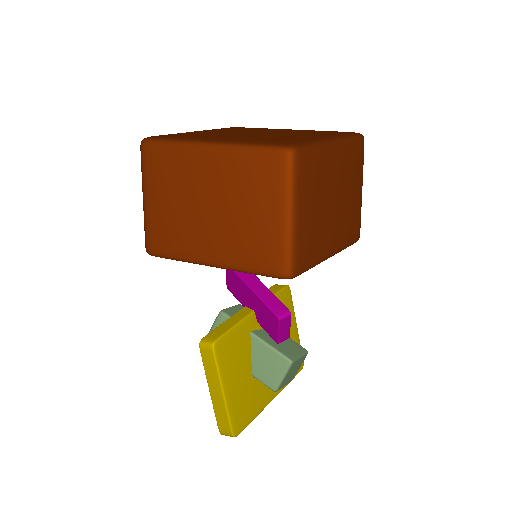}
    \end{subfigure}%
    \hfill%
    \begin{subfigure}[b]{0.20\linewidth}
	\centering
        \includegraphics[width=0.8\linewidth]{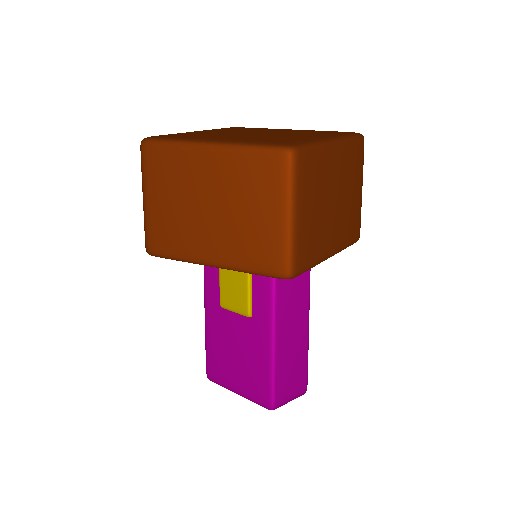}
    \end{subfigure}%
    \hfill%
    \begin{subfigure}[b]{0.20\linewidth}
	\centering
        \includegraphics[width=0.8\linewidth]{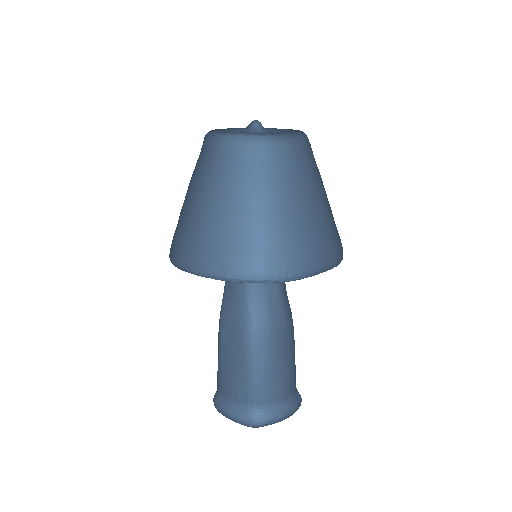}
    \end{subfigure}%
    \vskip\baselineskip%
    \vspace{-1.5em}
    \caption{{\bf Shape Completion Results on Lamps}. Starting from
    partial lamps, we show completions of our model,
    ATISS~\cite{Paschalidou2021NEURIPS} and PQ-NET~\cite{Wu2020CVPR}.}
    \label{fig:shapenet_qualitative_completion_comparison_lamps_supp}
\end{figure*}

%% file: fig/text_guided_generation_free_text_supp.tex
\begin{figure*}
    \centering
    \vspace{-1.2em}
    \vskip\baselineskip%
    \begin{subfigure}[b]{0.15\linewidth}
	\centering
	\includegraphics[trim={0 0 0 0cm},clip,width=\linewidth]{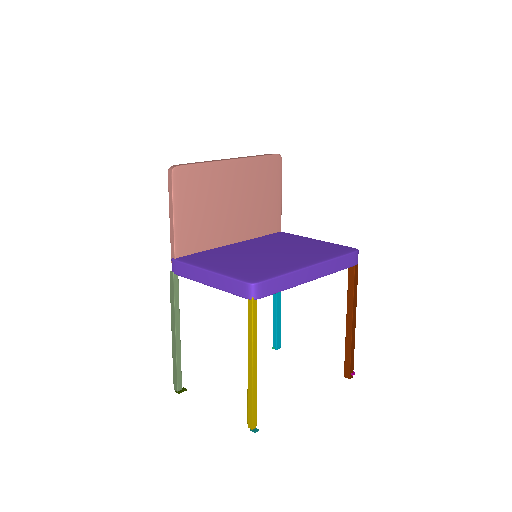}
    \end{subfigure}%
    \hfill%
     \begin{subfigure}[b]{0.15\linewidth}
	\centering
	\includegraphics[trim={0 0 0 0cm},clip,width=\linewidth]{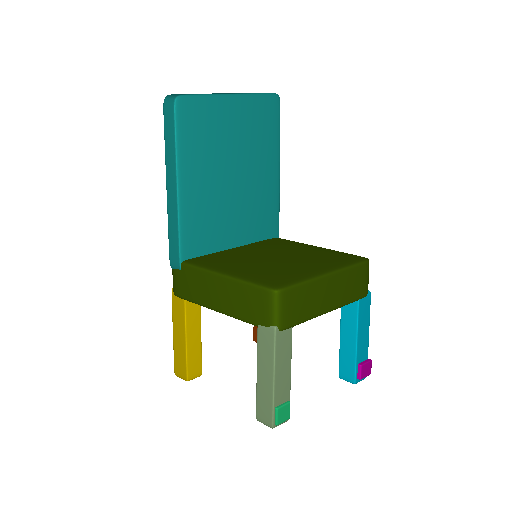}
    \end{subfigure}%
    \hfill%
    \begin{subfigure}[b]{0.15\linewidth}
        \centering
       \includegraphics[trim={0 0 0 0cm},clip,width=\linewidth]{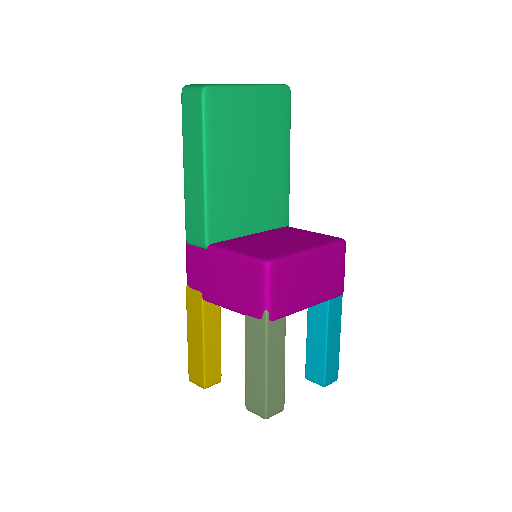}
    \end{subfigure}%
    \hfill%
    \begin{subfigure}[b]{0.15\linewidth}
	\centering
	\includegraphics[trim={0 0 0 0cm},clip,width=\linewidth]{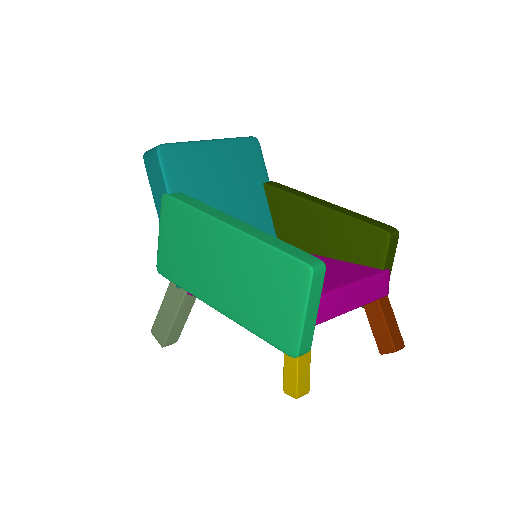}
    \end{subfigure}%
    \hfill%
     \begin{subfigure}[b]{0.15\linewidth}
	\centering
	\includegraphics[trim={0 0 0 0cm},clip,width=\linewidth]{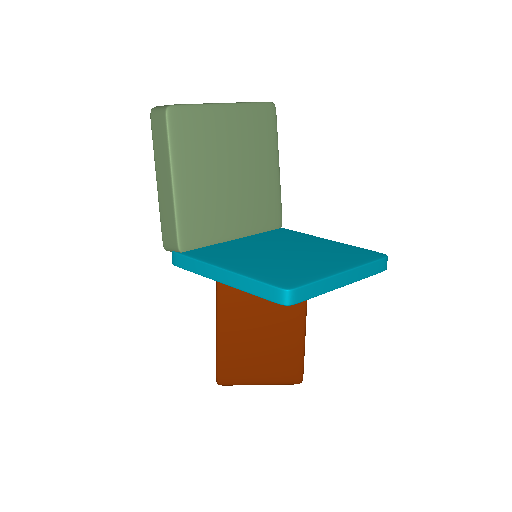}
    \end{subfigure}%
    \hfill%
    \begin{subfigure}[b]{0.15\linewidth}
        \centering
       \includegraphics[trim={0 0 0 0cm},clip,width=\linewidth]{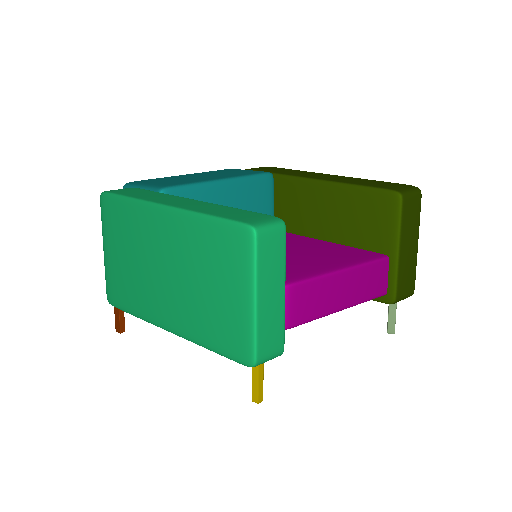}
    \end{subfigure}%
     \vskip\baselineskip%
    \vspace{-1.6em}
    \begin{subfigure}[b]{0.15\linewidth}
	\centering
	\includegraphics[trim={0 0 0 0cm},clip,width=\linewidth]{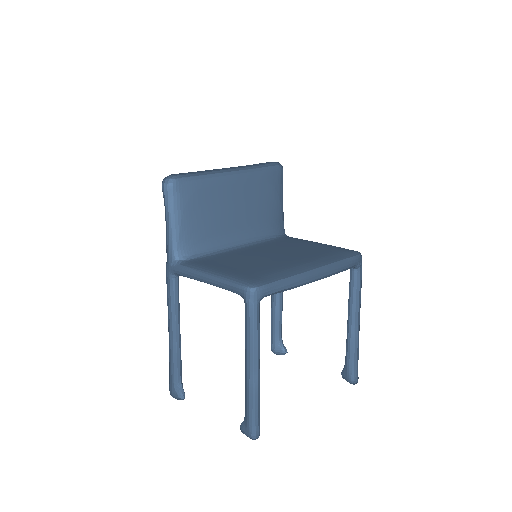}
    \end{subfigure}%
    \hfill%
     \begin{subfigure}[b]{0.15\linewidth}
	\centering
	\includegraphics[trim={0 0 0 0cm},clip,width=\linewidth]{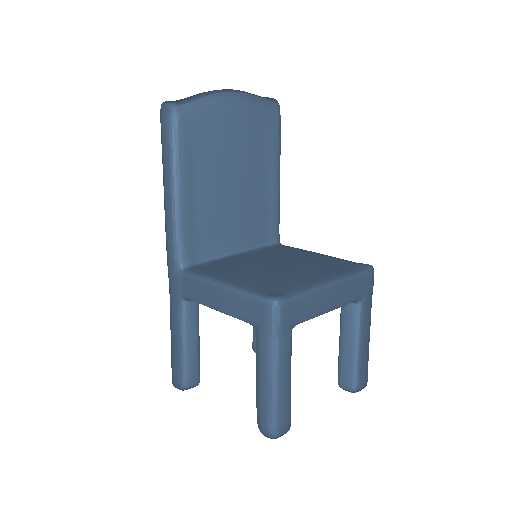}
    \end{subfigure}%
    \hfill%
    \begin{subfigure}[b]{0.15\linewidth}
        \centering
       \includegraphics[trim={0 0 0 0cm},clip,width=\linewidth]{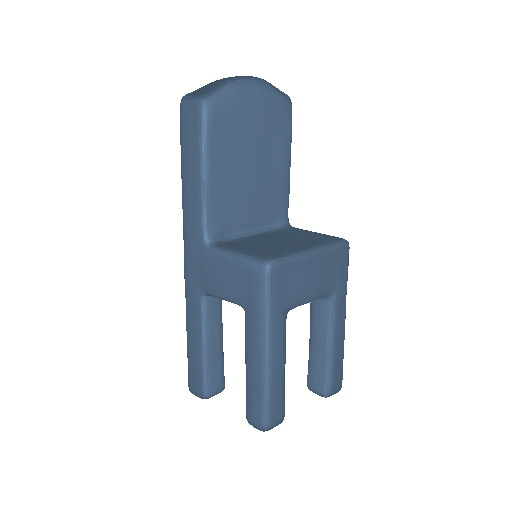}
    \end{subfigure}%
    \hfill%
    \begin{subfigure}[b]{0.15\linewidth}
	\centering
	\includegraphics[trim={0 0 0 0cm},clip,width=\linewidth]{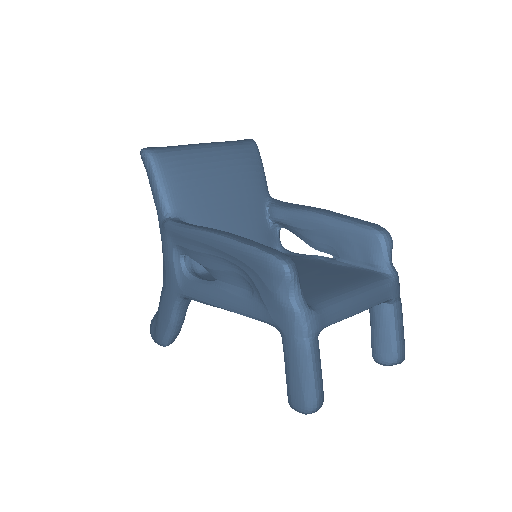}
    \end{subfigure}%
    \hfill%
     \begin{subfigure}[b]{0.15\linewidth}
	\centering
	\includegraphics[trim={0 0 0 0cm},clip,width=\linewidth]{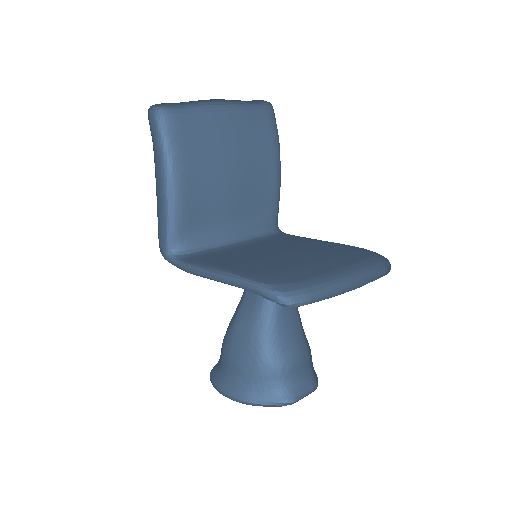}
    \end{subfigure}%
    \hfill%
    \begin{subfigure}[b]{0.15\linewidth}
        \centering
       \includegraphics[trim={0 0 0 0cm},clip,width=\linewidth]{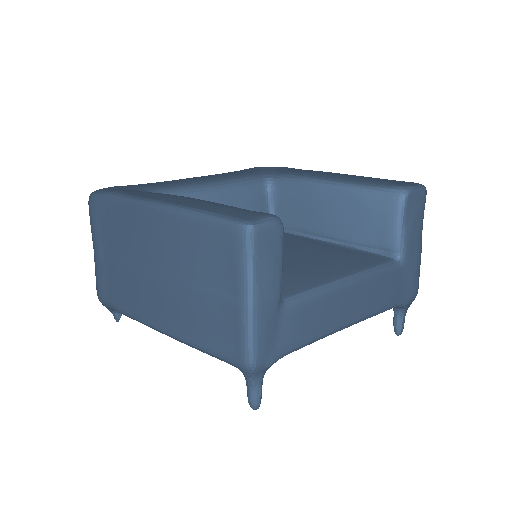}
    \end{subfigure}%
    \vskip\baselineskip%
    \vspace{-1.5em}
    \begin{subfigure}[t]{0.14\linewidth}
	\small{\textit{A short chair with four legs.}}
    \end{subfigure}%
    \hfill%
     \begin{subfigure}[t]{0.14\linewidth}
        \small{\textit{A tall chair with four legs.}}
    \end{subfigure}%
    \hfill%
    \begin{subfigure}[t]{0.14\linewidth}
        \small{\textit{A chair with four legs and a round back.}}
    \end{subfigure}%
    \hfill%
     \begin{subfigure}[t]{0.14\linewidth}
	\small{\textit{A chair with inclined back and four legs.}}
    \end{subfigure}%
    \hfill%
     \begin{subfigure}[t]{0.14\linewidth}
        \small{\textit{A chair with a pedestal, a seat, and a back.}}
    \end{subfigure}%
    \hfill%
    \begin{subfigure}[t]{0.14\linewidth}
        \small{\textit{A sofa.}}
    \end{subfigure}%
    \vspace{-1.2em}
    \vskip\baselineskip%
      \begin{subfigure}[b]{0.15\linewidth}
	\centering
	\includegraphics[trim={0 0 0 0cm},clip,width=\linewidth]{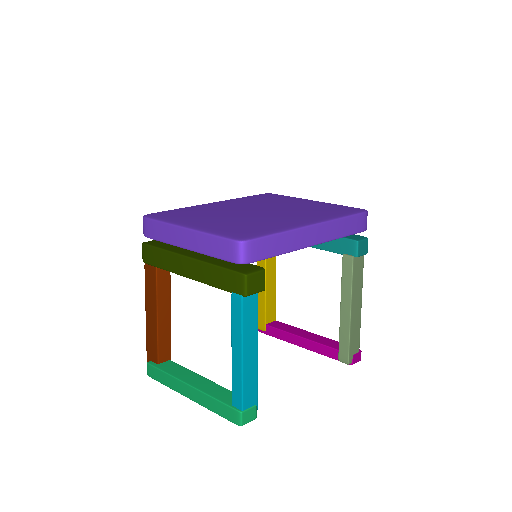}
    \end{subfigure}%
    \hfill%
     \begin{subfigure}[b]{0.15\linewidth}
	\centering
	\includegraphics[trim={0 0 0 0cm},clip,width=\linewidth]{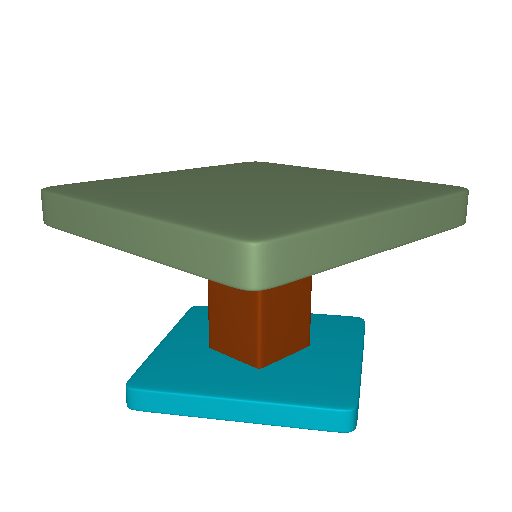}
    \end{subfigure}%
    \hfill%
    \begin{subfigure}[b]{0.15\linewidth}
        \centering
       \includegraphics[trim={0 0 0 0cm},clip,width=\linewidth]{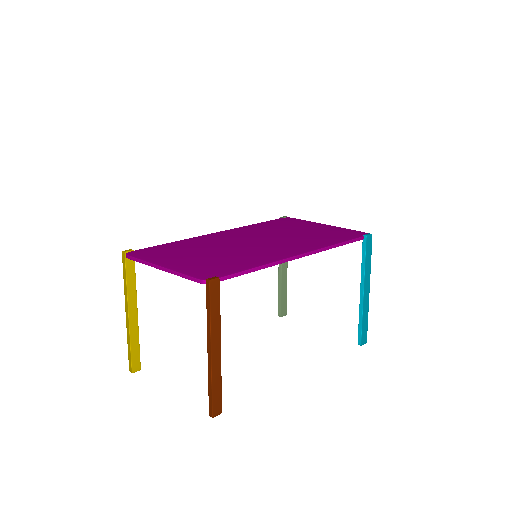}
    \end{subfigure}%
    \hfill%
    \begin{subfigure}[b]{0.15\linewidth}
	\centering
	\includegraphics[trim={0 0 0 0cm},clip,width=\linewidth]{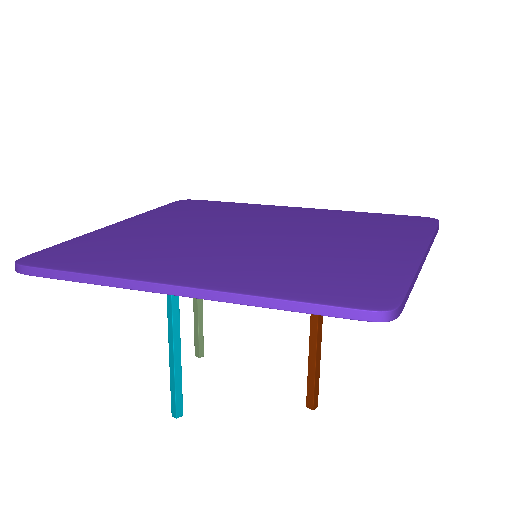}
    \end{subfigure}%
    \hfill%
     \begin{subfigure}[b]{0.15\linewidth}
	\centering
	\includegraphics[trim={0 0 0 0cm},clip,width=\linewidth]{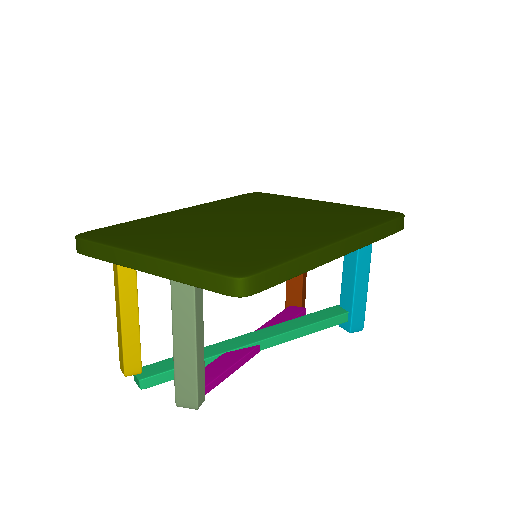}
    \end{subfigure}%
    \hfill%
    \begin{subfigure}[b]{0.15\linewidth}
        \centering
       \includegraphics[trim={0 0 0 0cm},clip,width=\linewidth]{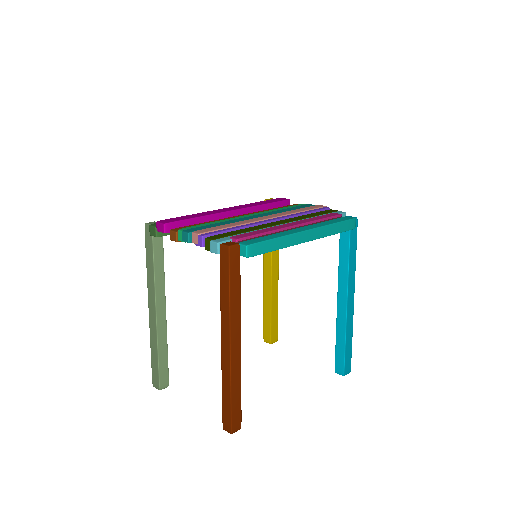}
    \end{subfigure}%
     \vskip\baselineskip%
    \vspace{-1.6em}
    \begin{subfigure}[b]{0.15\linewidth}
	\centering
	\includegraphics[trim={0 0 0 0cm},clip,width=\linewidth]{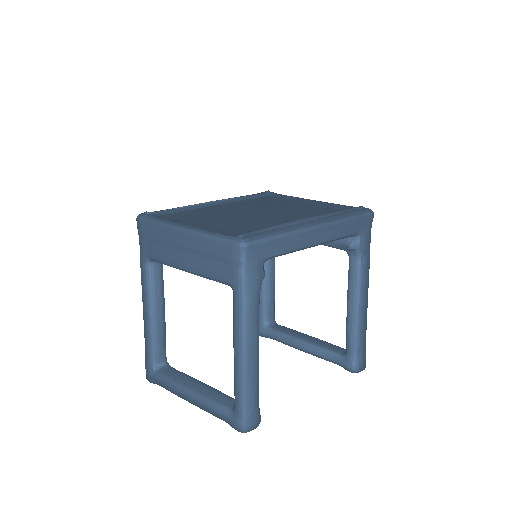}
    \end{subfigure}%
    \hfill%
     \begin{subfigure}[b]{0.15\linewidth}
	\centering
	\includegraphics[trim={0 0 0 0cm},clip,width=\linewidth]{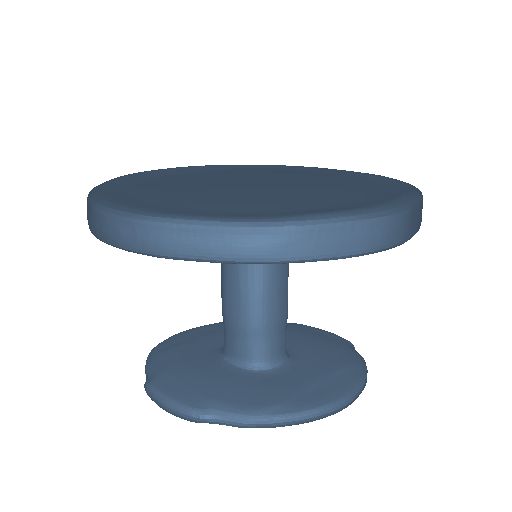}
    \end{subfigure}%
    \hfill%
    \begin{subfigure}[b]{0.15\linewidth}
        \centering
       \includegraphics[trim={0 0 0 0cm},clip,width=\linewidth]{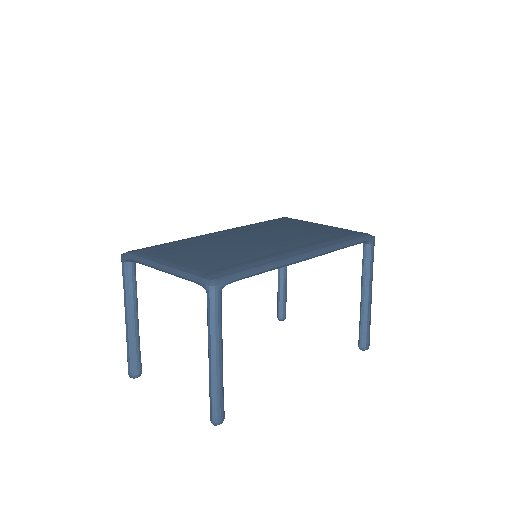}
    \end{subfigure}%
    \hfill%
    \begin{subfigure}[b]{0.15\linewidth}
	\centering
	\includegraphics[trim={0 0 0 0cm},clip,width=\linewidth]{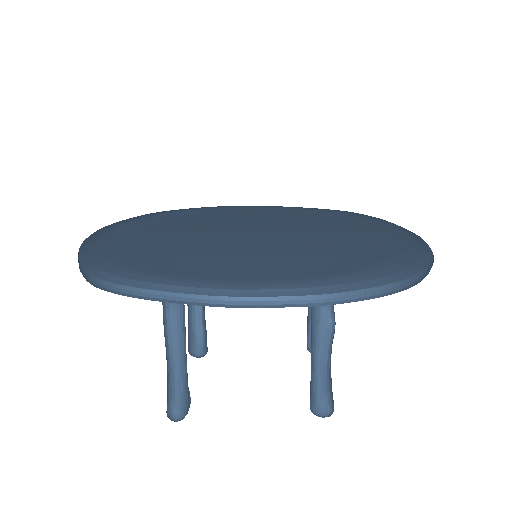}
    \end{subfigure}%
    \hfill%
     \begin{subfigure}[b]{0.15\linewidth}
	\centering
	\includegraphics[trim={0 0 0 0cm},clip,width=\linewidth]{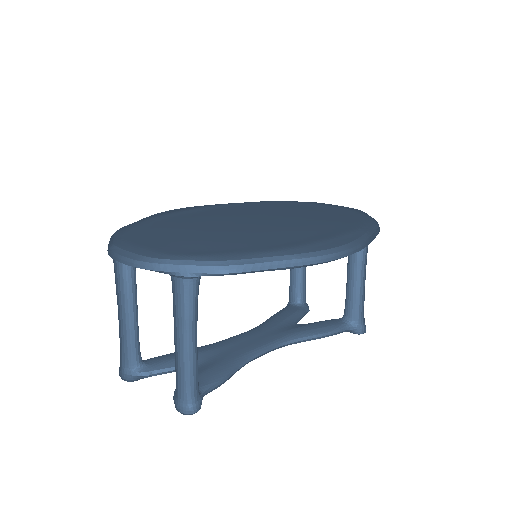}
    \end{subfigure}%
    \hfill%
    \begin{subfigure}[b]{0.15\linewidth}
        \centering
       \includegraphics[trim={0 0 0 0cm},clip,width=\linewidth]{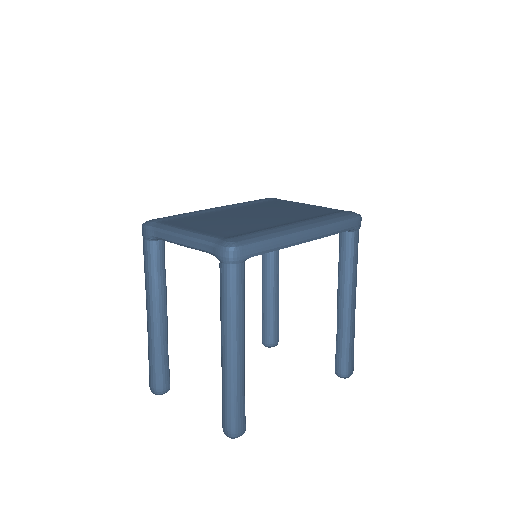}
    \end{subfigure}%
    \vskip\baselineskip%
    \vspace{-1.5em}
    \begin{subfigure}[t]{0.14\linewidth}
	\small{\textit{A narrow table with four legs and two runners.}}
    \end{subfigure}%
    \hfill%
     \begin{subfigure}[t]{0.14\linewidth}
        \small{\textit{A wide table with a central support and a board.}}
    \end{subfigure}%
    \hfill%
    \begin{subfigure}[t]{0.14\linewidth}
        \small{\textit{A rectangle table.}}
    \end{subfigure}%
    \hfill%
     \begin{subfigure}[t]{0.14\linewidth}
	\small{\textit{A table with four legs and a round tabletop.}}
    \end{subfigure}%
    \hfill%
     \begin{subfigure}[t]{0.14\linewidth}
        \small{\textit{A table with four legs, two runners, and one board.}}
    \end{subfigure}%
    \hfill%
    \begin{subfigure}[t]{0.14\linewidth}
        \small{\textit{A table with four legs and twelve bars.}}
    \end{subfigure}%
    \vspace{-1.2em}
    \vskip\baselineskip%
      \begin{subfigure}[b]{0.15\linewidth}
	\centering
	\includegraphics[trim={0 0 0 0cm},clip,width=\linewidth]{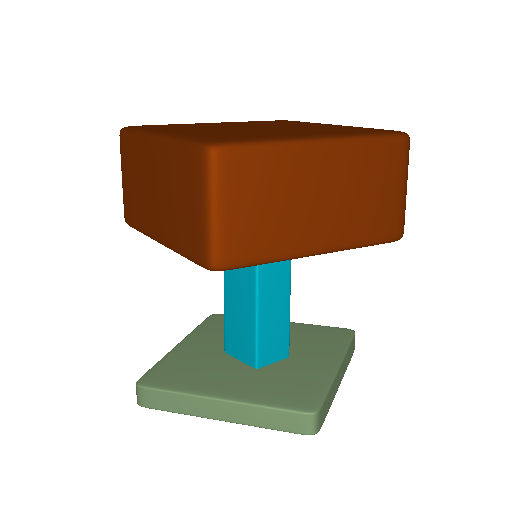}
    \end{subfigure}%
    \hfill%
     \begin{subfigure}[b]{0.15\linewidth}
	\centering
	\includegraphics[trim={0 0 0 0cm},clip,width=\linewidth]{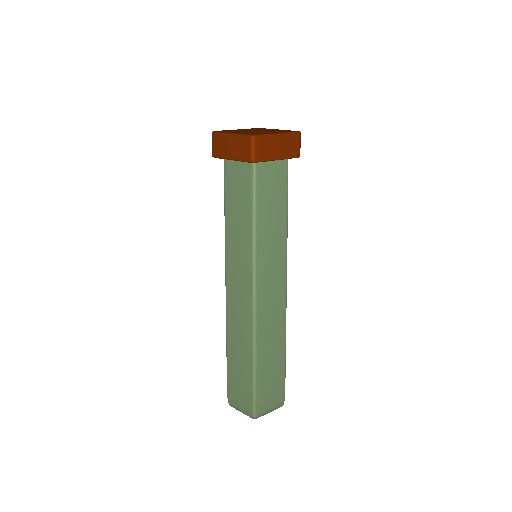}
    \end{subfigure}%
    \hfill%
    \begin{subfigure}[b]{0.15\linewidth}
        \centering
       \includegraphics[trim={0 0 0 0cm},clip,width=\linewidth]{pasta_size/lamps/completed/14242_050505.png}
    \end{subfigure}%
    \hfill%
    \begin{subfigure}[b]{0.15\linewidth}
	\centering
	\includegraphics[trim={0 0 0 0cm},clip,width=\linewidth]{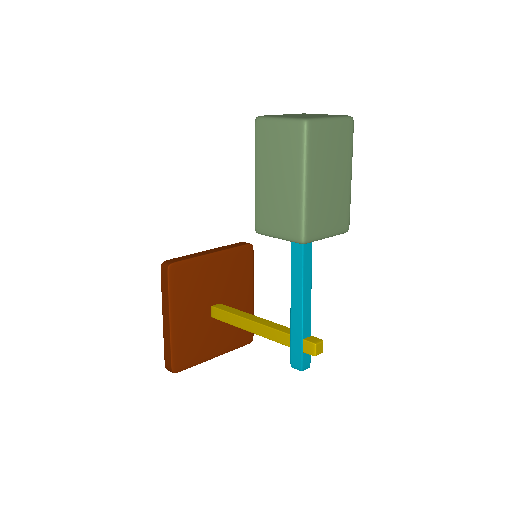}
    \end{subfigure}%
    \hfill%
     \begin{subfigure}[b]{0.15\linewidth}
	\centering
	\includegraphics[trim={0 0 0 0cm},clip,width=\linewidth]{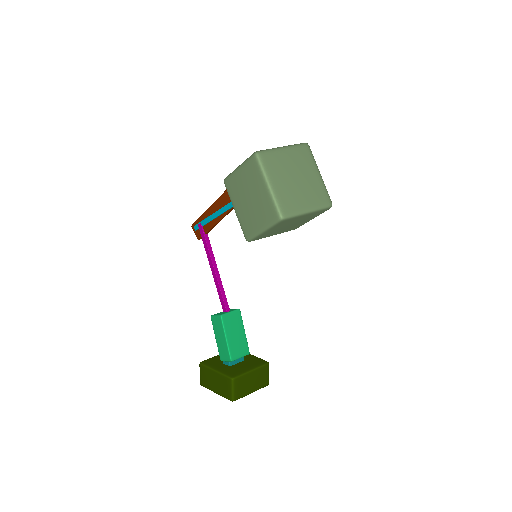}
    \end{subfigure}%
    \hfill%
    \begin{subfigure}[b]{0.15\linewidth}
        \centering
       \includegraphics[trim={0 0 0 0cm},clip,width=\linewidth]{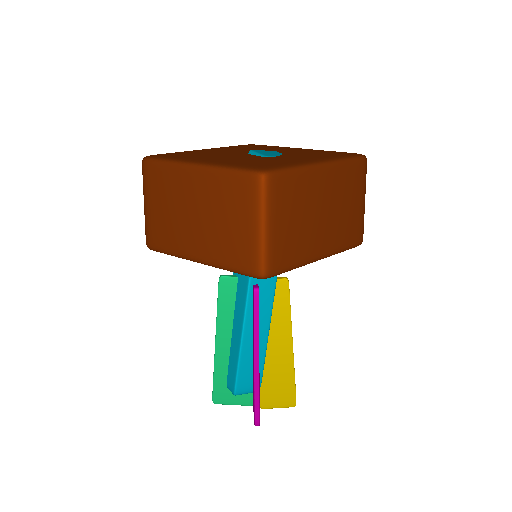}
    \end{subfigure}%
     \vskip\baselineskip%
    \vspace{-1.6em}
    \begin{subfigure}[b]{0.15\linewidth}
	\centering
	\includegraphics[trim={0 0 0 0cm},clip,width=\linewidth]{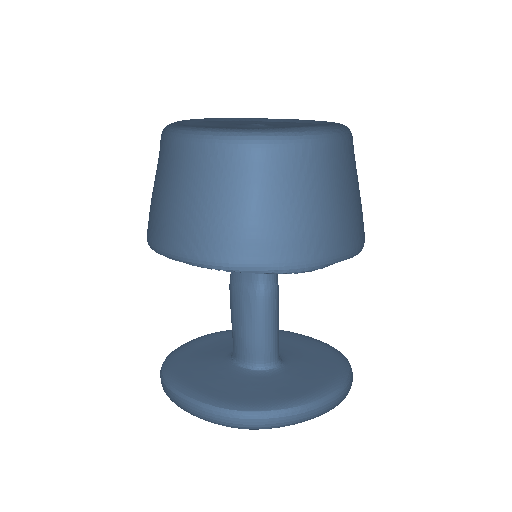}
    \end{subfigure}%
    \hfill%
     \begin{subfigure}[b]{0.15\linewidth}
	\centering
	\includegraphics[trim={0 0 0 0cm},clip,width=\linewidth]{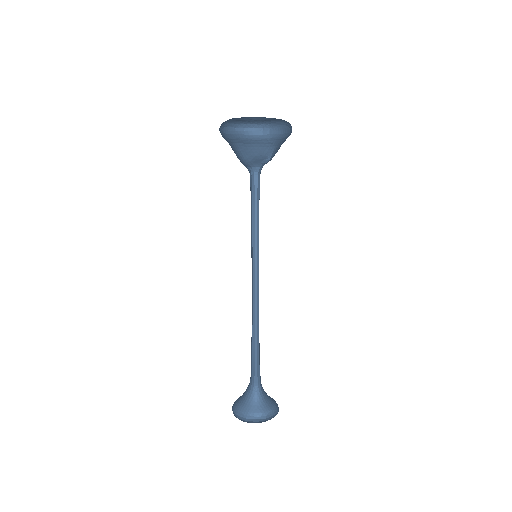}
    \end{subfigure}%
    \hfill%
    \begin{subfigure}[b]{0.15\linewidth}
        \centering
       \includegraphics[trim={0 0 0 0cm},clip,width=\linewidth]{pasta_size/lamps/decoder/14242_050505.png}
    \end{subfigure}%
    \hfill%
    \begin{subfigure}[b]{0.15\linewidth}
	\centering
	\includegraphics[trim={0 0 0 0cm},clip,width=\linewidth]{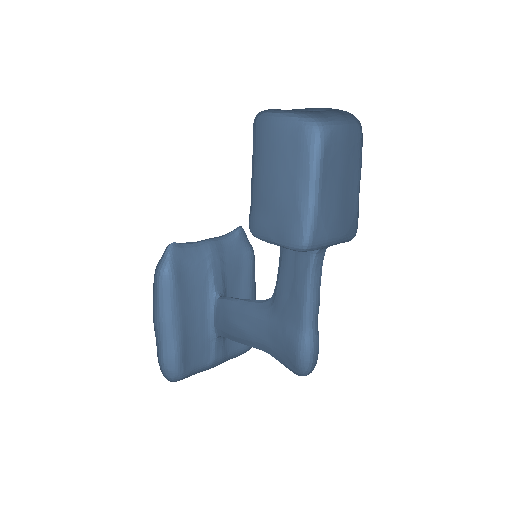}
    \end{subfigure}%
    \hfill%
     \begin{subfigure}[b]{0.15\linewidth}
	\centering
	\includegraphics[trim={0 0 0 0cm},clip,width=\linewidth]{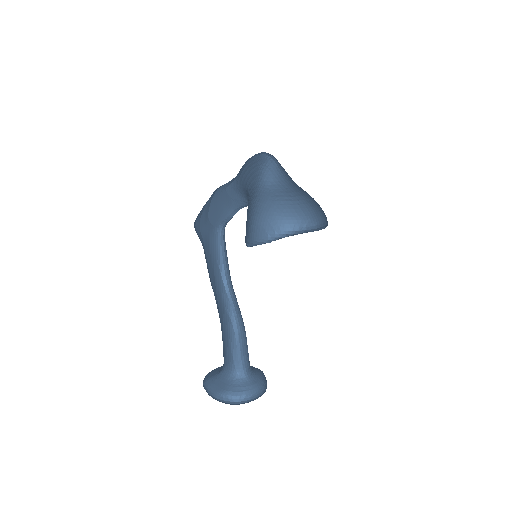}
    \end{subfigure}%
    \hfill%
    \begin{subfigure}[b]{0.15\linewidth}
        \centering
       \includegraphics[trim={0 0 0 0cm},clip,width=\linewidth]{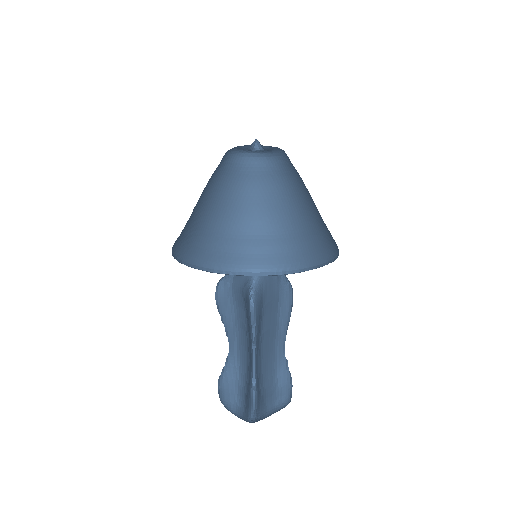}
    \end{subfigure}%
    \vskip\baselineskip%
    \vspace{-1.5em}
    \begin{subfigure}[t]{0.14\linewidth}
	\small{\textit{A chubby lamp with a round lamp shade.}}
    \end{subfigure}%
    \hfill%
     \begin{subfigure}[t]{0.14\linewidth}
        \small{\textit{A thin and tall lamp.}}
    \end{subfigure}%
    \hfill%
    \begin{subfigure}[t]{0.14\linewidth}
        \small{\textit{A floor lamp.}}
    \end{subfigure}%
    \hfill%
     \begin{subfigure}[t]{0.14\linewidth}
	\small{\textit{A door lamp extending out from a straight arm.}}
    \end{subfigure}%
    \hfill%
     \begin{subfigure}[t]{0.14\linewidth}
        \small{\textit{A table lamp with a bent body connected to a base.}}
    \end{subfigure}%
    \hfill%
    \begin{subfigure}[t]{0.14\linewidth}
        \small{\textit{A lamp with leaf-like base part.}}
    \end{subfigure}%
    \vskip\baselineskip%
    \vspace{-1.2em}
    \caption{{\bf Text-guided Shape Generation}. Given different text descriptions 
    our model can generate plausible 3D shapes of chairs, tables, and lamps.}
    \label{fig:text_guided_generation_supp}
\end{figure*}

%% file: fig/shape_generation_image_guided_supp.tex
\begin{figure*}
    \centering
    \begin{subfigure}[b]{0.11\linewidth}
	\centering
        Input Image
    \end{subfigure}%
    \hfill%
    \begin{subfigure}[b]{0.11\linewidth}
	\centering
        Ours-Parts
    \end{subfigure}%
    \hfill%
    \begin{subfigure}[b]{0.11\linewidth}
	\centering
        Ours
    \end{subfigure}%
    \hfill%
     \begin{subfigure}[b]{0.11\linewidth}
	\centering
        Input Image
    \end{subfigure}%
    \hfill%
    \begin{subfigure}[b]{0.11\linewidth}
	\centering
        Ours-Parts
    \end{subfigure}%
    \hfill%
    \begin{subfigure}[b]{0.11\linewidth}
	\centering
        Ours
    \end{subfigure}%
    \hfill%
     \begin{subfigure}[b]{0.11\linewidth}
	\centering
        Input Image
    \end{subfigure}%
    \hfill%
    \begin{subfigure}[b]{0.11\linewidth}
	\centering
        Ours-Parts
    \end{subfigure}%
    \hfill%
    \begin{subfigure}[b]{0.11\linewidth}
	\centering
        Ours
    \end{subfigure}%
    \vskip\baselineskip%
    \vspace{-1.0em}
    \begin{subfigure}[b]{0.11\linewidth}
        \centering
	\includegraphics[width=\linewidth]{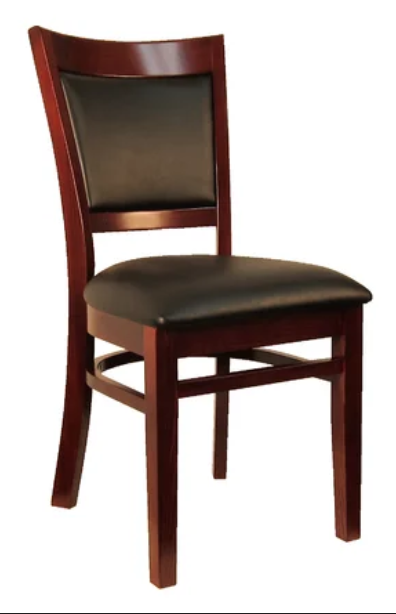}
    \end{subfigure}%
    \hfill%
    \begin{subfigure}[b]{0.11\linewidth}
        \centering
	\includegraphics[width=\linewidth]{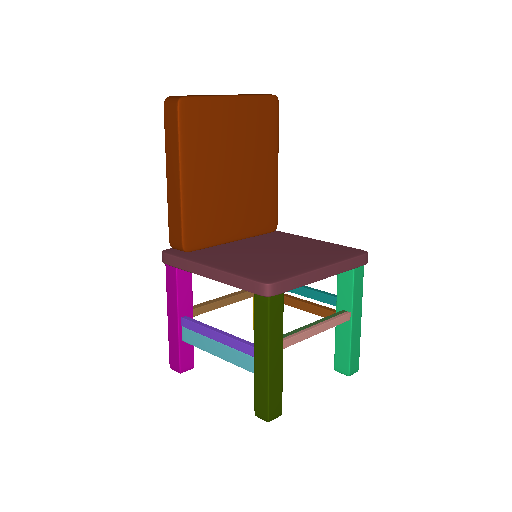}
    \end{subfigure}%
    \hfill%
    \begin{subfigure}[b]{0.11\linewidth}
	\centering
        \includegraphics[width=\linewidth]{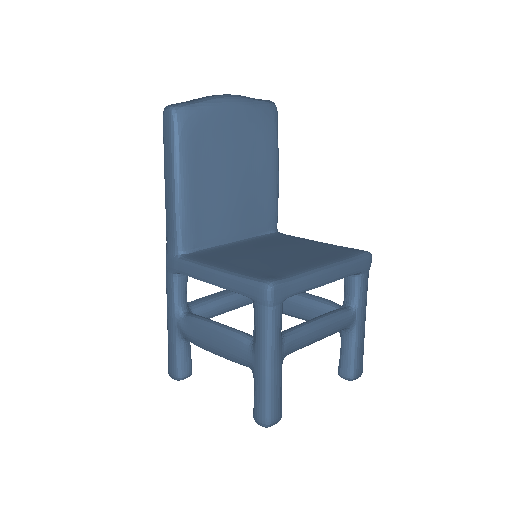}
    \end{subfigure}%
     \hfill%
    \begin{subfigure}[b]{0.11\linewidth}
        \centering
        \includegraphics[width=\linewidth]{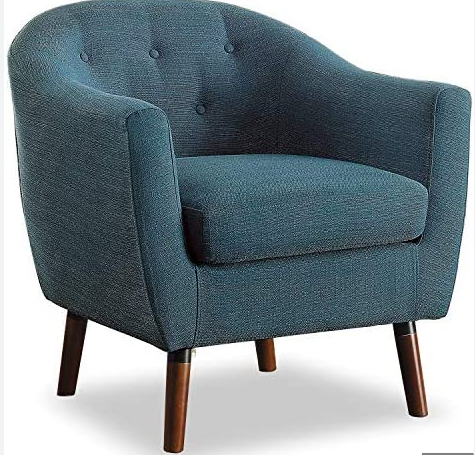}
    \end{subfigure}%
    \hfill%
    \begin{subfigure}[b]{0.11\linewidth}
        \centering
        \includegraphics[width=\linewidth]{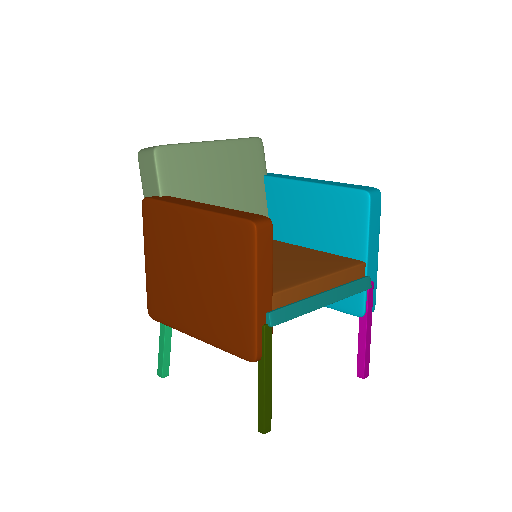}
    \end{subfigure}%
    \hfill%
    \begin{subfigure}[b]{0.11\linewidth}
	\centering
        \includegraphics[width=\linewidth]{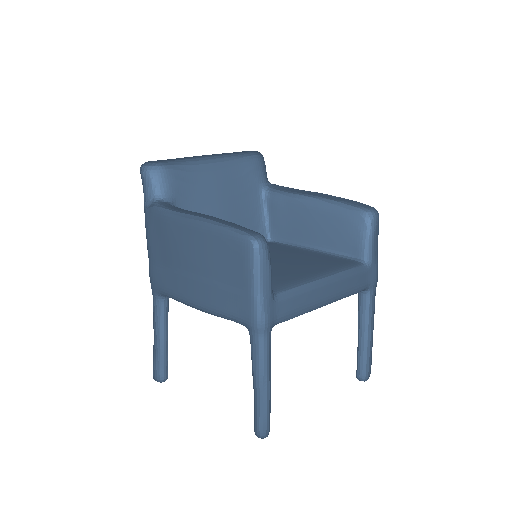}
    \end{subfigure}%
    \hfill%
    \begin{subfigure}[b]{0.11\linewidth}
        \centering
        \includegraphics[width=\linewidth]{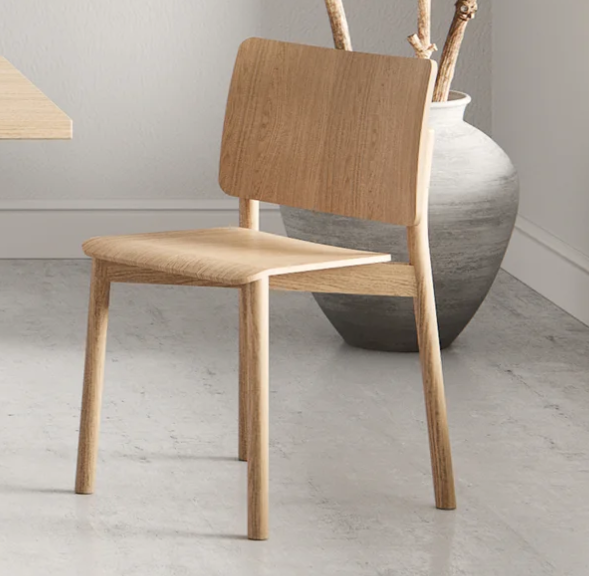}
    \end{subfigure}%
    \hfill%
    \begin{subfigure}[b]{0.11\linewidth}
        \centering
        \includegraphics[width=\linewidth]{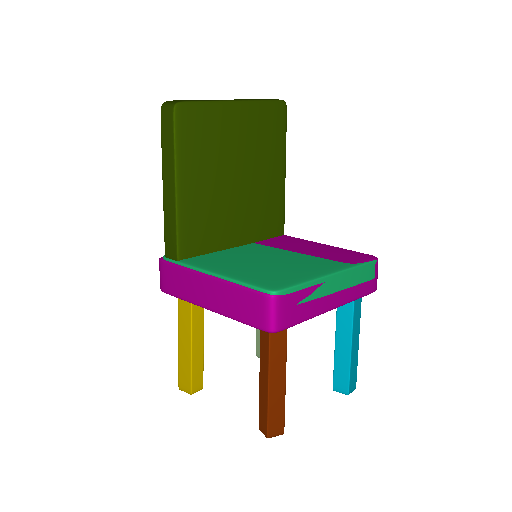}
    \end{subfigure}%
    \hfill%
    \begin{subfigure}[b]{0.11\linewidth}
	\centering
         \includegraphics[width=\linewidth]{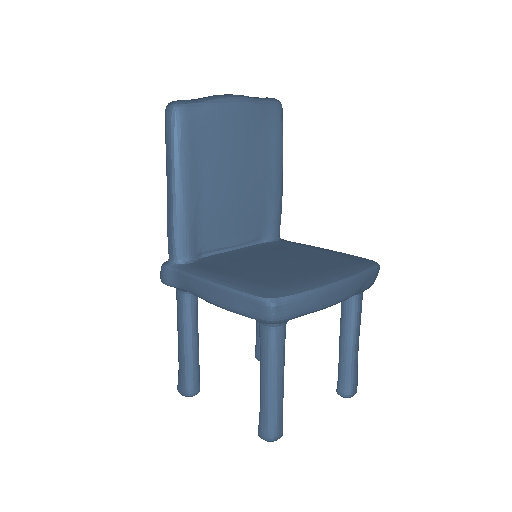}
    \end{subfigure}%
    \vskip\baselineskip%
    \begin{subfigure}[b]{0.11\linewidth}
        \centering
	\includegraphics[width=\linewidth]{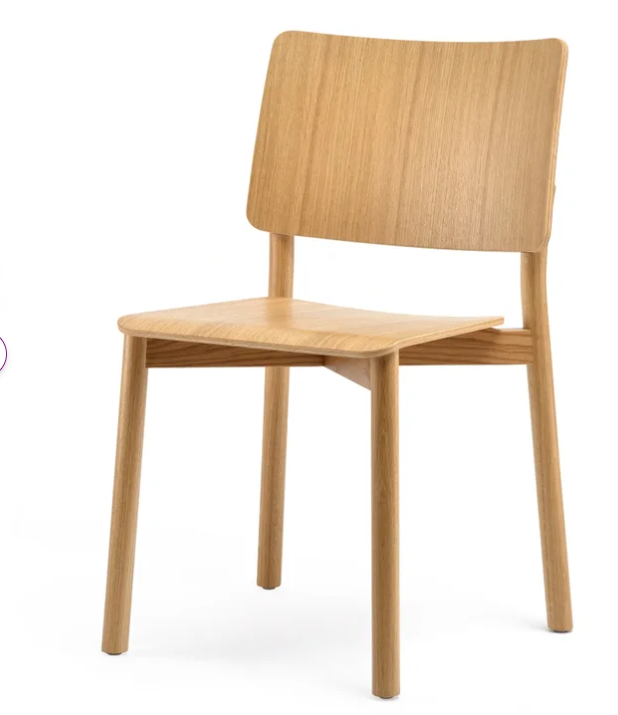}
    \end{subfigure}%
    \hfill%
    \begin{subfigure}[b]{0.11\linewidth}
        \centering
	\includegraphics[width=\linewidth]{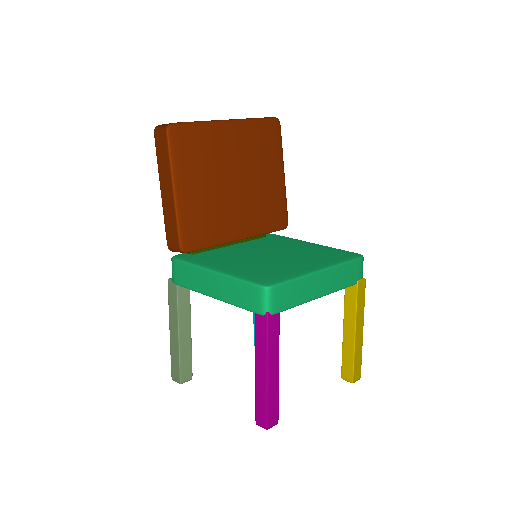}
    \end{subfigure}%
    \hfill%
    \begin{subfigure}[b]{0.11\linewidth}
	\centering
        \includegraphics[width=\linewidth]{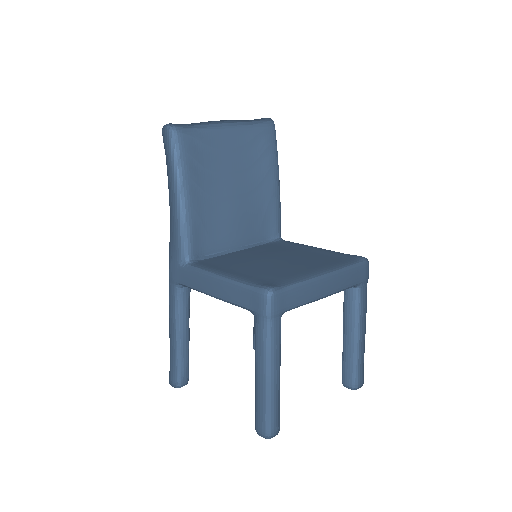}
    \end{subfigure}%
     \hfill%
    \begin{subfigure}[b]{0.11\linewidth}
        \centering
        \includegraphics[width=\linewidth]{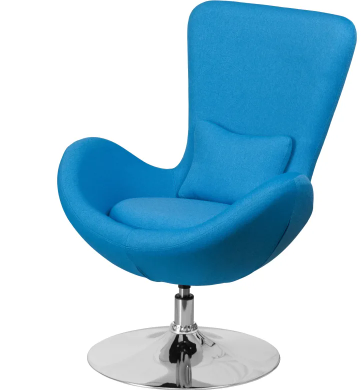}
    \end{subfigure}%
    \hfill%
    \begin{subfigure}[b]{0.11\linewidth}
        \centering
        \includegraphics[width=\linewidth]{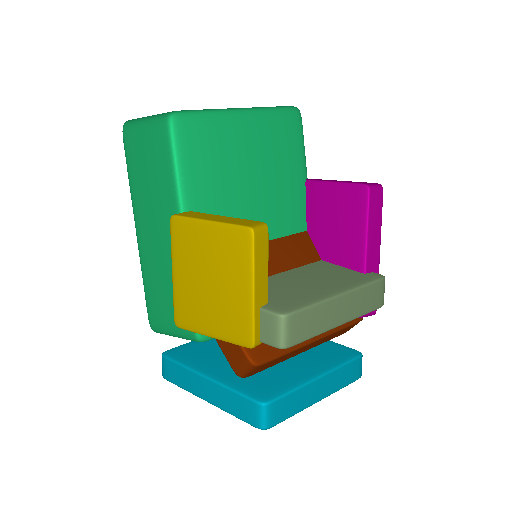}
    \end{subfigure}%
    \hfill%
    \begin{subfigure}[b]{0.11\linewidth}
	\centering
        \includegraphics[width=\linewidth]{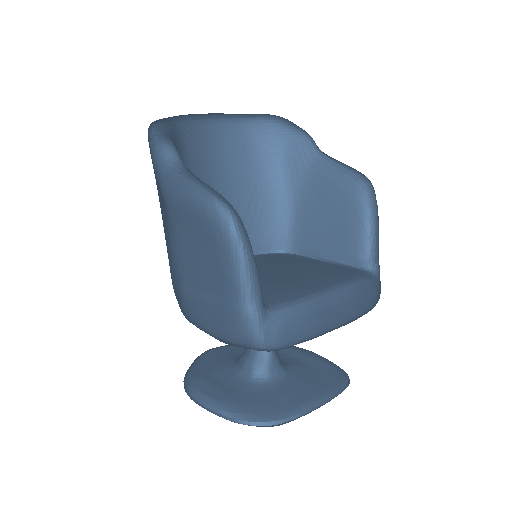}
    \end{subfigure}%
    \hfill%
    \begin{subfigure}[b]{0.11\linewidth}
        \centering
        \includegraphics[width=\linewidth]{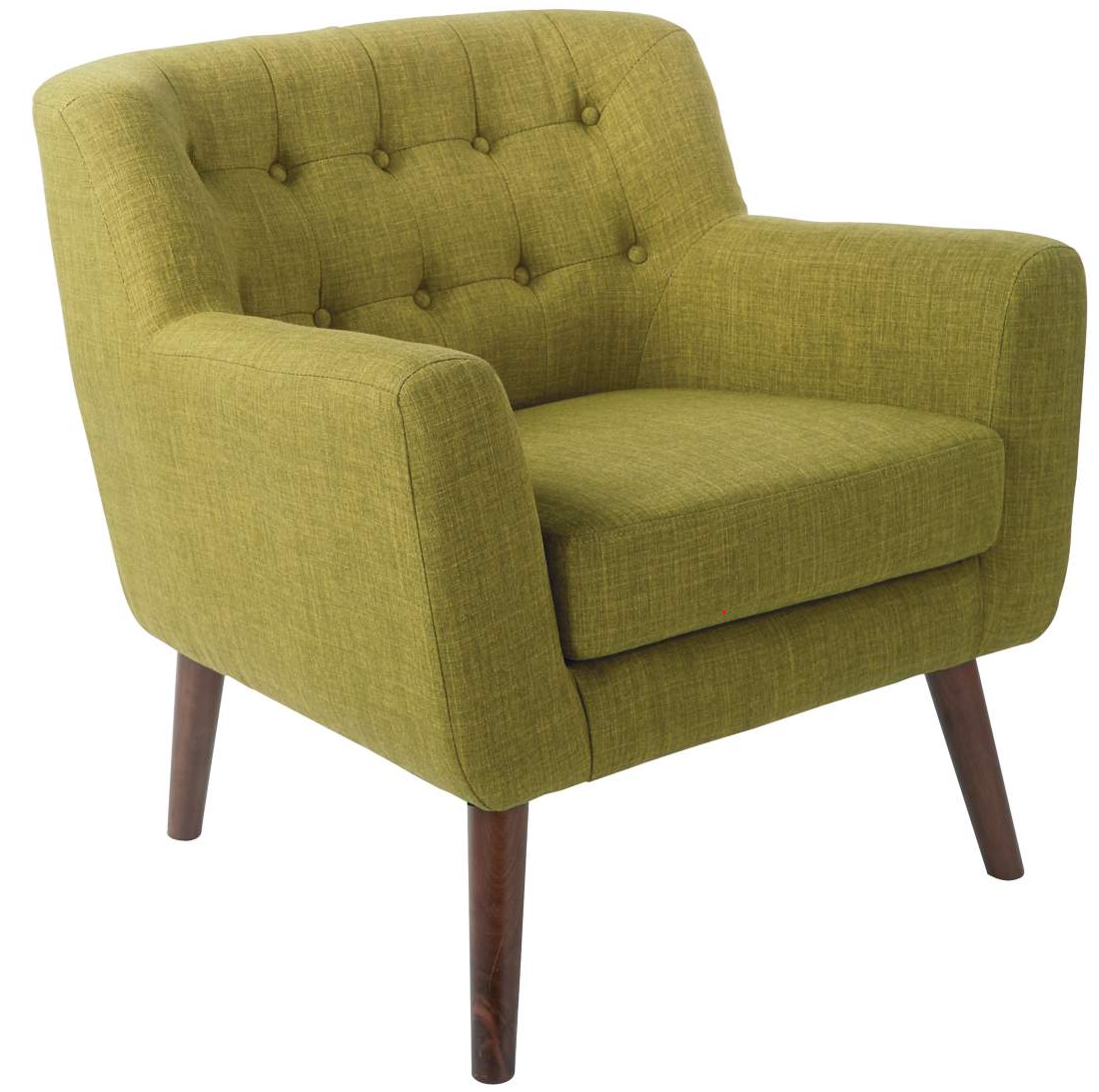}
    \end{subfigure}%
    \hfill%
    \begin{subfigure}[b]{0.11\linewidth}
        \centering
        \includegraphics[width=\linewidth]{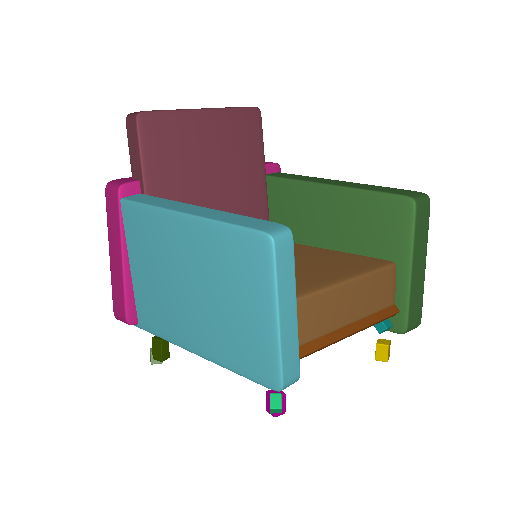}
    \end{subfigure}%
    \hfill%
    \begin{subfigure}[b]{0.11\linewidth}
	\centering
         \includegraphics[width=\linewidth]{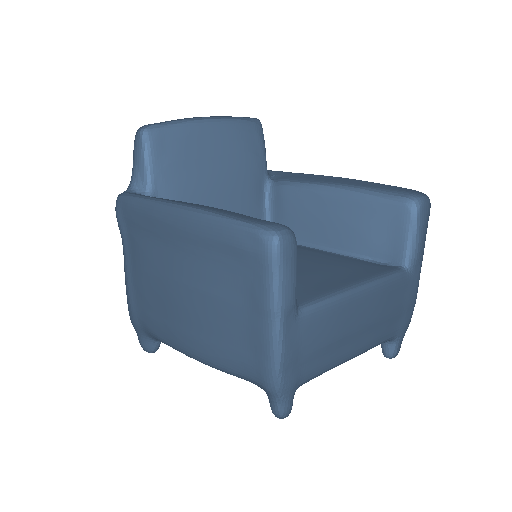}
    \end{subfigure}%
    \vskip\baselineskip%
    \begin{subfigure}[b]{0.11\linewidth}
        \centering
	\includegraphics[width=\linewidth]{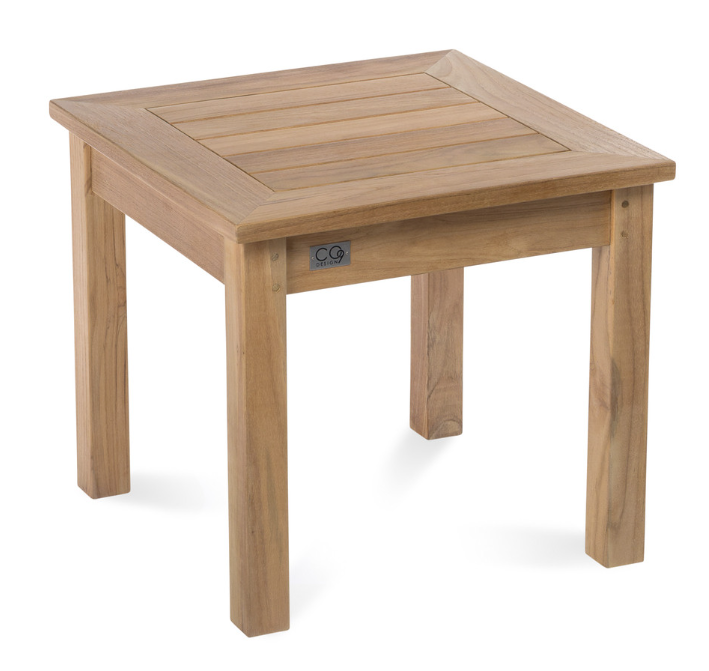}
    \end{subfigure}%
    \hfill%
    \begin{subfigure}[b]{0.11\linewidth}
        \centering
	\includegraphics[width=\linewidth]{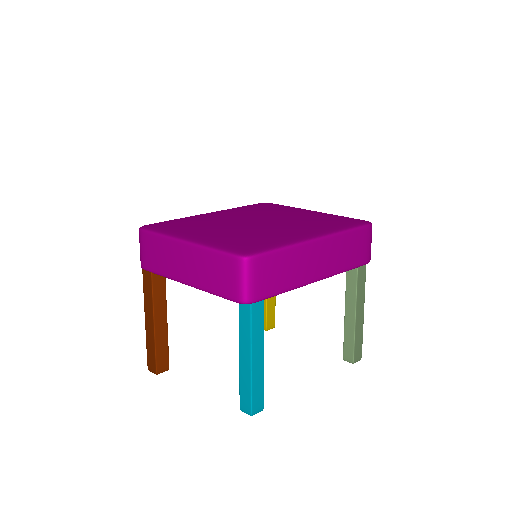}
    \end{subfigure}%
    \hfill%
    \begin{subfigure}[b]{0.11\linewidth}
	\centering
        \includegraphics[width=\linewidth]{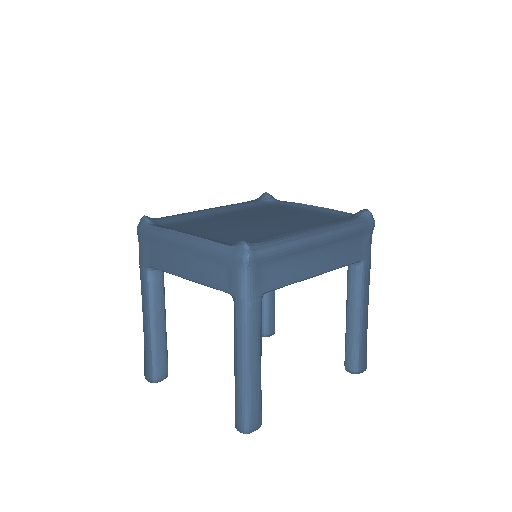}
    \end{subfigure}%
     \hfill%
    \begin{subfigure}[b]{0.11\linewidth}
        \centering
        \includegraphics[width=\linewidth]{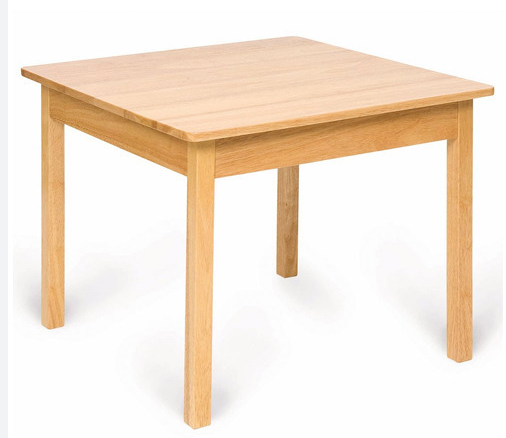}
    \end{subfigure}%
    \hfill%
    \begin{subfigure}[b]{0.11\linewidth}
        \centering
        \includegraphics[width=\linewidth]{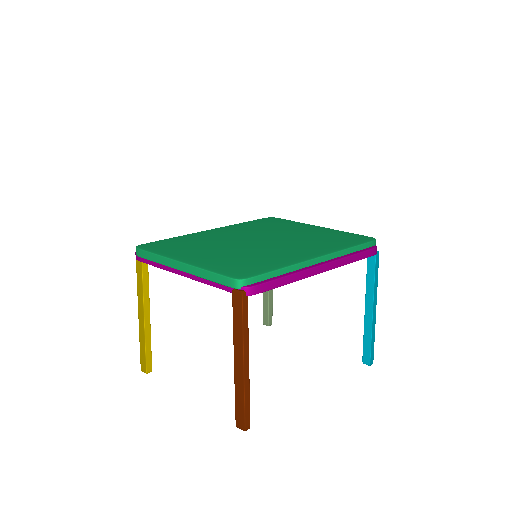}
    \end{subfigure}%
    \hfill%
    \begin{subfigure}[b]{0.11\linewidth}
	\centering
        \includegraphics[width=\linewidth]{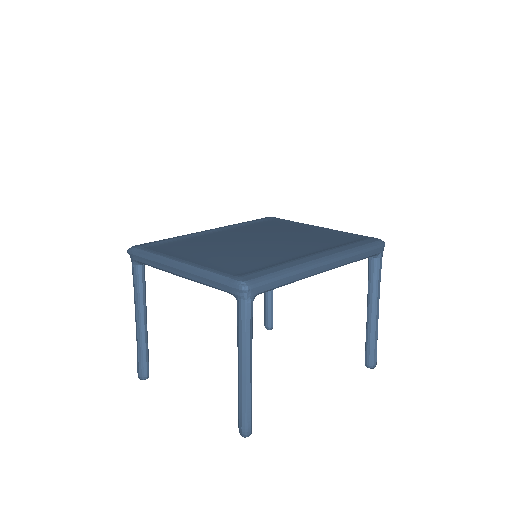}
    \end{subfigure}%
    \hfill%
    \begin{subfigure}[b]{0.11\linewidth}
        \centering
        \includegraphics[width=\linewidth]{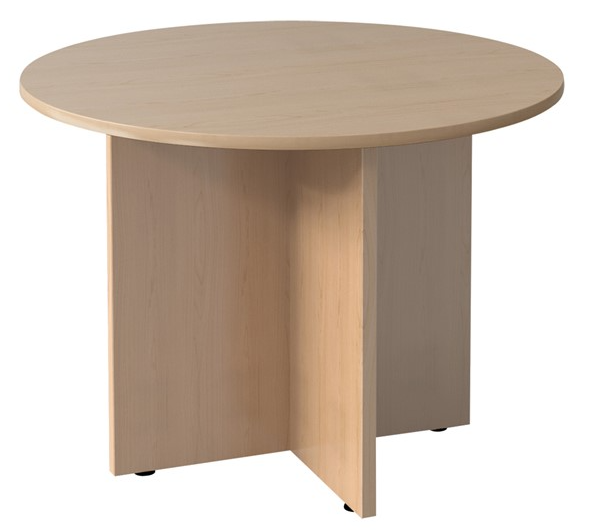}
    \end{subfigure}%
    \hfill%
    \begin{subfigure}[b]{0.11\linewidth}
        \centering
        \includegraphics[width=\linewidth]{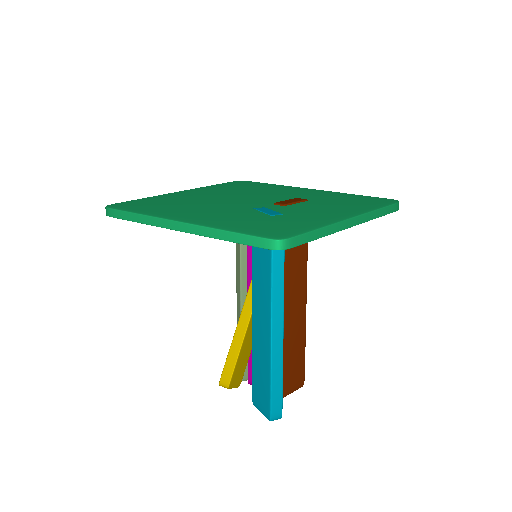}
    \end{subfigure}%
    \hfill%
    \begin{subfigure}[b]{0.11\linewidth}
	\centering
         \includegraphics[width=\linewidth]{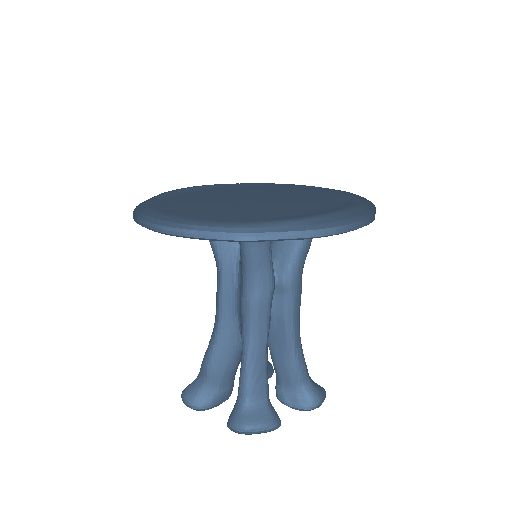}
    \end{subfigure}%
    \vskip\baselineskip%
    \begin{subfigure}[b]{0.11\linewidth}
        \centering
	\includegraphics[width=\linewidth]{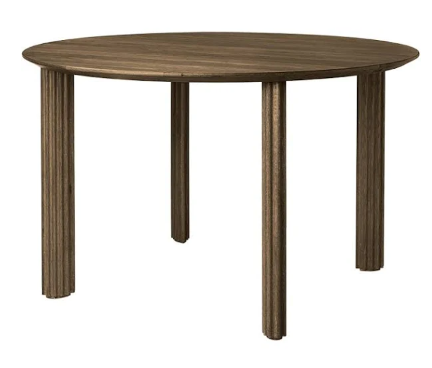}
    \end{subfigure}%
    \hfill%
    \begin{subfigure}[b]{0.11\linewidth}
        \centering
	\includegraphics[width=\linewidth]{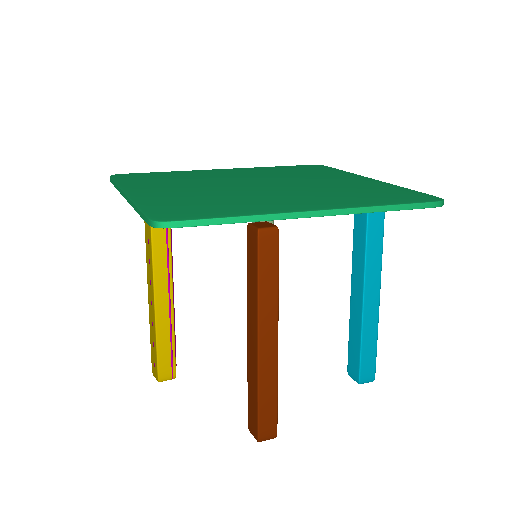}
    \end{subfigure}%
    \hfill%
    \begin{subfigure}[b]{0.11\linewidth}
	\centering
        \includegraphics[width=\linewidth]{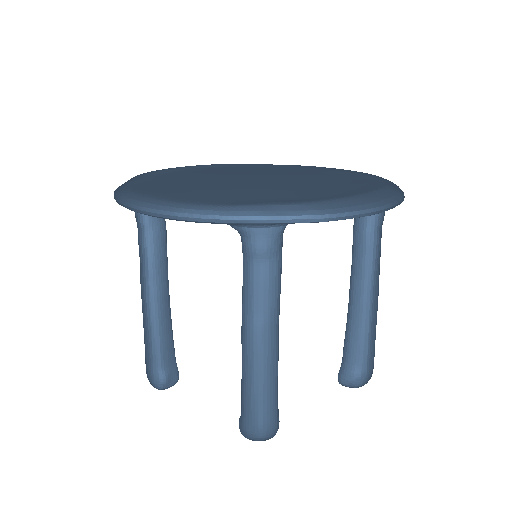}
    \end{subfigure}%
     \hfill%
    \begin{subfigure}[b]{0.11\linewidth}
        \centering
        \includegraphics[width=\linewidth]{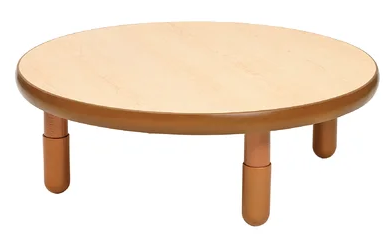}
    \end{subfigure}%
    \hfill%
    \begin{subfigure}[b]{0.11\linewidth}
        \centering
        \includegraphics[width=\linewidth]{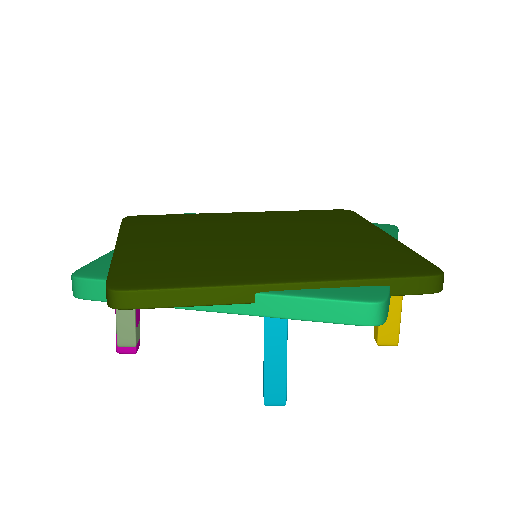}
    \end{subfigure}%
    \hfill%
    \begin{subfigure}[b]{0.11\linewidth}
	\centering
        \includegraphics[width=\linewidth]{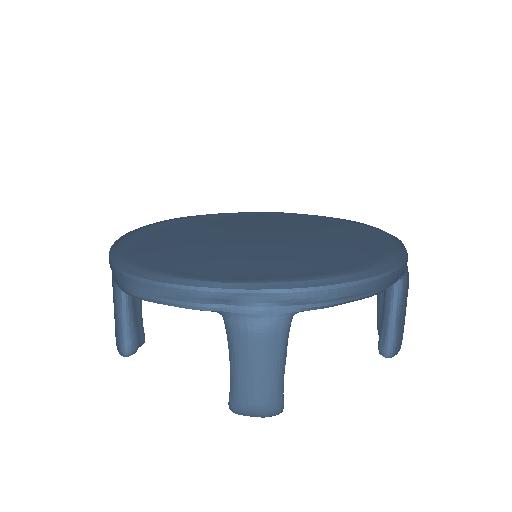}
    \end{subfigure}%
    \hfill%
    \begin{subfigure}[b]{0.11\linewidth}
        \centering
        \includegraphics[width=\linewidth]{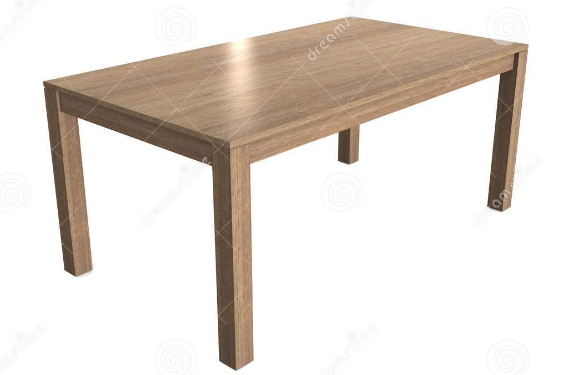}
    \end{subfigure}%
    \hfill%
    \begin{subfigure}[b]{0.11\linewidth}
        \centering
        \includegraphics[width=\linewidth]{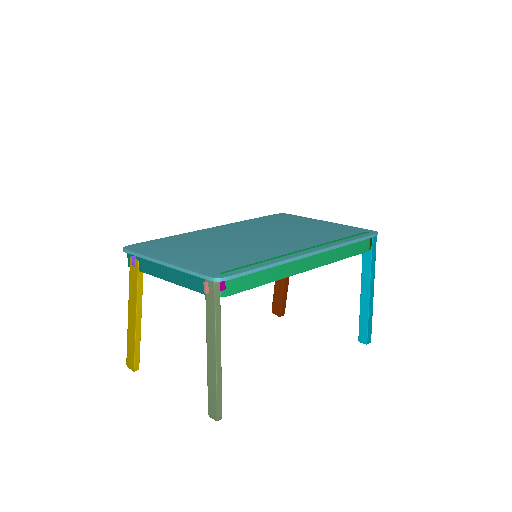}
    \end{subfigure}%
    \hfill%
    \begin{subfigure}[b]{0.11\linewidth}
	\centering
         \includegraphics[width=\linewidth]{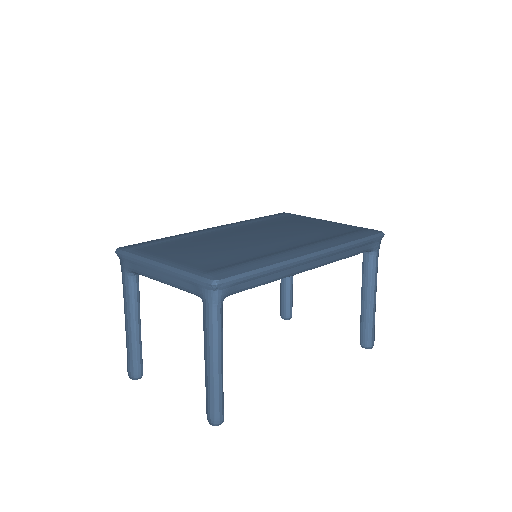}
    \end{subfigure}%
    \vskip\baselineskip%
    \begin{subfigure}[b]{0.11\linewidth}
        \centering
	\includegraphics[width=\linewidth]{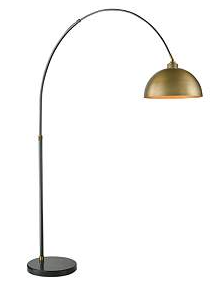}
    \end{subfigure}%
    \hfill%
    \begin{subfigure}[b]{0.11\linewidth}
        \centering
	\includegraphics[width=\linewidth]{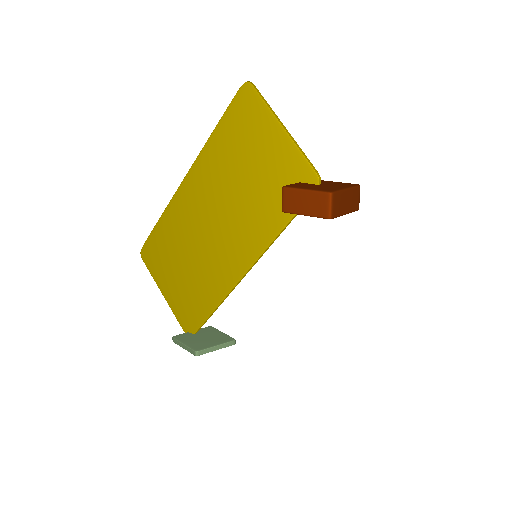}
    \end{subfigure}%
    \hfill%
    \begin{subfigure}[b]{0.11\linewidth}
	\centering
        \includegraphics[width=\linewidth]{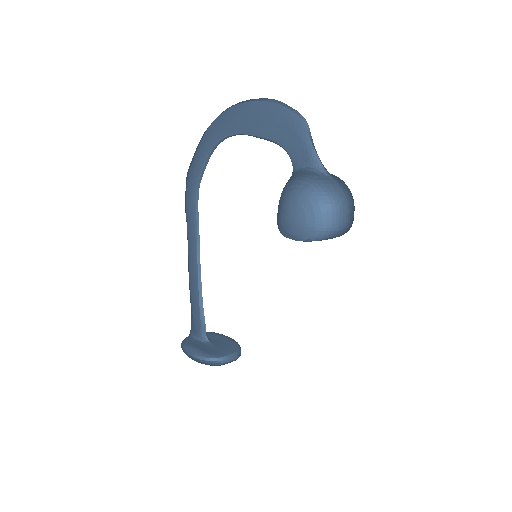}
    \end{subfigure}%
     \hfill%
    \begin{subfigure}[b]{0.11\linewidth}
        \centering
        \includegraphics[width=\linewidth]{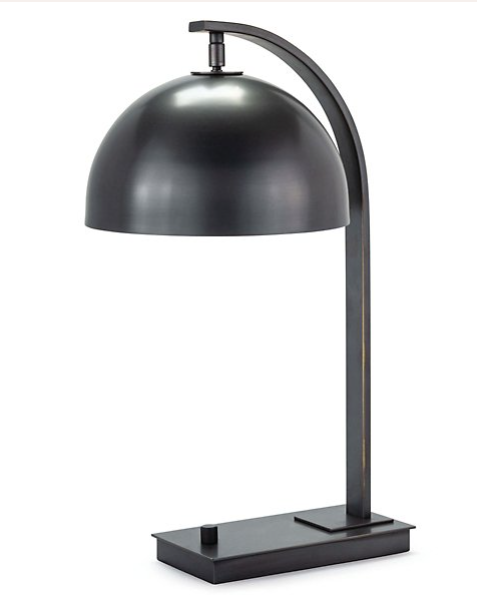}
    \end{subfigure}%
    \hfill%
    \begin{subfigure}[b]{0.11\linewidth}
        \centering
        \includegraphics[width=\linewidth]{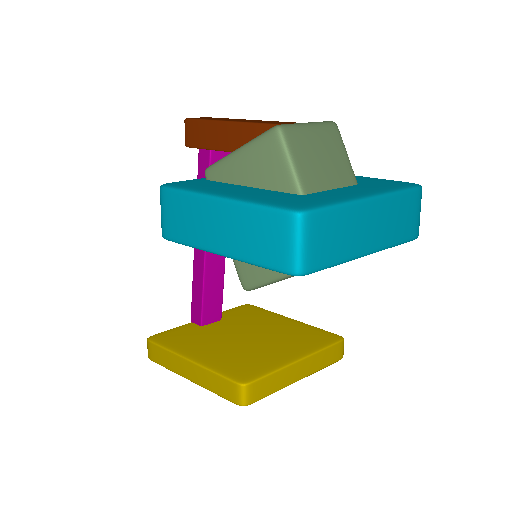}
    \end{subfigure}%
    \hfill%
    \begin{subfigure}[b]{0.11\linewidth}
	\centering
        \includegraphics[width=\linewidth]{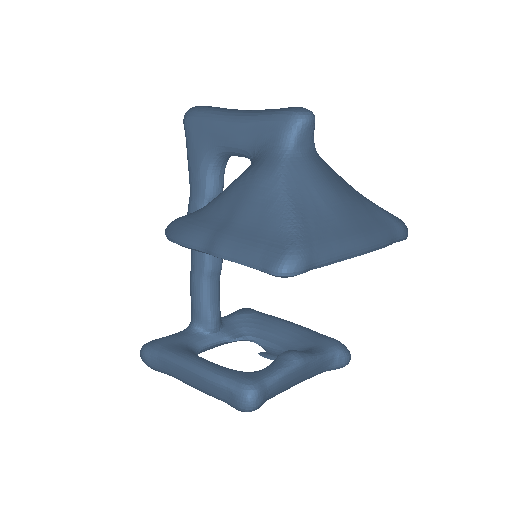}
    \end{subfigure}%
    \hfill%
    \begin{subfigure}[b]{0.11\linewidth}
        \centering
        \includegraphics[width=\linewidth]{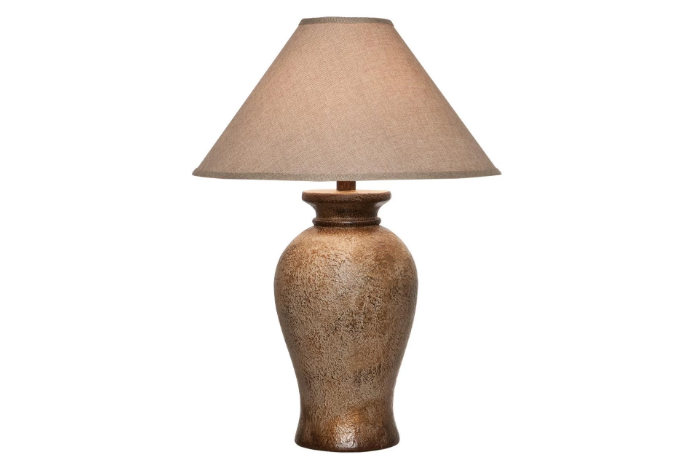}
    \end{subfigure}%
    \hfill%
    \begin{subfigure}[b]{0.11\linewidth}
        \centering
        \includegraphics[width=\linewidth]{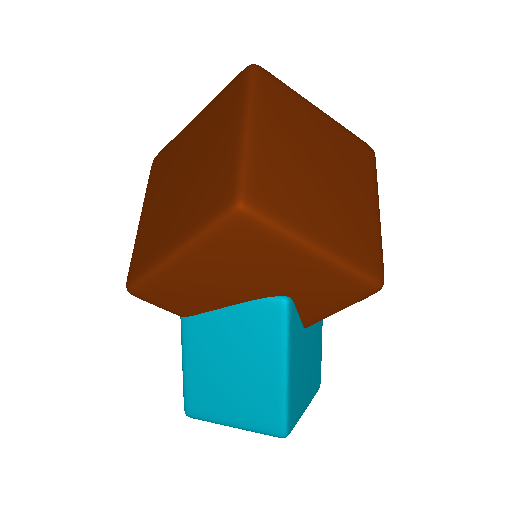}
    \end{subfigure}%
    \hfill%
    \begin{subfigure}[b]{0.11\linewidth}
	\centering
         \includegraphics[width=\linewidth]{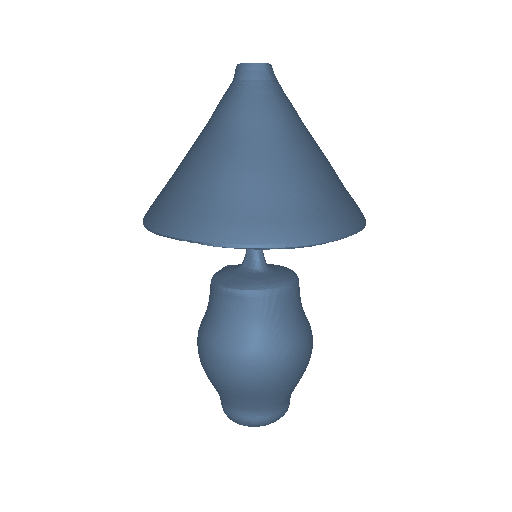}
    \end{subfigure}%
    \vskip\baselineskip%
    \begin{subfigure}[b]{0.11\linewidth}
        \centering
	\includegraphics[width=\linewidth]{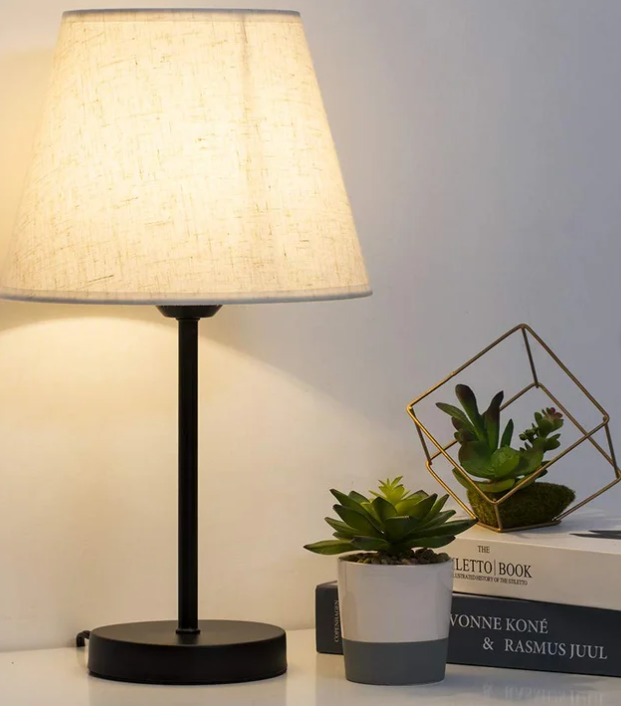}
    \end{subfigure}%
    \hfill%
    \begin{subfigure}[b]{0.11\linewidth}
        \centering
	\includegraphics[width=\linewidth]{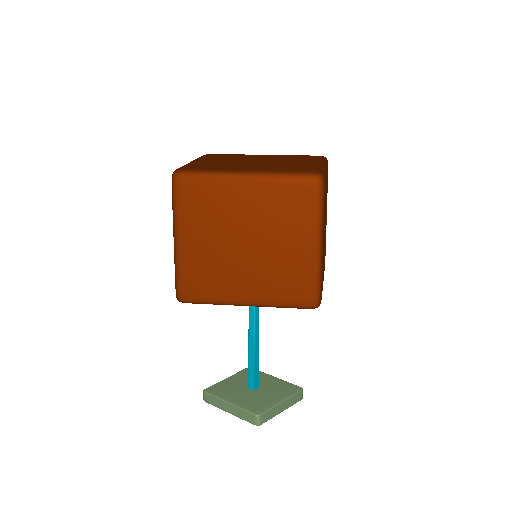}
    \end{subfigure}%
    \hfill%
    \begin{subfigure}[b]{0.11\linewidth}
	\centering
        \includegraphics[width=\linewidth]{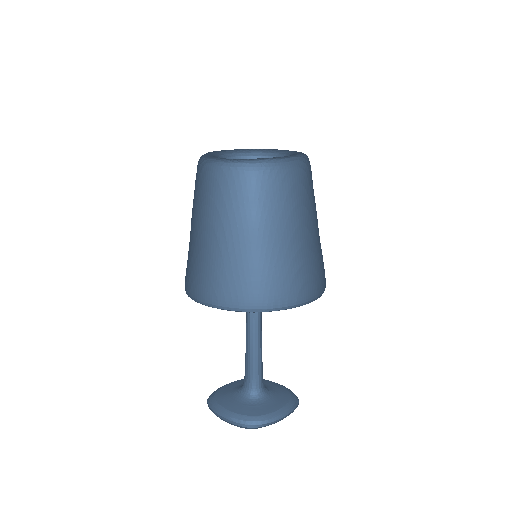}
    \end{subfigure}%
     \hfill%
    \begin{subfigure}[b]{0.11\linewidth}
        \centering
        \includegraphics[width=\linewidth]{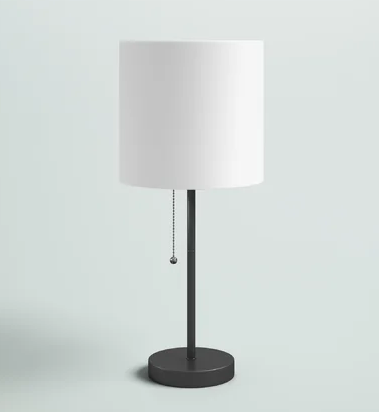}
    \end{subfigure}%
    \hfill%
    \begin{subfigure}[b]{0.11\linewidth}
        \centering
        \includegraphics[width=\linewidth]{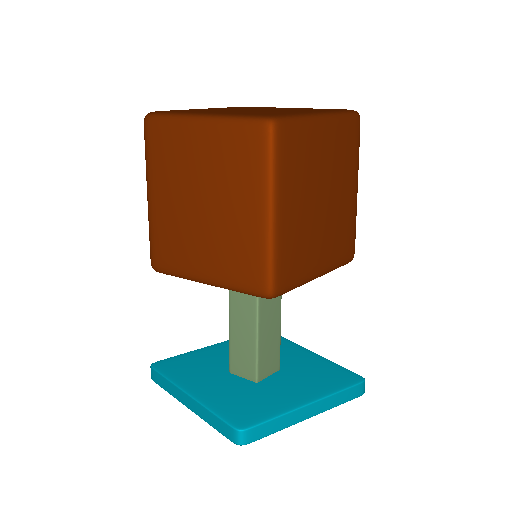}
    \end{subfigure}%
    \hfill%
    \begin{subfigure}[b]{0.11\linewidth}
	\centering
        \includegraphics[width=\linewidth]{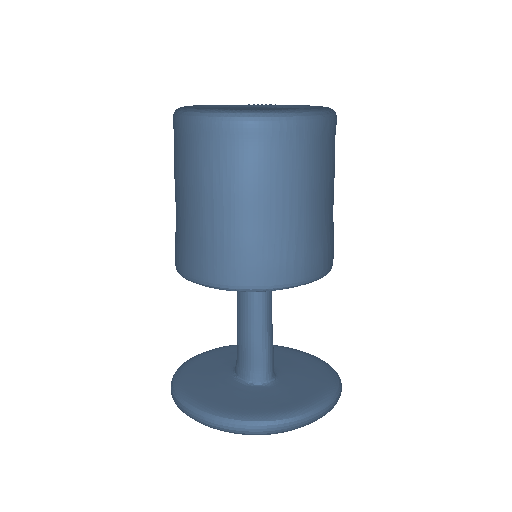}
    \end{subfigure}%
    \hfill%
    \begin{subfigure}[b]{0.11\linewidth}
        \centering
        \includegraphics[width=\linewidth]{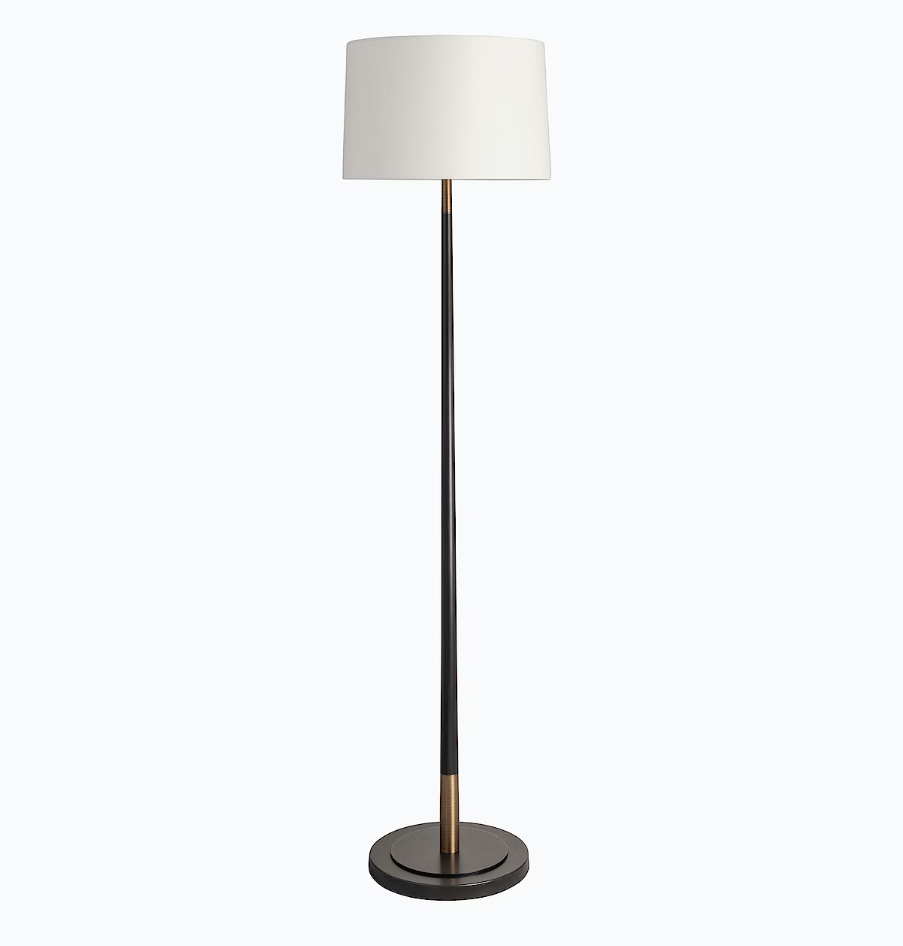}
    \end{subfigure}%
    \hfill%
    \begin{subfigure}[b]{0.11\linewidth}
        \centering
        \includegraphics[width=\linewidth]{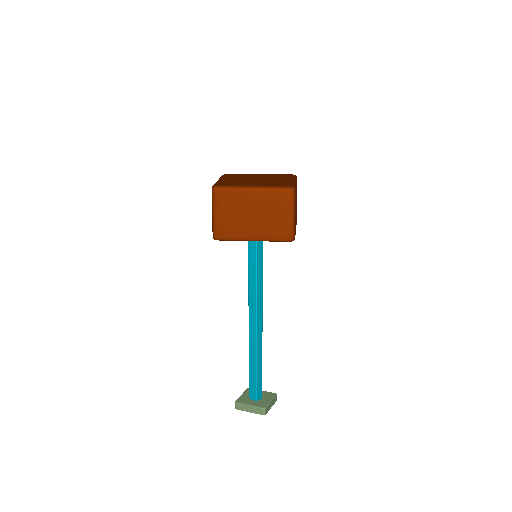}
    \end{subfigure}%
    \hfill%
    \begin{subfigure}[b]{0.11\linewidth}
	\centering
         \includegraphics[width=\linewidth]{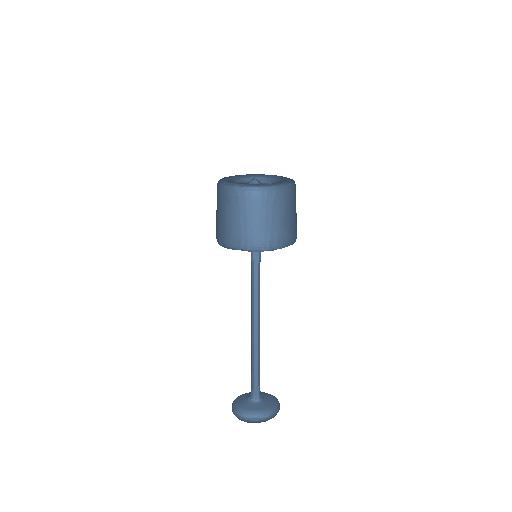}
    \end{subfigure}%
    \vskip\baselineskip%
    \caption{{\bf Image-guided Shape Generation}. Given different images
    our model can generate plausible 3D shapes of chairs, tables, and lamps. Note that for this
    experiment, we employ the variant of our model trained for language-guided
    training, without any re-training.}
    \label{fig:image_guided_generation_supp}
\end{figure*}

%% file: fig/part_variations.tex
\begin{figure*}
    \begin{subfigure}[t]{\linewidth}
    \centering
    \begin{subfigure}[b]{0.25\linewidth}
		\centering
		Input Shape
    \end{subfigure}%
    \begin{subfigure}[b]{0.25\linewidth}
        \centering
        Variation 1
    \end{subfigure}%
    \begin{subfigure}[b]{0.25\linewidth}
		\centering
       Variation 2
    \end{subfigure}%
    \begin{subfigure}[b]{0.25\linewidth}
        \centering
        Variation 3
    \end{subfigure}%
    \end{subfigure}
    \vspace{-1.5em}
    \begin{subfigure}[b]{0.25\linewidth}
        \centering
    \includegraphics[width=\linewidth]{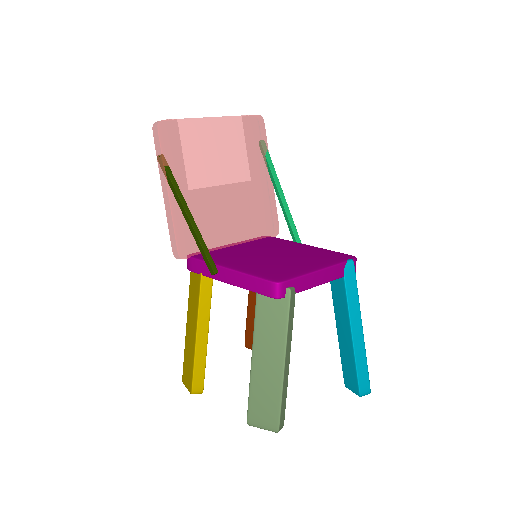}
    \end{subfigure}%
    \begin{subfigure}[b]{0.25\linewidth}
        \centering
    \includegraphics[width=\linewidth]{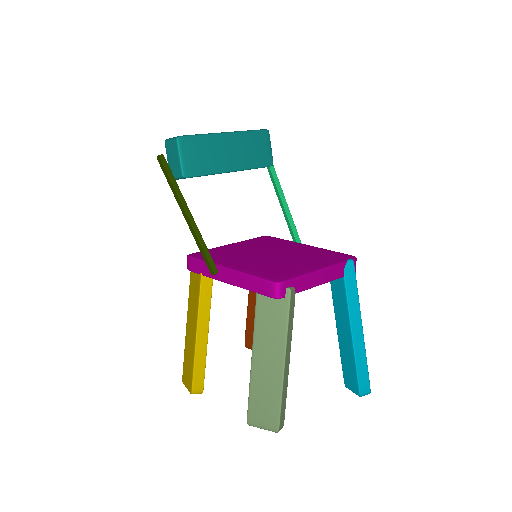}
    \end{subfigure}%
    \begin{subfigure}[b]{0.25\linewidth}
	\centering
        \includegraphics[width=\linewidth]{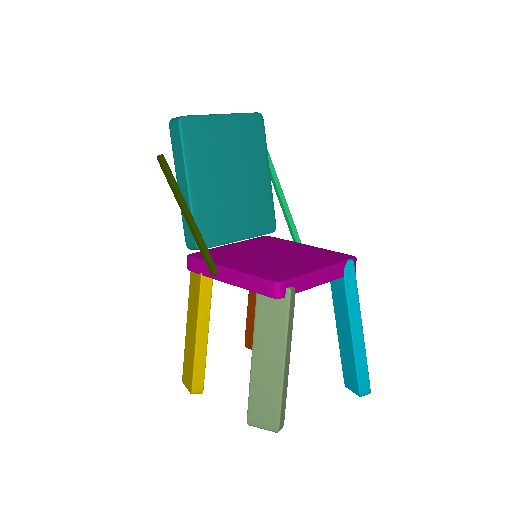}
    \end{subfigure}%
    \begin{subfigure}[b]{0.25\linewidth}
        \centering
        \includegraphics[width=\linewidth]{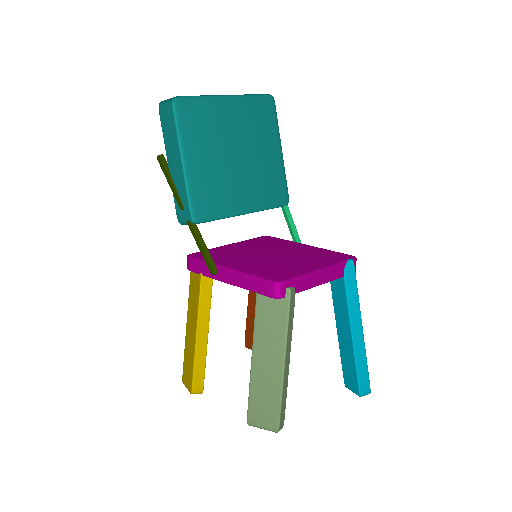}
    \end{subfigure}%
    \vskip\baselineskip%
    \begin{subfigure}[b]{0.25\linewidth}
        \centering
    \includegraphics[width=\linewidth]{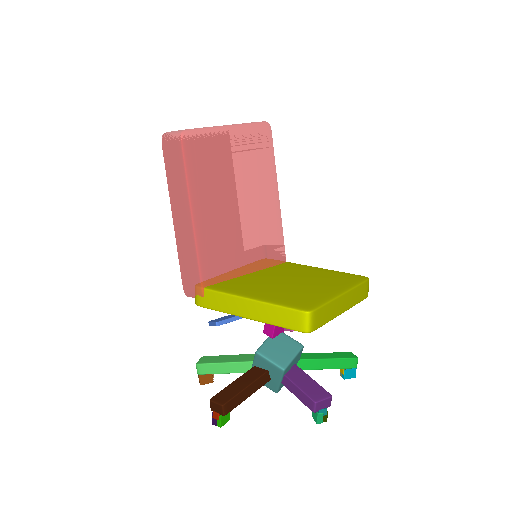}
    \end{subfigure}%
    \begin{subfigure}[b]{0.25\linewidth}
	\centering
        \includegraphics[width=\linewidth]{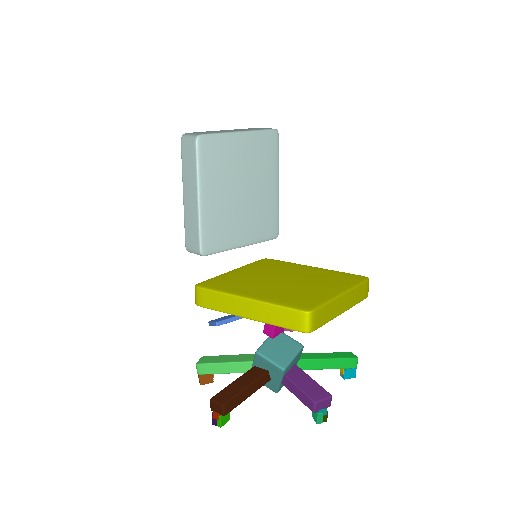}
    \end{subfigure}%
    \begin{subfigure}[b]{0.25\linewidth}
	\centering
        \includegraphics[width=\linewidth]{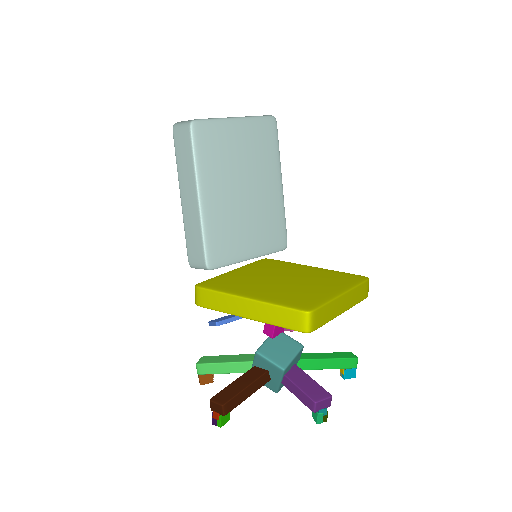}
    \end{subfigure}%
    \begin{subfigure}[b]{0.25\linewidth}
        \centering
        \includegraphics[width=\linewidth]{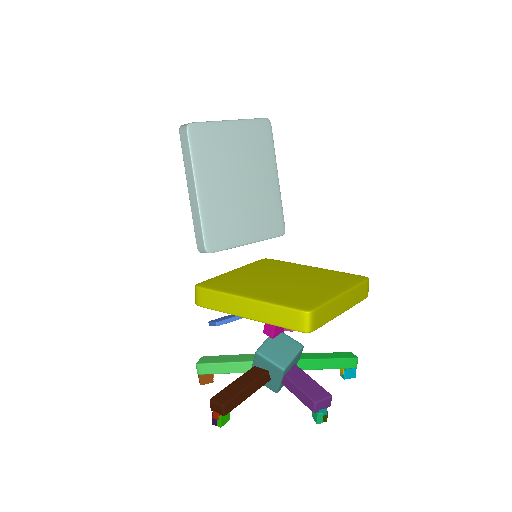}
    \end{subfigure}%
    \vskip\baselineskip%
    \vspace{-1.5em}
    \caption{{\bf Generating Part Variations} For the same part,
    highlighted with red, we showcase
    that our network can generate plausible part variations.}
    \label{fig:part_variation}
\end{figure*}

%% file: supp_discussion.tex
\section{Discussion and Limitations}
\label{sec:parts_discussion}

In this work, we devise PASTA, a part-aware generative model for 3D shapes.
Our architecture, consists of two main components: the \emph{object generator}
that autoregressively generates objects as sequences of labelled cuboids and
the \emph{blending network} that composes a sequence of cuboidal primitives and
synthesizes a high-quality implicit shape. We train the object generator to maximize the
log-likelihood of all part arrangements, in the dataset.  Unlike prior
part-based works \cite{Wu2020CVPR}, our model is simpler to train and only
requires part annotations in the form of cuboids and not the actual parts,
which are typically harder to acquire. Moreover, our blending network, which is implemented
with a transformer decoder, generates
3D shapes of high fidelity from the sequence of the generated cuboids. The supervision for training our blending
network comes in the form of watertight meshes. From our experimental
evaluation it becomes evident that our model outperforms existing part-based
and non-part based methods both on the task of shape generation and completion.
Furthermore, we showcase several applications of our model, such as language-
and image-guided shape generation.

Although, we believe that our model is an important step towards automating 3D
content creation, it has several limitations. Firstly, our model requires
part-supervision, which is difficulty to acquire, thus hindering applying 
our model to other data. Recent works
\cite{Hertz2022SIGGRAPH, Hao2020CVPR} proposed generative models that can be
trained without part-level annotations but as they are not autoregressive they
cannot be used for several completion tasks. We believe that an exciting direction
for future research is to explore whether we can learn an autoregressive
generative model of parts, without explicit part-level supervision. Note that this
task is not trivial, since training autoregressive models with teacher forcing
or schedule sampling requires part annotations.
Furthermore, while our model can generate plausible 3D geometries, in this work, we do not
consider the object's appearance. We believe that another interesting direction
for future research would be to explore learning a generative model of parts
with textures. This would unlock more editing operations both on the object's
geometry and appearance. In our current setup, this can be easily done, simply by
replacing our blending network with a NeRF-based decoder~\cite{Mildenhall2020ECCV}
that instead of only predicting occupancies, predicts colors and opacity values
for the set of query points.

\section{Potential Negative Impact on Society}
\label{sec:negative_societal_impact}

Our proposed model enables generating part-aware 3D shapes as well as several
editing operations such as size-, image- and language-guided generation. While, we
see our work as an important step towards automatic content creation and
enabling a multitude of editing functionalities, it can also lead to negative
consequences, when applied to sensitive data, such as human bodies. 
Therefore, we believe it is imperative to always check the license of any
publicly available 3D model. In addition, we see the development of techniques
for identifying real from synthetic data as an essential research direction that
could potential prevent deep fake. While throughout this work, we have only worked
with publicly available datasets, we recommend that future users that will train our model
on new data to remove biases from the training data in order to ensure that our model
can fairly capture the diversities in terms of shapes and sizes.